\documentclass[runningheads]{llncs}

\usepackage{eccv}

\usepackage{eccvabbrv}

\usepackage{graphicx}
\usepackage{booktabs}
\usepackage[acronym]{glossaries}
\usepackage{multirow}
\usepackage{relsize}
\usepackage[table]{xcolor}
\usepackage{comment}
\usepackage{wrapfig}
\usepackage{caption, copyrightbox}

\newcommand{\g}[1]{{\leavevmode\smaller[2]\color{gray}#1}}
\definecolor{green}{rgb}{0.0, 0.68, 0.33}
\definecolor{red}{rgb}{1.0, 0.0, 0.0}
\definecolor{blue}{rgb}{0.0, 0.44, 0.75}
\definecolor{orange}{rgb}{0.97, 0.58, 0.15}
\definecolor{purple}{rgb}{0.45, 0.19, 0.61}
\definecolor{grey}{rgb}{0.5, 0.5, 0.5}
\newcommand{\green}[1]{{\leavevmode\color{green}#1}}
\newcommand{\red}[1]{{\leavevmode\color{red}#1}}
\newcommand{\blue}[1]{{\leavevmode\color{blue}#1}}
\newcommand{\orange}[1]{{\leavevmode\color{orange}#1}}
\newcommand{\purple}[1]{{\leavevmode\color{purple}#1}}

\newacronym{nerf}{NeRF}{Neural Radiance Fields}
\newacronym{sdf}{SDF}{Signed Distance Field}
\newacronym{mvs}{MVS}{Multi-View Stereo}
\newacronym{nir}{NIR}{Near-Infrared}
\newacronym{ms}{MS}{Multispectral}
\newacronym{pol}{Pol}{Polarization}
\newacronym{tof}{ToF}{Time-of-Flight}
\newacronym{mlp}{MLP}{Multi-Layer Perceptron}
\newacronym{roi}{RoI}{Region of Interest}
\newacronym{multimodalstudio}{MMS}{MultimodalStudio}
\newacronym{mms-fw}{MMS-FW}{MultimodalStudio Framework}
\newacronym{nxdc}{NXDC}{Normalized Cross-Device Coordinates}
\newacronym{sfm}{SfM}{Structure-from-Motion}
\newacronym{gs}{GS}{Gaussian Splatting}
\newacronym{sh}{SH}{Spherical Harmonics}
\newacronym{brdf}{BRDF}{Bidirectional Reflectance Distribution Function}
\newacronym{aop}{AoP}{Angle of Polarization}
\newacronym{dop}{DoP}{Degree of Polarization}
\newacronym{method-name}{SPoILeR}{Spectral and Polarimetric Implicit Learned Representation}
\newacronym{dif}{DiF}{Dictionary Fields}
\newacronym{pt}{PT}{pre-training}
\newacronym{ft}{FT}{fine-tuning}
\newacronym{psnr}{PSNR}{Peak Signal-to-Noise Ratio}
\newacronym{ssim}{SSIM}{Structural Similarity Index Measure}
\newacronym{lpips}{LPIPS}{Learned Perceptual Image Patch Similarity}
\newacronym{hsi}{HS}{Hyperspectral}
\newacronym{vit}{ViT}{Vision Transformer}

\usepackage[accsupp]{axessibility}  

\usepackage[pagebackref,breaklinks,colorlinks,citecolor=eccvblue]{hyperref}

\usepackage{orcidlink}

\begin{document}

\title{Learning Spectral and Polarimetric Clues for One-to-Multimodal Novel View Synthesis} 

\titlerunning{Learning Spect. and Pol. Clues for One-to-Multimodal Novel View Synthesis}

\author{Federico Lincetto\inst{1}\orcidlink{0009-0002-9137-1482} \and
Gianluca Agresti\inst{2}\orcidlink{0000-0001-7072-0079} \and
Mattia Rossi\inst{2}\orcidlink{0000-0001-5158-2395} \and Piergiorgio Sartor\inst{2} \and Pietro Zanuttigh\inst{1}\orcidlink{0000-0002-9502-2389}}

\authorrunning{F.~Lincetto et al.}

\institute{MEDIA Lab, University of Padova, Padova, Italy\\
\email{\{federico.lincetto,zanuttigh\}@dei.unipd.it} \and
Sony EUISPC, Sony Semiconductor Solutions Europe, Stuttgart, Germany\\
\url{https://medialab.dei.unipd.it/paper_data/SPoILeR/}}

\maketitle

\begin{abstract}
Neural rendering techniques allow for accurate reconstruction of the geometry and color appearance of 3D scenes. Some methods have extended their use to additional imaging modalities, such as multispectral, infrared, or polarimetric data.
However, all of these approaches require expensive sensors and calibrated setups to capture new multimodal frames for each new scene. We propose \gls{method-name}, a novel method to obtain multi-view consistent renderings of unconventional modalities for scenes where either only RGB frames or very few of the additional modalities are available. Thanks to a multimodal pre-training phase, the model learns the mutual correlation between different modalities. This step allows predicting accurate renderings of unconventional modalities during a fine-tuning phase supervised only by RGB images. Experimental results show that the approach can accurately render infrared, polarimetric, and multispectral frames for scenes where no input sample captured by these types of sensors is provided.
\end{abstract}

\section{Introduction}
\label{sec:intro}
\begin{figure}[t]
    \centering
    \includegraphics[width=\linewidth]{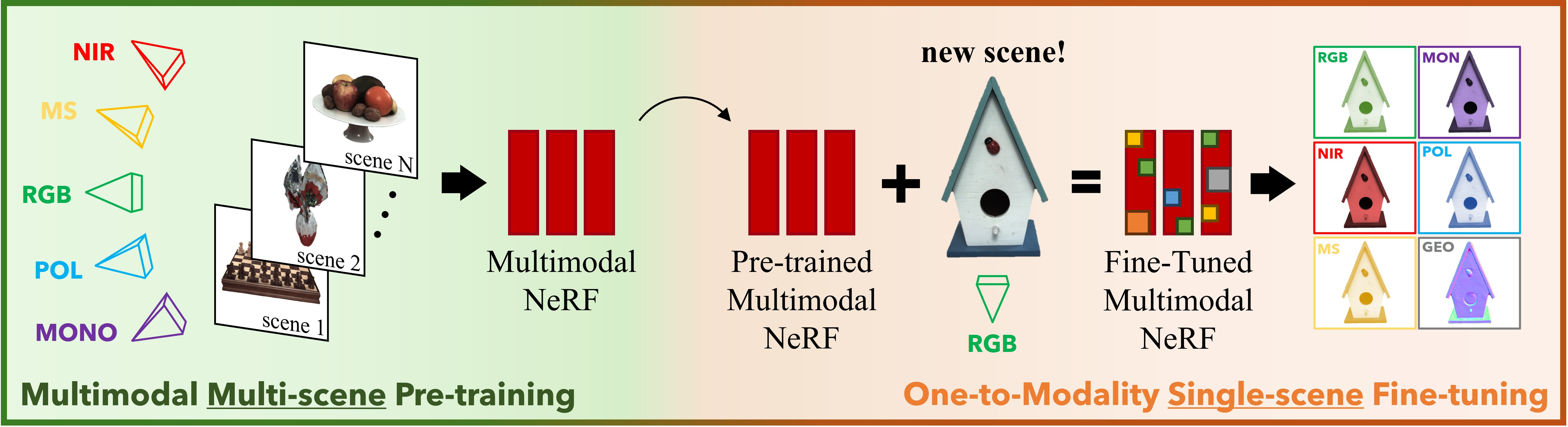}
    \caption{The proposed approach firstly learns the correlation across multiple modalities on a collection of scenes. Then, a single-scene fine-tuning on RGB data alone produces a model able to render views of arbitrary modalities for the considered scene. }
    \label{fig:abstract}
    \vspace{-0.5cm}
\end{figure}

\gls{nerf}~\cite{mildenhall2020nerf} represented a turning point for Novel View Synthesis by permitting the rendering of images with a photorealistic appearance and accurate multi-view consistency.
\gls{nerf} evolved rapidly to solve also related tasks, such as multi-view 3D geometry reconstruction~\cite{wang2021neus,yariv2021volume,lincetto2023emp,li2023neuralangelo,wang2023neus2,wang2024neurodin}, the inverse rendering problem~\cite{wu2025pbr,zhang2022iron,xu2023renerf,jin2023tensoir,boss2021nerd}, and 3D scene semantic segmentation~\cite{boss2021nerd,chen2022sem2nerf,chou2024gsnerf,kerr2023lerf,Liu_2023_ICCV}. 
\gls{gs}~\cite{kerbl20233dgs} further boosted rendering accuracy at the cost of higher memory requirements and promoted the revision of many tasks previously addressed with \gls{nerf}~\cite{huang20242d,dai2024high,zhang2025ref,liang2024gs,qin2024langsplat,qu2024goi}. 
However, the rapid development of NeRF and GS was possible due to the large amount of available RGB data, which is unquestionably the most popular imaging representation but definitely not the only one. There exists a long-standing interest in exploiting information captured by different imaging sensors to solve complex tasks or enhance result quality. For example, there exist \gls{ms} sensors that are sensitive to different bands of visible light, \gls{nir} sensors that allow sensing radiation with wavelength over 700 nm, and \gls{pol} sensors that can measure the polarization of light. Employing \gls{ms} and \gls{nir} data, it is possible to perform tasks such as physics-based rendering~\cite{darling2011real,kim2023neural}, image content enhancement~\cite{tsagaris2005multispectral,hashimoto2011multispectral,zhou2016fusion}, and solve ill-posed classification problems~\cite{grossmann2022improving,yin2007multispectral,saponaro2015material,wang2010improved}. Similarly, polarimetric data provide additional information on surface orientation and material properties, which is particularly helpful for the problems of shape-from-polarization~\cite{ba2020deep,lei2022shape}, material properties estimation~\cite{li2024neisf,li2025neisf++,dave2022pandora}, and reflection removal~\cite{lei2020polarized,yao2025polarfree}. There exist works that exploit data different from RGB for novel view synthesis, 3D reconstruction, or inverse rendering purposes, and that are based on \gls{nerf}~\cite{li2024spec,li2024spectralnerf,perez2025unmix,ma2024novel,ye2024thermal,ozer2024exploring,xu2024leveraging,li2024neisf,dave2022pandora,HASSAN2025103345} or \gls{gs}~\cite{thirgood2025hypergs,chen2024thermal3d,luthermalgaussian,han2025polgs,wang2025polarimetric}. However, these methods focus on either a single or a couple of modalities, while just a few works perform novel view synthesis by including a wider set of multiple imaging modalities~\cite{lincetto2025,poggi2022cross,kim2023neural,guo2025cross}. One limitation common to all of these works is that they require multiple multi-view consistent images of the same scene, captured by sensors that are more expensive and cumbersome to deploy compared to RGB cameras, thus limiting their practical applicability. Data availability is scarce, and this also prevents the training of large foundation models for these tasks. Moreover, methods that perform conversion from RGB to a different modality~\cite{cai2022mst++,zhang2025polaranything}  lack multi-view consistency.\\
\indent In this work, we build a multimodal neural representation that overcomes these limitations by learning the mutual correlation that exists between different modalities, given a generic set of scenes for which multimodal information is available. Once done, the learned knowledge can be exploited to render accurate and multi-view consistent novel views of unconventional modalities of a new scene for which either only RGB or a very limited number of other modality frames are available. For this purpose, we introduce several novel contributions:

\begin{itemize}
    \item A multimodal multi-scene \gls{pt} procedure able to encode the information captured by different sensors from different scenes in a multimodal latent space. The \gls{pt} also optimizes decoders for every sensor, capable of mapping the latent space into explicit multimodal radiance information.
    \item A \gls{ft} pipeline that can operate in a novel scene even if captured with a limited subset of sensors. The \gls{ft} phase, due to the \gls{pt} knowledge, estimates the radiance information of missing modalities, thus allowing the model to render photorealistic and multi-view consistent novel views without the need for modality-specific sensor captures of the new scene.
    \item A regularization procedure consisting of three ad-hoc loss functions that encourage the model to first build a robust and explainable multimodal latent space during the \gls{pt} and then to preserve the knowledge even with a lack of supervision from other modalities during the \gls{ft} phase.
    \item An extensive analysis of the model performance and capabilities with respect to previous multimodal state-of-the-art solutions for spectro-polarimetric modality-to-modality conversion~\cite{cai2022mst++,zhang2025polaranything} and novel view synthesis tasks~\cite{lincetto2025}.
\end{itemize}
\section{Related Work}
\label{sec:related}

\subsubsection{Multimodal Neural Rendering}
As anticipated in~\cref{sec:intro}, there exist methods that tackle the novel view synthesis task by involving multiple imaging modalities. X-NeRF~\cite{poggi2022cross} is a precursor in this field: it uses a traditional \gls{nerf}-based method to learn the radiance field of forward-facing scenes captured with RGB, \gls{nir}, and \gls{ms} sensors. 
Recently, a new interpretation based on \gls{gs} has been released, which reduces the rendering time and improves quality\cite{guo2025cross}. Another method for multimodal novel view synthesis is NeSpoF~\cite{kim2023neural}, which leverages images with a known polarization state for every band of the visible spectrum to learn a spectro-polarimetric radiance field. It models the dependence of polarization on spectral bands for novel view synthesis, but the data acquisition is very demanding in terms of hardware and time. Finally, MultimodalStudio~\cite{lincetto2025} is a recent \gls{nerf}-based method that enables novel view synthesis and 3D reconstruction from a wider set of heterogeneous modalities. However, all mentioned methods require capturing a complete set of multimodal images for every new scene to reconstruct. Generally, sensors different from RGB cameras are expensive, and the acquisition procedure may be more complex and time-consuming. With \gls{method-name}, we address this limitation by reducing the need for multimodal images to a one-off pre-training step that learns the mutual correlations between modalities. This allows one to perform fine-tuning trainings on novel scenes captured by a single modality (\eg, RGB) and still be able to render multimodal frames, thus eliminating the need for expensive multimodal acquisitions.

\vspace{-0.15cm}
\subsubsection{Modality-to-modality Conversion}
\label{sec:related_m2m}
Several feed-forward methods have been proposed to estimate different modalities from RGB images. 
Multi-MAE~\cite{bachmann2022multimae} and 4M~\cite{NEURIPS2023_b6446566,NEURIPS2024_71883294} are popular examples of models that can convert a single RGB frame to a variety of different semantic and geometric modalities. There exist also multi-view methods like RoRE~\cite{griffiths2026rore} that can achieve consistent novel view synthesis across multiple modalities. However, all these feed-forward models require extensive paired data for training, and thus, due to the lack of spectral and polarimetric data, they are not always viable solutions, particularly when dealing with custom sensors for which only small datasets exist. On the other hand, StylizedNeRF~\cite{stylizednerf} and StylizedGS~\cite{stylizedgs} enable the possibility of replicating the style of a single conditioning image on novel views. However, when using spectral or polarimetric images as conditioning, they cannot assign the physically correct spectral radiance to the scene objects.
Regarding RGB to \gls{hsi} conversion, some methods are based on CNN~\cite{shi2018hscnn+}, \gls{vit}~\cite{dosovitskiyimage,wu2024multistage}, or they recursively invoke a reconstruction module for a coarse-to-fine estimation~\cite{mengfrn}.
A popular robust baseline for RGB to \gls{hsi} conversion is MST++~\cite{cai2022mst++}, the winner of the NTIRE 2022 Spectral Recovery Challenge~\cite{arad2022ntire}.
It is based on a \gls{vit} that employs Spectral-wise Multi-head Self-attention blocks to learn the correlations between RGB and \gls{hsi} spatial and spectral information to perform spectral restoration. Considering polarimetric data, the recently published PolarAnything~\cite{zhang2025polaranything} work converts from RGB to \gls{pol} by estimating the \gls{aop} and \gls{dop}. For this purpose, the authors fine-tuned Stable Diffusion~\cite{rombach2022high} with synthetic polarimetric data. However, these methods share a common limitation: they predict information that is not multi-view consistent because they consider only a single RGB view at a time. Employing these frameworks to convert a set of multi-view RGB images to \gls{ms} or \gls{pol} data leads to results that exhibit clear inconsistencies between different views. Therefore, in practice, they are not reliable enough to support multimodal novel view synthesis. In contrast, \gls{method-name} can produce multimodal renderings that are multi-view consistent by construction, given that it optimizes a single 3D representation. In~\cref{sec:comparison_with_third_party}, we validate these claims by comparing \gls{method-name} with state-of-the-art modality conversion strategies such as MST++ and PolarAnything.

\vspace{-0.15cm}
\subsubsection{Cross-scene Novel View Synthesis}
\label{sec:related_multiscene}
Vanilla \gls{nerf} and \gls{gs} are generally optimized for single scenes. However, there exist methods that propose training on several scenes to enable zero-shot novel view synthesis of new scenes~\cite{varmaattention,zhao2025generalizable,hao2023cp-nerf,zhang2023mmnerf,charatan2024pixelsplat,caesarnerf,jain2021putting,golf-nrt}.  All these methods rely on CNN or \gls{vit} models which are known to be data hungry for training. On the other hand, \gls{dif}~\cite{Chen2023TOG} proposes an implicit representation decoupled into coefficient and basis modules, suitable for addressing the few-shot novel view rendering task as well. The key advantage of this approach is that its architecture explicitly induces a separation of cross-scene and scene-specific information, without requiring large amounts of data to train. In \gls{method-name}, we reinterpret this concept to promote the learning of shared multimodal knowledge. Unlike \gls{dif}, our \gls{ft} step is dense in terms of multi-view coverage; instead, we leverage the basis and coefficient architecture to address the much sparser modality coverage of the \gls{ft} with respect to the \gls{pt}. Moreover, unlike \gls{dif}, we explicitly define and regularize a multimodal latent space that learns to model the mutual correlations between modalities.
\section{Method}
\label{sec:method}

In this section, we present \gls{method-name}, a multimodal \gls{nerf}-based approach with generalization capabilities across different scenes and imaging modalities.
As shown in \cref{fig:abstract}, we assume to have a generic set of training scenes with multi-view images captured by different sensors. The goal is to learn a multimodal radiance field of a new scene for which only RGB or very few modalities are available. This allows for the recovery of an accurate and multi-view consistent multimodal representation that enables the photorealistic novel view synthesis of any modality included in the set of sensors. For this purpose, we first propose to learn multimodal knowledge during a \gls{pt} phase from a set of scenes. Then, a subsequent \gls{ft} phase adapts only part of the model to new scenes. 

\subsection{Model Architecture}
\label{sec:model_architecture}
The \gls{method-name} \gls{nerf}-based architecture is built on top of MultimodalStudio \cite{lincetto2025}, a state-of-the-art approach for multimodal neural rendering. As standard NeRF models, \gls{mms-fw} is scene-specific, thus it cannot transfer knowledge between different scenes.
\begin{table}[t]
    \begin{minipage}{\linewidth}
        \begin{minipage}{0.575\linewidth}
            \begin{flushleft}
                \includegraphics[width=\linewidth]{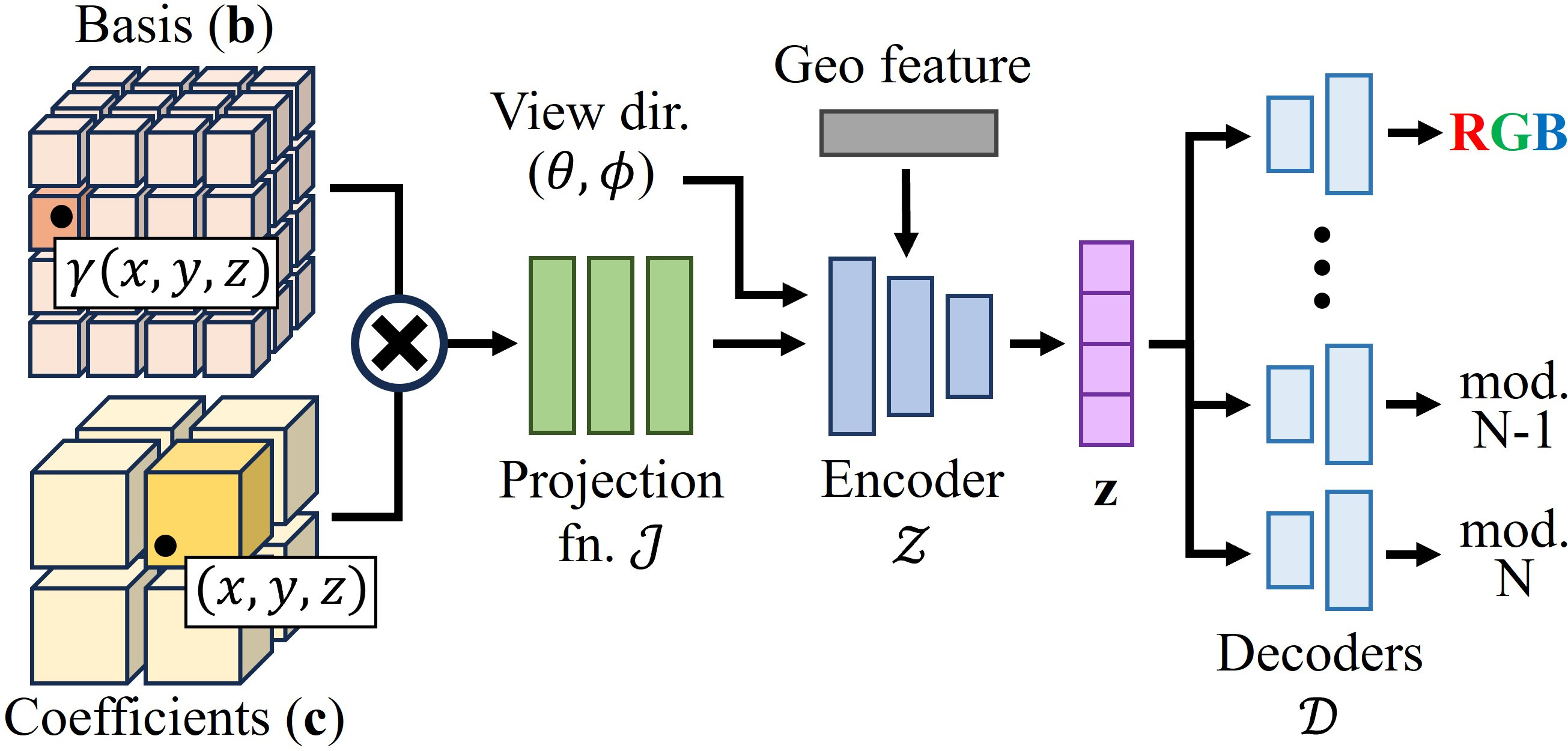}
                \vspace{0.1cm}
                \captionof{figure}{Radiance module architecture scheme: features from basis and coefficients are combined and decoded into different radiance modalities.}
                \label{fig:architecture}
            \end{flushleft}
        \end{minipage}
    \hfill
        \begin{minipage}{0.39\linewidth}
            \begin{flushright}
                \includegraphics[width=\linewidth]{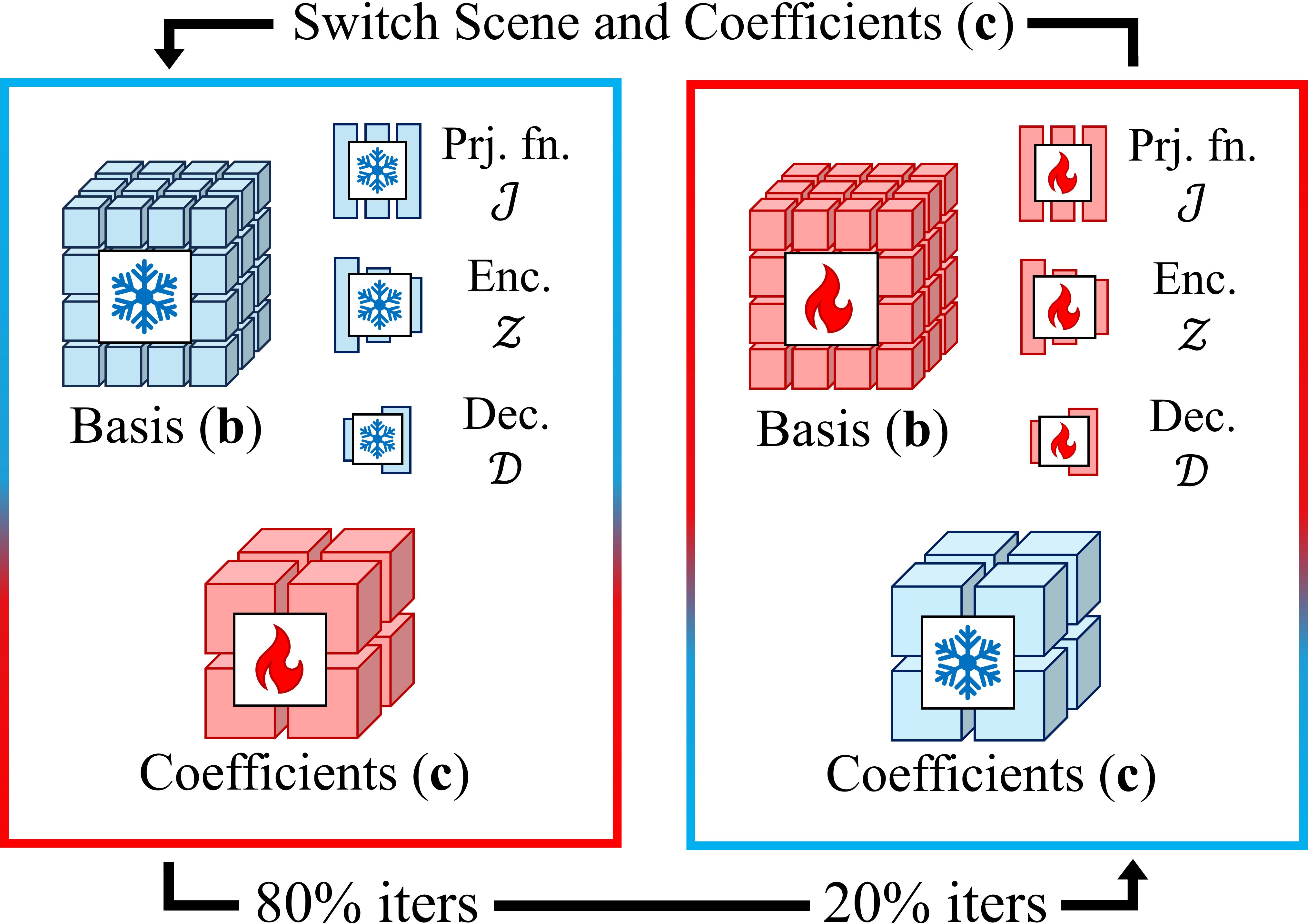}
                \vspace{0.1cm}
                \captionof{figure}{PT routine visualization: scene-specific and shared modules are alternatively optimized.}
                \label{fig:pt_schema}
            \end{flushright}
        \end{minipage}
    \end{minipage}
    \vspace{-0.8cm}
\end{table}
In contrast, our goal is to build a strategy that infers missing modalities even when the supervision is limited to a single modality. Therefore, we employ an architecture that allows us to store and preserve the learned multimodal priors, but is also flexible enough to estimate the geometry and spectral properties of specific scenes. We take inspiration from the \gls{dif} architecture~\cite{Chen2023TOG} and substitute the multi-resolution hash-grids adopted by \gls{mms-fw} with two sets of learnable feature grids that act as basis and coefficient fields, respectively, as shown in \cref{fig:architecture}.
Our model trains on frames of $N$ different modalities, whose channel sum is $C=\sum_{l=1}^{N}C_l$. Following the notation of~\cite{Chen2023TOG}, we estimate a radiance signal $\textbf{s}:\mathbb{R}^3\times\mathbb{S}^2\rightarrow\mathbb{R}^C$, with  $\textbf{r}(\theta, \phi)=(\sin\theta\cos\phi, \sin\theta\sin\phi, \cos\theta) $ and $\mathbb{S}^2=\left\{\textbf{r}(\theta, \phi) | 0 \le \theta < \pi, 0 \le \phi < 2\pi  \right\}$,  defined as follows:
\begin{equation}
    \textbf{s}(\textbf{x}, \textbf{v})=\mathcal{P}(\textbf{c}(x)^\top\textbf{b}(\gamma(\textbf{x})), \textbf{v}),
\end{equation}
where $\textbf{x}=(x,y,z)$ is the 3D position,  $\textbf{v}=\textbf{r}(\theta, \phi)\in \mathbb{S}^2$ the camera viewing direction,  $\textbf{c}:\mathbb{R}^3\rightarrow\mathbb{R}^D$ and $\textbf{b}:\mathbb{R}^{3\times D}\rightarrow\mathbb{R}^D$ the coefficient and basis fields, and $D$ the learnable feature dimension. The function $\gamma:\mathbb{R}^3\rightarrow\mathbb{R}^{3\times D}$ is a periodic coordinate mapping and $\mathcal{P}:\mathbb{R}^D\rightarrow\mathbb{R}^C$ is a function that predicts radiance values. The basis field $\textbf{b}$ and the function $\mathcal{P}$ are common to all scenes and responsible for learning the radiance correlation shared between different modalities. The coefficient field $\textbf{c}$ is optimized specifically for each scene to account for the scene-specific properties, such as textures and reflections.
The periodic mapping function $\gamma$, which is a multi-frequency sawtooth function, allows us to apply the same basis to different coordinates, thus promoting their generalization.
In our formulation, $\mathcal{P}$ is the combination of multiple functions: a projection function $\mathcal{J}:\mathbb{R}^D\rightarrow\mathbb{R}^D$, a latent encoder $\mathcal{Z}:\mathbb{R}^D\rightarrow\mathbb{R}^Z$, with  $Z$ as the latent space dimensionality, and a modality decoder $\mathcal{D}_l:\mathbb{R}^Z\rightarrow\mathbb{R}^{C_l}$ specific to each modality. 
The projection function $\mathcal{J}$ is responsible for manipulating the product of basis and coefficient features; the latent encoder $\mathcal{Z}$ receives the projection function output, concatenates it with the viewing direction $\textbf{v}$ and a geometry feature $\textbf{g}$ from the geometry module and returns the latent vector $\textbf{z}$; the decoders $\mathcal{D}_l$ decode \textbf{z} into the desired modality channels. Overall, $\mathcal{P}$ is defined as follows:
\begin{equation}
    \mathcal{P}(\textbf{f},\textbf{v}, \textbf{g}, l)=\mathcal{D}_l(\mathcal{Z}(\textbf{v}, \textbf{g}, \mathcal{J}(\textbf{f}))),
\end{equation}
with $\textbf{f}=\textbf{c}^\top\textbf{b}$. \gls{mms-fw} estimates geometry and radiance with two dedicated modules. 
The per-scene geometry module is based on \gls{sdf}~\cite{gibson1998using,osher2004level} and returns a density value and a geometry feature $\textbf{g}$. We maintain the geometry module unaltered. All the functions forming $\mathcal{P}$ are implemented as shallow \gls{mlp} in order to place most of the model capacity in the coefficient and basis fields. This is because we need a latent space that is not too compressed, as it must be easily and consistently estimable even when only a small subset of supervision modalities is available. One more reason is that we aim to split the information into either general or scene-specific: general information must be stored by the basis field to be available in any scenario, while scene-specific information must be stored only in the coefficient field, as it must be completely switched when the scene changes.\\
\indent The model is trained by minimizing an objective function that includes a photometric loss averaged between modalities $\mathcal{L}_{mod}$, the eikonal loss $\mathcal{L}_{eik}$, and curvature loss $\mathcal{L}_{curv}$ for geometry regularization~\cite{amos2020implicit,li2023neuralangelo}, and three new regularization losses ($\mathcal{L}_{lsg}$, $\mathcal{L}_{inv}$, and $\mathcal{L}_{m2l}$, detailed in~\cref{sec:losses}) that promote a more robust and explainable latent space. The full loss function is defined as follows:
\begin{equation}
    \mathcal{L}=\mathcal{L}_{mod}+\lambda_{eik}\mathcal{L}_{eik}+\lambda_{curv}\mathcal{L}_{curv}+\lambda_{lsg}\mathcal{L}_{lsg}+\lambda_{inv}\mathcal{L}_{inv}+\lambda_{m2l}\mathcal{L}_{m2l},
\end{equation}
where the $\lambda$ terms are the weighting coefficients (see \cref{sec:impl_details}).

\subsection{Pre-training Pipeline: Shared Modules Optimization}
\label{sec:pt_pipeline}
As anticipated previously, the model training is organized in two distinct phases: a \acrlong{pt} phase and a \acrlong{ft} phase. The \gls{pt} phase is meant to let the model learn a basis representation for the radiance, able to generalize well between different scenes, and the correlation that exists between different imaging modalities and different scene properties. For this purpose, the model is trained to estimate the radiance field of several scenes captured by the complete set of multimodal sensors. At the same time, the unique radiance and geometry properties of every scene are estimated by independent scene-specific coefficient fields and modules, respectively. All other modules, namely the projection function, the latent encoder, and the modality decoders, are shared between scenes. 
To promote an effective separation of shared and specific information, the training freezes shared and scene-specific modules alternately, with a fixed schedule, as shown in \cref{fig:pt_schema}. When loading a scene, the corresponding coefficient module is loaded too, and the model is trained for $B$ iterations. The scene-specific modules (namely, coefficients and geometry module) and the shared modules are trained for $80\%$ and $20\%$ of the $B$ iterations, respectively. Then, a new scene is loaded, coefficients are swapped, and the process is repeated. It is important to carefully consider the role of the geometry feature $\textbf{g}$ produced by the geometry model and passed as input to the latent encoder $\mathcal{Z}$, part of the radiance module. Previous works introduced this feature to improve geometry estimation~\cite{yariv2021volume}.
However, in our setting, there exists the possibility that part of the radiance information, that should be stored in the shared modules to remain available during the \gls{ft} process, is instead stored in the geometry feature, which is scene-specific.
We found that the best solution to prevent this is to employ a dropout of its gradient.
During the \gls{pt}, only the gradient of one random modality per iteration can back-propagate through the geometry feature.
We observe that this solution produces good geometry estimation while preserving the shared radiance information for the \gls{ft} step.

\subsection{Fine-tuning Pipeline: Coefficient Field Optimization}
In the \gls{ft} phase, we get a new scene, not available during the \gls{pt} phase, captured only by a subset of the imaging modalities. Leveraging the pre-trained model, the goal is to recover photorealistic renderings of modalities not available in the \gls{ft} phase but available in the \gls{pt}. 
For this reason, we need to exploit the knowledge stored in the pre-trained shared modules but, given that the multimodal supervision signal is now limited, we also need to preserve the learned correlations between modalities as much as possible to avoid catastrophic forgetting. At the same time, the model must be flexible to learn the new scene-specific geometry and radiance. Therefore, during \gls{ft} we freeze the shared modules and optimize only the scene-specific radiance coefficients and geometry field. As a result, the much smaller number of learnable parameters than during \gls{pt} leads to a much shorter training time than re-training a complete model from scratch on the new scene.
The training objective for the \gls{ft} matches the \gls{pt} one, except for $\mathcal{L}_{lsg}$ which is disabled since the latent space geometry must be preserved at this stage.

\subsection{Latent Regularization Strategies}
\label{sec:losses}

Despite the proposed architectural design and training schedule, as experienced through empirical evaluation, if the model is trained only on a subset of the modalities, the optimization is prone to diverge for the modalities not optimized. For this reason, we introduce 3 latent regularization losses: 1) the latent space geometry loss, 2) the inverse function loss, and 3) the modality-to-luma loss.

\vspace{-0.15cm}
\subsubsection{Latent Space Geometry Loss}

\begin{table}[t]
    \begin{minipage}{\linewidth}
        \begin{minipage}{0.53\linewidth}
            \begin{flushleft}
                \includegraphics[width=\textwidth]{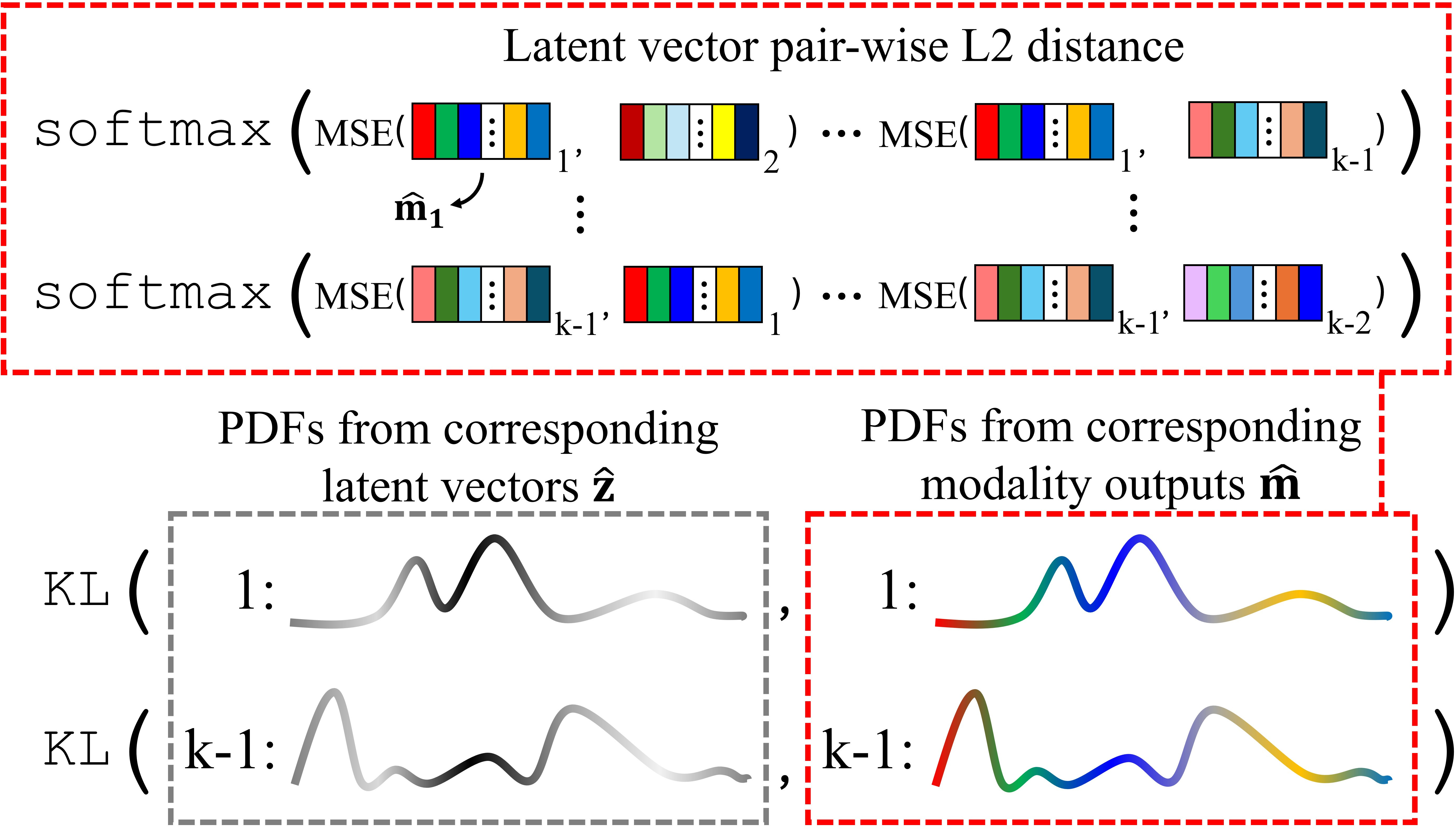}
                \vspace{0.11cm}
                \captionof{figure}{The latent space geometry loss encourages the model to produce latents whose pair-wise distance distribution matches the radiance pair-wise distance distribution.}
                \label{fig:latent_space_geo_loss}
            \end{flushleft}
        \end{minipage}
    \hfill
        \begin{minipage}{0.43\linewidth}
            \begin{flushright}
               \includegraphics[width=\linewidth]{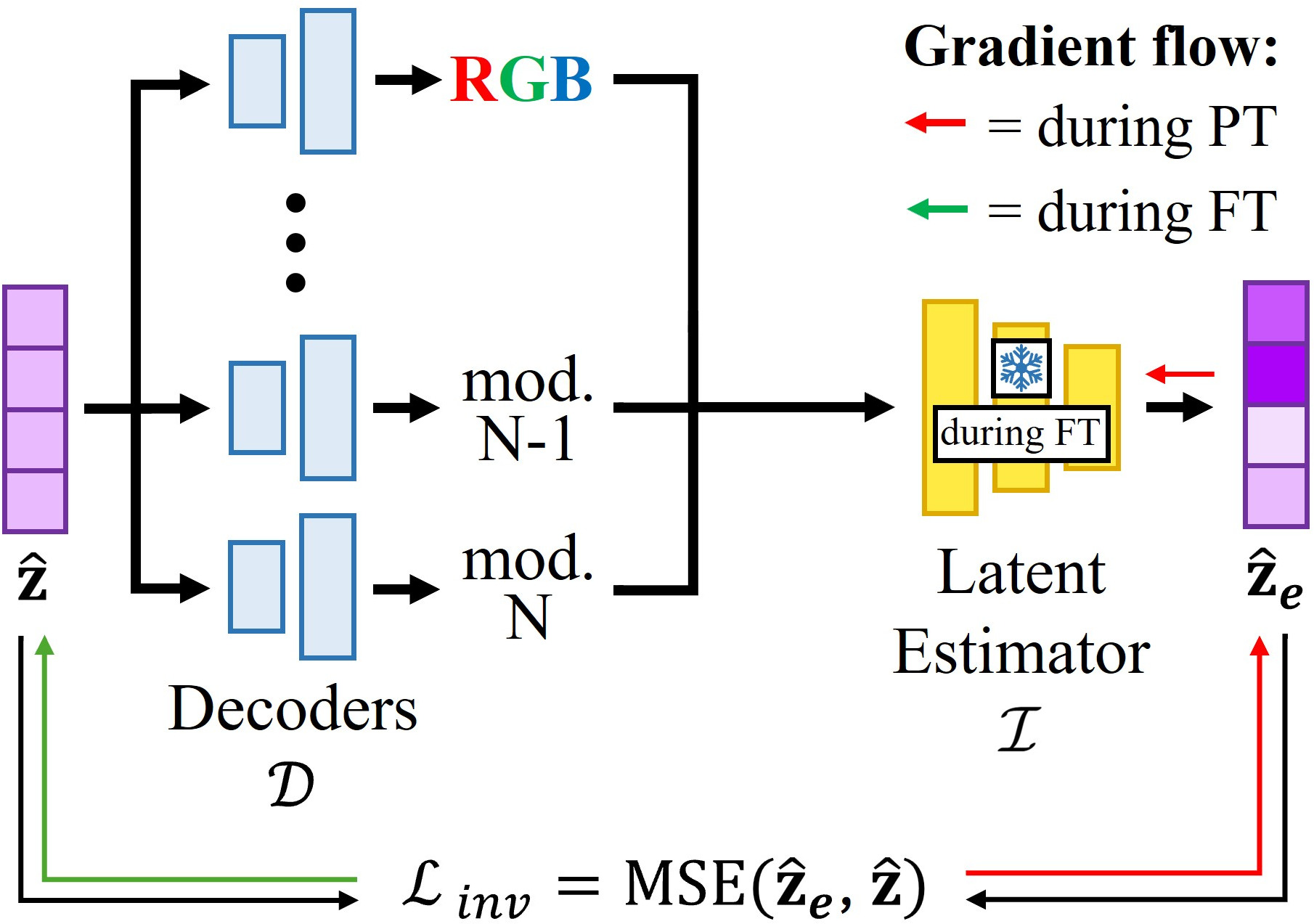}
                \vspace{0.1cm}
                \captionof{figure}{Scheme of the inverse function module $\mathcal{I}$. It predicts the estimated latent vector $\hat{\textbf{z}}_\textbf{e}$ from multimodal radiance values $\left\{\hat{\textbf{m}}_1, ..., \hat{\textbf{m}}_k\right\}$.}
                \label{fig:inverse_function_loss}
            \end{flushright}
        \end{minipage}
    \end{minipage}
    \vspace{-0.8cm}
\end{table}

This regularization loss aims to promote the latent space explainability. Considering that the multimodal supervision is strongly limited in the \gls{ft}, it is reasonable to observe the model estimating suboptimal latents. We observed that the decoding procedure is sensitive to small latent perturbations because similar latents are sometimes decoded into very different radiance values. 
This issue reduces the capability of the \gls{ft} phase to obtain accurate renderings of the missing modalities. To overcome this issue, inspired by the embedding loss proposed in~\cite{Camuffo2025}, we introduce the latent space geometry loss $\mathcal{L}_{lsg}$. The goal is to build a latent space where the distances between latent representations reflect the distances between the corresponding decoded modalities, as shown in \cref{fig:latent_space_geo_loss}. Unlike in~\cite{Camuffo2025} where the aim is to align two latent spaces, here we treat the modality space as if it were the latent space whose distribution has to be matched. Let's consider a set of $k$ latent vectors  $\left\{\hat{\textbf{z}}_1, ..., \hat{\textbf{z}}_k\right\}$ and the corresponding set of decoded modality channels $\left\{\hat{\textbf{m}}_1, ..., \hat{\textbf{m}}_k\right\}$, with $\hat{\textbf{m}}_k$ being the concatenation of all $N$ modality channels such as $\hat{\textbf{m}}_k=(C_{k1}, ..., C_{kN})$. 
We compute the pairwise Euclidean distance between all pairs of latents and modality channels independently, thus building two $(k-1)\times(k-1)$ matrices, $D_z$ and $D_m$. Then, we compute the softmax row-wise in order to extract two PDFs for each pair of latent vectors and the corresponding modality channels, and we minimize the difference between the two by applying the Kullback–Leibler (KL) divergence, as shown in \cref{fig:latent_space_geo_loss}. Therefore, $\mathcal{L}_{lsg}$ is defined as follows:
\begin{equation}
    \mathcal{L}_{lsg}=\texttt{KL}(\texttt{softmax}_{row}(-D_z), \texttt{softmax}(-D_m)).
\end{equation}
Note that the ``hat'' notation above disambiguates the latent vectors $\textbf{z}$, predicted for each ray sample, from the latent vector $\hat{\textbf{z}}$, which is the result of the integration of latent vectors along each ray. This solution is adopted to reduce the computational demand that otherwise would be unaffordable. Finally, we disable this loss during the \gls{ft}, as the latent space geometry is already formed.

\vspace{-0.15cm}
\subsubsection{Inverse Function Loss}
The second regularization loss that we propose aims to promote the latent space consistency between the \gls{pt} and the \gls{ft} phases. 
\begin{wrapfigure}[13]{tr}{0.55\linewidth}
  \centering
  \includegraphics[width=\linewidth]{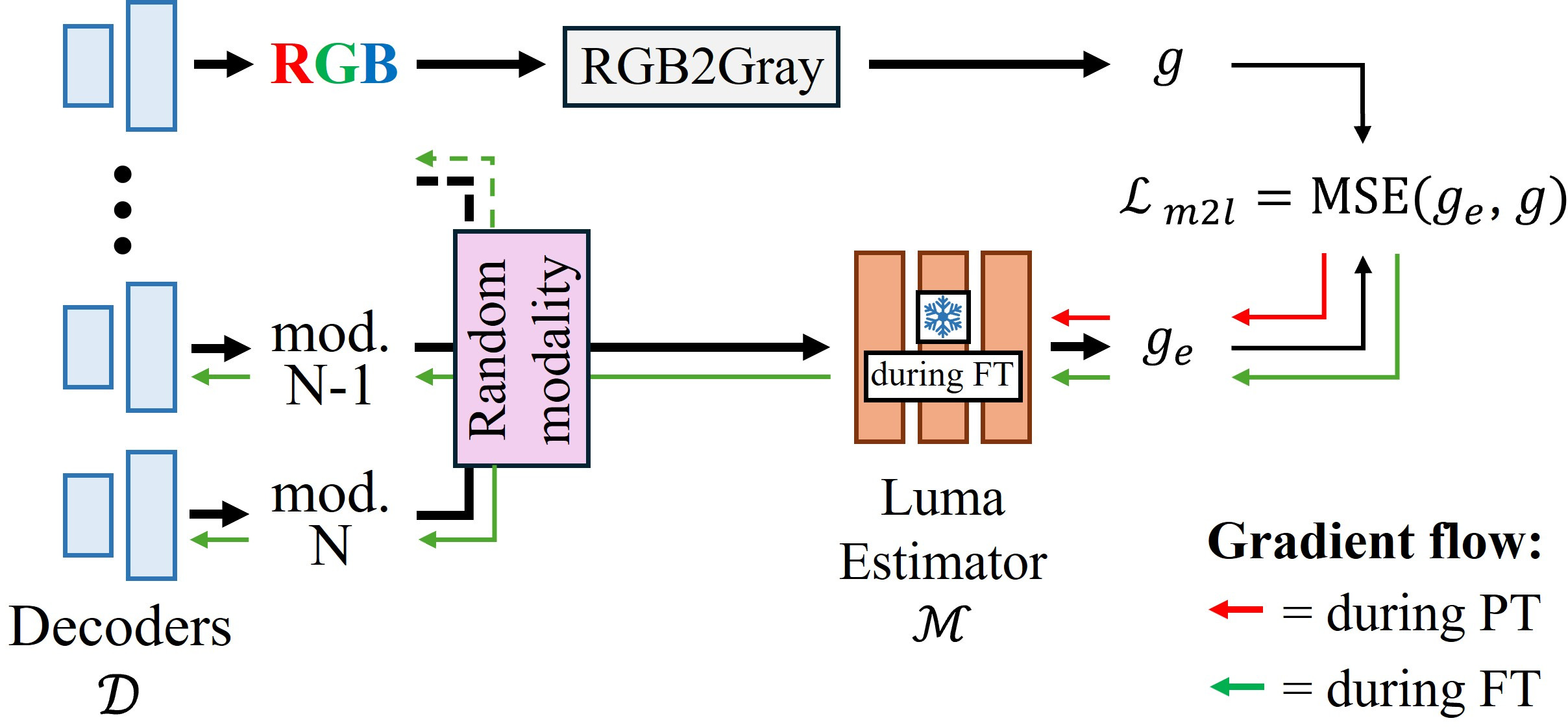}
  \caption{Scheme of the modality-to-luma module $\mathcal{M}$. It encourages consistency between the reference RGB luminance $g$ and the estimated luminance $g_e$ predicted from one random modality.}
  \label{fig:modality2luma}
\end{wrapfigure} 
As shown in \cref{fig:inverse_function_loss}, we train a \gls{mlp} module that, given the modality radiance values decoded from a latent vector, estimates the latent vector itself. We refer to it as inverse function because it performs the opposite operation done by the latent space decoders. Specifically, the inverse function $\mathcal{I}:\mathbb{R}^C\rightarrow\mathbb{R}^Z$ is defined as $\hat{\textbf{z}}_e=\mathcal{I}_\Theta(\hat{\textbf{m}})$.
This module is trained during the \gls{pt} phase to learn the mapping between modality space and latent space. In this phase, we prevent the gradient from back-propagating through the whole model to avoid influencing the latent space formation. Then, in the \gls{ft} phase, the inverse function module is frozen, and the estimated latent vector acts as a reference for the latent vector encoded by $\mathcal{Z}$. We compute the mean squared error between the actual and estimated latent space and back-propagate the gradient through the whole model to guide the optimization of the coefficients $\textbf{c}$. Therefore, the loss is defined as $\mathcal{L}_{inv}=\texttt{MSE}(\hat{\textbf{z}}_e,  \hat{\textbf{z}})$.\\
\indent The purpose of this loss is to encourage the \gls{ft} latent space to be coherent with the latent space learned during \gls{pt}; otherwise, it is likely that the latent space diverges and, consequently, the renderings of the estimated modalities.

\vspace{-0.15cm}
\subsubsection{Modality-to-luma Loss}
The last regularization loss is the modality-to-luma loss function. As shown in \cref{fig:modality2luma}, we train a shallow \gls{mlp} that estimates the function $\mathcal{M}:\mathbb{R}^{M_l}\rightarrow\mathbb{R}$, where $M_l$ refers to the number of channels of modality $l$. Given all channels of a selected modality $\hat{\mathbf{m}}_l$, this function returns a grayscale value $g_e$, defined as $ g_e=\mathcal{M}_\Theta(\hat{\textbf{m}}_l, l)$, that matches the luminance $g$ computed from the RGB channels.
Therefore, the modality-to-luma loss is computed as $\mathcal{L}_{m2l}=\texttt{MSE}(g_e, g)$.
In this case, as before, we train the module during the \gls{pt} by detaching the gradient, then use it frozen during  \gls{ft}.
In our case, all modalities integrate light from the visible spectrum; therefore, they are indeed correlated. However, some of them (\eg, near-infrared, mono, and polarization) also integrate part of the infrared spectrum. Because of this, we observed that the model tends to ignore their correlation with other modalities and decouple their representations into more orthogonal ones.  
The modality-to-luma loss encourages the model to preserve the modality correlation during \gls{ft}, when complete multimodal supervision is missing. However, its effectiveness is subject to the chosen modality set because it is not always the case that their values are correlated with RGB radiance (\eg, for thermal images). Considering that our modality set satisfies this constraint, we believe that its use is motivated in our scenario. Finally, it is worth noting that the inverse function and the modality-to-luma losses act as complementary regularizations: the inverse function loss promotes the preservation of latent space, particularly for well correlated modalities; the modality-to-luma loss acts as guidance for the less correlated ones.

\section{Experimental Evaluation}
\label{sec:experiments}
In this section, we present a set of experiments that demonstrate the effectiveness of \gls{method-name}. We recall that the main goal of this work is to build a method that can learn the mutual correlation between different imaging modalities in order to enable the generation of photorealistic renderings of any of them, even when the scene is captured by a single or a limited subset of sensors. For this reason, we focus on the \gls{ft} results while including most of the \gls{pt} results in the supplementary material.
In the next sections, we show that \gls{method-name} can produce accurate renderings of modalities with limited or no supervision. We also include comparisons with single-view RGB-to-modality methods such as MST++~\cite{cai2022mst++} and PolarAnything~\cite{zhang2025polaranything}, proving that \gls{method-name} can outperform them in terms of multi-view radiance consistency.

\subsection{Implementation Details}
\label{sec:impl_details}
We recall that the model is trained in 2 stages: a multi-scene pre-training followed by a scene-specific fine tuning.
The \gls{pt} runs for 1M iterations on a single NVIDIA RTX 6000 Ada. In the \gls{pt}, the parameter $B$ (\cref{sec:pt_pipeline}) is set to 15 iterations, while
$\lambda_{eik}$ and $\lambda_{curv}$ are set at 0.1 and 0.0005, respectively, as suggested in~\cite{li2023neuralangelo}. 
The regularization loss weights are  set to $\lambda_{lsg}=0.1$, $\lambda_{inv}=1.0$, and $\lambda_{m2l}=1.0$.
The scene-specific \gls{ft} runs for only $30k$ iterations using $\lambda_{inv}=0.01$ and $\lambda_{m2l}=0.02$, while the latent space geometry loss is disabled in this phase.\\
\indent To evaluate the results, we consider \gls{mms-fw} as a reference and compare with its metrics. However, it is important to mention that defining a fair comparison is not trivial. \gls{mms-fw} is trained from scratch on a single scene, optimized for 100k iterations, and its radiance module counts $\sim$12M parameters. On the other hand, our \gls{ft} model is optimized for 30k iterations and its radiance module has $\sim$4.1M parameters, but only $\sim$7\% of them are trainable ($\sim$300k parameters). The other $\sim$3.8M are frozen because optimized only during the \gls{pt} for $\sim$200k iterations on several different scenes. However, the \gls{pt} step must be completed just once, thus it is reasonable not to consider it when evaluating the cost of the per-scene optimization. The choice to use a smaller model with respect to~\cite{lincetto2025} is motivated by the fact that the \gls{pt} phase has high memory requirements in order to handle several scenes at the same time. Given the differences in architectures and training pipelines, it is impossible to align the model capacities and the number of training iterations. For these reasons, we opt to preserve the original architecture of \gls{mms-fw}, but it is worth noting that \gls{method-name} has $3$ times smaller capacity and the fine-tuning time is 4 times shorter.

\subsection{Datasets and Metrics}
\label{sec:dataset_metrics}
The experiments are conducted on MMS-DATA~\cite{lincetto2025}, a multi-view multimodal dataset encompassing 5 different modalities: \ie, RGB, \gls{nir}, Mono, \gls{pol}, and \gls{ms} images. It includes 32 object-centric scenes. The \gls{pt} is performed on 27 scenes, and the \gls{ft} on the remaining 5 scenes. 
Given that \gls{method-name} is built on MultimodalStudio~\cite{lincetto2025}, we follow the same strategy of training our model with MMS-DATA raw mosaicked frames and computing the metrics only on masked foreground regions.
To evaluate the quality of all rendering results, we use \gls{psnr} as the main metric. In addition, we also show \gls{ssim} for single channel or demosaicked images, since its use is not appropriate with mosaicked data \cite{lincetto2025}.

\begin{figure*}[t]
    \centering
    \begin{minipage}{\linewidth}
        \centering
        \hfill
        \begin{minipage}{0.96\linewidth}
            \begin{minipage}{0.195\linewidth}
                \centering
                RGB
            \end{minipage}
            \begin{minipage}{0.195\linewidth}
                \centering
                NIR
            \end{minipage}
            \begin{minipage}{0.195\linewidth}
                \centering
                Mono
            \end{minipage}
            \begin{minipage}{0.195\linewidth}
                \centering
                Pol
            \end{minipage}
            \begin{minipage}{0.195\linewidth}
                \centering
                MS
            \end{minipage}
        \end{minipage}
        \vspace{2pt}
    \end{minipage}
    \begin{minipage}{\linewidth}
        \begin{minipage}{0.025\linewidth}
            \vfill
            \rotatebox[origin=cb]{90}{GT}
            \vfill
        \end{minipage}
        \begin{minipage}{0.97\linewidth}
            \begin{subfigure}{0.195\linewidth}
                \centering
                \includegraphics[width=\linewidth]{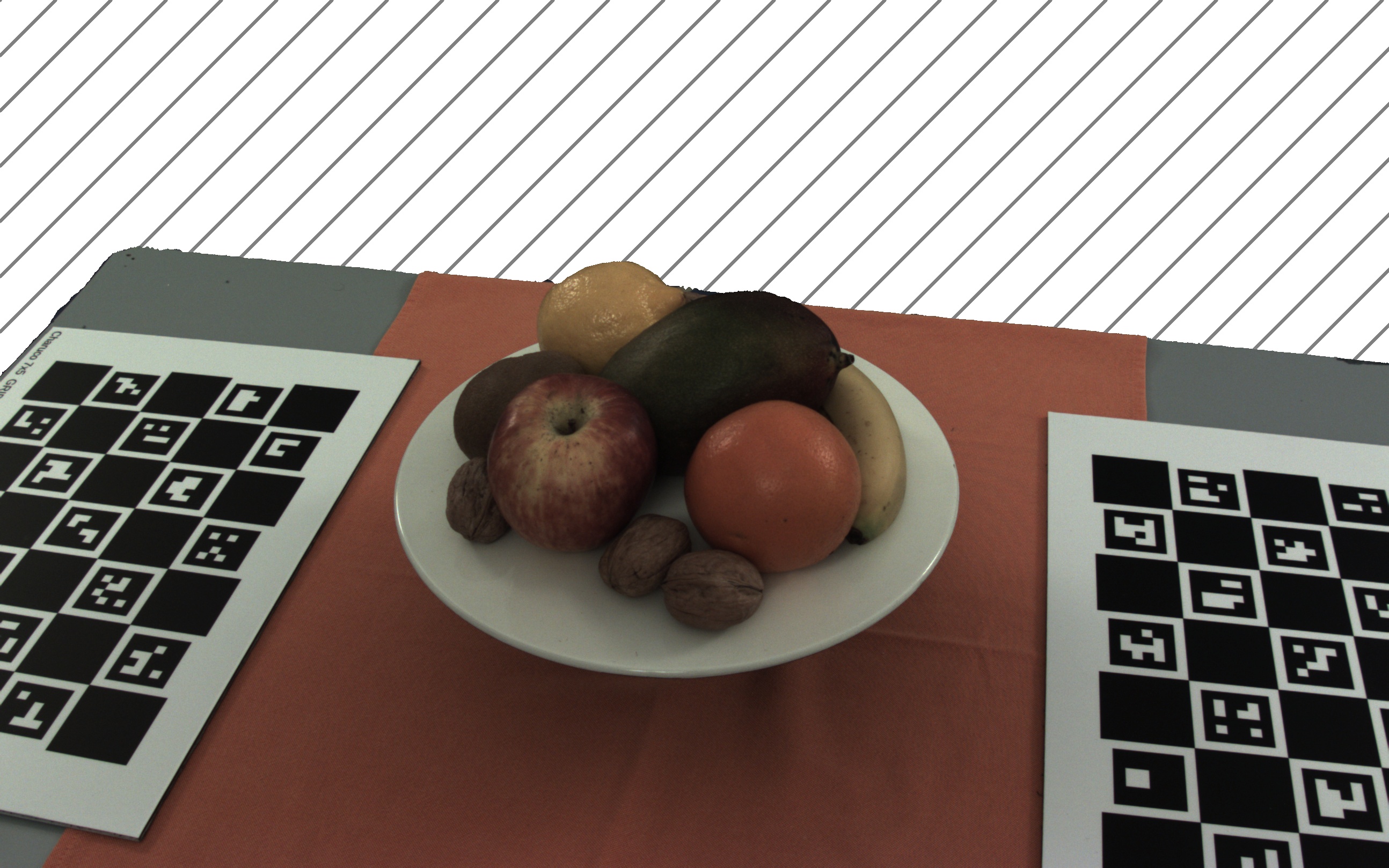}
            \end{subfigure}
            \begin{subfigure}{0.195\linewidth}
                \centering
                \includegraphics[width=\linewidth]{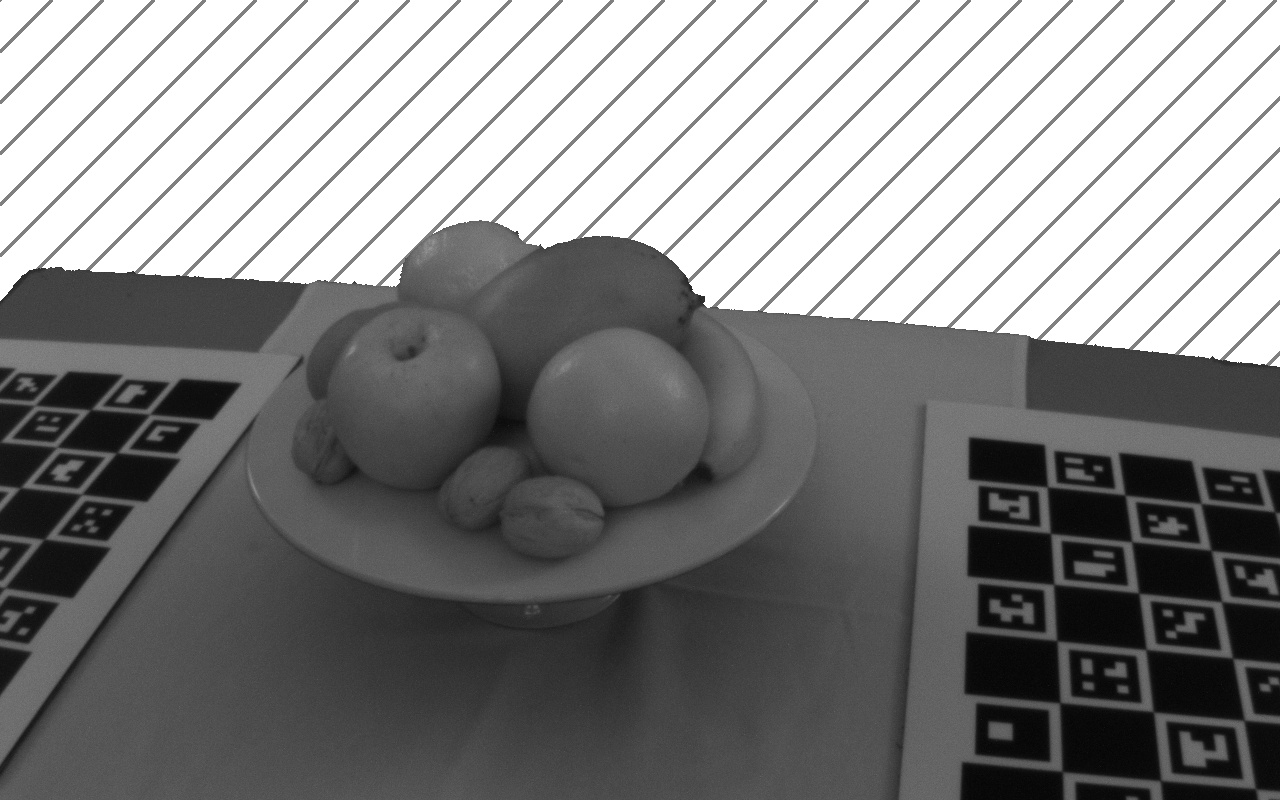}
            \end{subfigure}
            \begin{subfigure}{0.195\linewidth}
                \centering
                \includegraphics[width=\linewidth]{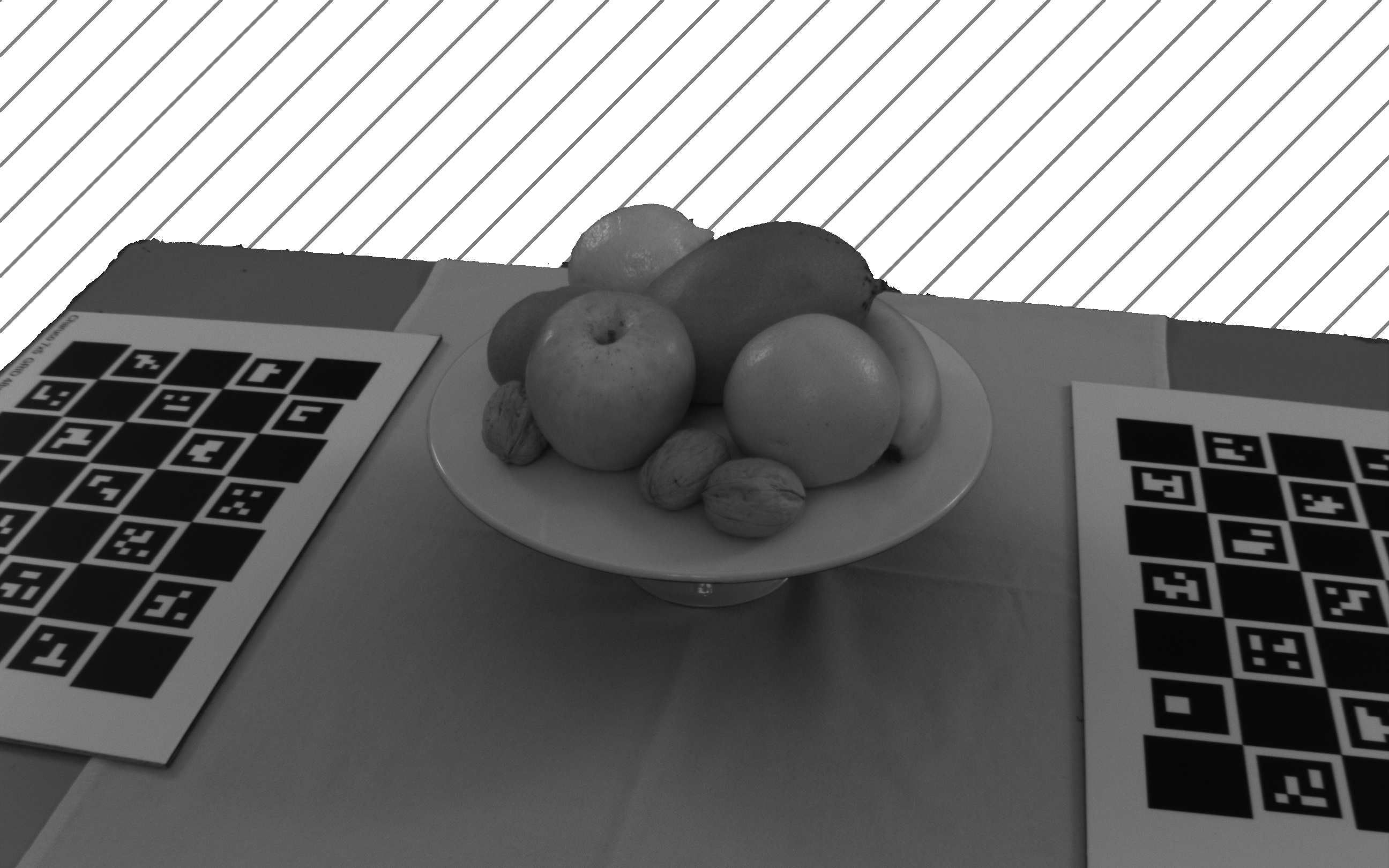}
            \end{subfigure}
            \begin{subfigure}{0.195\linewidth}
                \centering
                \includegraphics[width=\linewidth]{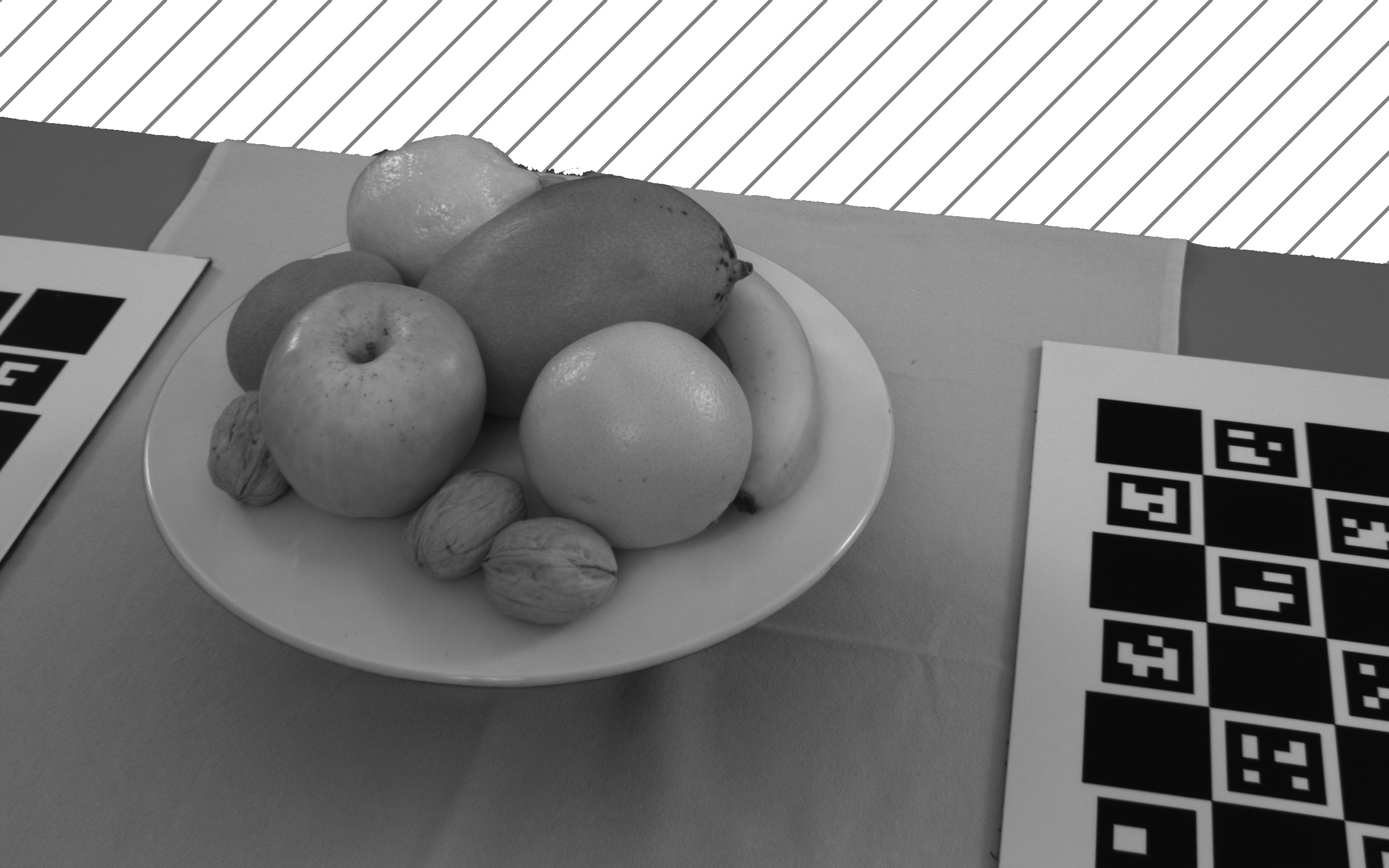}
            \end{subfigure}
            \begin{subfigure}{0.195\linewidth}
                \centering
                \includegraphics[width=\linewidth]{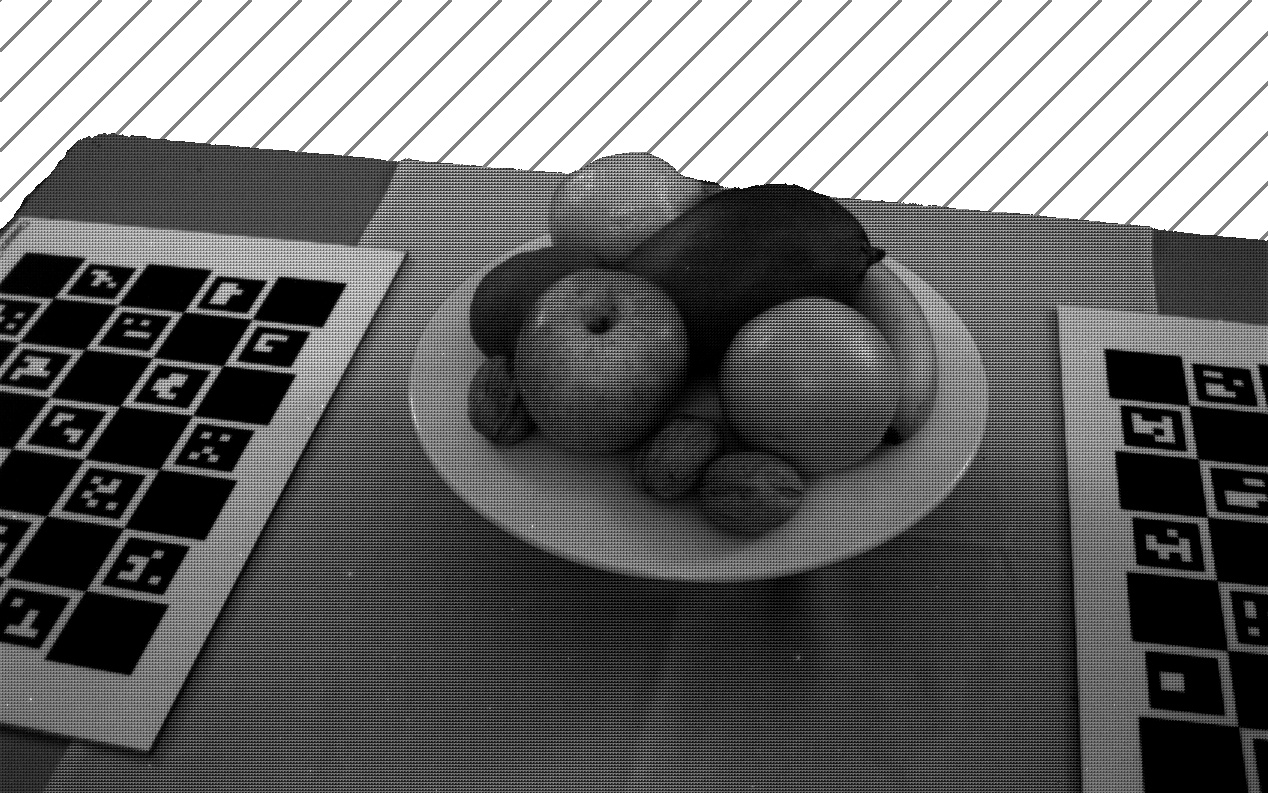}
            \end{subfigure}
        \end{minipage}
    \end{minipage}
    \begin{minipage}{\linewidth}
        \begin{minipage}{0.025\linewidth}
            \vfill
            \rotatebox[origin=cb]{90}{Ours}
            \vfill
        \end{minipage}
        \begin{minipage}{0.97\linewidth}
            \begin{subfigure}{0.195\linewidth}
                \centering
                \includegraphics[width=\linewidth]{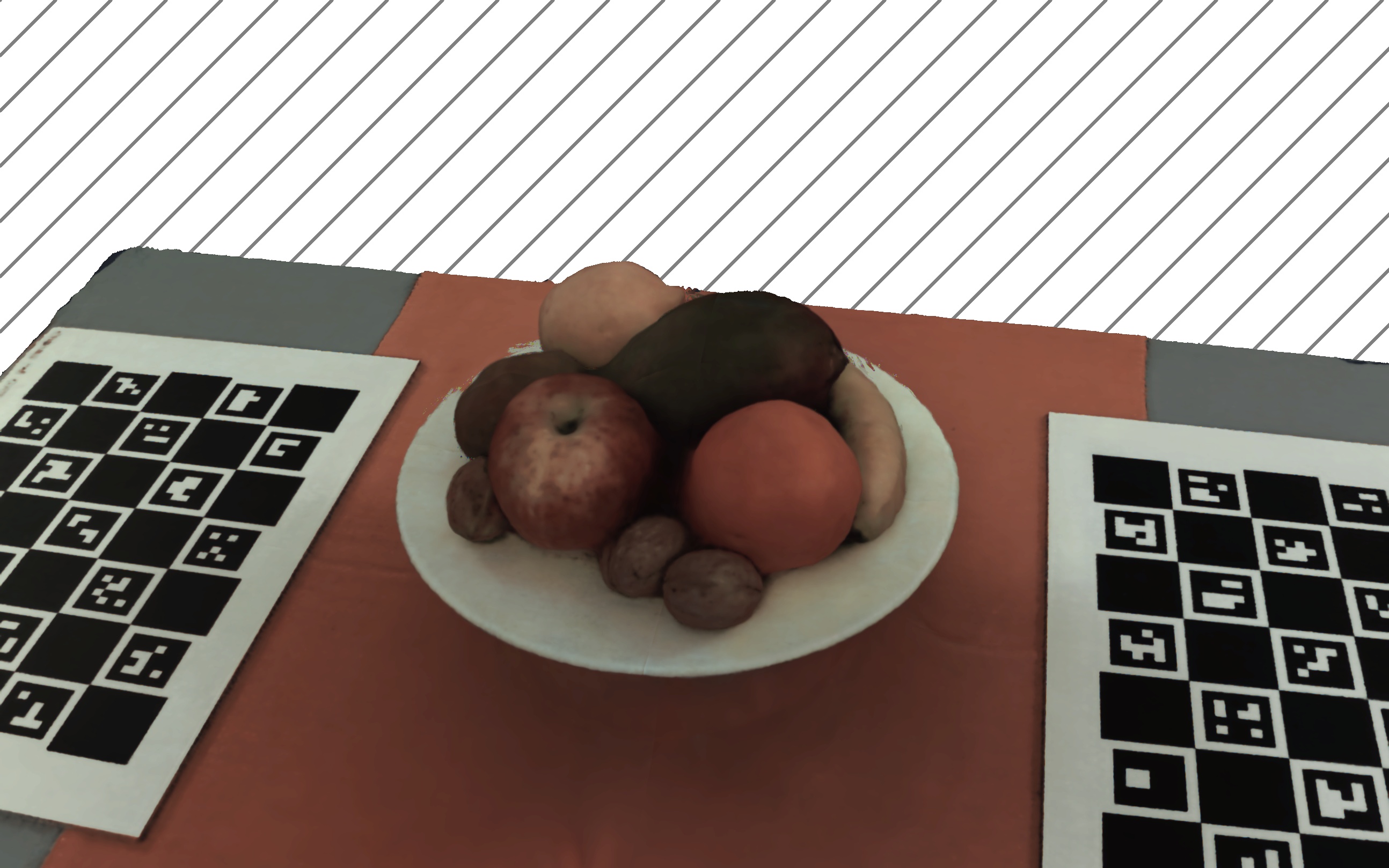}
            \end{subfigure}
            \begin{subfigure}{0.195\linewidth}
                \centering
                \includegraphics[width=\linewidth]{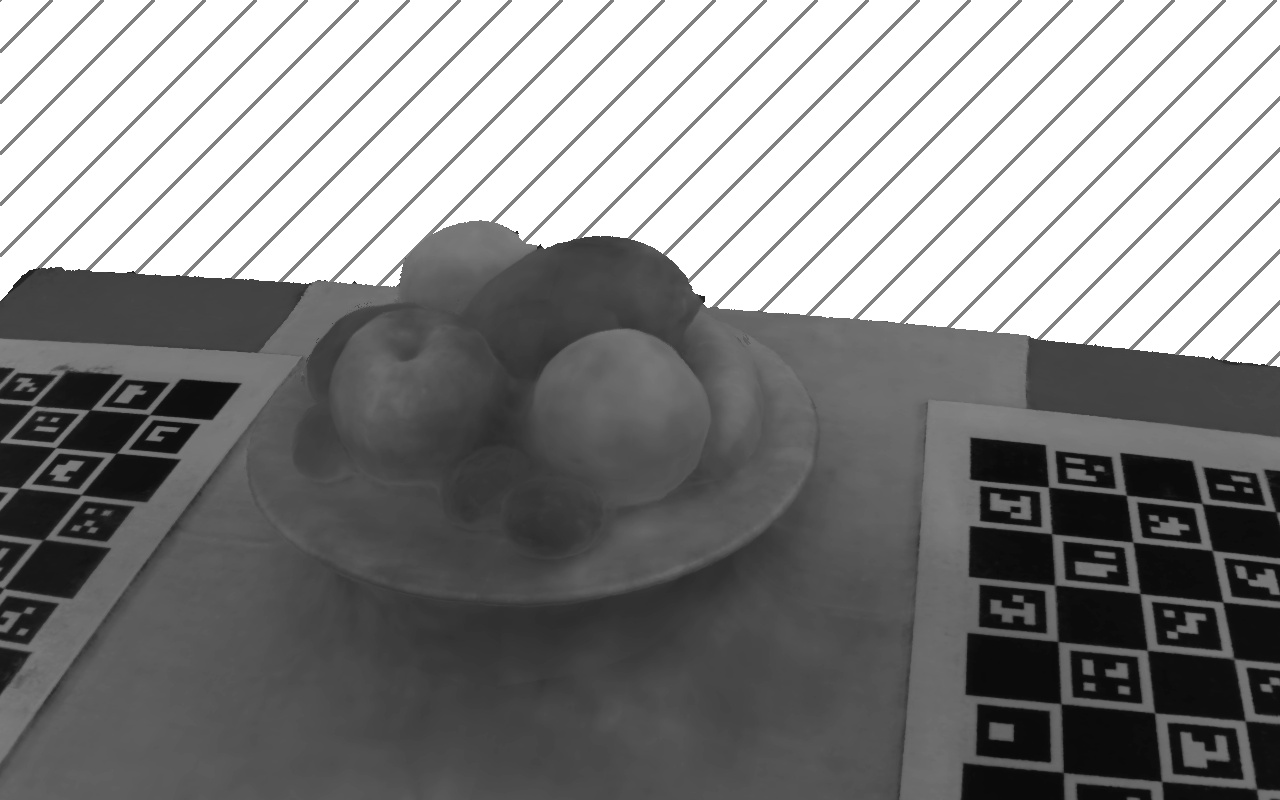}
            \end{subfigure}
            \begin{subfigure}{0.195\linewidth}
                \centering
                \includegraphics[width=\linewidth]{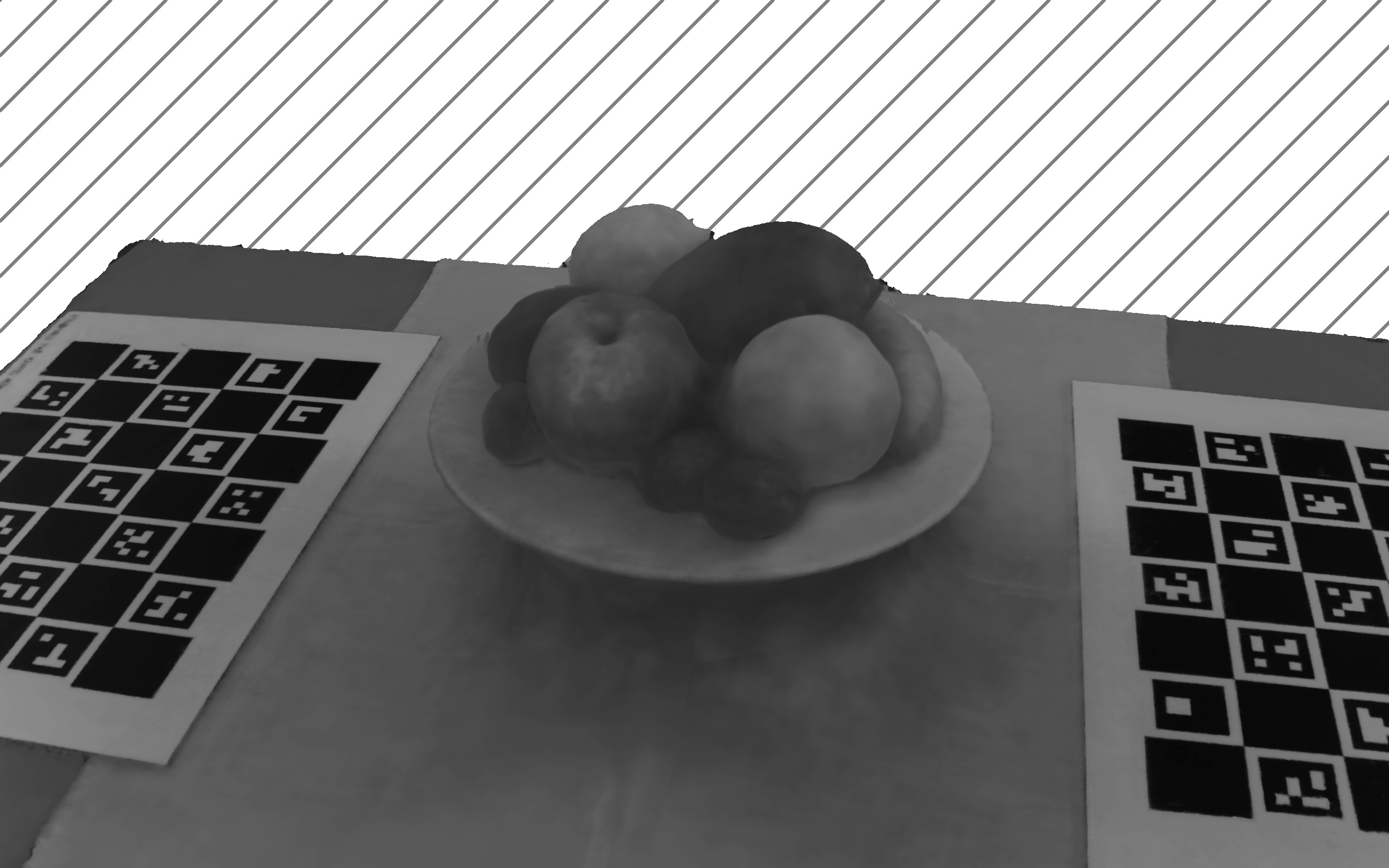}
            \end{subfigure}
            \begin{subfigure}{0.195\linewidth}
                \centering
                \includegraphics[width=\linewidth]{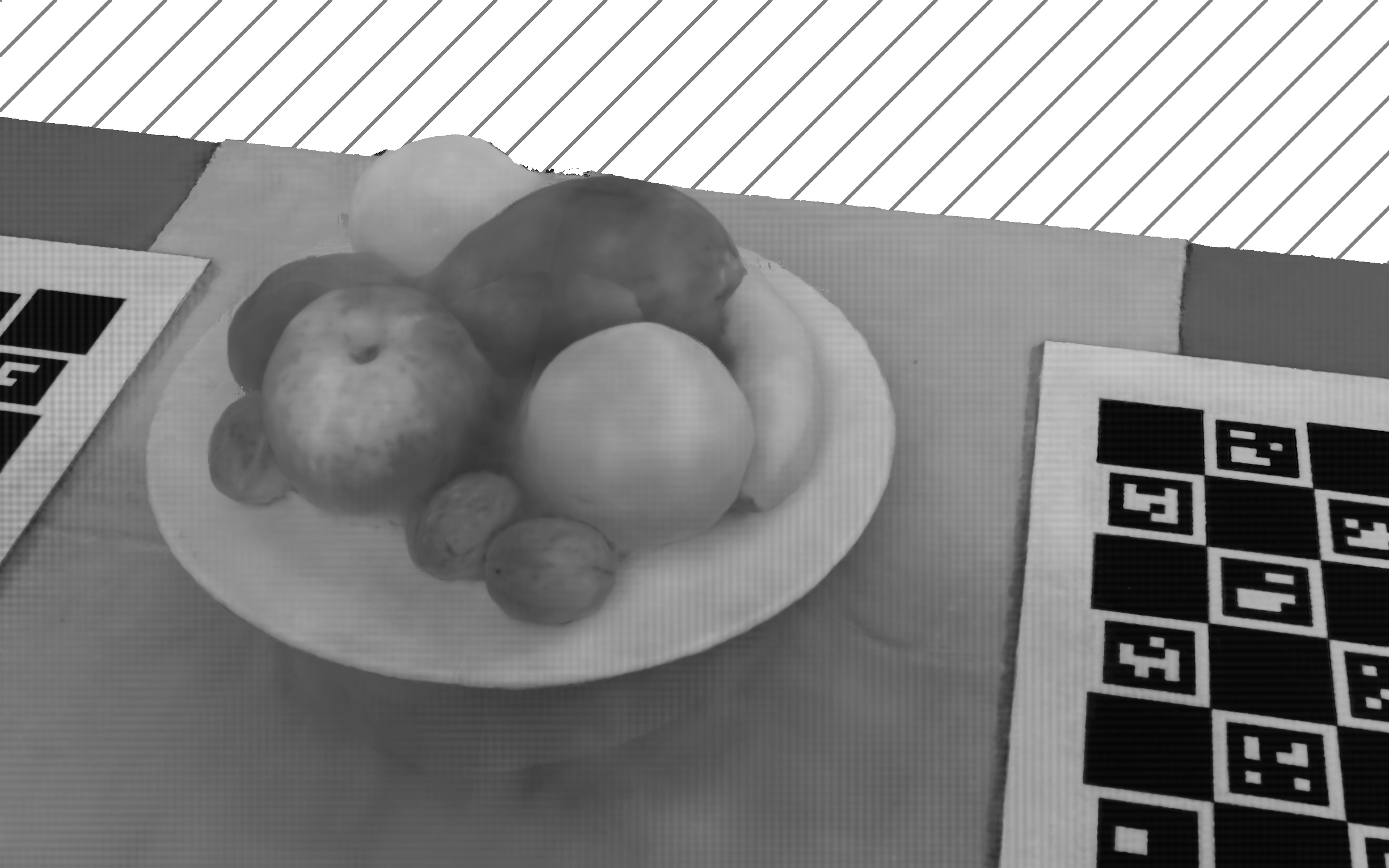}
            \end{subfigure}
            \begin{subfigure}{0.195\linewidth}
                \centering
                \includegraphics[width=\linewidth]{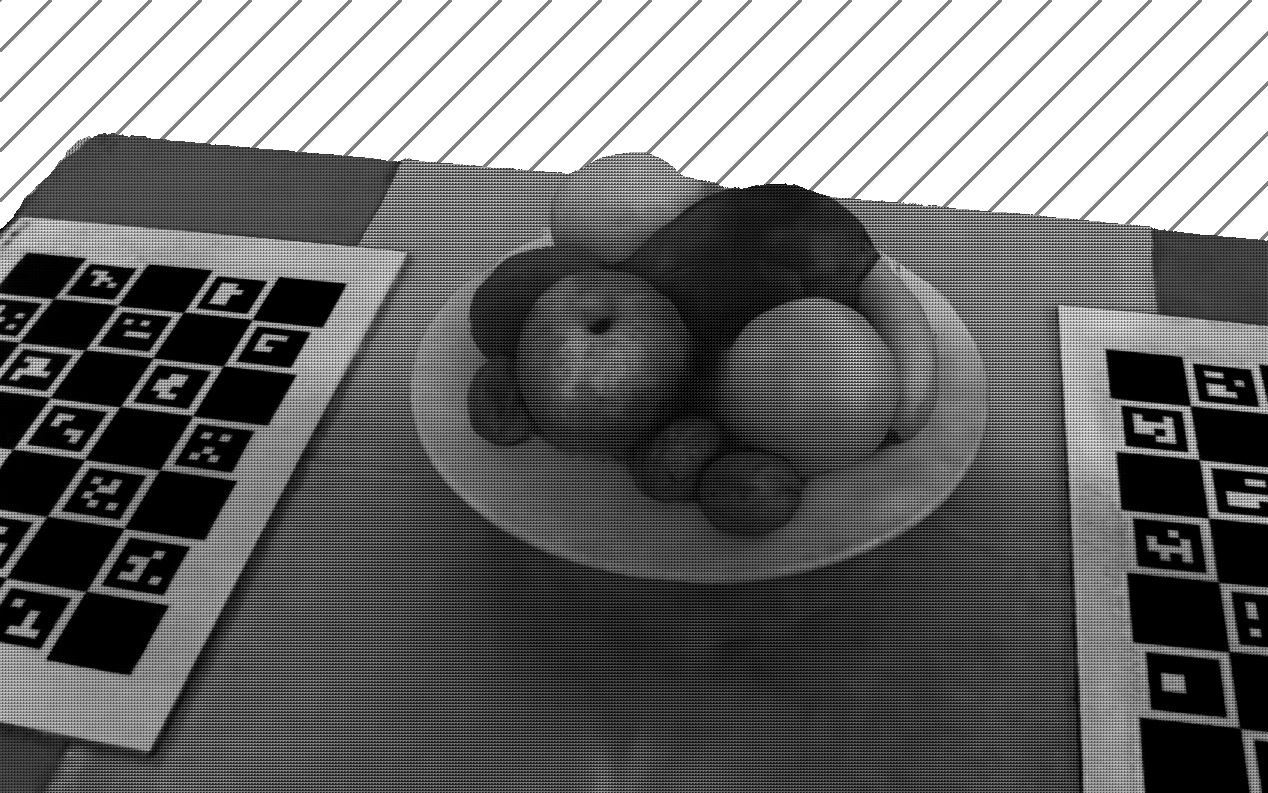}
            \end{subfigure}
        \end{minipage}
    \end{minipage}
    \begin{minipage}{\linewidth}
        \begin{minipage}{0.025\linewidth}
            \vfill
            \rotatebox[origin=cb]{90}{Error}
            \vfill
        \end{minipage}
        \begin{minipage}{0.97\linewidth}
            \begin{subfigure}{0.195\linewidth}
                \centering
                \includegraphics[width=\linewidth]{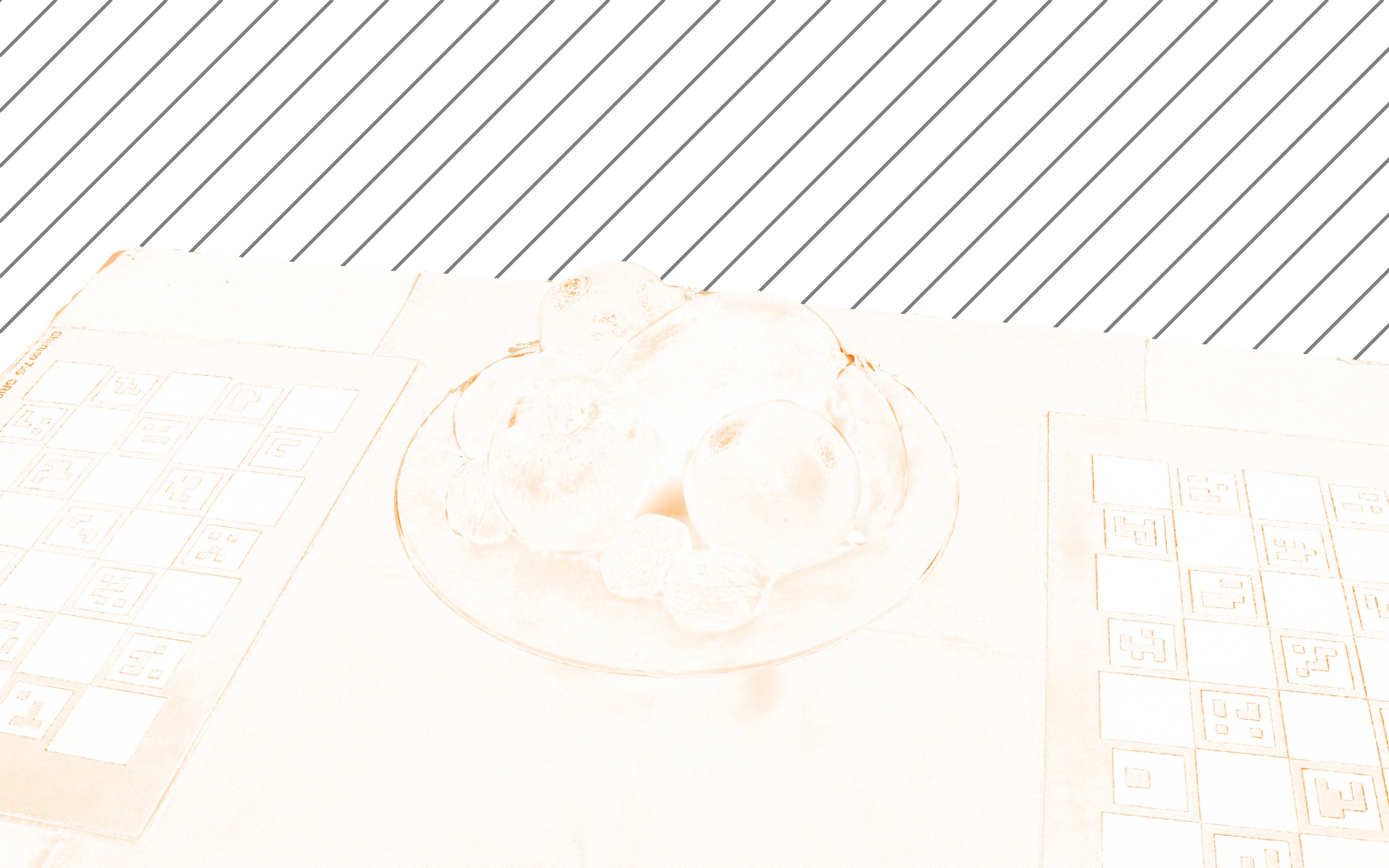}
            \end{subfigure}
            \begin{subfigure}{0.195\linewidth}
                \centering
                \includegraphics[width=\linewidth]{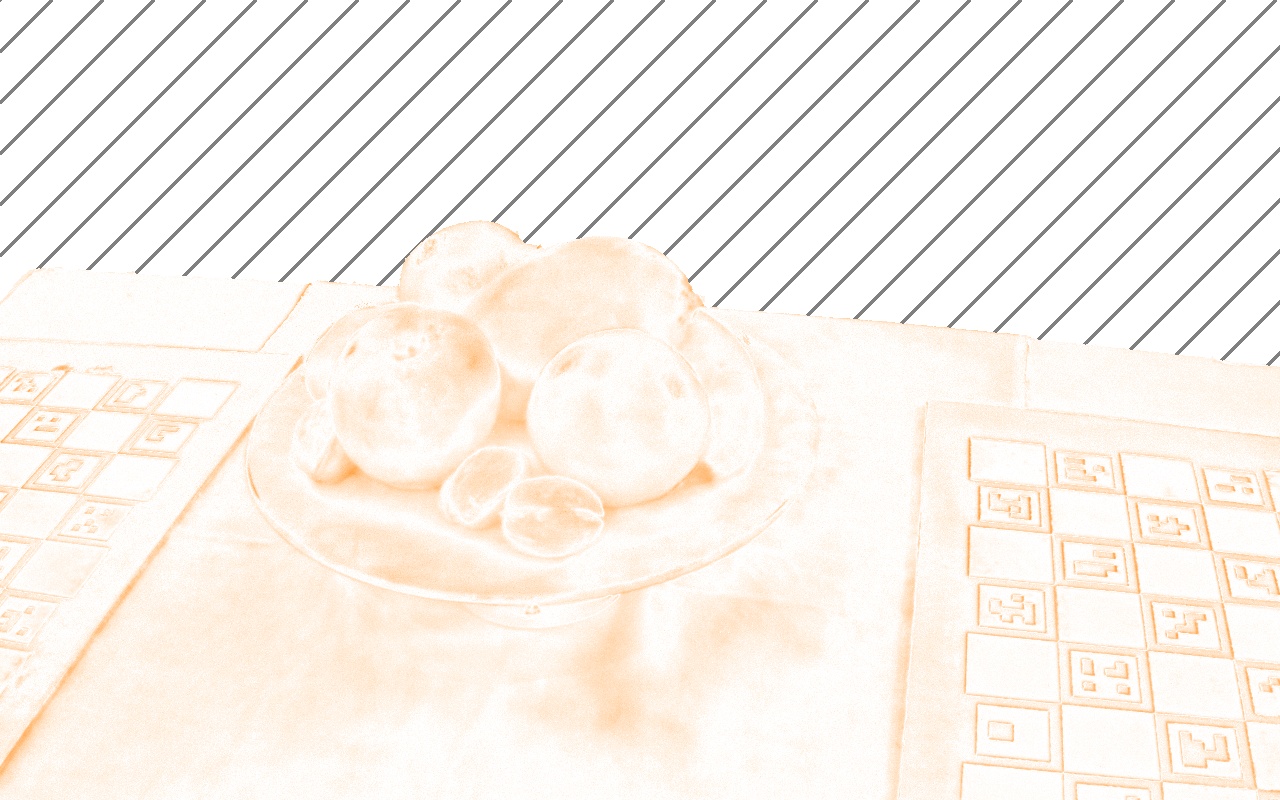}
            \end{subfigure}
            \begin{subfigure}{0.195\linewidth}
                \centering
                \includegraphics[width=\linewidth]{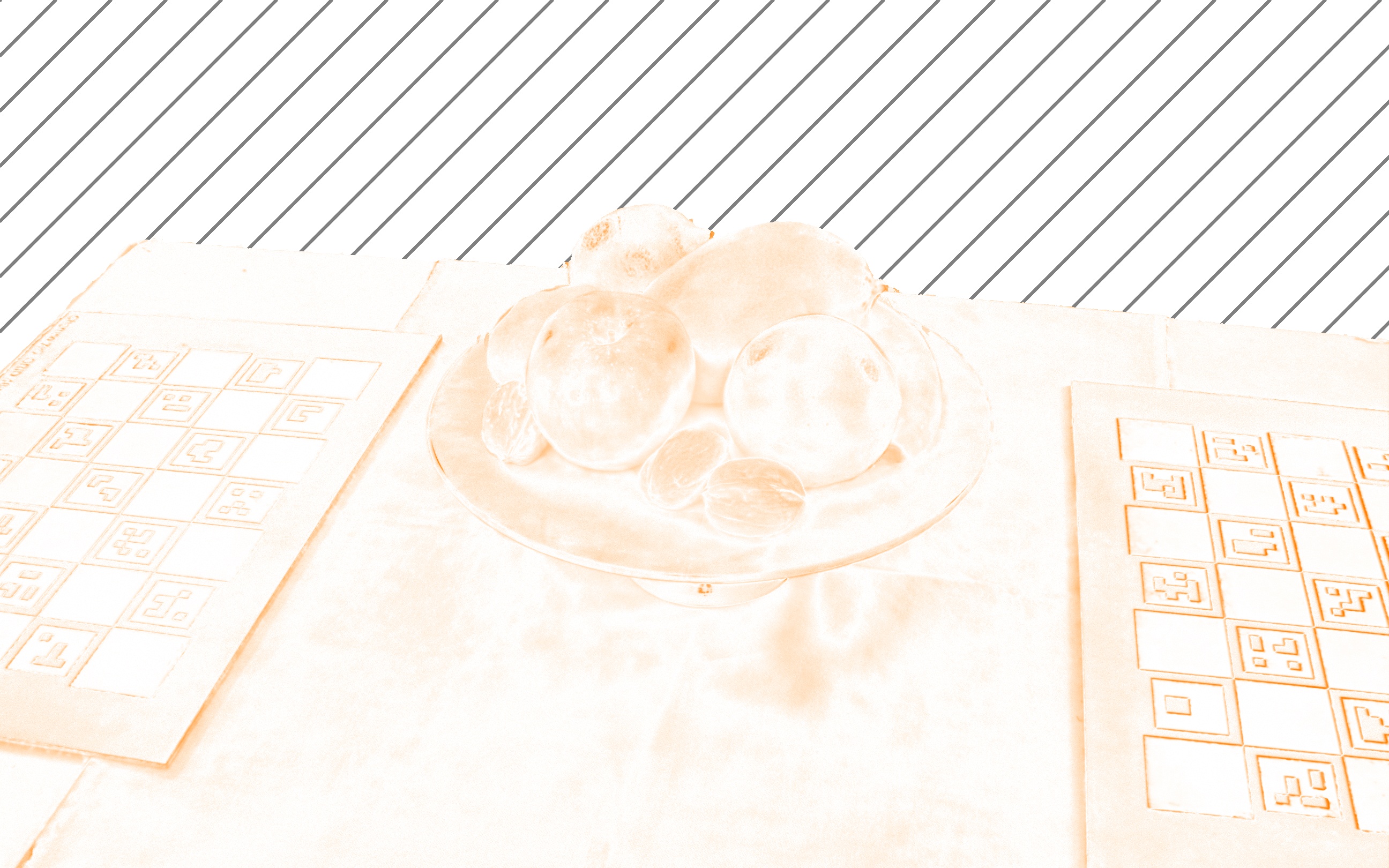}
            \end{subfigure}
            \begin{subfigure}{0.195\linewidth}
                \centering
                \includegraphics[width=\linewidth]{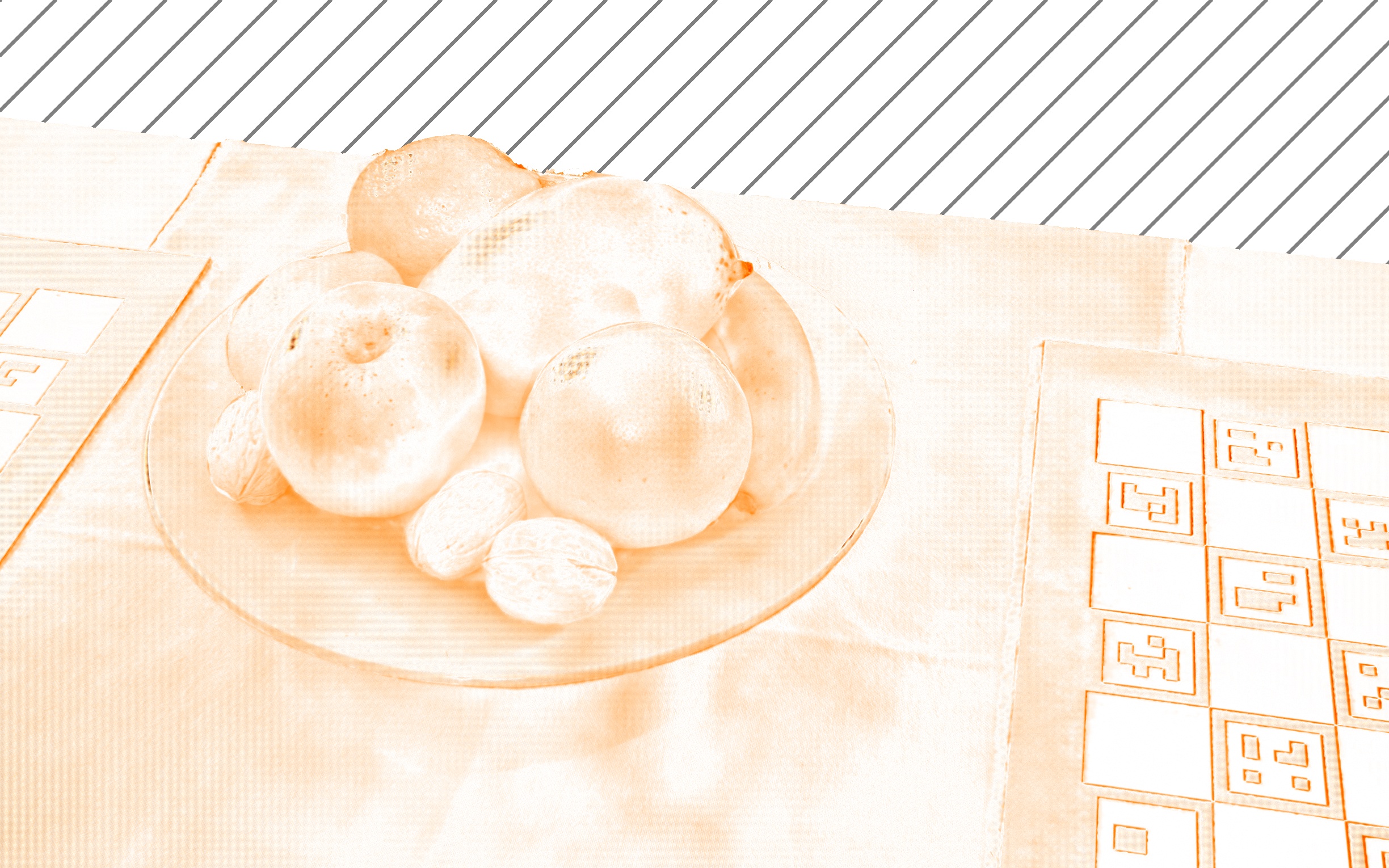}
            \end{subfigure}
            \begin{subfigure}{0.195\linewidth}
                \centering
                \includegraphics[width=\linewidth]{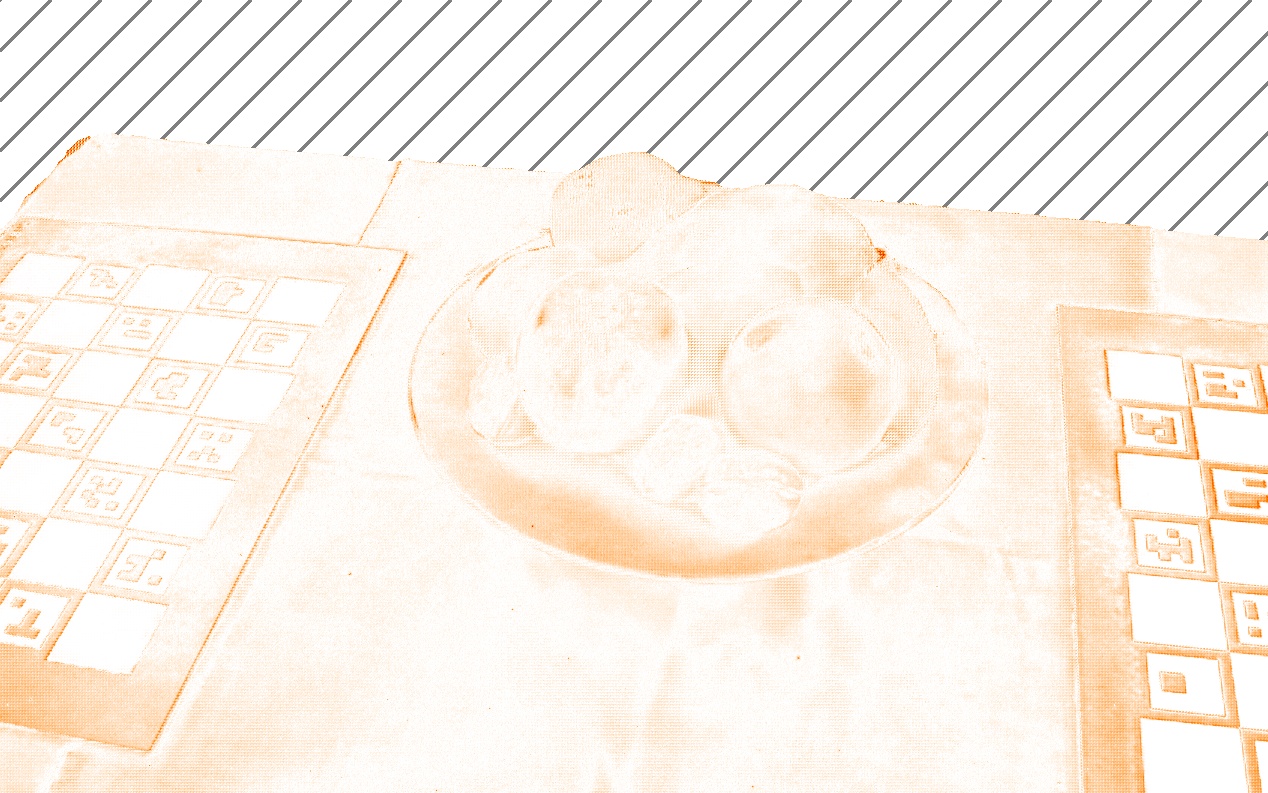}
            \end{subfigure}
        \end{minipage}
        \begin{center}
            \vspace{-0.25cm}
            \hspace{0.2cm}
            \includegraphics[width=0.5\linewidth]{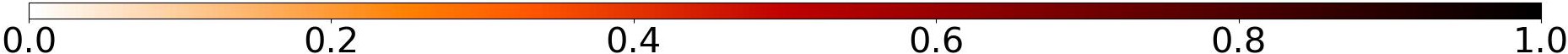}
        \end{center}
    \end{minipage}
    \caption{Qualitative samples from the ``Fruits'' scene \gls{ft}  performed with only RGB data}
    \label{fig:qualitative}
    \vspace{-0.5cm}
\end{figure*}
\begin{table}[t]
    \begin{minipage}{0.49\linewidth}
    \centering
    \caption{Results of \gls{method-name} (Ours) FT and of MMS-FW~\cite{lincetto2025}, averaged on the five \gls{ft} scenes. Standard case: both models are trained only with RGB frames. Upper bounds: both models are supervised with frames of all modalities.}
    \label{tab:pt_ft_results}
    \begin{tabular*}{\linewidth}{@{\extracolsep{\fill}}l|ccccc}
        \toprule
            && Train& Test& \multirow{2}{*}{PSNR$\uparrow$}&\multirow{2}{*}{SSIM$\uparrow$}\\
            && Mod.& Mod.& &\\
        \midrule
             \multirow{6}{0.5cm}{\centering\rotatebox[origin=c]{90}{Standard case}}&\multirow{5}{*}{Ours}& \multirow{5}{*}{\green{RGB}}& \green{RGB} & 30.07&-\\
        \arrayrulecolor{black!30}\cline{4-6}
             && & \purple{Mono} & 25.78&0.88\\
        \cline{4-6}
             && & \red{NIR} & 26.55&0.87\\
        \cline{4-6}
             && & \blue{Pol} & 24.25&-\\ 
        \cline{4-6}
             && & \orange{MS}& 25.45&-\\ 
        \arrayrulecolor{black}\cmidrule{2-6}
            &MMS& \green{RGB} & \green{RGB}& 29.53&-\\
        \midrule
            \multirow{10}{0.5cm}{\centering\rotatebox[origin=c]{90}{Upper bounds}}&\multirow{5}{*}{Ours}&  \multirow{5}{*}{ALL}& \green{RGB}& 30.54& -\\
        \arrayrulecolor{black!30}\cline{4-6}
            && & \purple{Mono}& 29.21&0.93\\
        \cline{4-6}
            && & \red{NIR}& 32.12&0.92\\
        \cline{4-6}
            && & \blue{Pol}& 29.32&-\\
        \cline{4-6}
            && & \orange{MS}& 28.46&-\\
        \arrayrulecolor{black}\cmidrule{2-6}
        &\multirow{5}{*}{MMS}&  \multirow{5}{*}{ALL}& \green{RGB}& 32.77& -\\
        \arrayrulecolor{black!30}\cline{4-6}
            && & \purple{Mono}& 32.98&0.94\\
        \cline{4-6}
            && & \red{NIR}& 34.25&0.93\\
        \cline{4-6}
            && & \blue{Pol}& 30.66&-\\
        \cline{4-6}
            && & \orange{MS}& 31.42&-\\
        \arrayrulecolor{black}\bottomrule
    \end{tabular*}
    \end{minipage}
    \hfill
    \begin{minipage}{0.49\linewidth}
    \begin{center}
        \includegraphics[width=\linewidth]{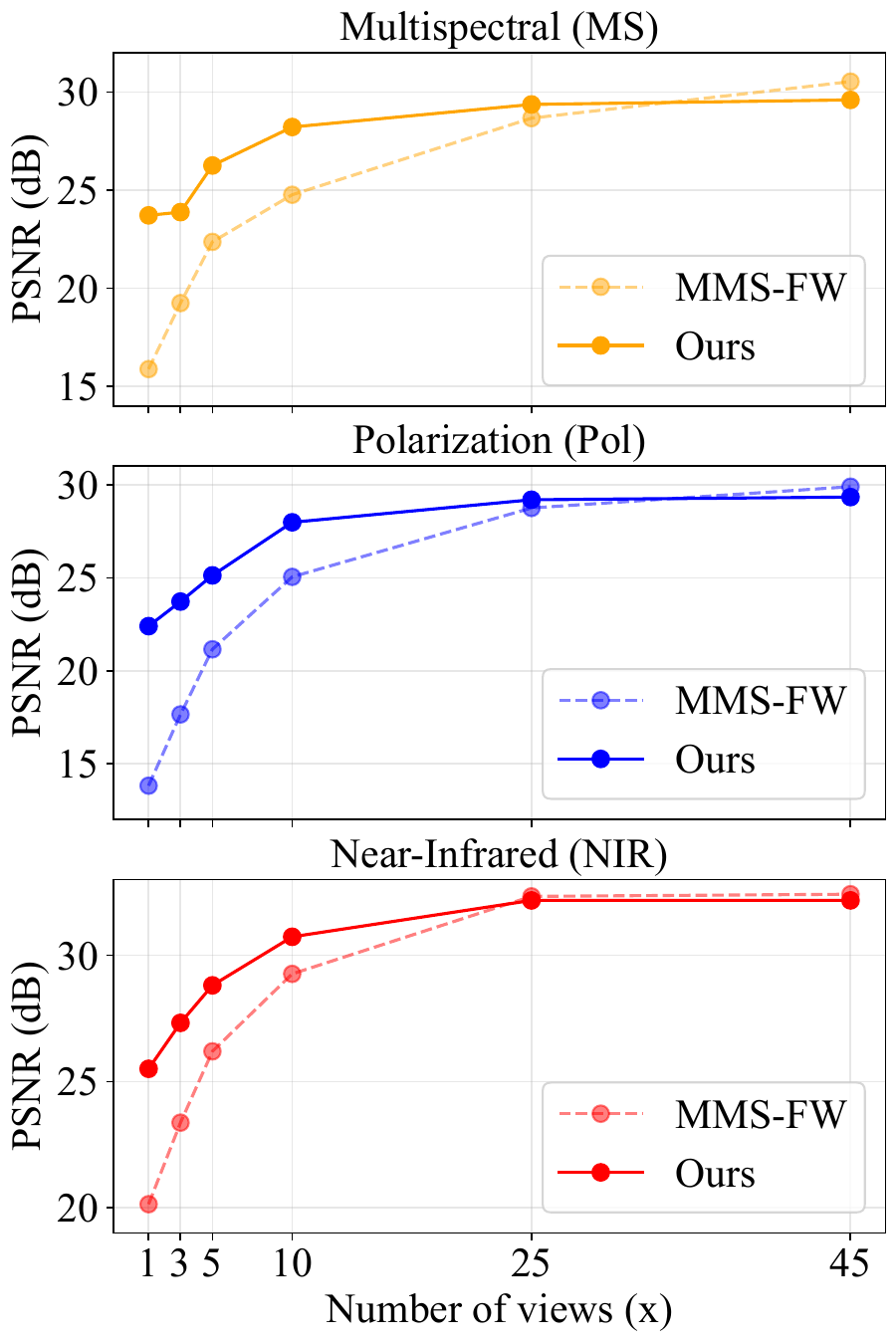}
        \captionof{figure}{Tests with an unbalanced combination of modalities. X axis refers to the number of additional modality (MS, Pol, or NIR) frames. Comparison between \gls{method-name} (Ours) and MMS-FW. }
        \label{fig:unbalanced_plots}
    \end{center}
    \end{minipage}
    \vspace{-0.5cm}
\end{table}

\subsection{Multi-scene Pre-training and Single-scene Fine-tuning}
\label{sec:pt_ft_results}
As described in~\cref{sec:method}, \gls{method-name} is trained in two stages: firstly, a multi-scene \gls{pt} phase and then a single-scene \gls{ft} phase.
The output of the \gls{ft} stage is the final outcome that we discuss in this section. 
The main purpose of the \gls{pt} is to learn the scene-independent mutual correlations between modalities to effectively support the lack of multimodal data during the \gls{ft}, rather than accurately reconstructing the pre-training scenes.
The \gls{pt} rendering accuracy, reported in the supplementary material, is slightly lower than \gls{ft} results, and this is reasonable as, unlike single-scene methods, its focus is more on generalization.
Considering the \gls{ft} metrics shown in~\cref{tab:pt_ft_results}, it is possible to compare the results achieved by \gls{method-name} and \gls{mms-fw}. When supervised by a single modality, namely RGB, it is interesting to note that \gls{method-name} achieves a higher \gls{psnr} than \gls{mms-fw}.
This result highlights that the \gls{pt} knowledge, built across several modalities and scenes, helps to improve the quality also of supervised modalities with respect to a model trained from scratch on a single scene.
More importantly, it also allows rendering additional modalities not part of the \gls{ft} training images, which is impossible for \gls{mms-fw} and for  standard novel view synthesis approaches.\\
\indent In~\cref{tab:pt_ft_results}, we also show the results achieved by \gls{method-name} and \gls{mms-fw} when supervised by all modalities. These results  serve as upper bounds for our \gls{ft} model with single-modality supervision. As expected, \gls{mms-fw} outperforms \gls{method-name}, due to its larger capacity and the training on a single scene from scratch. We observe that the improvement driven by these factors can be quantified in $\sim$2 dB of \gls{psnr}, while \gls{ssim} is perfectly comparable.
In \cref{fig:qualitative}, we show some qualitative results; additional results are in the supplementary material.

\subsection{Unbalanced Combination of Modalities}
\label{sec:unbalanced_modalities}
In this section, we investigate the results achieved by \gls{mms-fw} when trained with an unbalanced combination of modalities. This test is introduced by  \cite{lincetto2025} and involves scenarios where many frames of one modality but only a few of a second one are available, thus the model receives stronger supervision for the first modality and limited guidance for the second. This scenario represents situations where, due to deployment  or  cost constraints, it is impossible to capture the same amount of frames for every modality. At the same time, even when it is possible, it demonstrates that having full scene coverage for all modalities is not mandatory. To prove this claim, we perform some tests involving RGB-MS, RGB-Pol, and RGB-NIR frames. Following the idea that RGB cameras are cheaper and more widely accessible, all RGB frames (45) are always available, while the second-modality frames are reduced to 1, 3, 5, 10, and 25. In~\cref{fig:unbalanced_plots}, we show the \gls{psnr} trend of the rendered second-modality frames as a function of the number of available frames. We observe that \gls{method-name} clearly outperforms \gls{mms-fw}, particularly in scenarios with few second-modality frames available. When only one MS, Pol, or NIR frame is available, \gls{method-name} achieves higher \gls{psnr}, with gains of $\sim$8, $\sim$7, and $\sim$6 dB, respectively. With 10 second-modality frames, \gls{method-name} still gains $\sim$2.5 dB on average, and  \gls{mms-fw} achieves similar performance as our approach only with 25 frames. These results confirm the positive effect of the \gls{pt} phase: by leveraging this knowledge, the \gls{ft} can more easily infer the second-modality radiance even with very few available frames. In these tests, for the sake of fair comparison, both \gls{method-name} fine-tuning and \gls{mms-fw} are trained for 100k iterations.
Note that the only other public multi-view multimodal dataset that could be used, \ie, X-NeRF, is not very informative since all scenes are forward-facing and even a single view encompasses almost the whole scene, thus the benefits of the \gls{pt} are not appreciable. However, we replicated the test on X-NeRF: as expected, \gls{method-name} performs on par with \gls{mms-fw}. We include results on X-NeRF data in the supplementary material.

\subsection{Comparisons with Third-party Modality Conversion Methods}
\label{sec:comparison_with_third_party}
\begin{table}[t]
    \begin{minipage}{0.4\linewidth}
    \centering
    \caption{Comparison between \gls{method-name} and \gls{mms-fw} on MS frames. \gls{mms-fw} is trained with \gls{ms} frames estimated from RGB using MST++, denoted as MS*.}
    \label{tab:mst++}
    \begin{tabular*}{\linewidth}{@{\extracolsep{\fill}}cccc}
        \toprule
            & Train& Test& PSNR$\uparrow$\\
        \midrule
            MMS& \orange{MS*}& \orange{MS}& 21.99\\
        \arrayrulecolor{black!30}\midrule
            Ours& \green{RGB}& \orange{MS}& \textbf{25.45}\\
        \arrayrulecolor{black}\bottomrule
    \end{tabular*}
    \end{minipage}
    \hfill
    \begin{minipage}{0.57\linewidth}
    \caption{{Comparison between \gls{method-name} and \gls{mms-fw} on \gls{pol} rendering quality. \gls{mms-fw} is trained with \gls{pol} frames estimated from RGB by PolarAnything, denoted as Pol*. \gls{aop} MAngE ranges in $[$0°, 90°$]$, \gls{dop} MAbsE ranges in $[0,1]$.}}
    \label{tab:polaranything}
    \centering
    \begin{tabular*}{\linewidth}{@{\extracolsep{\fill}}ccccc}
        \toprule
            & Train& Test& MAngE(°)$\downarrow$&MAbsE$\downarrow$\\
        \midrule
               MMS&\blue{Pol*}& \blue{Pol}& 44.13&0.089\\
        \arrayrulecolor{black!30}\midrule
            Ours& \green{RGB}& \blue{Pol}& \textbf{30.05}&\textbf{0.045}\\
        \arrayrulecolor{black}\bottomrule
    \end{tabular*}
    \end{minipage}
    \vspace{-0.4cm}
\end{table}

In literature, there exist many solutions to predict multimodal data starting from  RGB images, thus a viable strategy for multimodal novel view synthesis is to use them together with a standard NeRF/GS model. As discussed in~\cref{sec:related_m2m}, it is possible to employ methods such as MST++~\cite{cai2022mst++} and PolarAnything~\cite{zhang2025polaranything} that infer \gls{ms} and \gls{pol} information, respectively, from RGB frames with a feed-forward single shot approach. 
However, these methods work on a single view and could produce results that lack multi-view consistency. To prove this claim, a possible test is to convert single images from RGB to other modalities using MST++ or PolarAnything and then feed the output to a NeRF model, such as \gls{mms-fw}. 
We performed some additional tests to compare this approach with  \gls{method-name} 
on  \gls{ms} or \gls{pol} frames.
We start by using MST++ to estimate \gls{ms} data. 
Then, we train \gls{mms-fw} on these scenes with only the converted \gls{ms} images and render the evaluation frames that are then compared with the GT.
This allows us to compute the metrics between the renderings and the GT frames. Results are in~\cref{tab:mst++}: we observe that \gls{method-name} outperforms \gls{mms-fw} coupled with MST++ in terms of \gls{psnr} by 3.46 dB. This happens because, as expected, MST++ does not produce multi-view radiance consistent frames. Therefore, inaccurate information is propagated and averaged through the 3D volume by the NeRF, reducing the accuracy of the final rendering.\
 We perform analogous tests for \gls{pol} frames converted from RGB with PolarAnything. 
As proposed by~\cite{zhang2025polaranything}, the polarization can be evaluated by considering the \gls{aop} and \gls{dop} in terms of mean angle error (MAngE) and mean absolute error (MAbsE).
In~\cref{tab:polaranything} we compare the results achieved by \gls{method-name} fine-tuning with those of \gls{mms-fw} using PolarAnything data. Our model outperforms the competitor in terms of both MAngE and MAbsE, by 14.08° and 0.044, respectively. In this case, PolarAnything cannot generate multi-view consistent \gls{pol} frames because it was trained on single-view images. Therefore, combining the contributions of frames from different views leads to very inaccurate results when training a NeRF model. In contrast, \gls{method-name} relies on the \gls{pt} knowledge, built across multiple different 3D scenes, thus it is more robust when predicting different views of one missing modality. In conclusion, \gls{method-name} confirms its reliability and proves to be the current state-of-the-art in terms of multi-view consistency for the task of modality-to-modality conversion. Details about the conversion procedure and qualitative results by MST++ and PolarAnything are in the supplementary material.

\subsection{Ablation Study}
\label{sec:ablation_study}

\begin{table}[t]
    \begin{minipage}{0.55\linewidth}
    \centering
    \caption{Ablation study PSNR results.  Between brackets: difference with respect to the full model. Results of FT on RGB frames only.}
    \begin{tabular*}{\linewidth}{@{\extracolsep{\fill}}cc|c|c|c}
        \toprule
        Mod.&Ours&w/o $\mathcal{L}_{lsg}$&  w/o $\mathcal{L}_{inv}$ &  w/o $\mathcal{L}_{m2l}$\\
        \midrule
        \green{RGB}&\textbf{30.07}& 29.90\g{(-0.17)}& 29.96\g{(-0.11)}& 29.95\g{(-0.12)}\\
        \purple{Mono}& \textbf{25.78}& 25.44\g{(-0.34)}& 25.54\g{(-0.25)}& 21.04\g{(-4.38)}\\
        \red{NIR}&\textbf{26.55}& 26.30\g{(-0.24)}& 25.96\g{(-0.58)}& 22.91\g{(-3.64)}\\
        \blue{Pol}& \textbf{24.25}& 23.08\g{(-1.18)}& 24.01\g{(-0.24)}& 23.73\g{(-0.52)}\\
        \orange{MS}& \textbf{25.45}& 23.14\g{(-2.31)}& 25.28\g{(-0.17)}& 25.36\g{(-0.09)}\\
        \bottomrule
    \end{tabular*}
    \label{tab:ablation}
    \end{minipage}
    \hfill
    \begin{minipage}{0.38\linewidth}
        \centering
        \includegraphics[width=\linewidth]{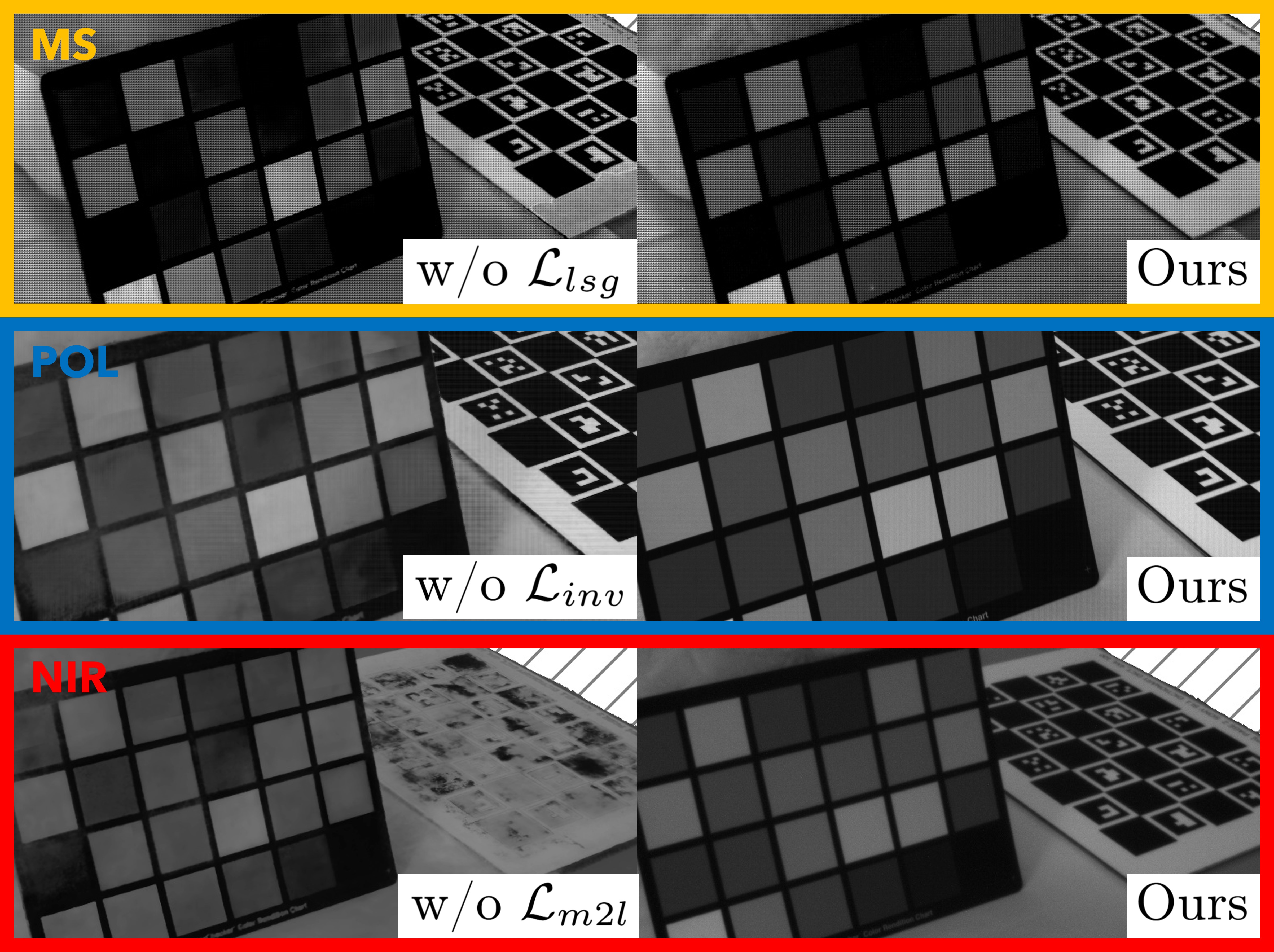}
        \vspace{0.1pt}
        \captionof{figure}{MS, Pol, and NIR renderings with different loss ablations}
    \end{minipage}
    \vspace{-0.8cm}
\end{table}

Finally, we perform an ablation study on the model components. In~\cref{tab:ablation}, we compare the results of the \gls{ft} with the results of the model when ablating each of the loss functions introduced in~\cref{sec:losses}. Firstly, we observe that every ablation leads to worse results in terms of \gls{psnr}. Looking at the results, the modality-to-luma $\mathcal{L}_{m2l}$ loss appears to be the most impactful, particularly for Mono and \gls{nir} data, even though \gls{pol} is also slightly affected.
One likely reason is that these three sensors also measure part of the infrared spectrum, albeit with different sensitivities. Therefore, the luminance information carried by these frames may diverge from what measured by the RGB or \gls{ms} sensor, thus the model is likely to decouple the information into more orthogonal latent representations. Therefore, when supervising only with RGB frames, the model struggles more to recover the correct Mono, \gls{nir}, and \gls{pol} information. In contrast, by applying the $\mathcal{L}_{m2l}$ loss, we make the correlation between all modalities more explicit, thus encouraging the model to learn a shared mutual representation.\\
\indent Similar considerations can be made for the inverse function loss $\mathcal{L}_{inv}$. As mentioned in \cref{sec:losses}, the purpose of the loss is complementary to the previous one. The inverse function encourages the model to maintain the latent space geometry learned during the \gls{pt} stable. In contrast to the $\mathcal{L}_{m2l}$ that pulls modalities' radiance towards RGB luma, the $\mathcal{L}_{inv}$ aims at preserving the mutual correlations learned over several scenes and when all the modalities are supervised. In other words, it acts as an antagonist to the $\mathcal{L}_{m2l}$ to promote latent space stability.\\
\indent Finally, we observe that the lack of latent space geometry loss $\mathcal{L}_{lsg}$ produces worse results for \gls{pol} and \gls{ms} renderings, while the other modalities are less affected. The purpose of this loss is to make the latent space more explainable, as it encourages the model to have a latent space that shares consistent distance measures with the corresponding explicit radiance space. In other words, the decoded radiance values are less sensitive to small shifts in their latent representation. \gls{ms} and \gls{pol} are the modalities with the highest information content, and the channels of each of them are highly correlated. Therefore, it is sufficient that only part of the \gls{ms} or \gls{pol} channels is well correlated with the supervised modality to predict a latent vector that, even if sub-optimal, thanks to the regularized latent space geometry, is decoded to an approximately correct radiance.

\section{Limitations}
\label{sec:limitations}
\gls{method-name} demonstrates the capability of rendering accurate and multi-view consistent spectral and polarimetric frames from a subset of modalities. However, its design inherently carries some limitations worth mentioning. For example, the estimation of spectral properties is reliable only for materials similar to those observed during the \gls{pt}.
Furthermore, a new \gls{pt} stage must be performed to predict new modalities. Finally, for consistent predictions, the \gls{pt} images must be captured by devices with fixed exposure and white balancing, and with controlled illumination.
Future work will focus on relaxing these constraints by making the model robust to variations of the illumination spectrum or of the sensor properties.
\section{Conclusions}
\label{sec:conclusions}
In this paper, we propose a novel multimodal neural rendering method that can decouple shared multimodal information from scene-specific content. This allows learning mutual correlations among different modalities from a set of scenes and then exploiting this information to render novel multimodal views of new scenes for which only RGB information is available.
The experimental results demonstrate the effectiveness of our method and show that it outperforms both multimodal NeRFs and two-stage approaches that employ third-party methods for modality conversion.

\section*{Acknowledgment}

This collaborative work is funded by Sony Europe Limited. 
The work of P. Zanuttigh is also partially supported by the Italian Ministry of University and Research (MUR) under the PRIN 2022 SEQUOIA project (code 2022KCKYA2).

%
%
\bibliographystyle{splncs04}
\bibliography{main}

\title{Learning Spectral and Polarimetric Clues for One-to-Multimodal Novel View Synthesis}
\subtitle{Supplementary Material}

\titlerunning{Learning Spect. and Pol. Clues for One-to-Multimodal Novel View Synthesis}

\author{Federico Lincetto\inst{1}\orcidlink{0009-0002-9137-1482} \and
Gianluca Agresti\inst{2}\orcidlink{0000-0001-7072-0079} \and
Mattia Rossi\inst{2}\orcidlink{0000-0001-5158-2395} \and Piergiorgio Sartor\inst{2} \and Pietro Zanuttigh\inst{1}\orcidlink{0000-0002-9502-2389}}

\authorrunning{F.~Lincetto et al.}

\institute{MEDIA Lab, University of Padova, Padova, Italy\\
\email{\{federico.lincetto,zanuttigh\}@dei.unipd.it} \and
Sony EUISPC, Sony Semiconductor Solutions Europe, Stuttgart, Germany\\
\url{https://medialab.dei.unipd.it/paper_data/SPoILeR/}}

\maketitlesupplementary
\thispagestyle{empty}
\renewcommand\thesection{\Alph{section}}
\setcounter{section}{0}

\noindent In this document, we present some additional experimental results that were not possible to include in the main document due to space constraints. In particular, we show some additional qualitative results of the \acrfull{ft} step in \cref{sec:ft}, a more detailed explanation of the comparison procedure with third-party modality conversion methods in \cref{sec:3rd}, some results on the X-NeRF dataset in \cref{sec:xnerf}, and finally, some qualitative and quantitative results of the \acrfull{pt} step in \cref{sec:pt}.
\section{Additional Fine-tuning Results}
\label{sec:ft}

\begin{figure*}[t]
    \centering
    \begin{minipage}{0.97\linewidth}
        \begin{minipage}{\linewidth}
            \centering
            \hfill
            \begin{minipage}{0.96\linewidth}
                \begin{minipage}{0.325\linewidth}
                    \centering
                    Ours
                \end{minipage}
                \begin{minipage}{0.325\linewidth}
                    \centering
                    GT
                \end{minipage}
                \begin{minipage}{0.325\linewidth}
                    \centering
                    Error
                \end{minipage}
            \end{minipage}
            \vspace{2pt}
        \end{minipage}
        \begin{minipage}{\linewidth}
            \begin{minipage}{0.025\linewidth}
                \vfill
                \rotatebox[origin=cb]{90}{RGB}
                \vfill
            \end{minipage}
            \begin{minipage}{0.97\linewidth}
                \begin{subfigure}{0.325\linewidth}
                    \centering
                    \includegraphics[width=\linewidth]{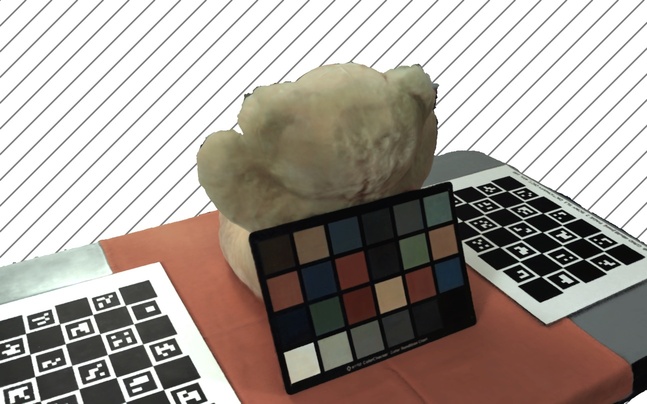}
                \end{subfigure}
                \begin{subfigure}{0.325\linewidth}
                    \centering
                    \includegraphics[width=\linewidth]{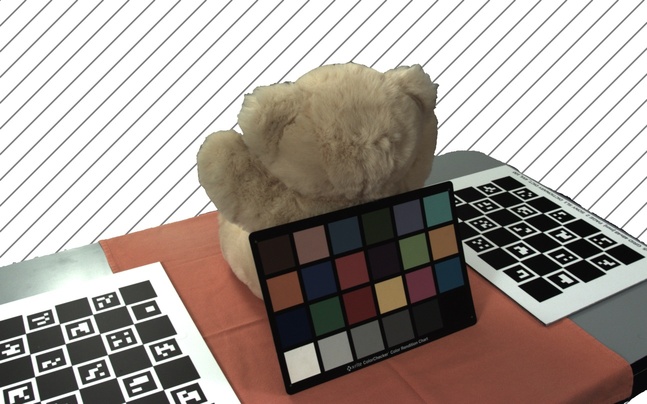}
                \end{subfigure}
                \begin{subfigure}{0.325\linewidth}
                    \centering
                    \includegraphics[width=\linewidth]{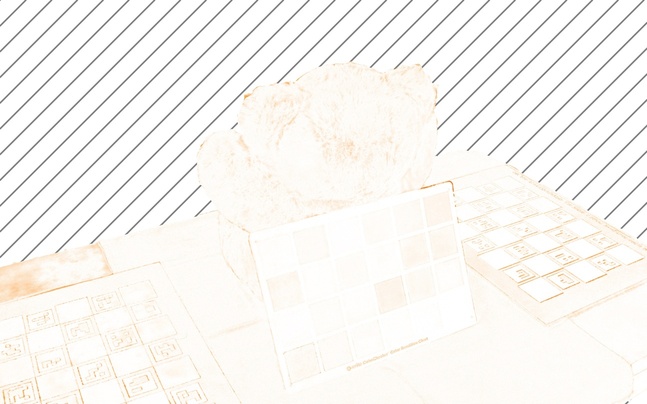}
                \end{subfigure}
            \end{minipage}
        \end{minipage}
    \end{minipage}
    \begin{minipage}{0.02\linewidth}
        \rotatebox[origin=cb]{270}{\hspace{0.3cm}Supervised}
    \end{minipage}
    \vspace{1pt}
    \rule{\linewidth}{2pt}
    \vspace{1pt}
    \begin{minipage}{0.97\linewidth}
        \begin{minipage}{\linewidth}
            \begin{minipage}{0.025\linewidth}
                \vfill
                \rotatebox[origin=cb]{90}{NIR}
                \vfill
            \end{minipage}
            \begin{minipage}{0.97\linewidth}
                \begin{subfigure}{0.325\linewidth}
                    \centering
                    \includegraphics[width=\linewidth]{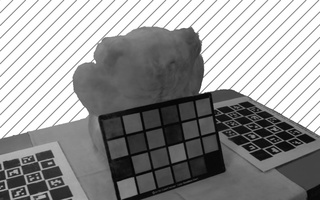}
                \end{subfigure}
                \begin{subfigure}{0.325\linewidth}
                    \centering
                    \includegraphics[width=\linewidth]{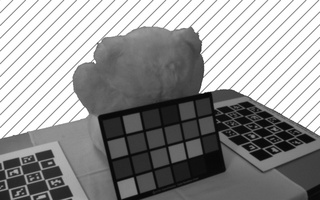}
                \end{subfigure}
                \begin{subfigure}{0.325\linewidth}
                    \centering
                    \includegraphics[width=\linewidth]{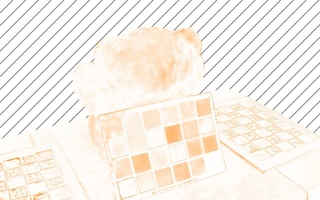}
                \end{subfigure}
            \end{minipage}
        \end{minipage}
        \begin{minipage}{\linewidth}
            \begin{minipage}{0.025\linewidth}
                \vfill
                \rotatebox[origin=cb]{90}{Mono}
                \vfill
            \end{minipage}
            \begin{minipage}{0.97\linewidth}
                \begin{subfigure}{0.325\linewidth}
                    \centering
                    \includegraphics[width=\linewidth]{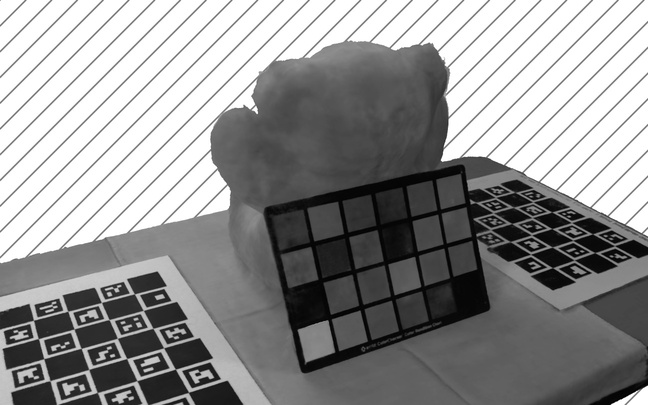}
                \end{subfigure}
                \begin{subfigure}{0.325\linewidth}
                    \centering
                    \includegraphics[width=\linewidth]{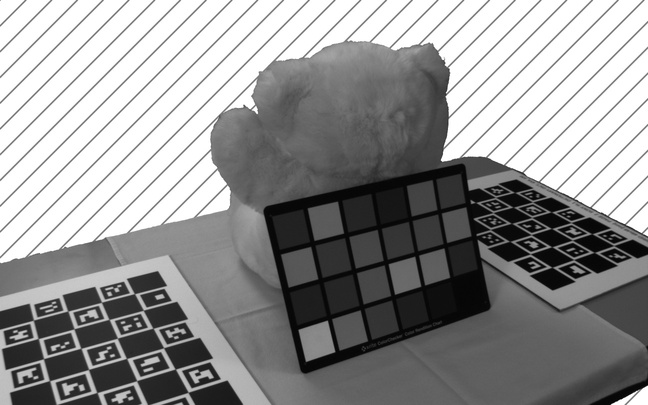}
                \end{subfigure}
                \begin{subfigure}{0.325\linewidth}
                    \centering
                    \includegraphics[width=\linewidth]{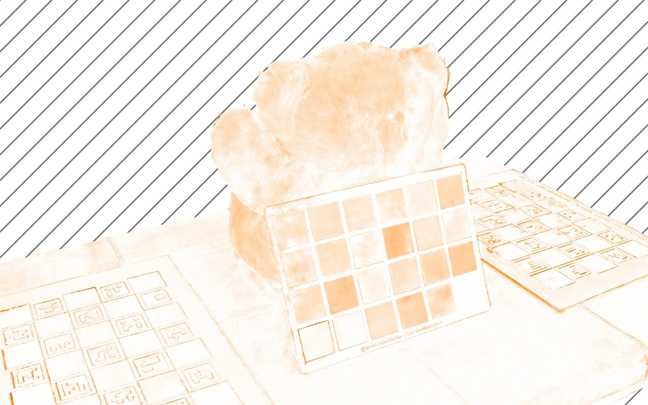}
                \end{subfigure}
            \end{minipage}
        \end{minipage}
        \begin{minipage}{\linewidth}
            \begin{minipage}{0.025\linewidth}
                \vfill
                \rotatebox[origin=cb]{90}{Pol}
                \vfill
            \end{minipage}
            \begin{minipage}{0.97\linewidth}
                \begin{subfigure}{0.325\linewidth}
                    \centering
                    \includegraphics[width=\linewidth]{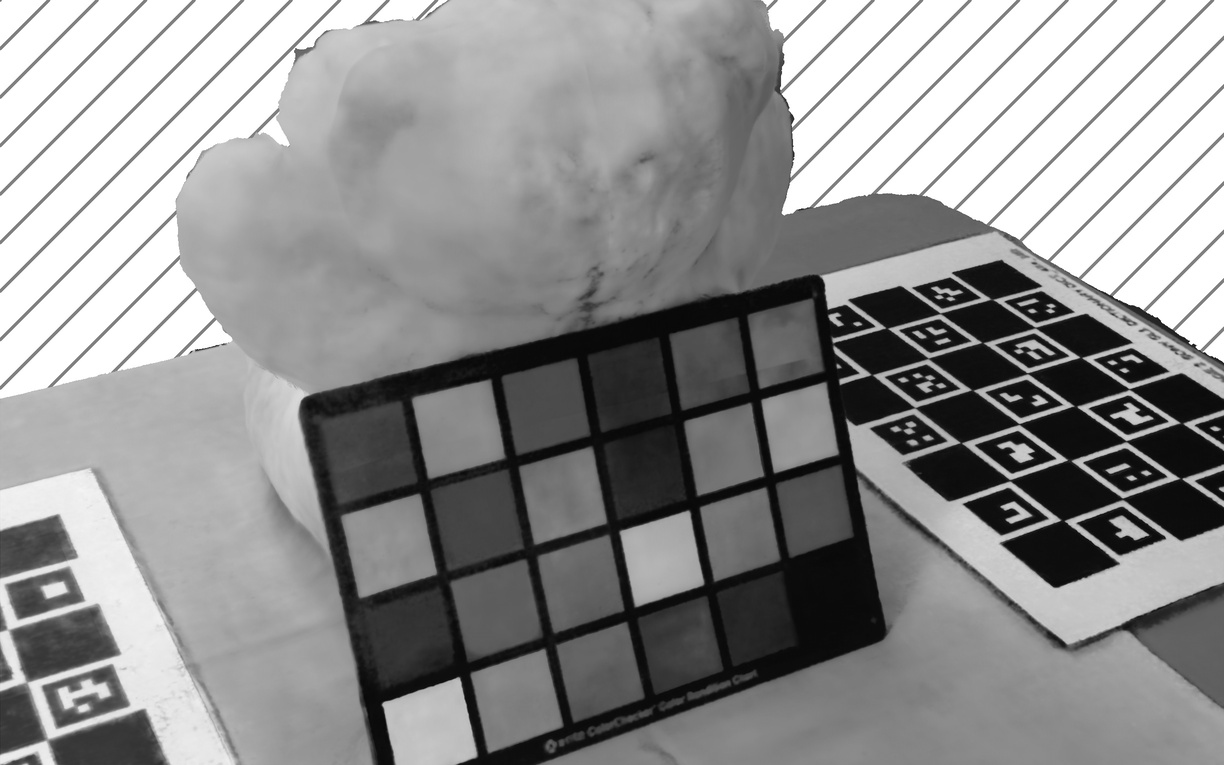}
                \end{subfigure}
                \begin{subfigure}{0.325\linewidth}
                    \centering
                    \includegraphics[width=\linewidth]{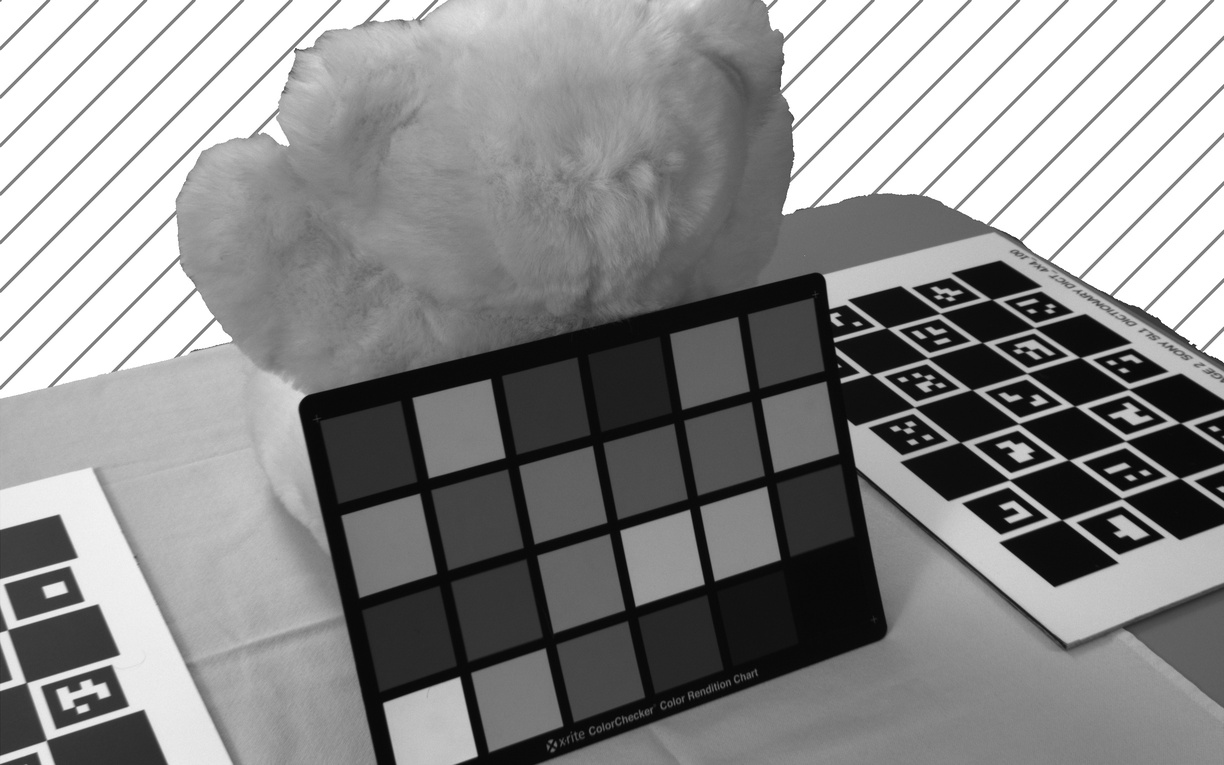}
                \end{subfigure}
                \begin{subfigure}{0.325\linewidth}
                    \centering
                    \includegraphics[width=\linewidth]{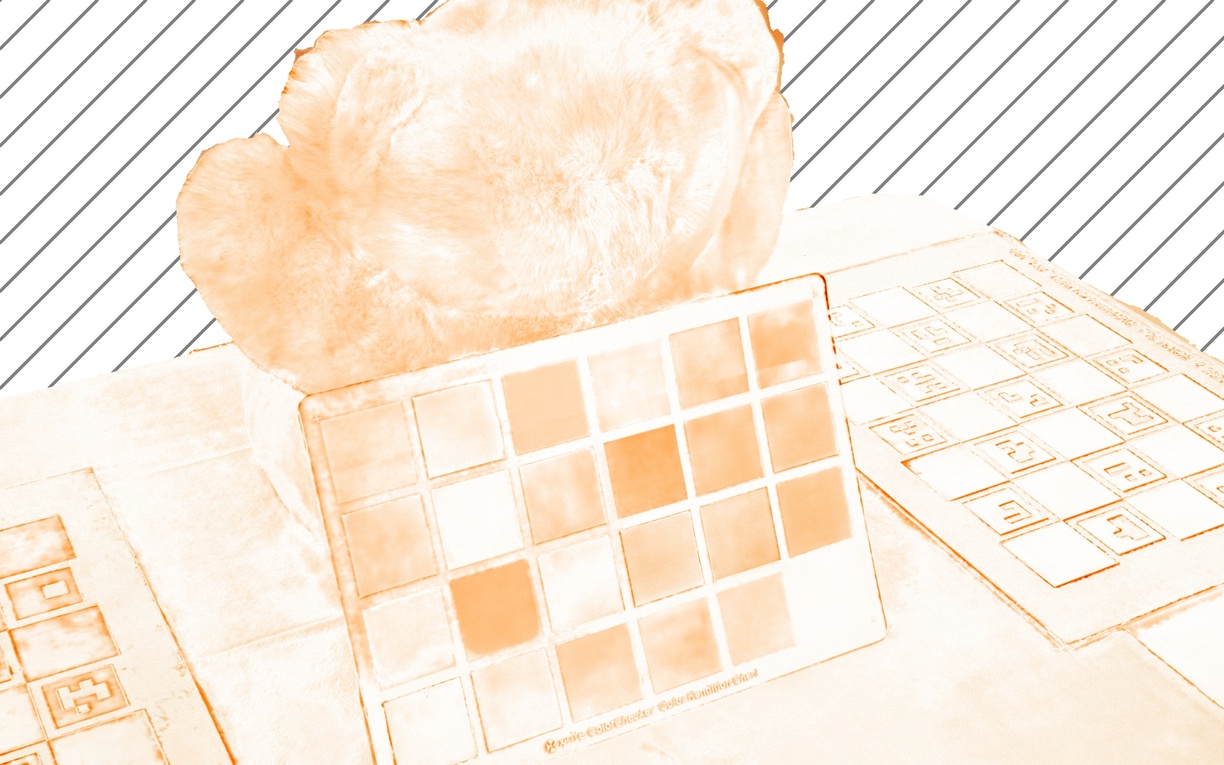}
                \end{subfigure}
            \end{minipage}
        \end{minipage}
        \begin{minipage}{\linewidth}
            \begin{minipage}{0.025\linewidth}
                \vfill
                \rotatebox[origin=cb]{90}{MS}
                \vfill
            \end{minipage}
            \begin{minipage}{0.97\linewidth}
                \begin{subfigure}{0.325\linewidth}
                    \centering
                    \includegraphics[width=\linewidth]{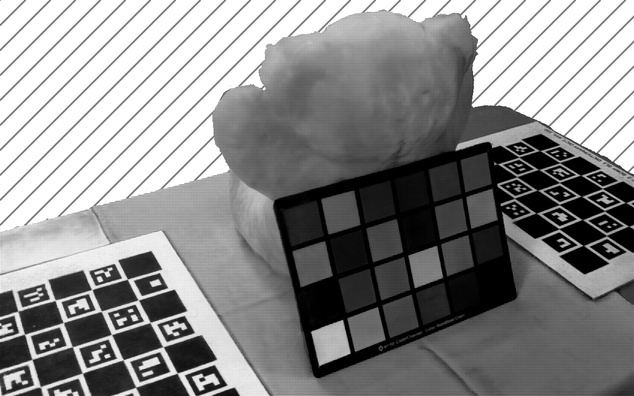}
                \end{subfigure}
                \begin{subfigure}{0.325\linewidth}
                    \centering
                    \includegraphics[width=\linewidth]{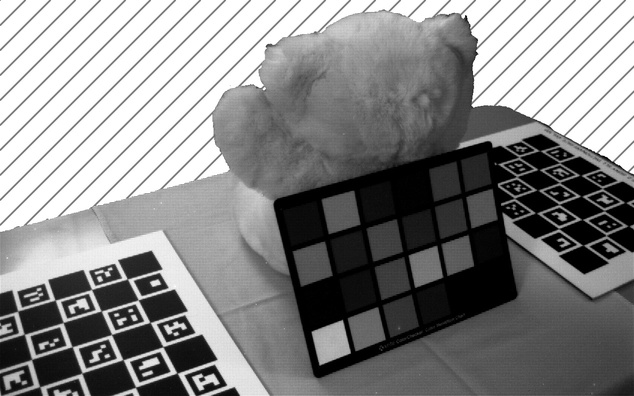}
                \end{subfigure}
                \begin{subfigure}{0.325\linewidth}
                    \centering
                    \includegraphics[width=\linewidth]{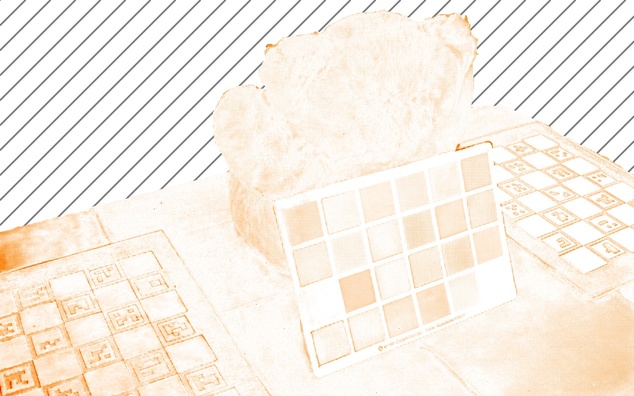}
                \end{subfigure}
            \end{minipage}
        \end{minipage}
    \end{minipage}
    \begin{minipage}{0.02\linewidth}
        \rotatebox[origin=cb]{270}{No supervision}
    \end{minipage}
    \caption{Qualitative renderings of the ``Teddybear'' scene from the \gls{ft} step supervised with only RGB data. All frames are mosaicked, except RGB frames. RGB is demosaicked only for visualization purposes.}
    \label{fig:qualitative_teddybear}
    \vspace{-0.5cm}
\end{figure*}

\begin{figure*}[t]
    \centering
    \begin{minipage}{0.97\linewidth}
        \begin{minipage}{\linewidth}
            \centering
            \hfill
            \begin{minipage}{0.96\linewidth}
                \begin{minipage}{0.325\linewidth}
                    \centering
                    Ours
                \end{minipage}
                \begin{minipage}{0.325\linewidth}
                    \centering
                    GT
                \end{minipage}
                \begin{minipage}{0.325\linewidth}
                    \centering
                    Error
                \end{minipage}
            \end{minipage}
            \vspace{2pt}
        \end{minipage}
        \begin{minipage}{\linewidth}
            \begin{minipage}{0.025\linewidth}
                \vfill
                \rotatebox[origin=cb]{90}{RGB}
                \vfill
            \end{minipage}
            \begin{minipage}{0.97\linewidth}
                \begin{subfigure}{0.325\linewidth}
                    \centering
                    \includegraphics[width=\linewidth]{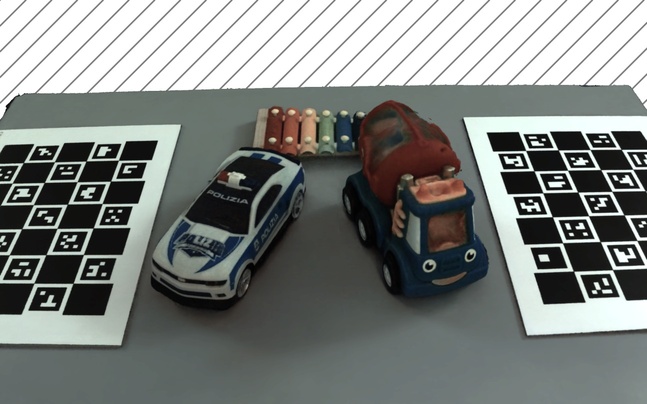}
                \end{subfigure}
                \begin{subfigure}{0.325\linewidth}
                    \centering
                    \includegraphics[width=\linewidth]{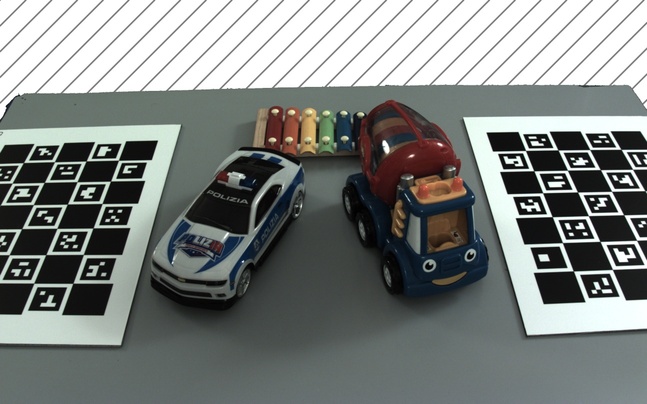}
                \end{subfigure}
                \begin{subfigure}{0.325\linewidth}
                    \centering
                    \includegraphics[width=\linewidth]{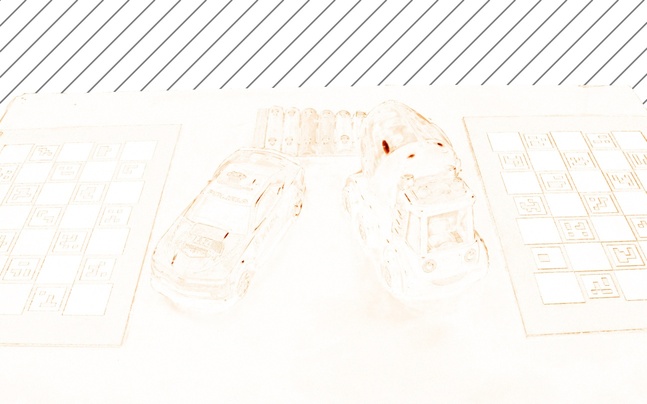}
                \end{subfigure}
            \end{minipage}
        \end{minipage}
    \end{minipage}
    \begin{minipage}{0.02\linewidth}
        \rotatebox[origin=cb]{270}{\hspace{0.3cm}Supervised}
    \end{minipage}
    \vspace{1pt}
    \rule{\linewidth}{2pt}
    \vspace{1pt}
    \begin{minipage}{0.97\linewidth}
        \begin{minipage}{\linewidth}
            \begin{minipage}{0.025\linewidth}
                \vfill
                \rotatebox[origin=cb]{90}{NIR}
                \vfill
            \end{minipage}
            \begin{minipage}{0.97\linewidth}
                \begin{subfigure}{0.325\linewidth}
                    \centering
                    \includegraphics[width=\linewidth]{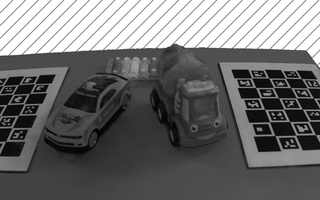}
                \end{subfigure}
                \begin{subfigure}{0.325\linewidth}
                    \centering
                    \includegraphics[width=\linewidth]{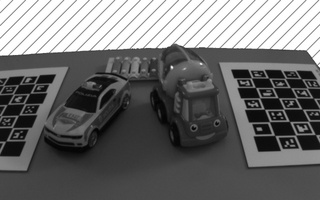}
                \end{subfigure}
                \begin{subfigure}{0.325\linewidth}
                    \centering
                    \includegraphics[width=\linewidth]{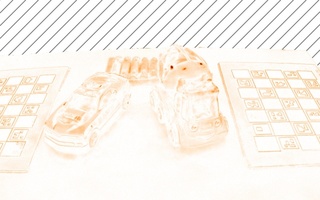}
                \end{subfigure}
            \end{minipage}
        \end{minipage}
        \begin{minipage}{\linewidth}
            \begin{minipage}{0.025\linewidth}
                \vfill
                \rotatebox[origin=cb]{90}{Mono}
                \vfill
            \end{minipage}
            \begin{minipage}{0.97\linewidth}
                \begin{subfigure}{0.325\linewidth}
                    \centering
                    \includegraphics[width=\linewidth]{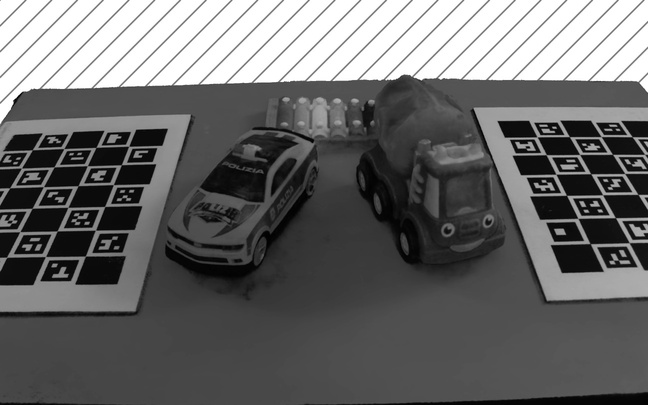}
                \end{subfigure}
                \begin{subfigure}{0.325\linewidth}
                    \centering
                    \includegraphics[width=\linewidth]{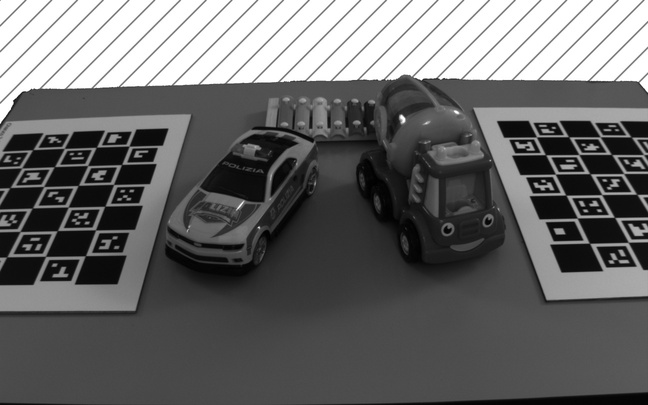}
                \end{subfigure}
                \begin{subfigure}{0.325\linewidth}
                    \centering
                    \includegraphics[width=\linewidth]{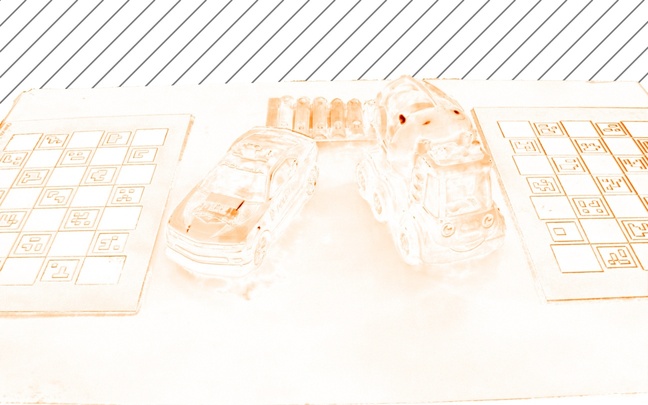}
                \end{subfigure}
            \end{minipage}
        \end{minipage}
        \begin{minipage}{\linewidth}
            \begin{minipage}{0.025\linewidth}
                \vfill
                \rotatebox[origin=cb]{90}{Pol}
                \vfill
            \end{minipage}
            \begin{minipage}{0.97\linewidth}
                \begin{subfigure}{0.325\linewidth}
                    \centering
                    \includegraphics[width=\linewidth]{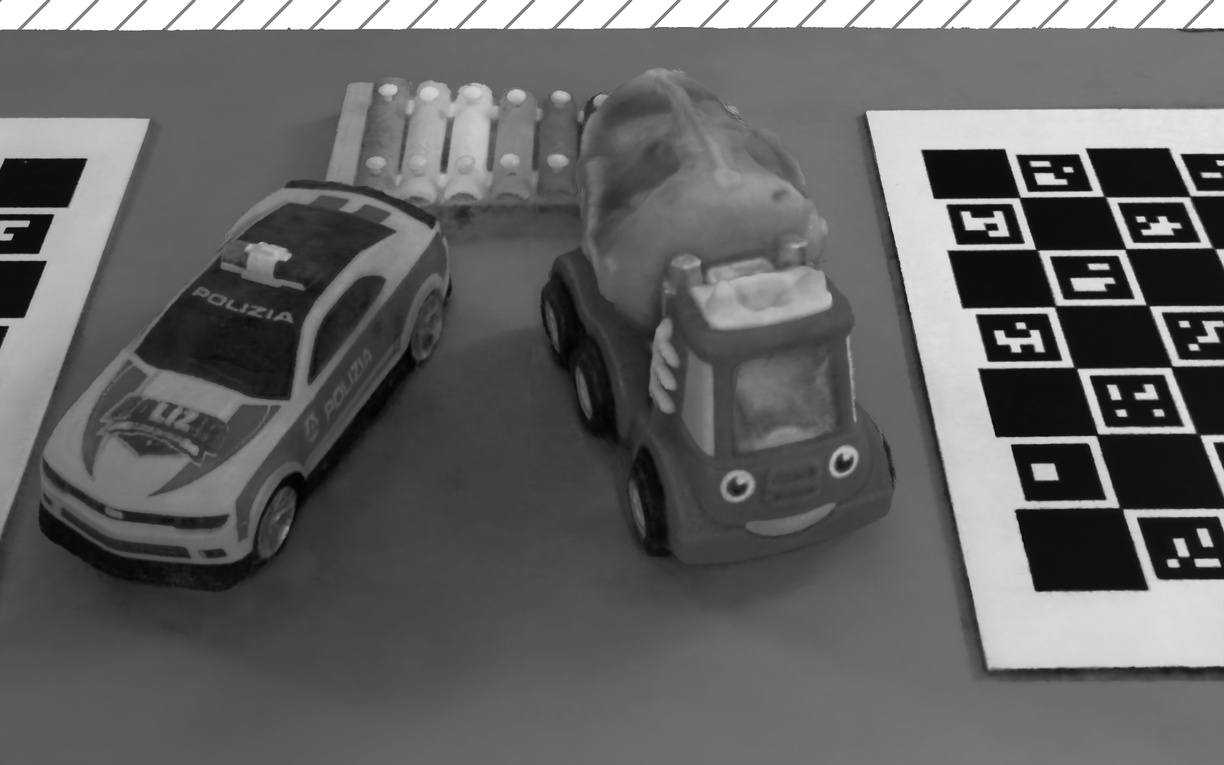}
                \end{subfigure}
                \begin{subfigure}{0.325\linewidth}
                    \centering
                    \includegraphics[width=\linewidth]{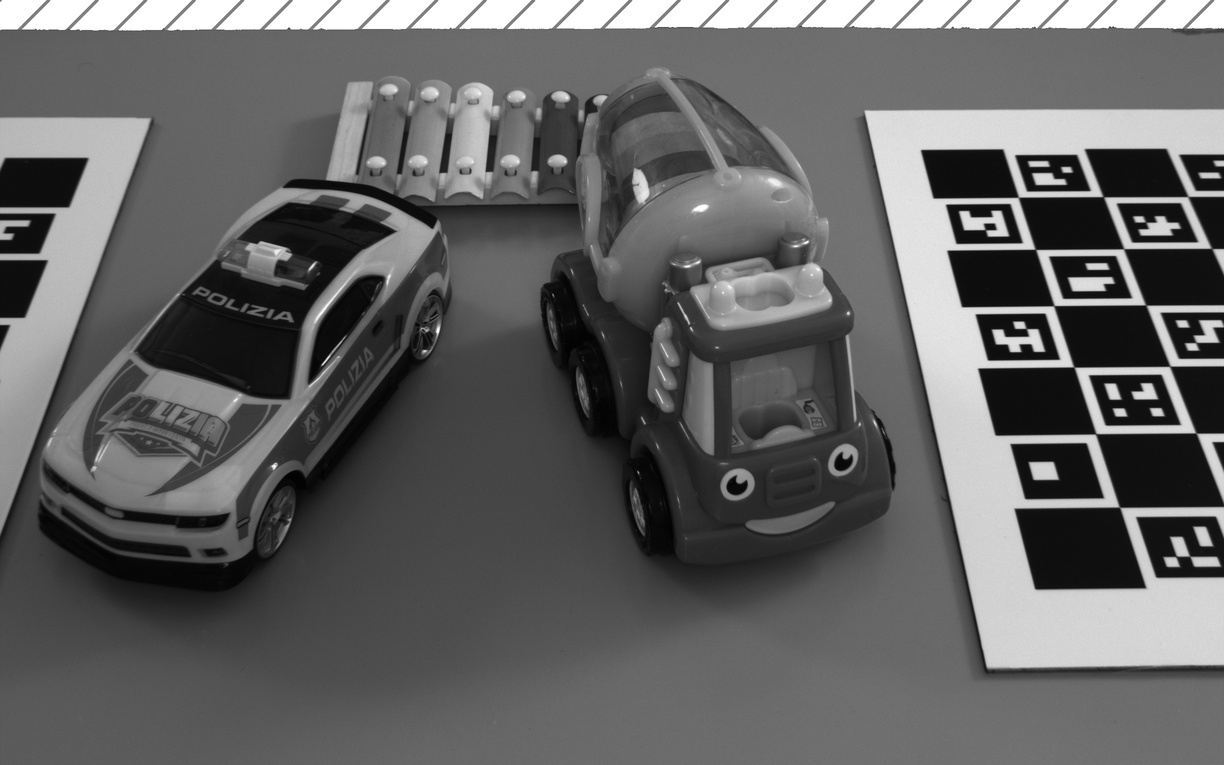}
                \end{subfigure}
                \begin{subfigure}{0.325\linewidth}
                    \centering
                    \includegraphics[width=\linewidth]{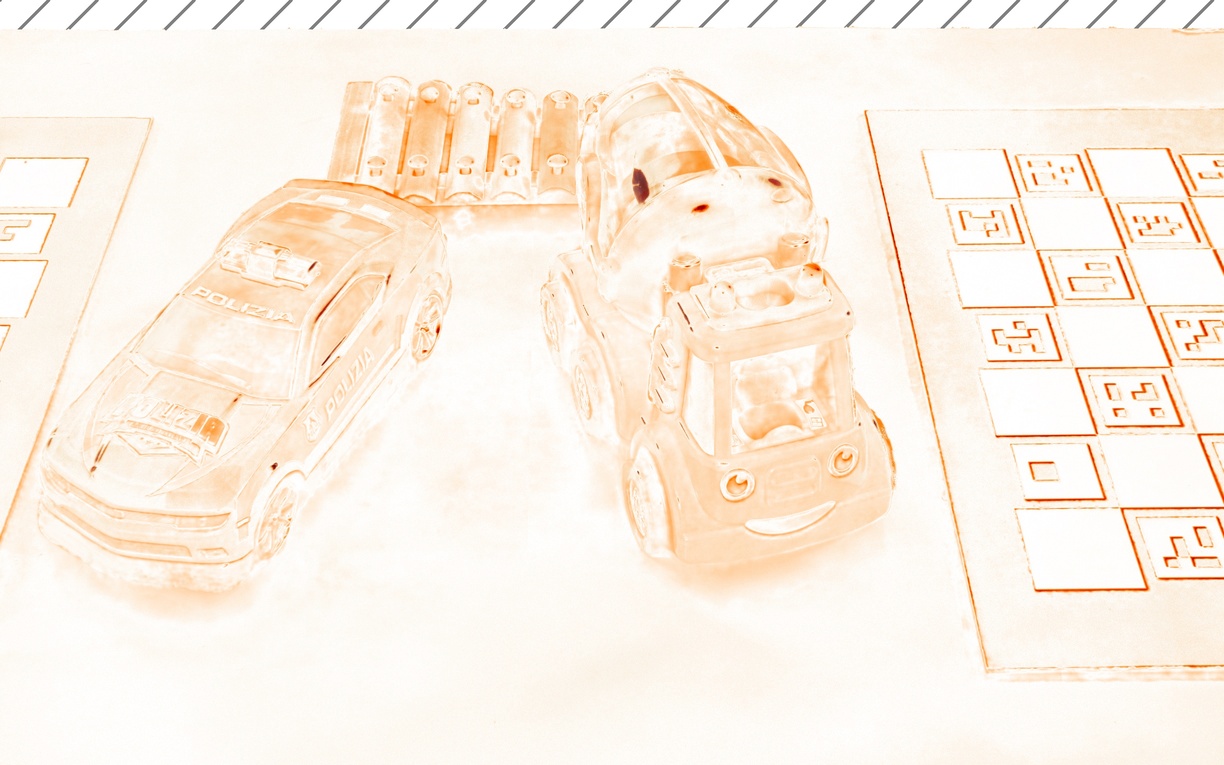}
                \end{subfigure}
            \end{minipage}
        \end{minipage}
        \begin{minipage}{\linewidth}
            \begin{minipage}{0.025\linewidth}
                \vfill
                \rotatebox[origin=cb]{90}{MS}
                \vfill
            \end{minipage}
            \begin{minipage}{0.97\linewidth}
                \begin{subfigure}{0.325\linewidth}
                    \centering
                    \includegraphics[width=\linewidth]{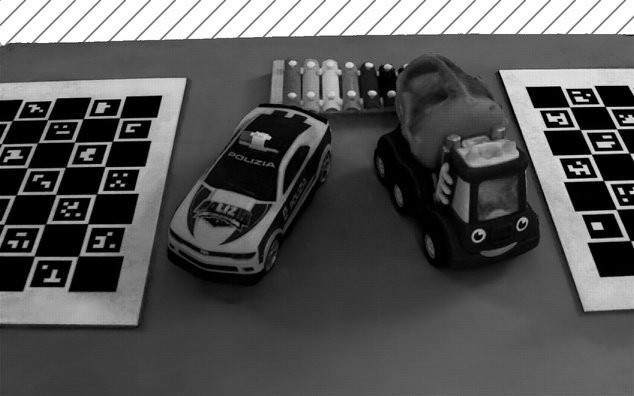}
                \end{subfigure}
                \begin{subfigure}{0.325\linewidth}
                    \centering
                    \includegraphics[width=\linewidth]{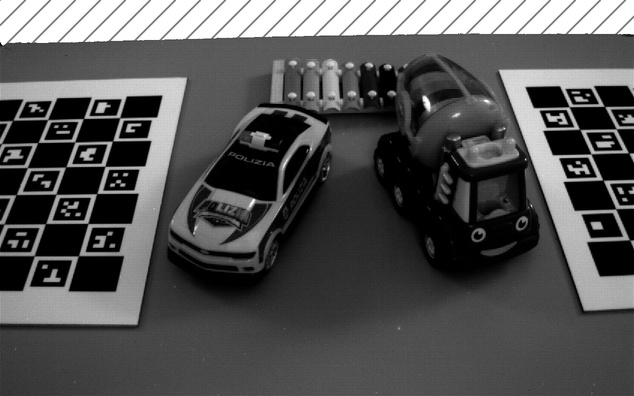}
                \end{subfigure}
                \begin{subfigure}{0.325\linewidth}
                    \centering
                    \includegraphics[width=\linewidth]{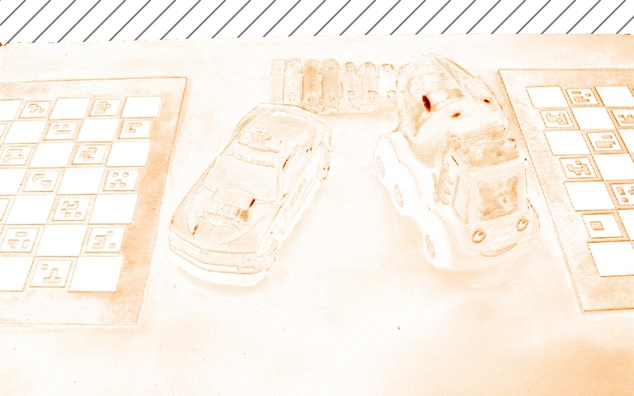}
                \end{subfigure}
            \end{minipage}
        \end{minipage}
    \end{minipage}
    \begin{minipage}{0.02\linewidth}
        \rotatebox[origin=cb]{270}{No supervision}
    \end{minipage}
    \caption{Qualitative renderings of the ``Toys'' scene from the \gls{ft} step  supervised with only RGB data. All frames are mosaicked, except RGB frames. RGB is demosaicked only for visualization purposes.}
    \label{fig:qualitative_toys}
    \vspace{-0.5cm}
\end{figure*}

\begin{figure*}[t]
    \centering
    \begin{minipage}{0.97\linewidth}
        \begin{minipage}{\linewidth}
            \centering
            \hfill
            \begin{minipage}{0.96\linewidth}
                \begin{minipage}{0.325\linewidth}
                    \centering
                    Ours
                \end{minipage}
                \begin{minipage}{0.325\linewidth}
                    \centering
                    GT
                \end{minipage}
                \begin{minipage}{0.325\linewidth}
                    \centering
                    Error
                \end{minipage}
            \end{minipage}
            \vspace{2pt}
        \end{minipage}
        \begin{minipage}{\linewidth}
            \begin{minipage}{0.025\linewidth}
                \vfill
                \rotatebox[origin=cb]{90}{RGB}
                \vfill
            \end{minipage}
            \begin{minipage}{0.97\linewidth}
                \begin{subfigure}{0.325\linewidth}
                    \centering
                    \includegraphics[width=\linewidth]{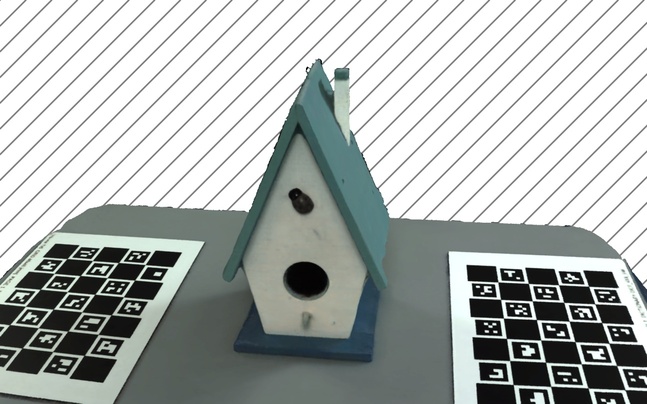}
                \end{subfigure}
                \begin{subfigure}{0.325\linewidth}
                    \centering
                    \includegraphics[width=\linewidth]{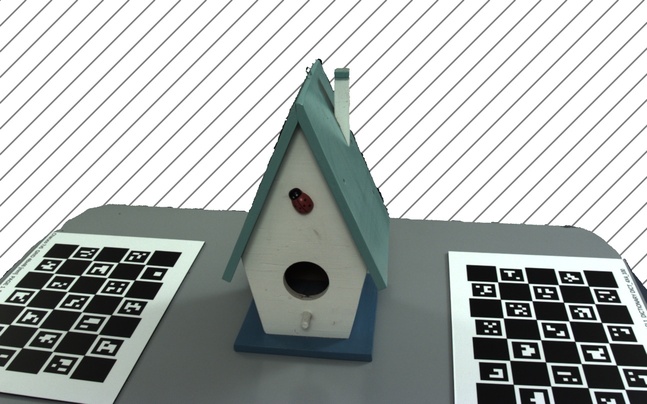}
                \end{subfigure}
                \begin{subfigure}{0.325\linewidth}
                    \centering
                    \includegraphics[width=\linewidth]{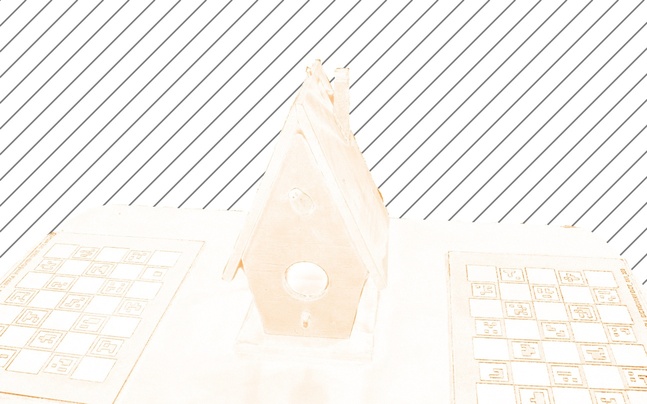}
                \end{subfigure}
            \end{minipage}
        \end{minipage}
    \end{minipage}
    \begin{minipage}{0.02\linewidth}
        \rotatebox[origin=cb]{270}{\hspace{0.3cm}Supervised}
    \end{minipage}
    \vspace{1pt}
    \rule{\linewidth}{2pt}
    \vspace{1pt}
    \begin{minipage}{0.97\linewidth}
        \begin{minipage}{\linewidth}
            \begin{minipage}{0.025\linewidth}
                \vfill
                \rotatebox[origin=cb]{90}{NIR}
                \vfill
            \end{minipage}
            \begin{minipage}{0.97\linewidth}
                \begin{subfigure}{0.325\linewidth}
                    \centering
                    \includegraphics[width=\linewidth]{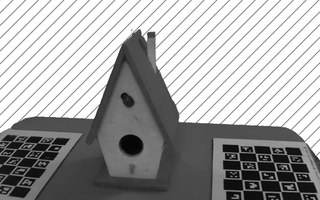}
                \end{subfigure}
                \begin{subfigure}{0.325\linewidth}
                    \centering
                    \includegraphics[width=\linewidth]{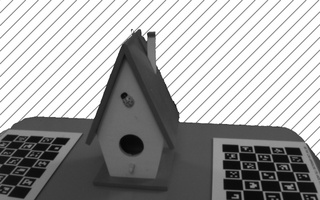}
                \end{subfigure}
                \begin{subfigure}{0.325\linewidth}
                    \centering
                    \includegraphics[width=\linewidth]{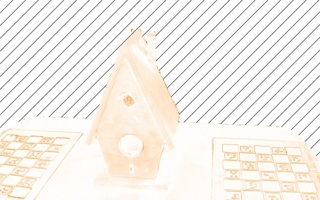}
                \end{subfigure}
            \end{minipage}
        \end{minipage}
        \begin{minipage}{\linewidth}
            \begin{minipage}{0.025\linewidth}
                \vfill
                \rotatebox[origin=cb]{90}{Mono}
                \vfill
            \end{minipage}
            \begin{minipage}{0.97\linewidth}
                \begin{subfigure}{0.325\linewidth}
                    \centering
                    \includegraphics[width=\linewidth]{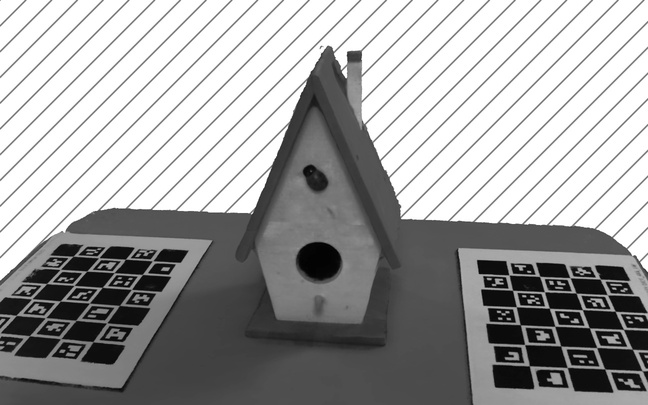}
                \end{subfigure}
                \begin{subfigure}{0.325\linewidth}
                    \centering
                    \includegraphics[width=\linewidth]{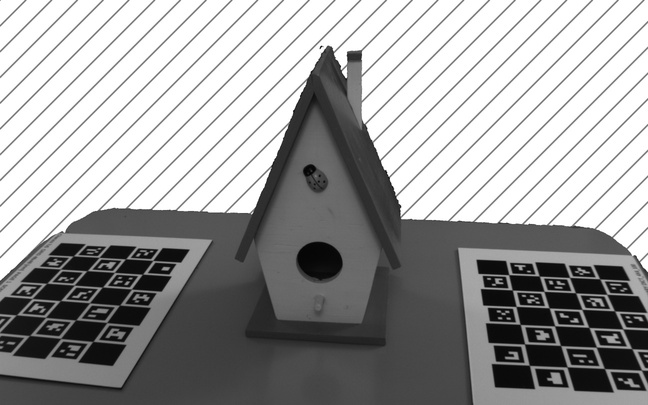}
                \end{subfigure}
                \begin{subfigure}{0.325\linewidth}
                    \centering
                    \includegraphics[width=\linewidth]{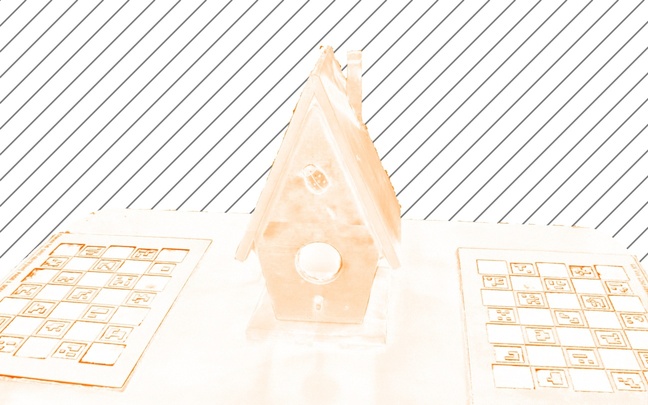}
                \end{subfigure}
            \end{minipage}
        \end{minipage}
        \begin{minipage}{\linewidth}
            \begin{minipage}{0.025\linewidth}
                \vfill
                \rotatebox[origin=cb]{90}{Pol}
                \vfill
            \end{minipage}
            \begin{minipage}{0.97\linewidth}
                \begin{subfigure}{0.325\linewidth}
                    \centering
                    \includegraphics[width=\linewidth]{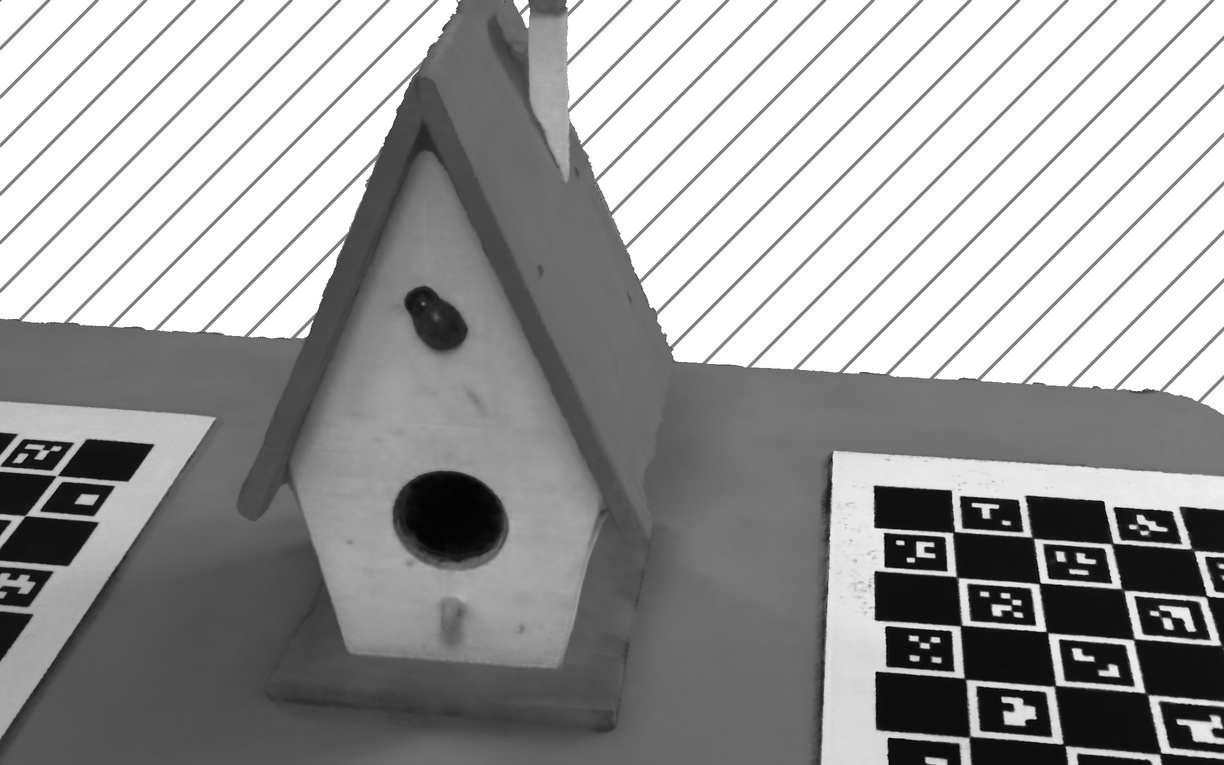}
                \end{subfigure}
                \begin{subfigure}{0.325\linewidth}
                    \centering
                    \includegraphics[width=\linewidth]{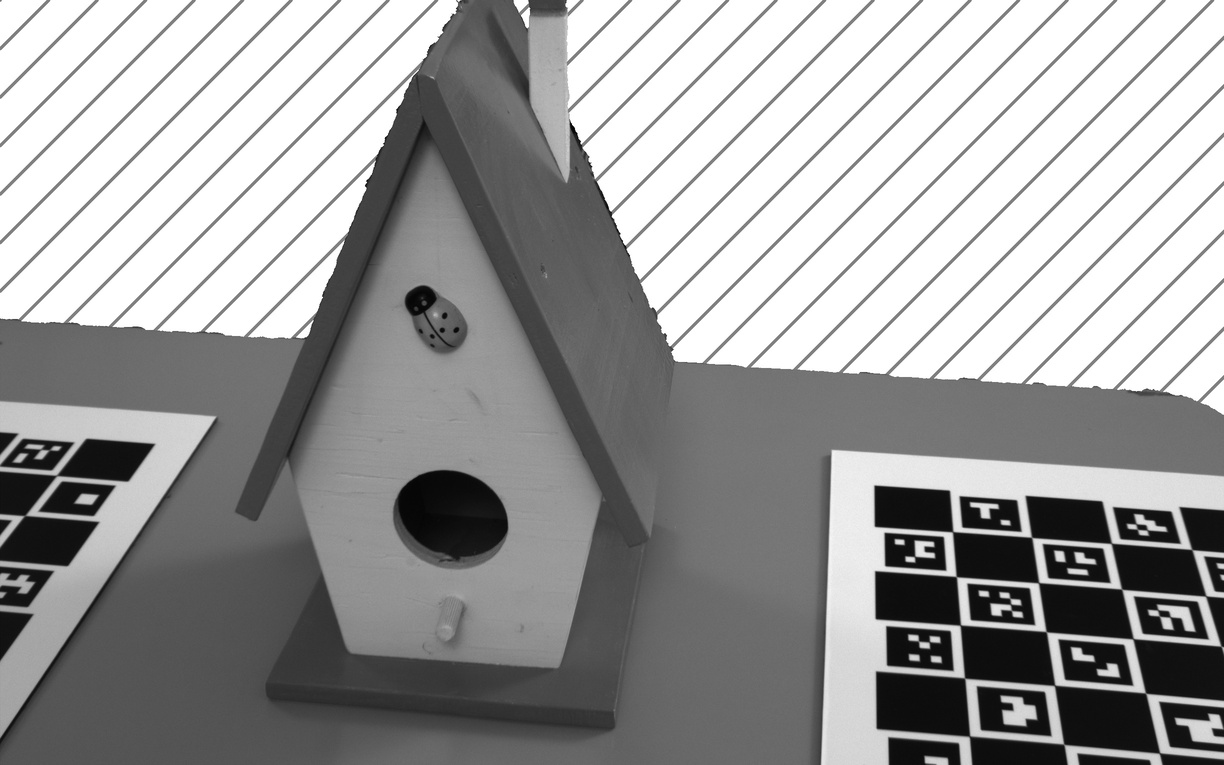}
                \end{subfigure}
                \begin{subfigure}{0.325\linewidth}
                    \centering
                    \includegraphics[width=\linewidth]{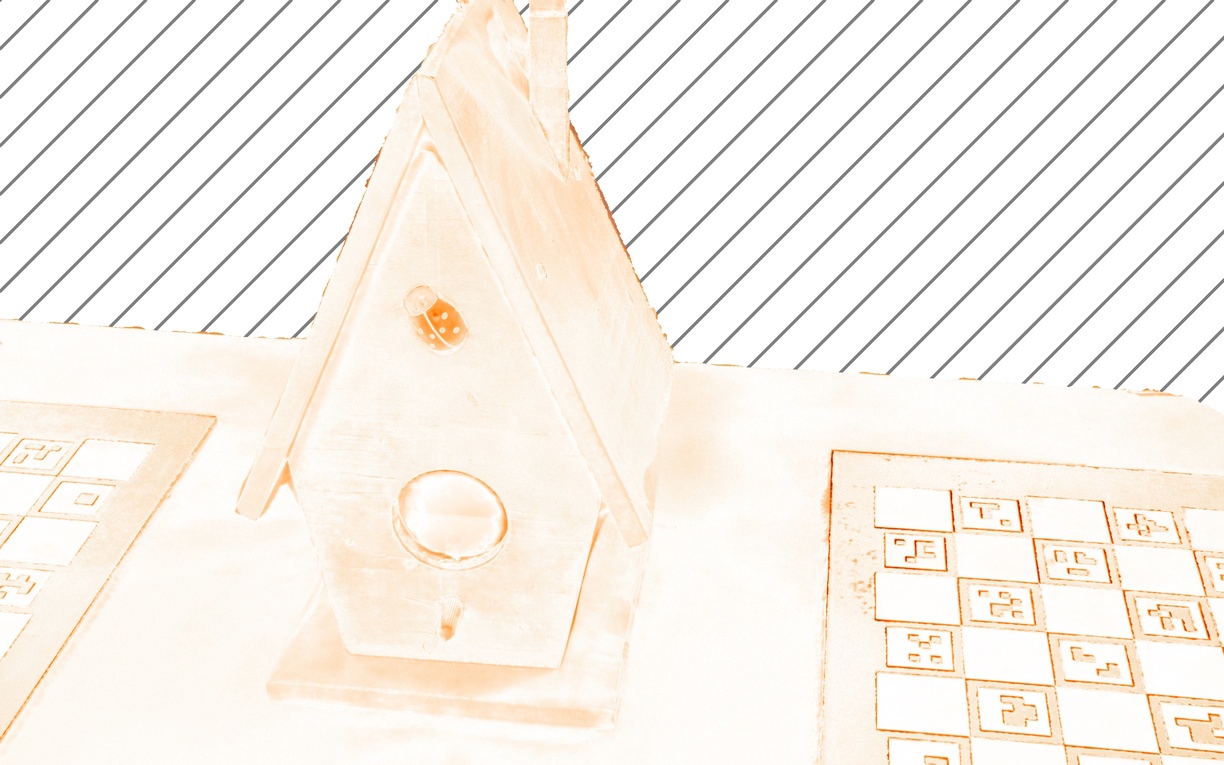}
                \end{subfigure}
            \end{minipage}
        \end{minipage}
        \begin{minipage}{\linewidth}
            \begin{minipage}{0.025\linewidth}
                \vfill
                \rotatebox[origin=cb]{90}{MS}
                \vfill
            \end{minipage}
            \begin{minipage}{0.97\linewidth}
                \begin{subfigure}{0.325\linewidth}
                    \centering
                    \includegraphics[width=\linewidth]{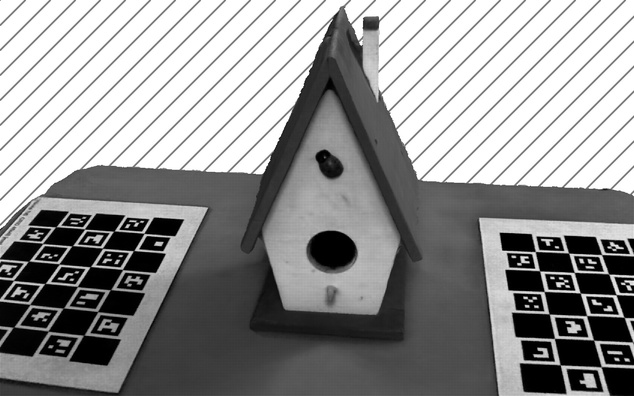}
                \end{subfigure}
                \begin{subfigure}{0.325\linewidth}
                    \centering
                    \includegraphics[width=\linewidth]{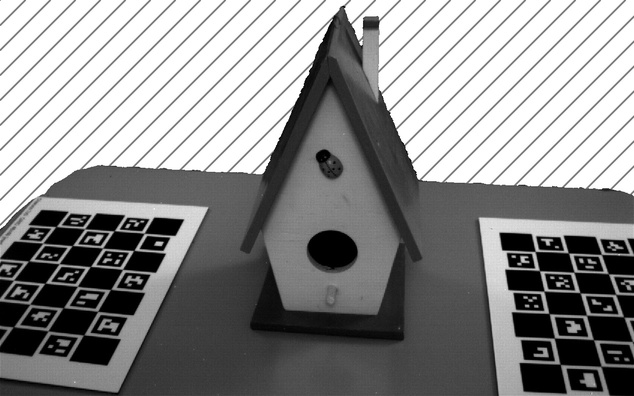}
                \end{subfigure}
                \begin{subfigure}{0.325\linewidth}
                    \centering
                    \includegraphics[width=\linewidth]{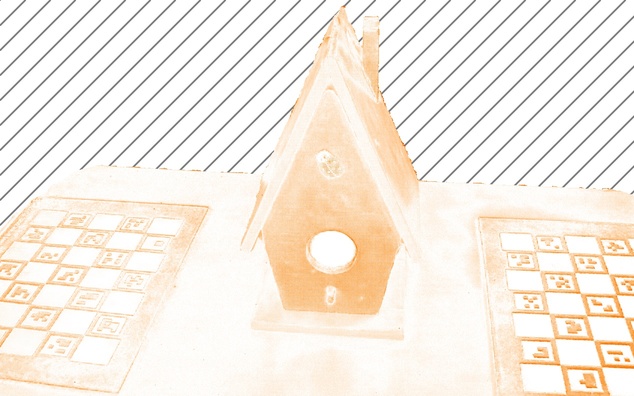}
                \end{subfigure}
            \end{minipage}
        \end{minipage}
    \end{minipage}
    \begin{minipage}{0.02\linewidth}
        \rotatebox[origin=cb]{270}{No supervision}
    \end{minipage}
    \caption{Qualitative renderings of the ``Birdhouse'' scene from the \gls{ft} step supervised with only RGB data. All frames are mosaicked, except RGB frames. RGB is demosaicked only for visualization purposes.}
    \label{fig:qualitative_birdhouse}
    \vspace{-0.5cm}
\end{figure*}

\begin{figure*}[t]
    \centering
    \begin{minipage}{0.97\linewidth}
        \begin{minipage}{\linewidth}
            \centering
            \hfill
            \begin{minipage}{0.96\linewidth}
                \begin{minipage}{0.325\linewidth}
                    \centering
                    Ours
                \end{minipage}
                \begin{minipage}{0.325\linewidth}
                    \centering
                    GT
                \end{minipage}
                \begin{minipage}{0.325\linewidth}
                    \centering
                    Error
                \end{minipage}
            \end{minipage}
            \vspace{2pt}
        \end{minipage}
        \begin{minipage}{\linewidth}
            \begin{minipage}{0.025\linewidth}
                \vfill
                \rotatebox[origin=cb]{90}{RGB}
                \vfill
            \end{minipage}
            \begin{minipage}{0.97\linewidth}
                \begin{subfigure}{0.325\linewidth}
                    \centering
                    \includegraphics[width=\linewidth]{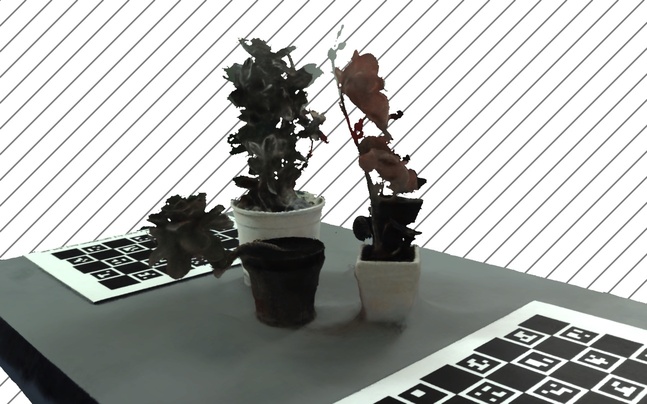}
                \end{subfigure}
                \begin{subfigure}{0.325\linewidth}
                    \centering
                    \includegraphics[width=\linewidth]{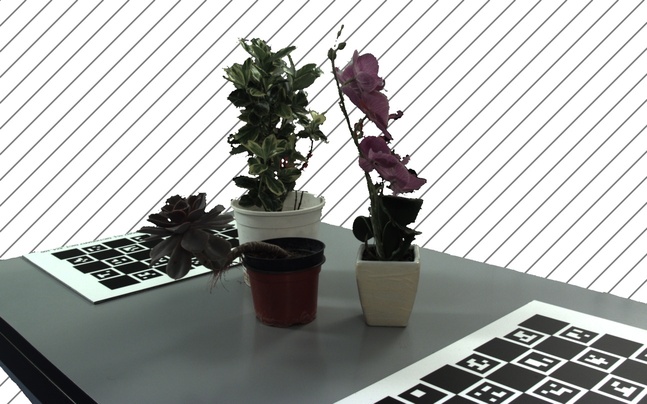}
                \end{subfigure}
                \begin{subfigure}{0.325\linewidth}
                    \centering
                    \includegraphics[width=\linewidth]{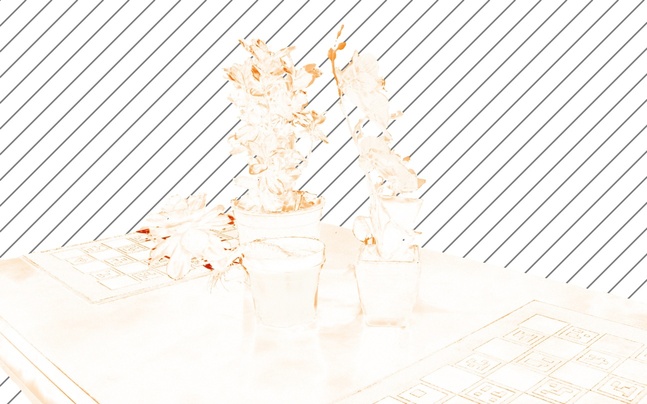}
                \end{subfigure}
            \end{minipage}
        \end{minipage}
    \end{minipage}
    \begin{minipage}{0.02\linewidth}
        \rotatebox[origin=cb]{270}{\hspace{0.3cm}Supervised}
    \end{minipage}
    \vspace{1pt}
    \rule{\linewidth}{2pt}
    \vspace{1pt}
    \begin{minipage}{0.97\linewidth}
        \begin{minipage}{\linewidth}
            \begin{minipage}{0.025\linewidth}
                \vfill
                \rotatebox[origin=cb]{90}{NIR}
                \vfill
            \end{minipage}
            \begin{minipage}{0.97\linewidth}
                \begin{subfigure}{0.325\linewidth}
                    \centering
                    \includegraphics[width=\linewidth]{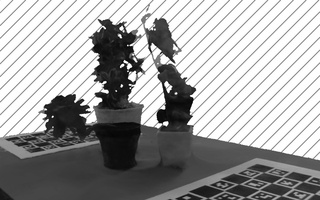}
                \end{subfigure}
                \begin{subfigure}{0.325\linewidth}
                    \centering
                    \includegraphics[width=\linewidth]{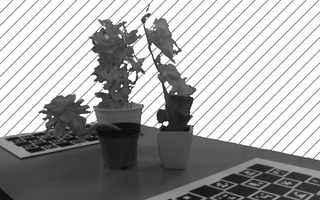}
                \end{subfigure}
                \begin{subfigure}{0.325\linewidth}
                    \centering
                    \includegraphics[width=\linewidth]{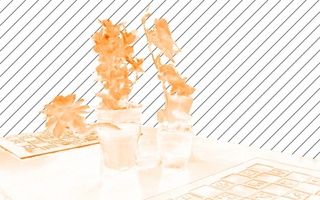}
                \end{subfigure}
            \end{minipage}
        \end{minipage}
        \begin{minipage}{\linewidth}
            \begin{minipage}{0.025\linewidth}
                \vfill
                \rotatebox[origin=cb]{90}{Mono}
                \vfill
            \end{minipage}
            \begin{minipage}{0.97\linewidth}
                \begin{subfigure}{0.325\linewidth}
                    \centering
                    \includegraphics[width=\linewidth]{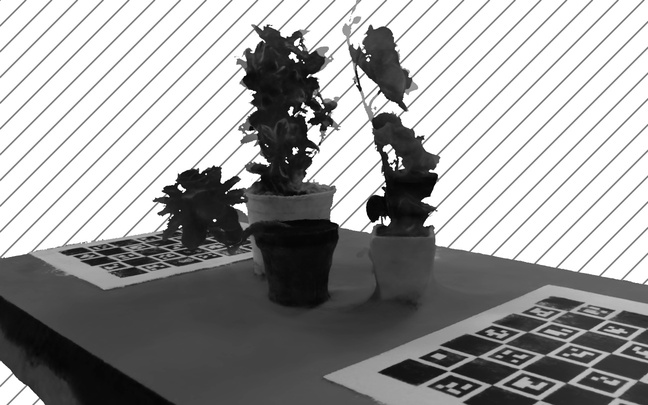}
                \end{subfigure}
                \begin{subfigure}{0.325\linewidth}
                    \centering
                    \includegraphics[width=\linewidth]{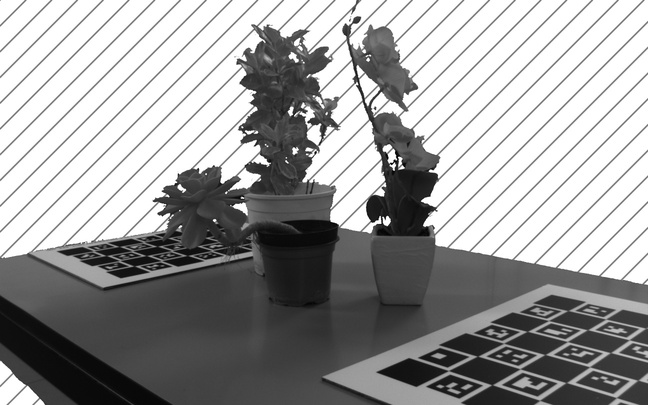}
                \end{subfigure}
                \begin{subfigure}{0.325\linewidth}
                    \centering
                    \includegraphics[width=\linewidth]{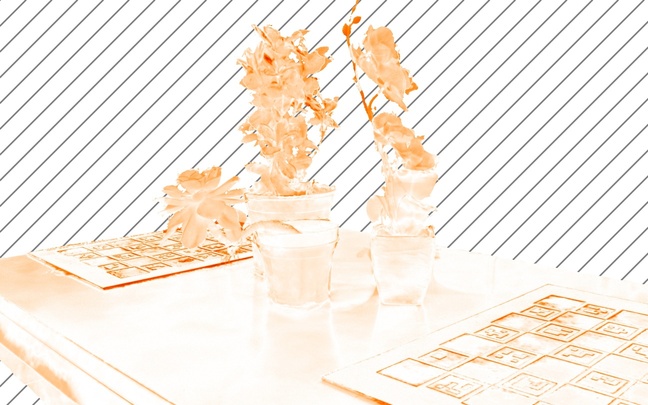}
                \end{subfigure}
            \end{minipage}
        \end{minipage}
        \begin{minipage}{\linewidth}
            \begin{minipage}{0.025\linewidth}
                \vfill
                \rotatebox[origin=cb]{90}{Pol}
                \vfill
            \end{minipage}
            \begin{minipage}{0.97\linewidth}
                \begin{subfigure}{0.325\linewidth}
                    \centering
                    \includegraphics[width=\linewidth]{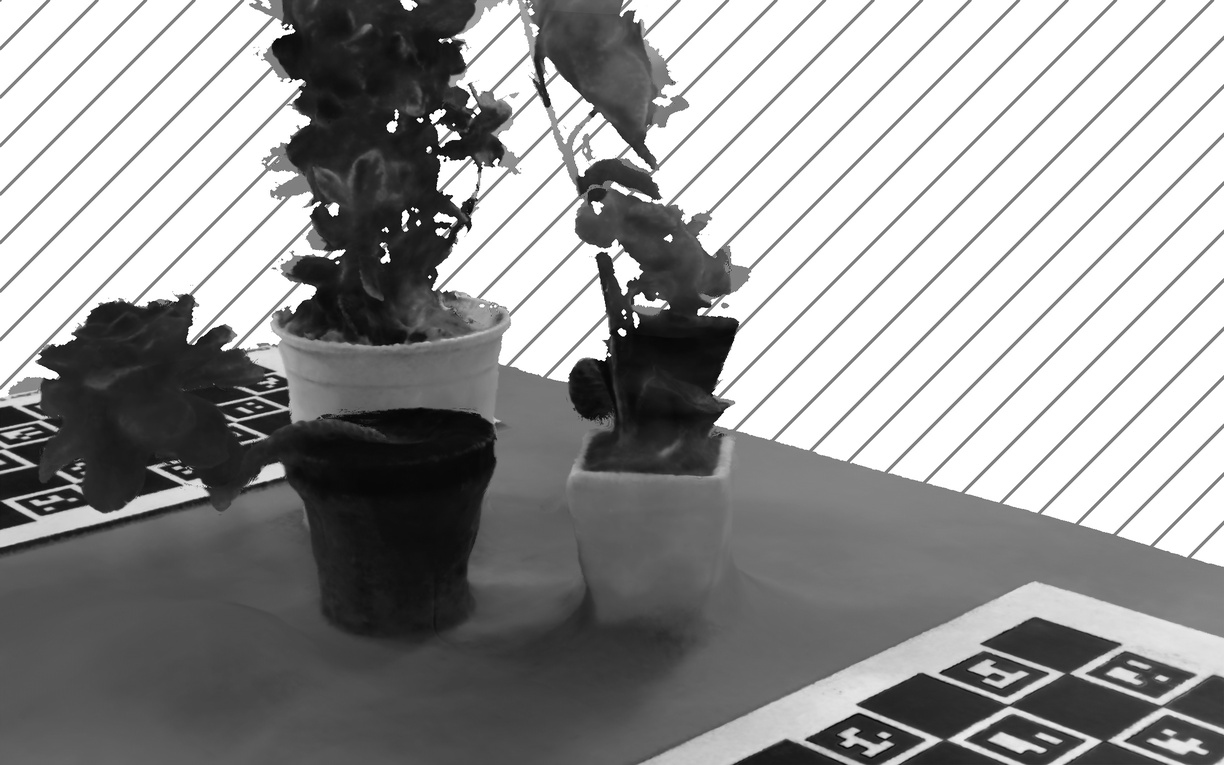}
                \end{subfigure}
                \begin{subfigure}{0.325\linewidth}
                    \centering
                    \includegraphics[width=\linewidth]{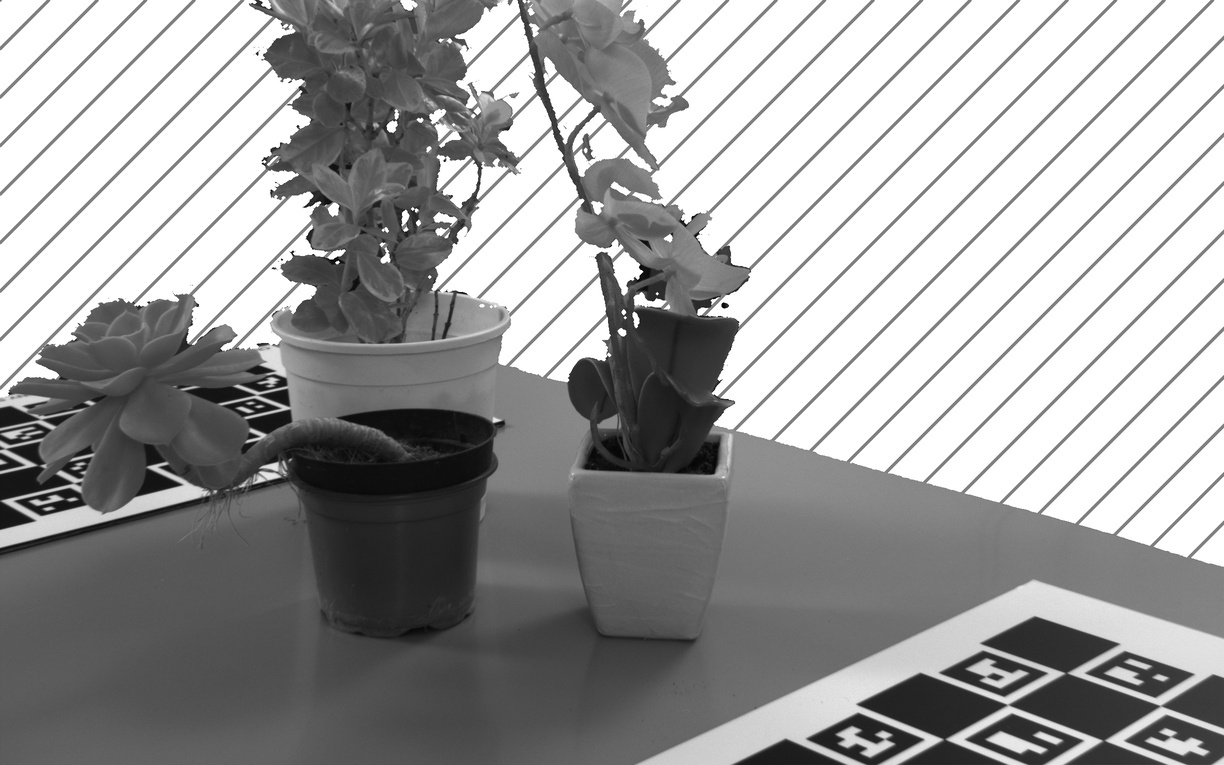}
                \end{subfigure}
                \begin{subfigure}{0.325\linewidth}
                    \centering
                    \includegraphics[width=\linewidth]{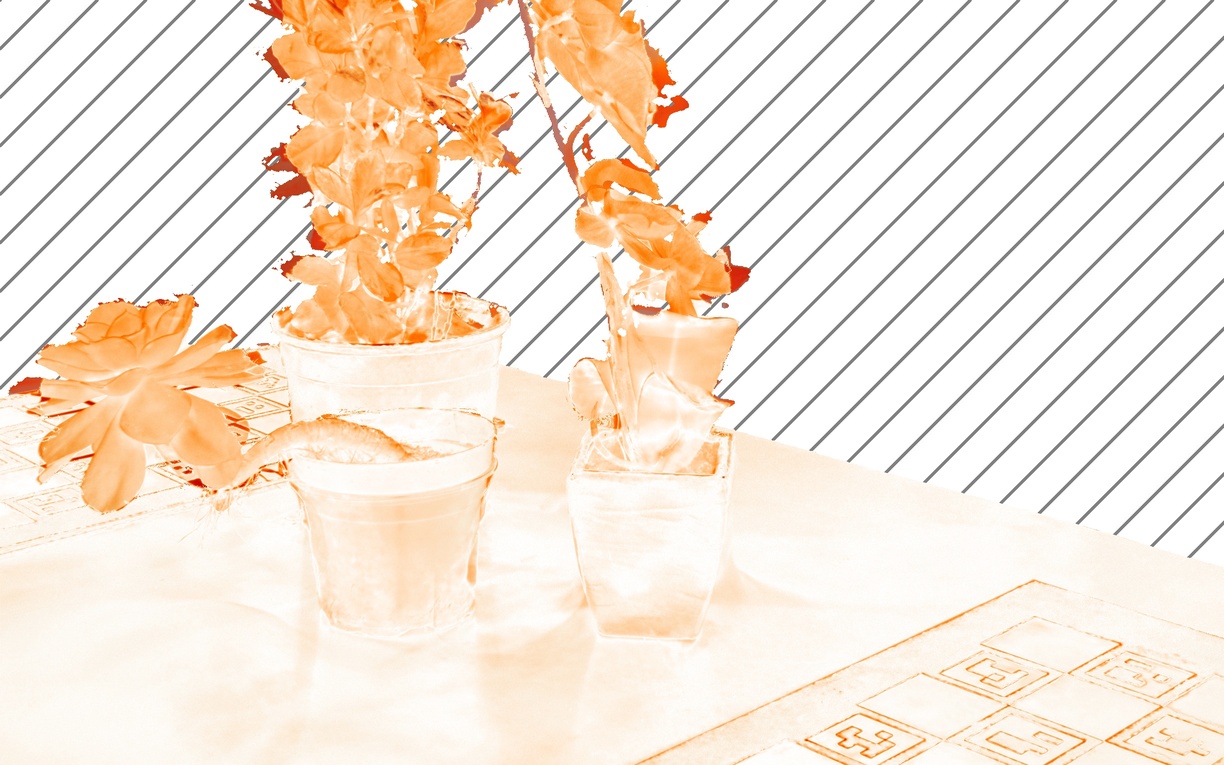}
                \end{subfigure}
            \end{minipage}
        \end{minipage}
        \begin{minipage}{\linewidth}
            \begin{minipage}{0.025\linewidth}
                \vfill
                \rotatebox[origin=cb]{90}{MS}
                \vfill
            \end{minipage}
            \begin{minipage}{0.97\linewidth}
                \begin{subfigure}{0.325\linewidth}
                    \centering
                    \includegraphics[width=\linewidth]{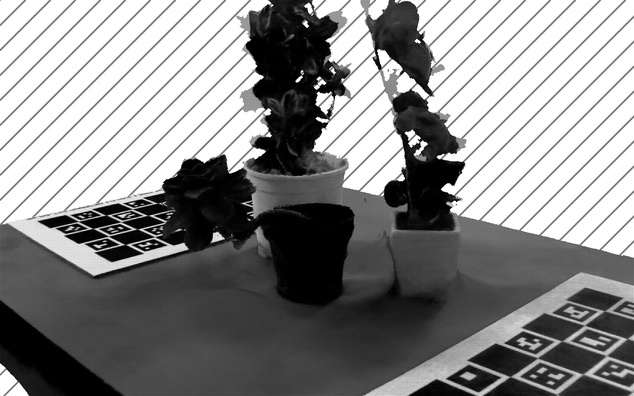}
                \end{subfigure}
                \begin{subfigure}{0.325\linewidth}
                    \centering
                    \includegraphics[width=\linewidth]{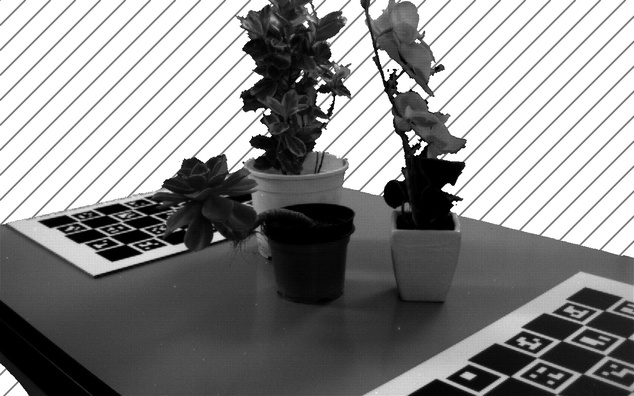}
                \end{subfigure}
                \begin{subfigure}{0.325\linewidth}
                    \centering
                    \includegraphics[width=\linewidth]{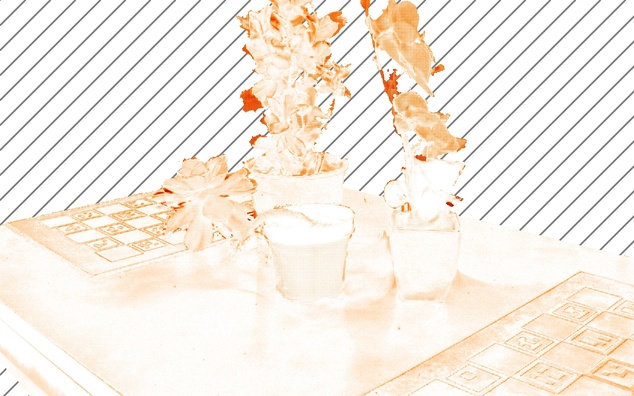}
                \end{subfigure}
            \end{minipage}
        \end{minipage}
    \end{minipage}
    \begin{minipage}{0.02\linewidth}
        \rotatebox[origin=cb]{270}{No supervision}
    \end{minipage}
    \caption{Qualitative renderings of the ``Bouquet'' scene from the \gls{ft} step supervised with only RGB data. All frames are mosaicked, except RGB frames. RGB is demosaicked only for visualization purposes.}
    \label{fig:qualitative_bouquet}
    \vspace{-0.5cm}
\end{figure*}

In \cref{fig:qualitative_birdhouse,fig:qualitative_bouquet,fig:qualitative_teddybear,fig:qualitative_toys}, we include some additional qualitative results of the fine-tuning step. The selected \gls{ft} scenes of MMS-DATA are ``Fruits'', ``Teddybear'', ``Toys'', ``Birdhouse'', and ``Bouquet''. We choose these scenes to cover a wide range of colors and different materials. 
Following  the evaluation setting used in MultimodalStudio~\cite{lincetto2025}, the test views of every scene are view 9, 19, 29, 39, and 49.  The figures show that \acrfull{method-name} can produce accurate reconstruction not only for the supervised RGB data but also for modalities for which no supervision frames are provided. Considering the error maps, we observe that the error is generally small and mostly concentrated on surfaces with challenging reflection properties (\eg, the leaves of the plant in \cref{fig:qualitative_bouquet}). Moreover, in \cref{tab:ft_modality_combinations}, we report additional results obtained by performing the \gls{ft} with various combinations of modalities. When using only MS images, the metrics of Mono, NIR, and Pol are comparable to a FT with only RGB data, as they do not benefit from more complete spectral information in the visible range. Additionally, RGB quality decreases since MS frames have a much lower resolution. When supervised also with high-res Mono, the FT accuracy increases by up to 3.5~dB. MS frames contribute color information, Mono contributes fine details, and RGB contributes both. By combining RGB with other modalities, the metrics consistently benefit from the complementary information carried by each of them.

\section{Modality Conversion with Third-party Methods}
\label{sec:3rd}

\begin{table*}[t]
    \centering
    \caption{Results of SPoILeR fine-tuning in terms of PSNR, averaged on the considered 5 FT scenes. The model is fine-tuned by enabling the supervision of various combinations of modalities, to analyze the contributions of complementary information.}
    \begin{tabular*}{0.8\linewidth}{@{\extracolsep{\fill}}cccccccc}
        \toprule
        &\multirow{2}{*}{Train Modalities} & \multicolumn{6}{c}{Test Modalities}\\
        \cmidrule{3-8}
        &&\green{RGB}& \purple{Mono}& \red{NIR}& \blue{Pol}&\orange{MS}&\\
        \midrule
        &\orange{MS}& 26.67& 24.78& 26.45& 25.24&28.15&\\
        \arrayrulecolor{black!30}\midrule
        &\orange{MS} - \purple{Mono}& 28.52& 30.04& 29.94& 26.44&28.99&\\
        \arrayrulecolor{black}\midrule
        &\green{RGB} - \red{NIR}& 30.25& 28.87& 31.95& 25.95&26.77&\\
        \arrayrulecolor{black!30}\midrule
        &\green{RGB} - \red{NIR} - \blue{Pol}& 30.43& 29.44& 31.92& 28.93&27.76&\\
        \arrayrulecolor{black}\bottomrule
    \end{tabular*}
    \label{tab:ft_modality_combinations}
\end{table*}

\begin{figure*}[t]
    \centering
    \begin{minipage}{\linewidth}
        \centering
        \hfill
        \begin{minipage}{0.96\linewidth}
            \begin{minipage}{0.495\linewidth}
                \centering
                Rendering
            \end{minipage}
            \begin{minipage}{0.495\linewidth}
                \centering
                Error
            \end{minipage}
        \end{minipage}
        \vspace{2pt}
    \end{minipage}
    \begin{minipage}{\linewidth}
        \begin{minipage}{0.025\linewidth}
            \vfill
            \rotatebox[origin=cb]{90}{Ours}
            \vfill
        \end{minipage}
        \begin{minipage}{0.97\linewidth}
            \begin{subfigure}{0.495\linewidth}
                \centering
                \includegraphics[width=\linewidth]{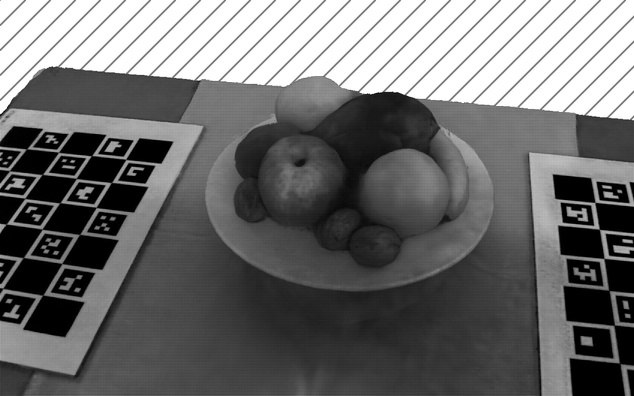}
            \end{subfigure}
            \begin{subfigure}{0.495\linewidth}
                \centering
                \includegraphics[width=\linewidth]{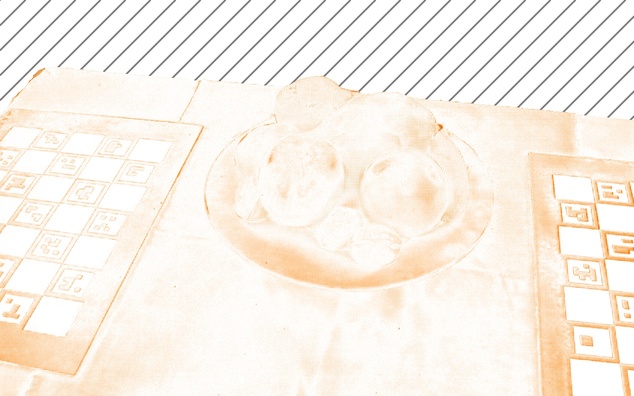}
            \end{subfigure}
        \end{minipage}
    \end{minipage}
    \begin{minipage}{\linewidth}
        \begin{minipage}{0.025\linewidth}
            \vfill
            \rotatebox[origin=cb]{90}{MST++ before}
            \vfill
        \end{minipage}
        \begin{minipage}{0.97\linewidth}
            \begin{subfigure}{0.495\linewidth}
                \centering
                \includegraphics[width=\linewidth]{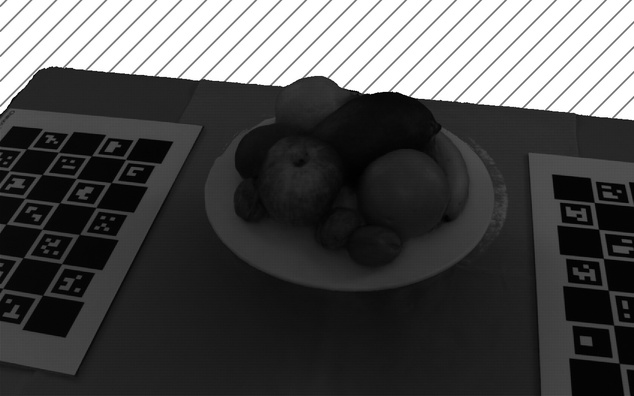}
            \end{subfigure}
            \begin{subfigure}{0.495\linewidth}
                \centering
                \includegraphics[width=\linewidth]{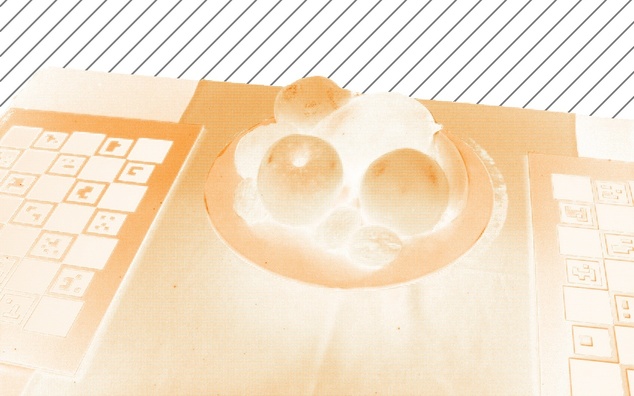}
            \end{subfigure}
        \end{minipage}
    \end{minipage}
    \begin{minipage}{\linewidth}
        \begin{minipage}{0.025\linewidth}
            \vfill
            \rotatebox[origin=cb]{90}{MST++ after}
            \vfill
        \end{minipage}
        \begin{minipage}{0.97\linewidth}
            \begin{subfigure}{0.495\linewidth}
                \centering
                \includegraphics[width=\linewidth]{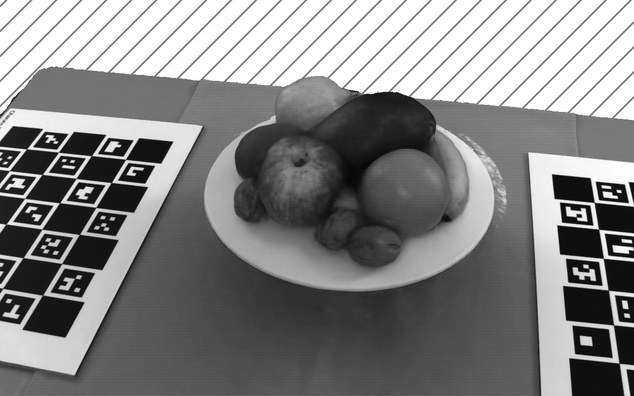}
            \end{subfigure}
            \begin{subfigure}{0.495\linewidth}
                \centering
                \includegraphics[width=\linewidth]{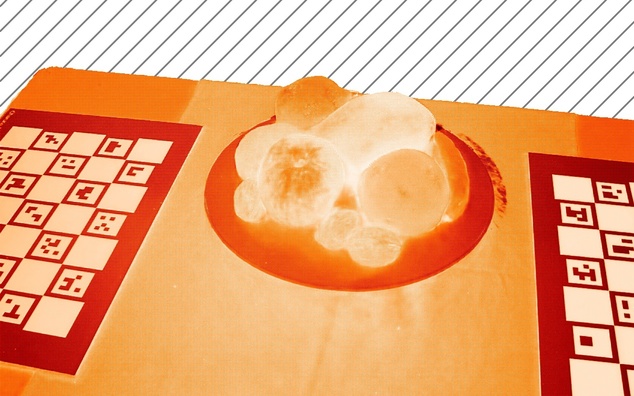}
            \end{subfigure}
        \end{minipage}
    \end{minipage}
    \begin{minipage}{\linewidth}
        \begin{minipage}{0.025\linewidth}
            \vfill
            \rotatebox[origin=cb]{90}{GT}
            \vfill
        \end{minipage}
        \begin{minipage}{0.97\linewidth}
            \begin{subfigure}{0.495\linewidth}
                \centering
                \includegraphics[width=\linewidth]{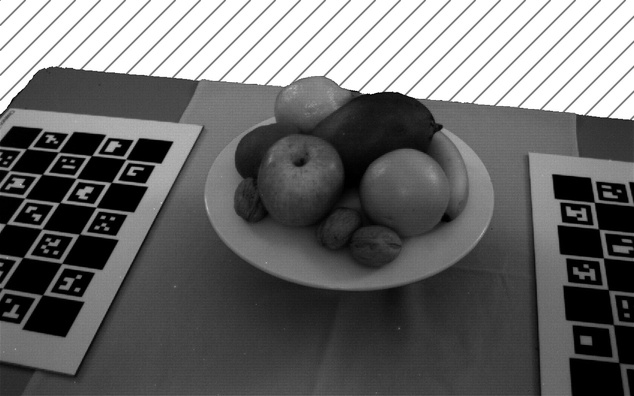}
            \end{subfigure}
            \begin{subfigure}{0.495\linewidth}
                \centering
                \hfill
            \end{subfigure}
        \end{minipage}
    \end{minipage}
    \caption{Qualitative comparison between \gls{method-name} and \gls{mms-fw} $+$ MST++ in terms of recovered multispectral radiance on the ``Bouquet'' scene. All frames are mosaicked. ``before'' and ``after'' refer to whether the RGB-to-MS conversion in performed before or after the training of \gls{mms-fw}.}
    \label{fig:mst++_fruits}
    \vspace{-0.5cm}
\end{figure*}

\begin{figure*}[t]
    \centering
    \begin{minipage}{\linewidth}
        \centering
        \hfill
        \begin{minipage}{0.96\linewidth}
            \begin{minipage}{0.495\linewidth}
                \centering
                Rendering
            \end{minipage}
            \begin{minipage}{0.495\linewidth}
                \centering
                Error
            \end{minipage}
        \end{minipage}
        \vspace{2pt}
    \end{minipage}
    \begin{minipage}{\linewidth}
        \begin{minipage}{0.025\linewidth}
            \vfill
            \rotatebox[origin=cb]{90}{Ours}
            \vfill
        \end{minipage}
        \begin{minipage}{0.97\linewidth}
            \begin{subfigure}{0.495\linewidth}
                \centering
                \includegraphics[width=\linewidth]{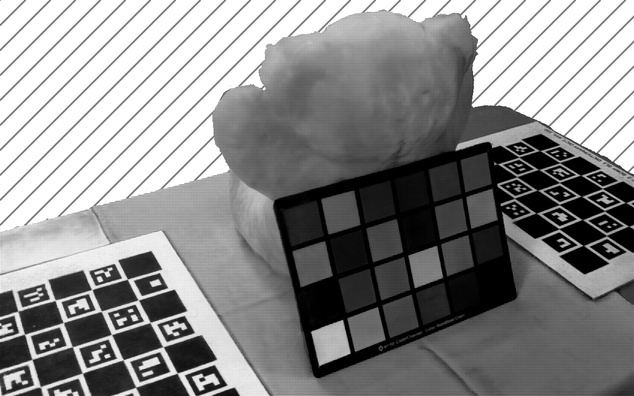}
            \end{subfigure}
            \begin{subfigure}{0.495\linewidth}
                \centering
                \includegraphics[width=\linewidth]{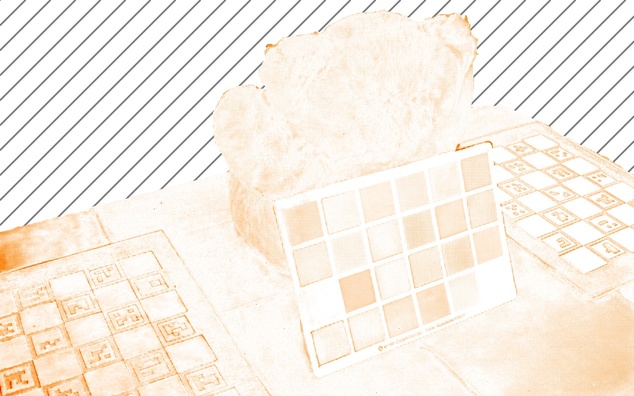}
            \end{subfigure}
        \end{minipage}
    \end{minipage}
    \begin{minipage}{\linewidth}
        \begin{minipage}{0.025\linewidth}
            \vfill
            \rotatebox[origin=cb]{90}{MST++ before}
            \vfill
        \end{minipage}
        \begin{minipage}{0.97\linewidth}
            \begin{subfigure}{0.495\linewidth}
                \centering
                \includegraphics[width=\linewidth]{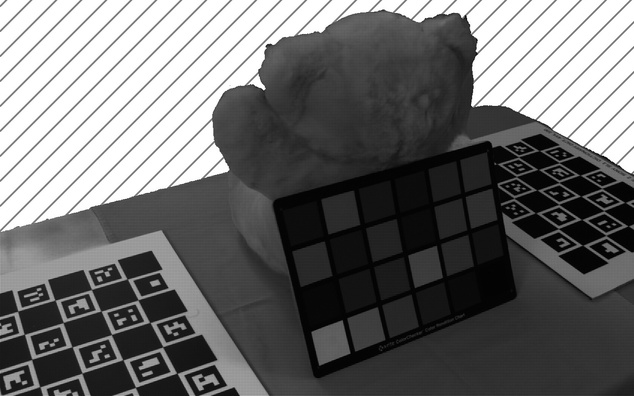}
            \end{subfigure}
            \begin{subfigure}{0.495\linewidth}
                \centering
                \includegraphics[width=\linewidth]{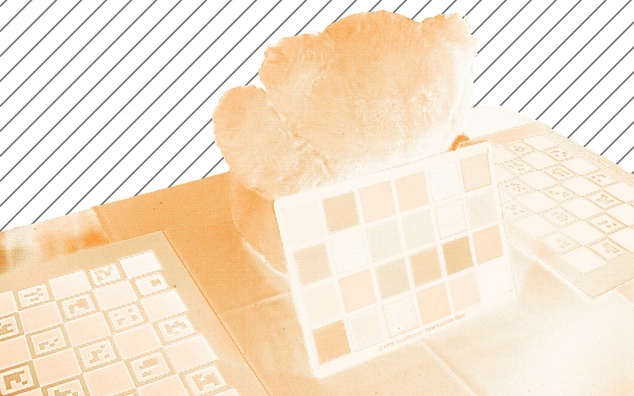}
            \end{subfigure}
        \end{minipage}
    \end{minipage}
    \begin{minipage}{\linewidth}
        \begin{minipage}{0.025\linewidth}
            \vfill
            \rotatebox[origin=cb]{90}{MST++ after}
            \vfill
        \end{minipage}
        \begin{minipage}{0.97\linewidth}
            \begin{subfigure}{0.495\linewidth}
                \centering
                \includegraphics[width=\linewidth]{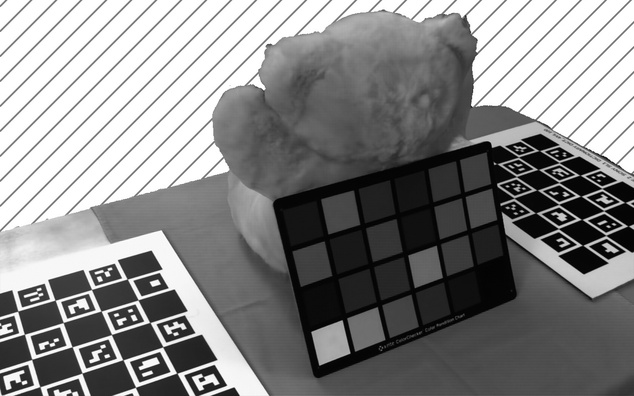}
            \end{subfigure}
            \begin{subfigure}{0.495\linewidth}
                \centering
                \includegraphics[width=\linewidth]{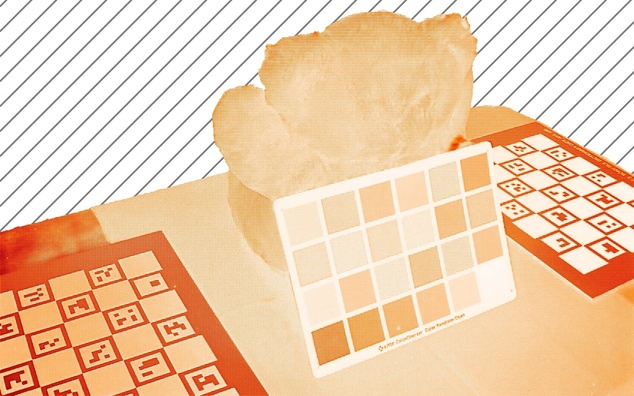}
            \end{subfigure}
        \end{minipage}
    \end{minipage}
    \begin{minipage}{\linewidth}
        \begin{minipage}{0.025\linewidth}
            \vfill
            \rotatebox[origin=cb]{90}{GT}
            \vfill
        \end{minipage}
        \begin{minipage}{0.97\linewidth}
            \begin{subfigure}{0.495\linewidth}
                \centering
                \includegraphics[width=\linewidth]{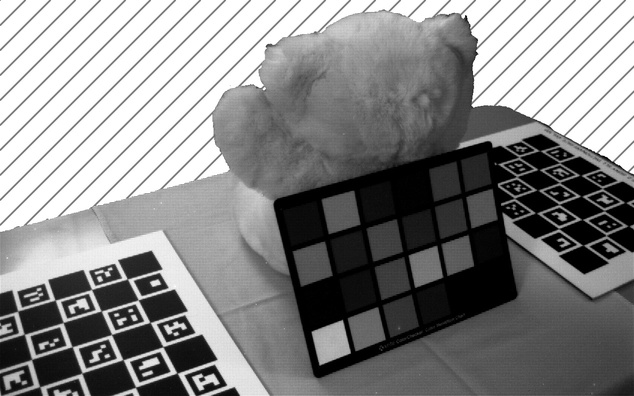}
            \end{subfigure}
            \begin{subfigure}{0.495\linewidth}
                \centering
                \hfill
            \end{subfigure}
        \end{minipage}
    \end{minipage}
    \caption{Qualitative comparison between \gls{method-name} and \gls{mms-fw} $+$ MST++ in terms of recovered multispectral radiance on the ``Teddybear'' scene. All frames are mosaicked. ``before'' and ``after'' refer to whether the RGB-to-MS conversion in performed before or after the training of \gls{mms-fw}.}
    \label{fig:mst++_teddybear}
    \vspace{-0.5cm}
\end{figure*}

\begin{figure*}[t]
    \centering
    \begin{minipage}{\linewidth}
        \centering
        \hfill
        \begin{minipage}{0.96\linewidth}
            \begin{minipage}{0.495\linewidth}
                \centering
                Rendering
            \end{minipage}
            \begin{minipage}{0.495\linewidth}
                \centering
                Error
            \end{minipage}
        \end{minipage}
        \vspace{2pt}
    \end{minipage}
    \begin{minipage}{\linewidth}
        \begin{minipage}{0.025\linewidth}
            \vfill
            \rotatebox[origin=cb]{90}{Ours}
            \vfill
        \end{minipage}
        \begin{minipage}{0.97\linewidth}
            \begin{subfigure}{0.495\linewidth}
                \centering
                \includegraphics[width=\linewidth]{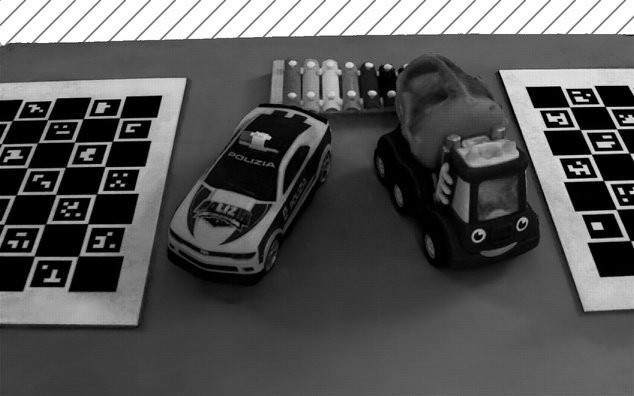}
            \end{subfigure}
            \begin{subfigure}{0.495\linewidth}
                \centering
                \includegraphics[width=\linewidth]{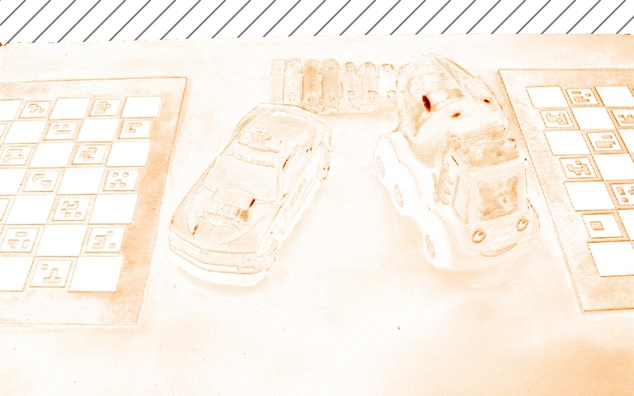}
            \end{subfigure}
        \end{minipage}
    \end{minipage}
    \begin{minipage}{\linewidth}
        \begin{minipage}{0.025\linewidth}
            \vfill
            \rotatebox[origin=cb]{90}{MST++ before}
            \vfill
        \end{minipage}
        \begin{minipage}{0.97\linewidth}
            \begin{subfigure}{0.495\linewidth}
                \centering
                \includegraphics[width=\linewidth]{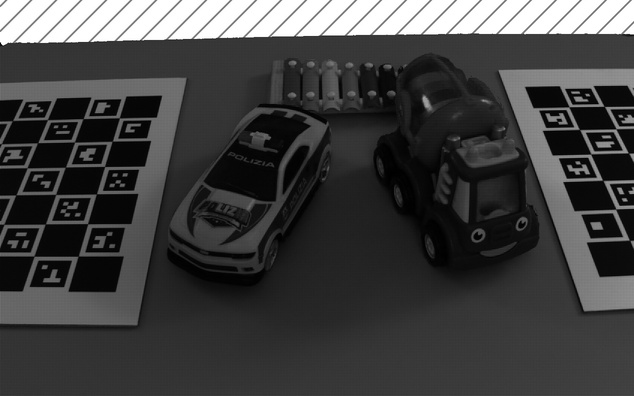}
            \end{subfigure}
            \begin{subfigure}{0.495\linewidth}
                \centering
                \includegraphics[width=\linewidth]{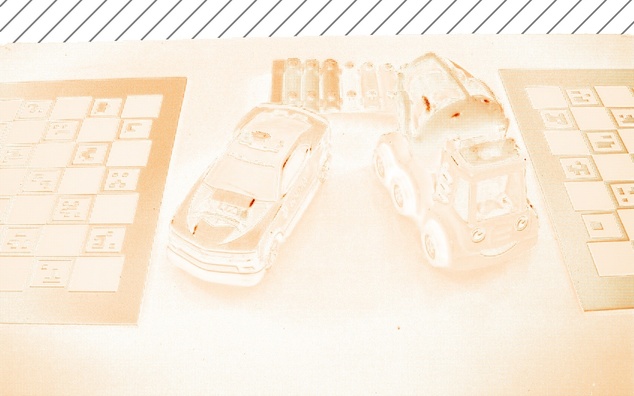}
            \end{subfigure}
        \end{minipage}
    \end{minipage}
    \begin{minipage}{\linewidth}
        \begin{minipage}{0.025\linewidth}
            \vfill
            \rotatebox[origin=cb]{90}{MST++ after}
            \vfill
        \end{minipage}
        \begin{minipage}{0.97\linewidth}
            \begin{subfigure}{0.495\linewidth}
                \centering
                \includegraphics[width=\linewidth]{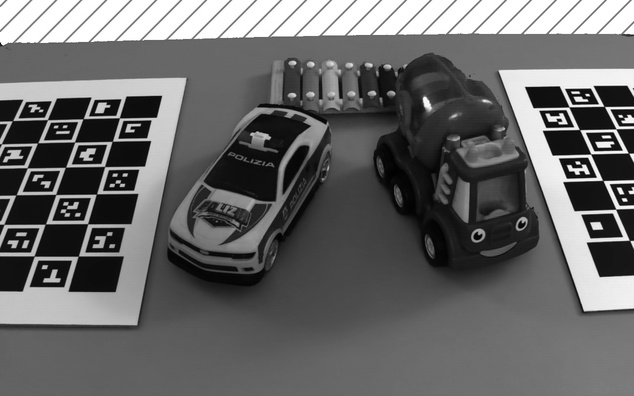}
            \end{subfigure}
            \begin{subfigure}{0.495\linewidth}
                \centering
                \includegraphics[width=\linewidth]{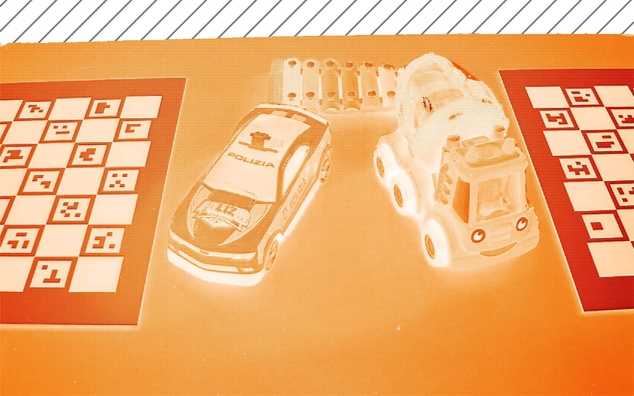}
            \end{subfigure}
        \end{minipage}
    \end{minipage}
    \begin{minipage}{\linewidth}
        \begin{minipage}{0.025\linewidth}
            \vfill
            \rotatebox[origin=cb]{90}{GT}
            \vfill
        \end{minipage}
        \begin{minipage}{0.97\linewidth}
            \begin{subfigure}{0.495\linewidth}
                \centering
                \includegraphics[width=\linewidth]{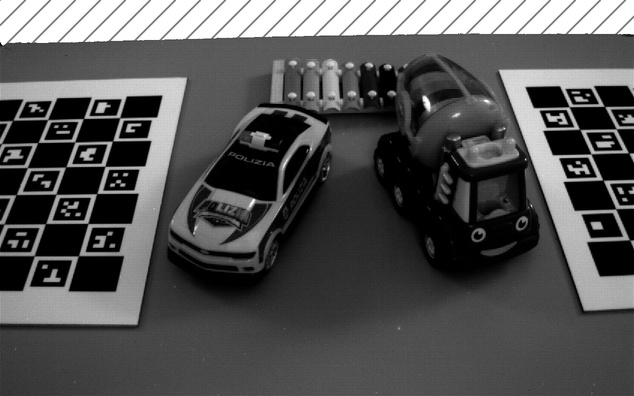}
            \end{subfigure}
            \begin{subfigure}{0.495\linewidth}
                \centering
                \hfill
            \end{subfigure}
        \end{minipage}
    \end{minipage}
    \caption{Qualitative comparison between \gls{method-name} and \gls{mms-fw} $+$ MST++ in terms of recovered multispectral radiance on the ``Toys'' scene. All frames are mosaicked. ``before'' and ``after'' refer to whether the RGB-to-MS conversion in performed before or after the training of \gls{mms-fw}.}
    \label{fig:mst++_toys}
    \vspace{-0.5cm}
\end{figure*}

\begin{figure*}[t]
    \centering
    \begin{minipage}{\linewidth}
        \centering
        \hfill
        \begin{minipage}{0.96\linewidth}
            \begin{minipage}{0.495\linewidth}
                \centering
                Rendering
            \end{minipage}
            \begin{minipage}{0.495\linewidth}
                \centering
                Error
            \end{minipage}
        \end{minipage}
        \vspace{2pt}
    \end{minipage}
    \begin{minipage}{\linewidth}
        \begin{minipage}{0.025\linewidth}
            \vfill
            \rotatebox[origin=cb]{90}{Ours}
            \vfill
        \end{minipage}
        \begin{minipage}{0.97\linewidth}
            \begin{subfigure}{0.495\linewidth}
                \centering
                \includegraphics[width=\linewidth]{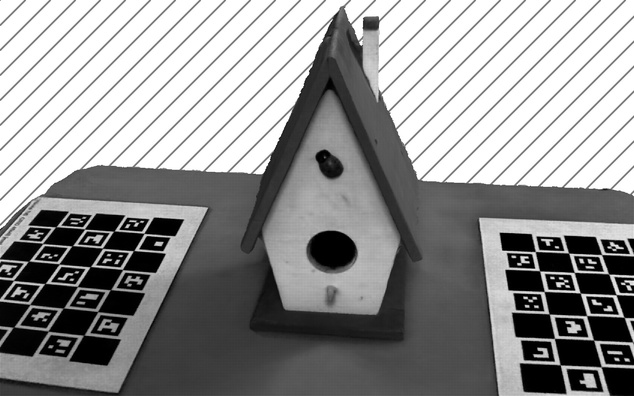}
            \end{subfigure}
            \begin{subfigure}{0.495\linewidth}
                \centering
                \includegraphics[width=\linewidth]{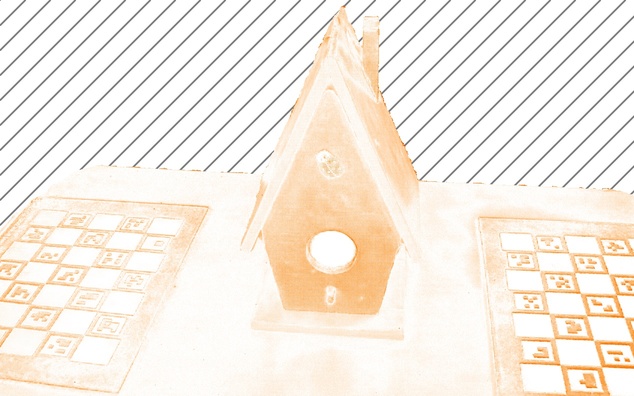}
            \end{subfigure}
        \end{minipage}
    \end{minipage}
    \begin{minipage}{\linewidth}
        \begin{minipage}{0.025\linewidth}
            \vfill
            \rotatebox[origin=cb]{90}{MST++ before}
            \vfill
        \end{minipage}
        \begin{minipage}{0.97\linewidth}
            \begin{subfigure}{0.495\linewidth}
                \centering
                \includegraphics[width=\linewidth]{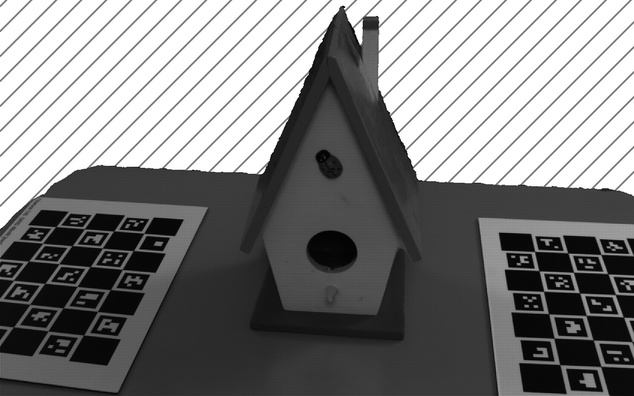}
            \end{subfigure}
            \begin{subfigure}{0.495\linewidth}
                \centering
                \includegraphics[width=\linewidth]{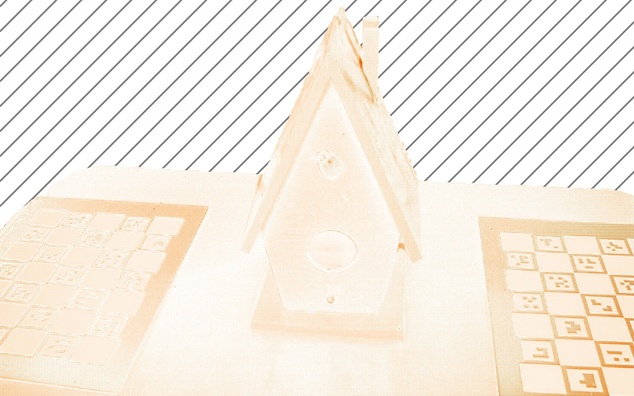}
            \end{subfigure}
        \end{minipage}
    \end{minipage}
    \begin{minipage}{\linewidth}
        \begin{minipage}{0.025\linewidth}
            \vfill
            \rotatebox[origin=cb]{90}{MST++ after}
            \vfill
        \end{minipage}
        \begin{minipage}{0.97\linewidth}
            \begin{subfigure}{0.495\linewidth}
                \centering
                \includegraphics[width=\linewidth]{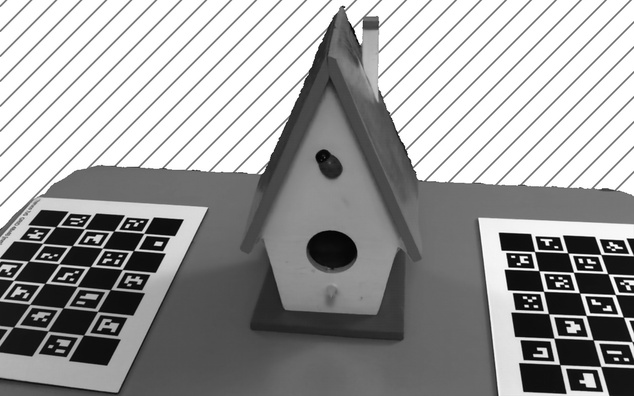}
            \end{subfigure}
            \begin{subfigure}{0.495\linewidth}
                \centering
                \includegraphics[width=\linewidth]{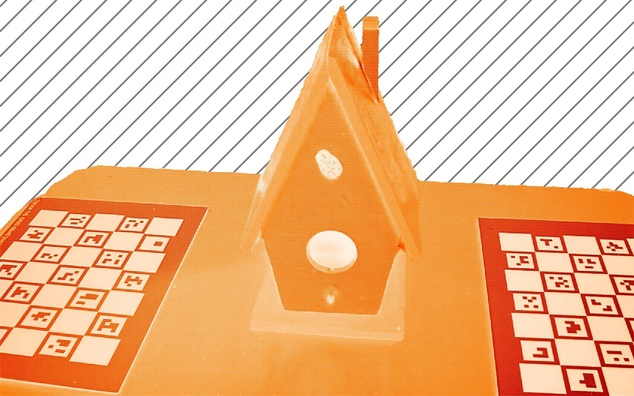}
            \end{subfigure}
        \end{minipage}
    \end{minipage}
    \begin{minipage}{\linewidth}
        \begin{minipage}{0.025\linewidth}
            \vfill
            \rotatebox[origin=cb]{90}{GT}
            \vfill
        \end{minipage}
        \begin{minipage}{0.97\linewidth}
            \begin{subfigure}{0.495\linewidth}
                \centering
                \includegraphics[width=\linewidth]{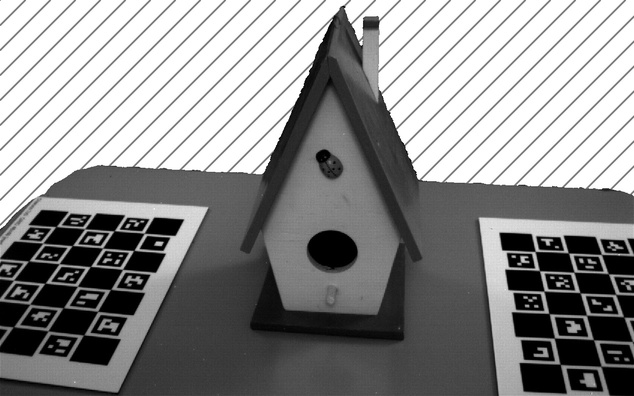}
            \end{subfigure}
            \begin{subfigure}{0.495\linewidth}
                \centering
                \hfill
            \end{subfigure}
        \end{minipage}
    \end{minipage}
    \caption{Qualitative comparison between \gls{method-name} and \gls{mms-fw} $+$ MST++ in terms of recovered multispectral radiance on the ``Birdhouse'' scene. All frames are mosaicked. ``before'' and ``after'' refer to whether the RGB-to-MS conversion in performed before or after the training of \gls{mms-fw}.}
    \label{fig:mst++_birdhouse}
    \vspace{-0.5cm}
\end{figure*}

\begin{figure*}[t]
    \centering
    \begin{minipage}{\linewidth}
        \centering
        \hfill
        \begin{minipage}{0.96\linewidth}
            \begin{minipage}{0.495\linewidth}
                \centering
                Rendering
            \end{minipage}
            \begin{minipage}{0.495\linewidth}
                \centering
                Error
            \end{minipage}
        \end{minipage}
        \vspace{2pt}
    \end{minipage}
    \begin{minipage}{\linewidth}
        \begin{minipage}{0.025\linewidth}
            \vfill
            \rotatebox[origin=cb]{90}{Ours}
            \vfill
        \end{minipage}
        \begin{minipage}{0.97\linewidth}
            \begin{subfigure}{0.495\linewidth}
                \centering
                \includegraphics[width=\linewidth]{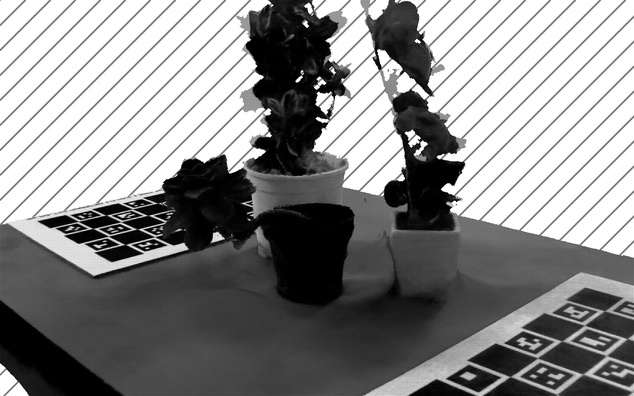}
            \end{subfigure}
            \begin{subfigure}{0.495\linewidth}
                \centering
                \includegraphics[width=\linewidth]{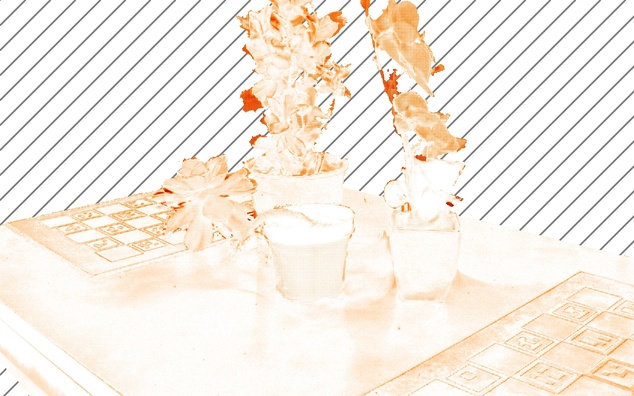}
            \end{subfigure}
        \end{minipage}
    \end{minipage}
    \begin{minipage}{\linewidth}
        \begin{minipage}{0.025\linewidth}
            \vfill
            \rotatebox[origin=cb]{90}{MST++ before}
            \vfill
        \end{minipage}
        \begin{minipage}{0.97\linewidth}
            \begin{subfigure}{0.495\linewidth}
                \centering
                \includegraphics[width=\linewidth]{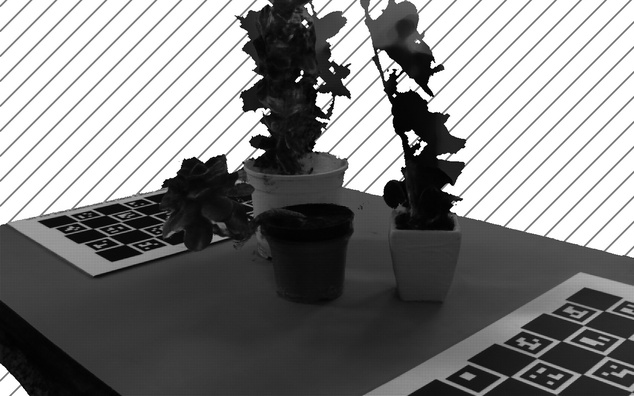}
            \end{subfigure}
            \begin{subfigure}{0.495\linewidth}
                \centering
                \includegraphics[width=\linewidth]{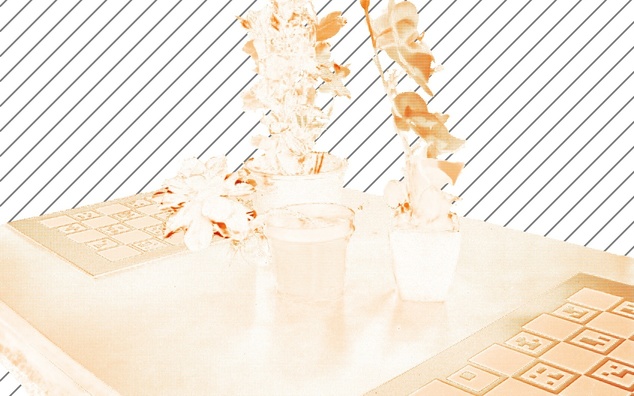}
            \end{subfigure}
        \end{minipage}
    \end{minipage}
    \begin{minipage}{\linewidth}
        \begin{minipage}{0.025\linewidth}
            \vfill
            \rotatebox[origin=cb]{90}{MST++ after}
            \vfill
        \end{minipage}
        \begin{minipage}{0.97\linewidth}
            \begin{subfigure}{0.495\linewidth}
                \centering
                \includegraphics[width=\linewidth]{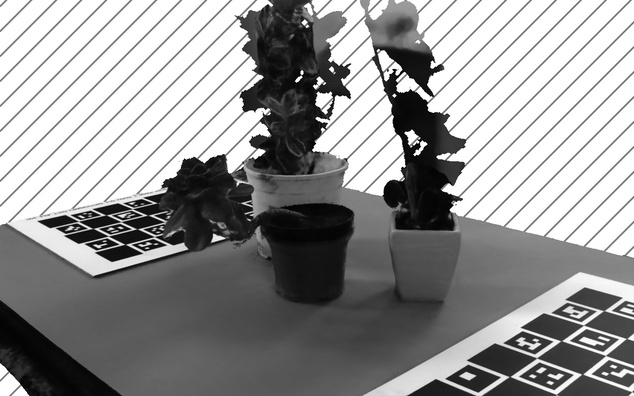}
            \end{subfigure}
            \begin{subfigure}{0.495\linewidth}
                \centering
                \includegraphics[width=\linewidth]{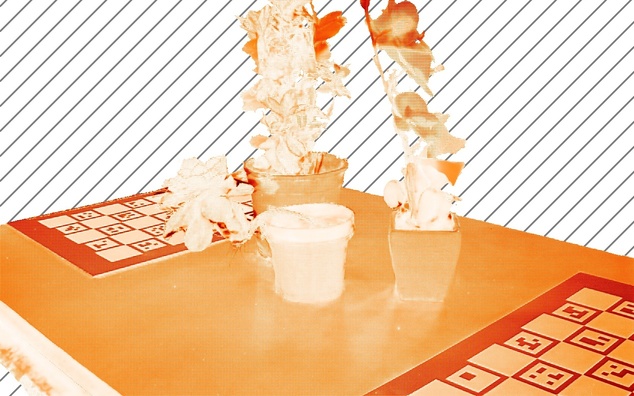}
            \end{subfigure}
        \end{minipage}
    \end{minipage}
    \begin{minipage}{\linewidth}
        \begin{minipage}{0.025\linewidth}
            \vfill
            \rotatebox[origin=cb]{90}{GT}
            \vfill
        \end{minipage}
        \begin{minipage}{0.97\linewidth}
            \begin{subfigure}{0.495\linewidth}
                \centering
                \includegraphics[width=\linewidth]{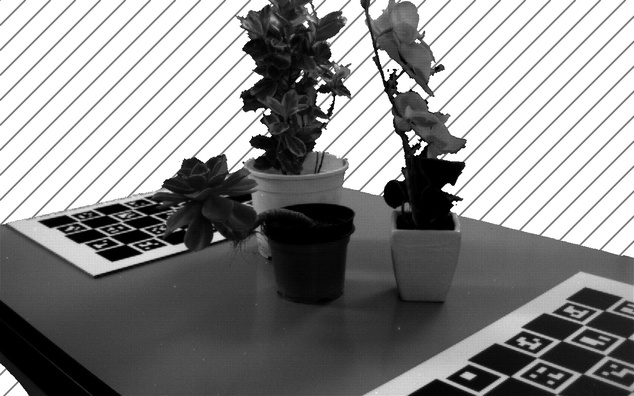}
            \end{subfigure}
            \begin{subfigure}{0.495\linewidth}
                \centering
                \hfill
            \end{subfigure}
        \end{minipage}
    \end{minipage}
    \caption{Qualitative comparison between \gls{method-name} and \gls{mms-fw} $+$ MST++ in terms of recovered multispectral radiance on the ``Bouquet'' scene. All frames are mosaicked. ``before'' and ``after'' refer to whether the RGB-to-MS conversion in performed before or after the training of \gls{mms-fw}.}
    \label{fig:mst++_bouquet}
    \vspace{-0.5cm}
\end{figure*}

\begin{figure*}[t]
    \centering
    \begin{minipage}{\linewidth}
        \centering
        \hfill
        \begin{minipage}{0.96\linewidth}
            \begin{minipage}{0.245\linewidth}
                \centering
                AoP
            \end{minipage}
            \begin{minipage}{0.245\linewidth}
                \centering
                MAngE
            \end{minipage}
            \begin{minipage}{0.245\linewidth}
                \centering
                DoP
            \end{minipage}
            \begin{minipage}{0.245\linewidth}
                \centering
                MAbsE
            \end{minipage}
        \end{minipage}
        \vspace{2pt}
    \end{minipage}
    \begin{minipage}{\linewidth}
        \begin{minipage}{0.025\linewidth}
            \vfill
            \rotatebox[origin=cb]{90}{Ours}
            \vfill
        \end{minipage}
        \begin{minipage}{0.97\linewidth}
            \begin{subfigure}{0.245\linewidth}
                \centering
                \includegraphics[width=\linewidth]{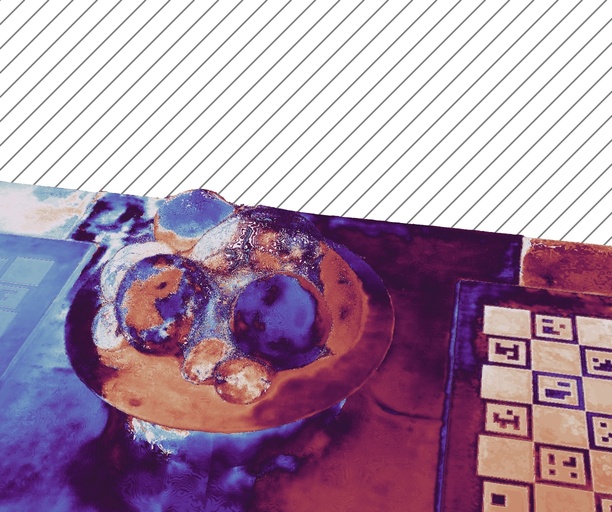}
            \end{subfigure}
            \begin{subfigure}{0.245\linewidth}
                \centering
                \includegraphics[width=\linewidth]{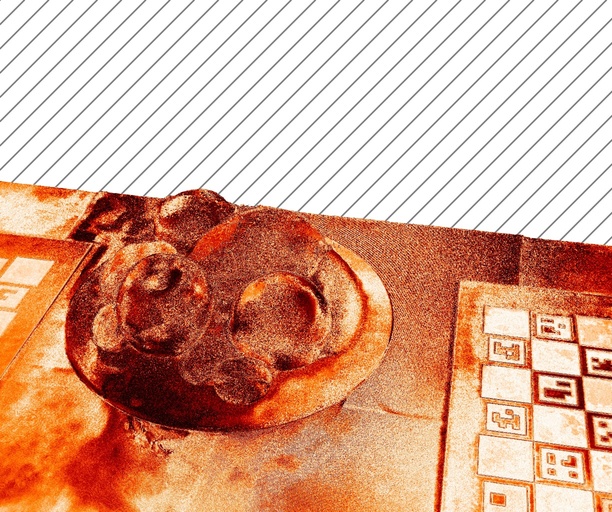}
            \end{subfigure}
            \begin{subfigure}{0.245\linewidth}
                \centering
                \includegraphics[width=\linewidth]{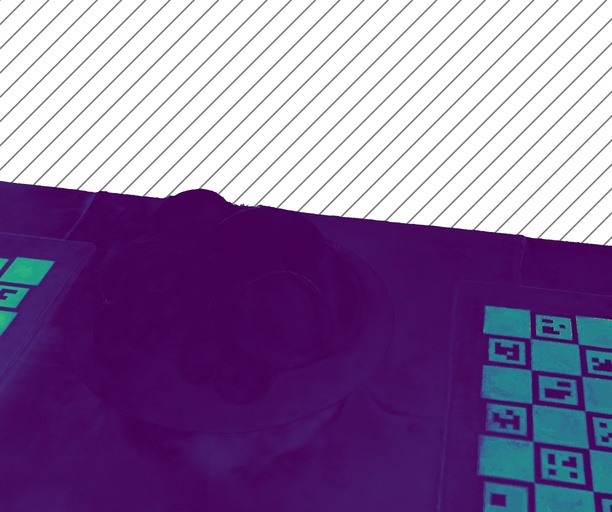}
            \end{subfigure}
            \begin{subfigure}{0.245\linewidth}
                \centering
                \includegraphics[width=\linewidth]{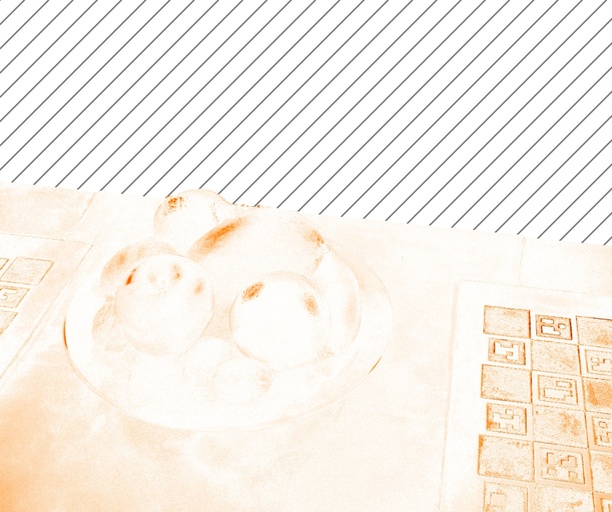}
            \end{subfigure}
        \end{minipage}
    \end{minipage}
    \begin{minipage}{\linewidth}
        \begin{minipage}{0.025\linewidth}
            \vfill
            \rotatebox[origin=cb]{90}{PA before}
            \vfill
        \end{minipage}
        \begin{minipage}{0.97\linewidth}
            \begin{subfigure}{0.245\linewidth}
                \centering
                \includegraphics[width=\linewidth]{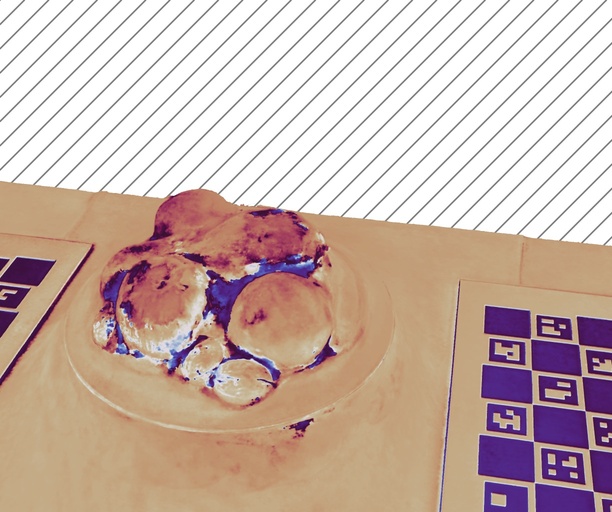}
            \end{subfigure}
            \begin{subfigure}{0.245\linewidth}
                \centering
                \includegraphics[width=\linewidth]{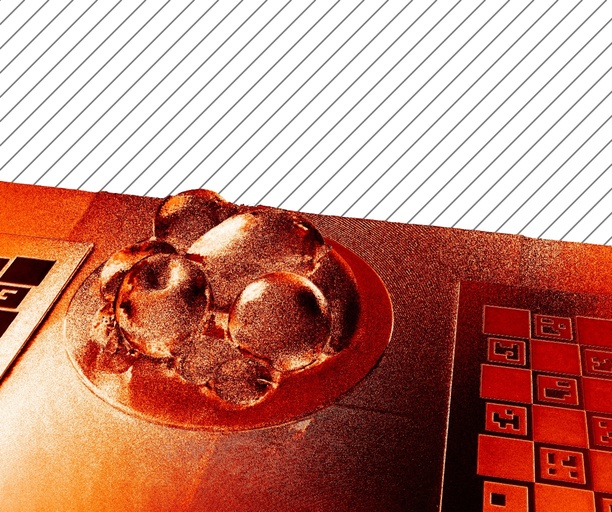}
            \end{subfigure}
            \begin{subfigure}{0.245\linewidth}
                \centering
                \includegraphics[width=\linewidth]{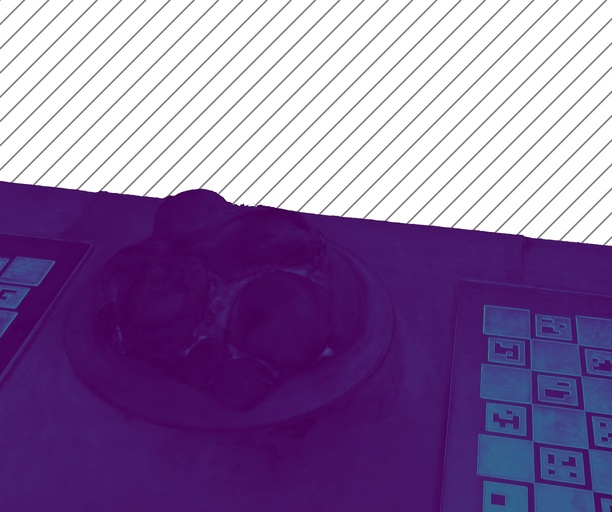}
            \end{subfigure}
            \begin{subfigure}{0.245\linewidth}
                \centering
                \includegraphics[width=\linewidth]{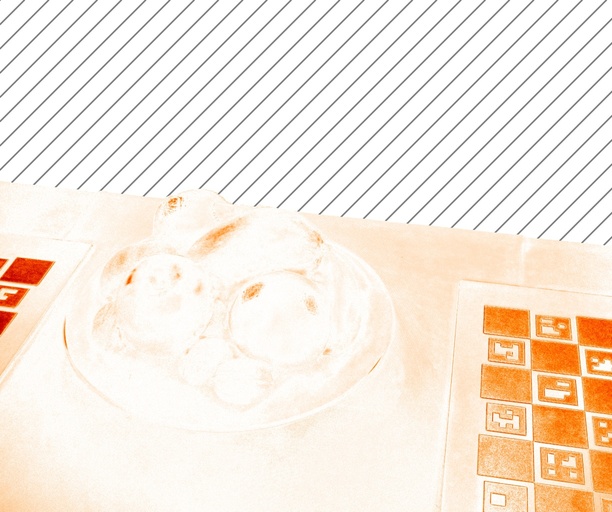}
            \end{subfigure}
        \end{minipage}
    \end{minipage}
    \begin{minipage}{\linewidth}
        \begin{minipage}{0.025\linewidth}
            \vfill
            \rotatebox[origin=cb]{90}{PA after}
            \vfill
        \end{minipage}
        \begin{minipage}{0.97\linewidth}
            \begin{subfigure}{0.245\linewidth}
                \centering
                \includegraphics[width=\linewidth]{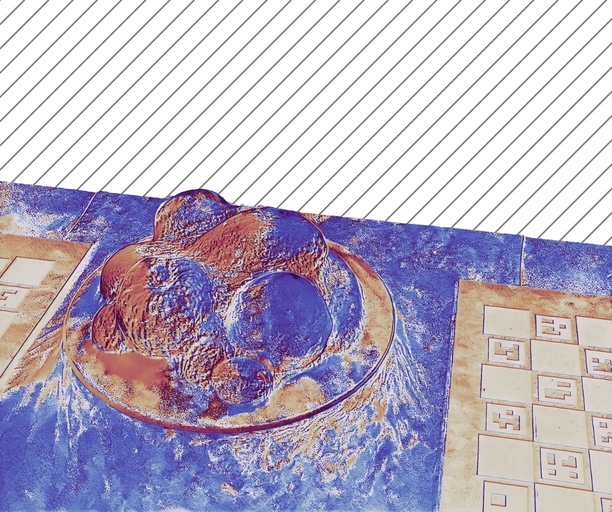}
            \end{subfigure}
            \begin{subfigure}{0.245\linewidth}
                \centering
                \includegraphics[width=\linewidth]{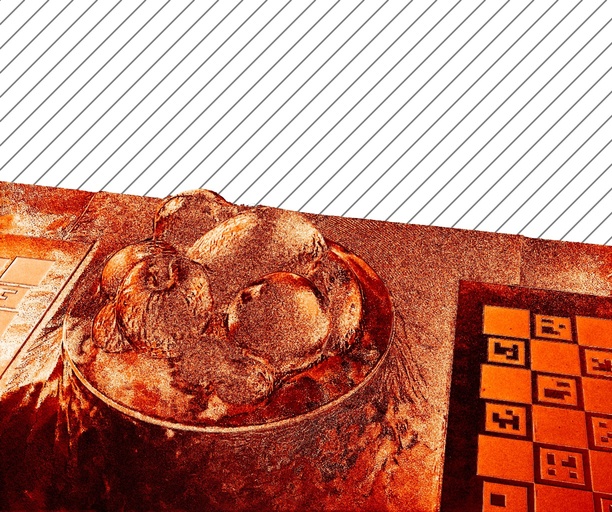}
            \end{subfigure}
            \begin{subfigure}{0.245\linewidth}
                \centering
                \includegraphics[width=\linewidth]{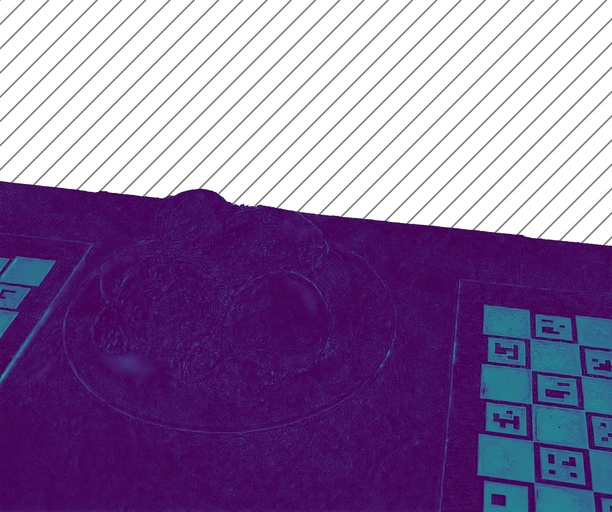}
            \end{subfigure}
            \begin{subfigure}{0.245\linewidth}
                \centering
                \includegraphics[width=\linewidth]{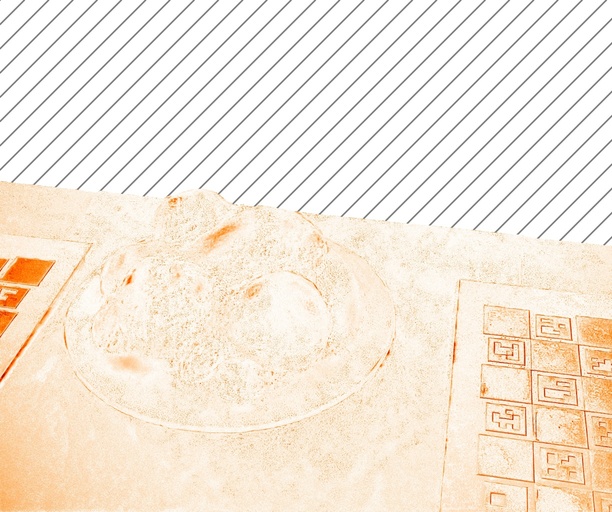}
            \end{subfigure}
        \end{minipage}
    \end{minipage}
    \begin{minipage}{\linewidth}
        \begin{minipage}{0.025\linewidth}
            \vfill
            \rotatebox[origin=cb]{90}{GT}
            \vfill
        \end{minipage}
        \begin{minipage}{0.97\linewidth}
            \begin{subfigure}{0.245\linewidth}
                \centering
                \includegraphics[width=\linewidth]{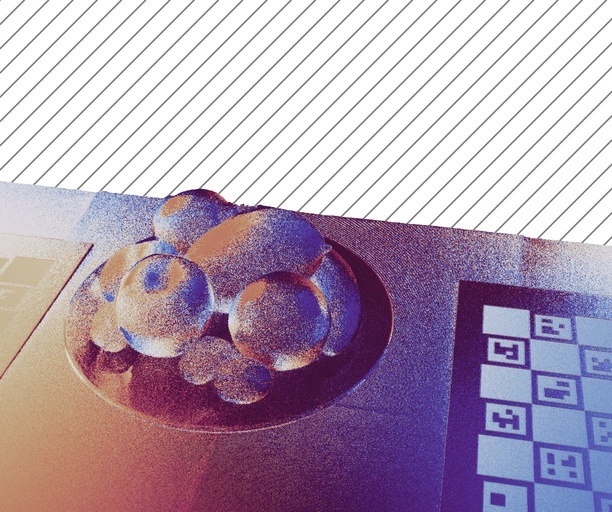}
            \end{subfigure}
            \begin{subfigure}{0.245\linewidth}
                \centering
                \hfill
            \end{subfigure}
            \begin{subfigure}{0.245\linewidth}
                \centering
                \includegraphics[width=\linewidth]{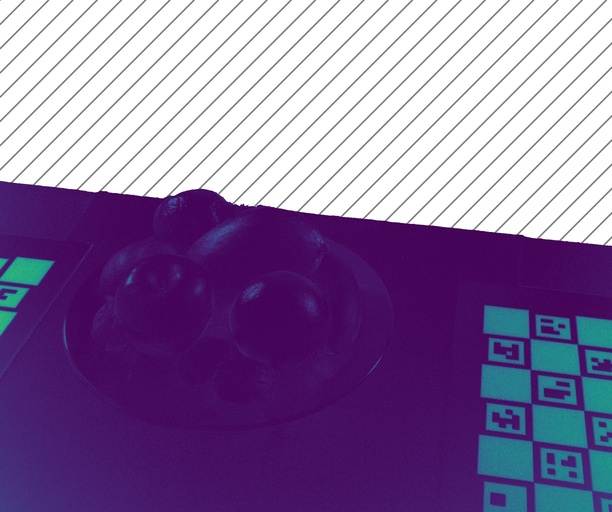}
            \end{subfigure}
            \begin{subfigure}{0.245\linewidth}
                \centering
                \hfill
            \end{subfigure}
        \end{minipage}
    \end{minipage}
    \caption{Qualitative comparison between \gls{method-name} and \gls{mms-fw} $+$ PolarAnything (PA) in terms of recovered polarization on the ``Fruits'' scene. ``before'' and ``after'' refer to whether the RGB-to-Pol conversion in performed before or after the training of \gls{mms-fw}.}
    \label{fig:polar_fruits}
    \vspace{-0.5cm}
\end{figure*}

\begin{figure*}[t]
    \centering
    \begin{minipage}{\linewidth}
        \centering
        \hfill
        \begin{minipage}{0.96\linewidth}
            \begin{minipage}{0.245\linewidth}
                \centering
                AoP
            \end{minipage}
            \begin{minipage}{0.245\linewidth}
                \centering
                MAngE
            \end{minipage}
            \begin{minipage}{0.245\linewidth}
                \centering
                DoP
            \end{minipage}
            \begin{minipage}{0.245\linewidth}
                \centering
                MAbsE
            \end{minipage}
        \end{minipage}
        \vspace{2pt}
    \end{minipage}
    \begin{minipage}{\linewidth}
        \begin{minipage}{0.025\linewidth}
            \vfill
            \rotatebox[origin=cb]{90}{Ours}
            \vfill
        \end{minipage}
        \begin{minipage}{0.97\linewidth}
            \begin{subfigure}{0.245\linewidth}
                \centering
                \includegraphics[width=\linewidth]{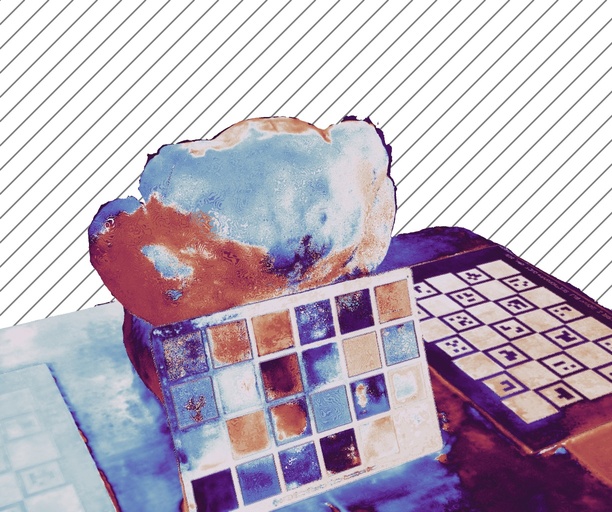}
            \end{subfigure}
            \begin{subfigure}{0.245\linewidth}
                \centering
                \includegraphics[width=\linewidth]{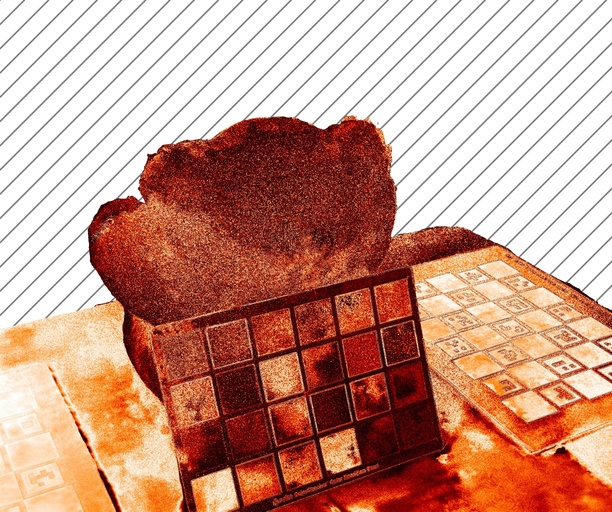}
            \end{subfigure}
            \begin{subfigure}{0.245\linewidth}
                \centering
                \includegraphics[width=\linewidth]{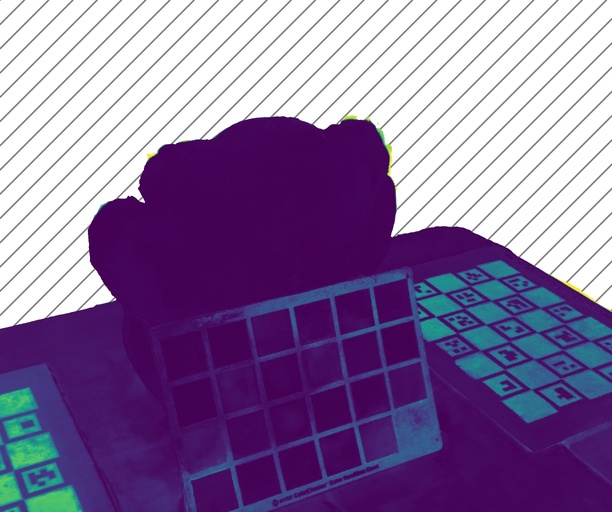}
            \end{subfigure}
            \begin{subfigure}{0.245\linewidth}
                \centering
                \includegraphics[width=\linewidth]{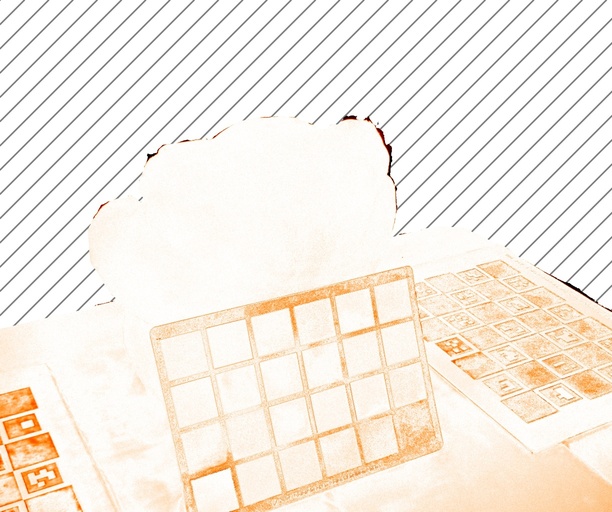}
            \end{subfigure}
        \end{minipage}
    \end{minipage}
    \begin{minipage}{\linewidth}
        \begin{minipage}{0.025\linewidth}
            \vfill
            \rotatebox[origin=cb]{90}{PA before}
            \vfill
        \end{minipage}
        \begin{minipage}{0.97\linewidth}
            \begin{subfigure}{0.245\linewidth}
                \centering
                \includegraphics[width=\linewidth]{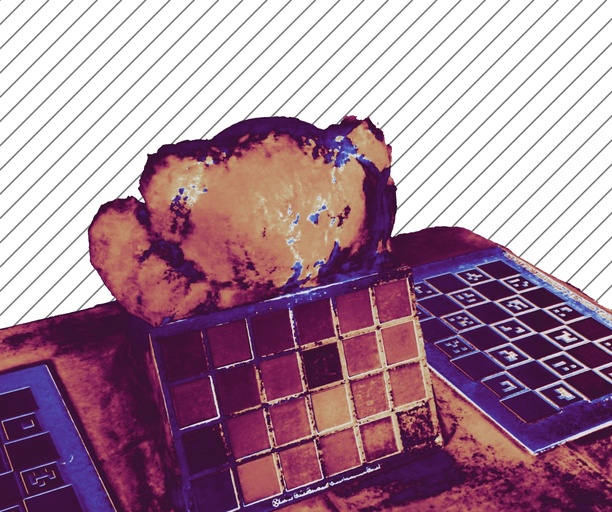}
            \end{subfigure}
            \begin{subfigure}{0.245\linewidth}
                \centering
                \includegraphics[width=\linewidth]{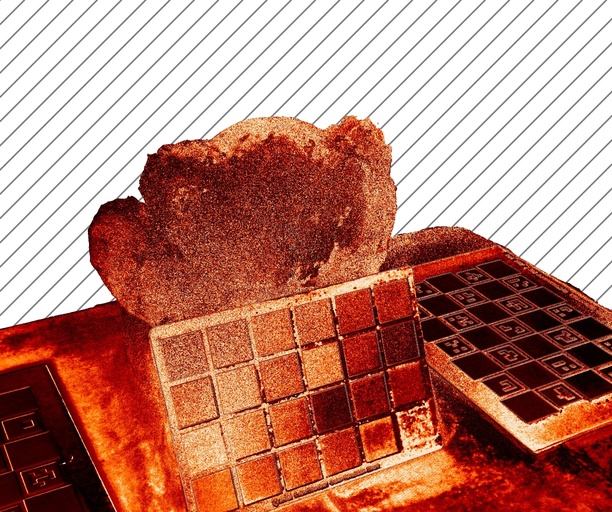}
            \end{subfigure}
            \begin{subfigure}{0.245\linewidth}
                \centering
                \includegraphics[width=\linewidth]{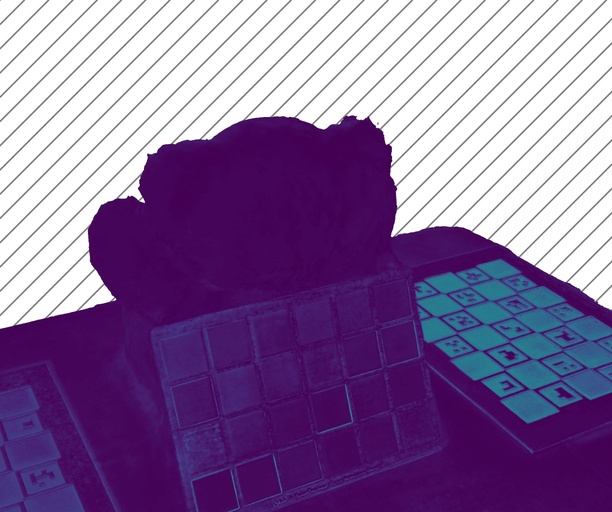}
            \end{subfigure}
            \begin{subfigure}{0.245\linewidth}
                \centering
                \includegraphics[width=\linewidth]{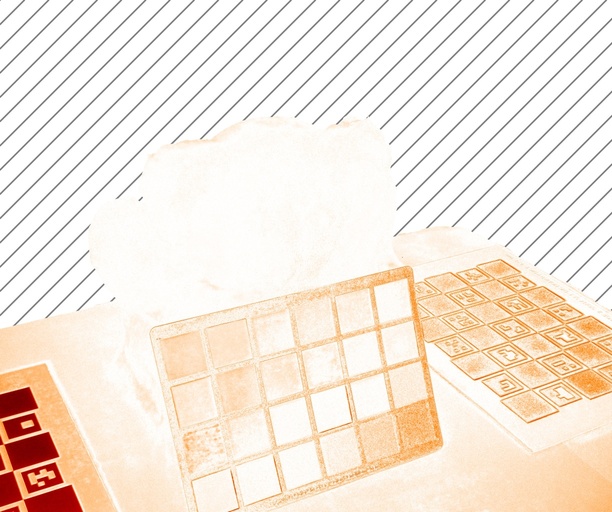}
            \end{subfigure}
        \end{minipage}
    \end{minipage}
    \begin{minipage}{\linewidth}
        \begin{minipage}{0.025\linewidth}
            \vfill
            \rotatebox[origin=cb]{90}{PA after}
            \vfill
        \end{minipage}
        \begin{minipage}{0.97\linewidth}
            \begin{subfigure}{0.245\linewidth}
                \centering
                \includegraphics[width=\linewidth]{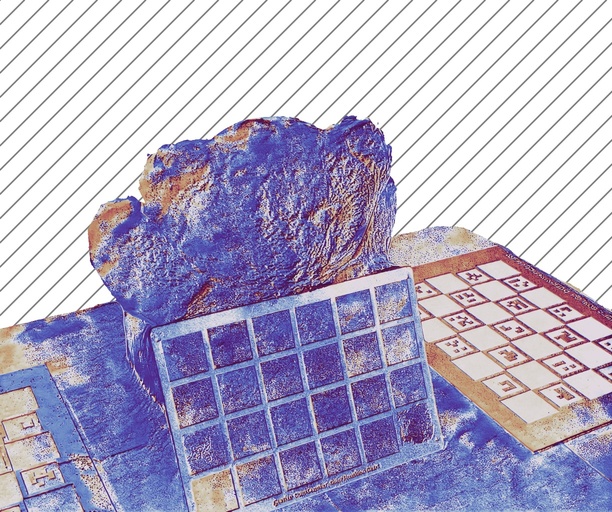}
            \end{subfigure}
            \begin{subfigure}{0.245\linewidth}
                \centering
                \includegraphics[width=\linewidth]{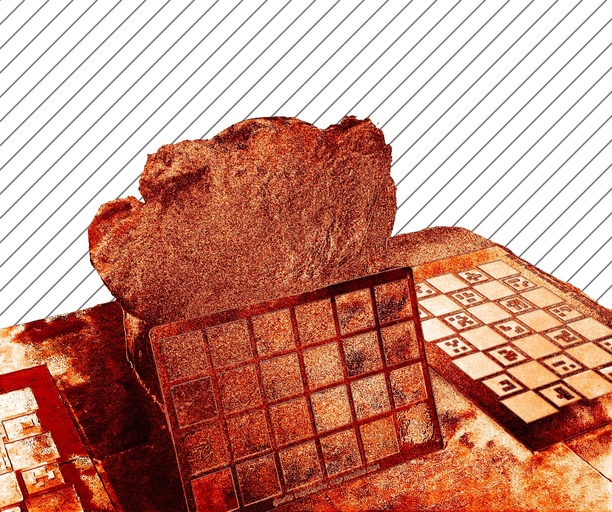}
            \end{subfigure}
            \begin{subfigure}{0.245\linewidth}
                \centering
                \includegraphics[width=\linewidth]{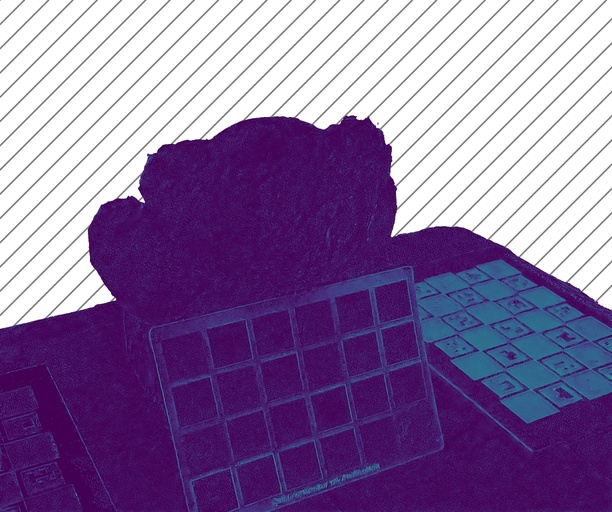}
            \end{subfigure}
            \begin{subfigure}{0.245\linewidth}
                \centering
                \includegraphics[width=\linewidth]{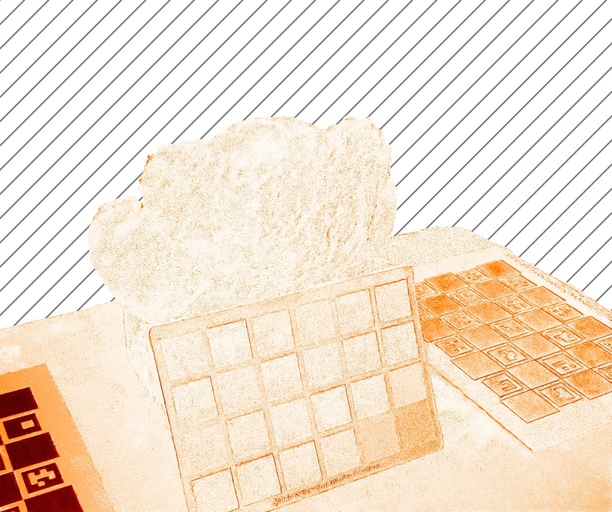}
            \end{subfigure}
        \end{minipage}
    \end{minipage}
    \begin{minipage}{\linewidth}
        \begin{minipage}{0.025\linewidth}
            \vfill
            \rotatebox[origin=cb]{90}{GT}
            \vfill
        \end{minipage}
        \begin{minipage}{0.97\linewidth}
            \begin{subfigure}{0.245\linewidth}
                \centering
                \includegraphics[width=\linewidth]{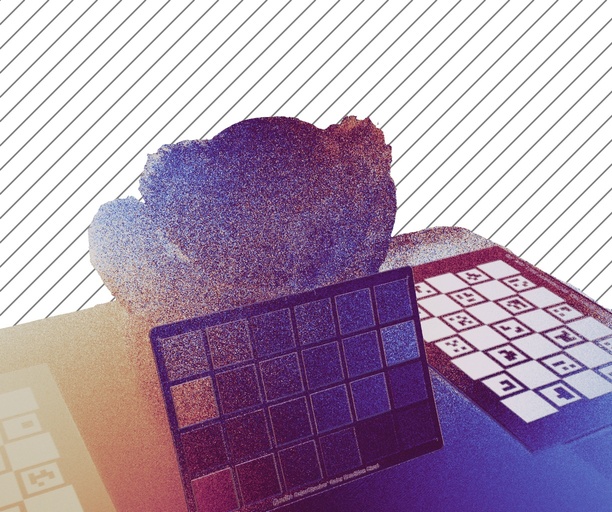}
            \end{subfigure}
            \begin{subfigure}{0.245\linewidth}
                \centering
                \hfill
            \end{subfigure}
            \begin{subfigure}{0.245\linewidth}
                \centering
                \includegraphics[width=\linewidth]{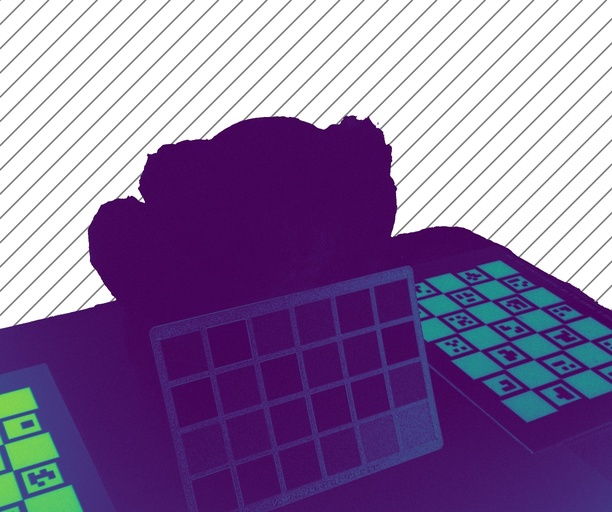}
            \end{subfigure}
            \begin{subfigure}{0.245\linewidth}
                \centering
                \hfill
            \end{subfigure}
        \end{minipage}
    \end{minipage}
    \caption{Qualitative comparison between \gls{method-name} and \gls{mms-fw} $+$ PolarAnything (PA) in terms of recovered polarization on the ``Teddybear'' scene. ``before'' and ``after'' refer to whether the RGB-to-Pol conversion in performed before or after the training of \gls{mms-fw}.}
    \label{fig:polar_teddybear}
    \vspace{-0.5cm}
\end{figure*}

\begin{figure*}[t]
    \centering
    \begin{minipage}{\linewidth}
        \centering
        \hfill
        \begin{minipage}{0.96\linewidth}
            \begin{minipage}{0.245\linewidth}
                \centering
                AoP
            \end{minipage}
            \begin{minipage}{0.245\linewidth}
                \centering
                MAngE
            \end{minipage}
            \begin{minipage}{0.245\linewidth}
                \centering
                DoP
            \end{minipage}
            \begin{minipage}{0.245\linewidth}
                \centering
                MAbsE
            \end{minipage}
        \end{minipage}
        \vspace{2pt}
    \end{minipage}
    \begin{minipage}{\linewidth}
        \begin{minipage}{0.025\linewidth}
            \vfill
            \rotatebox[origin=cb]{90}{Ours}
            \vfill
        \end{minipage}
        \begin{minipage}{0.97\linewidth}
            \begin{subfigure}{0.245\linewidth}
                \centering
                \includegraphics[width=\linewidth]{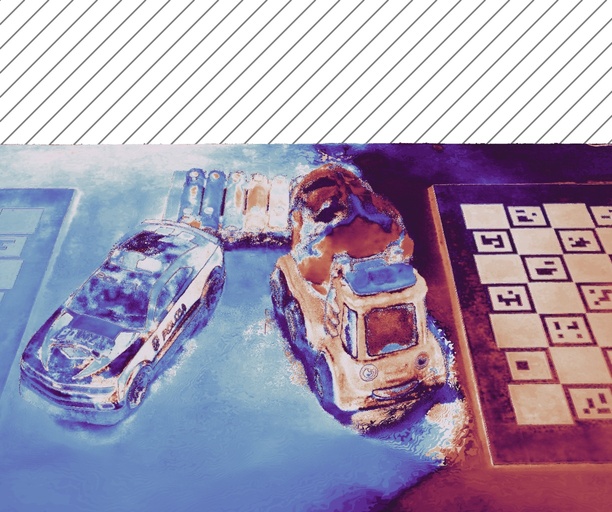}
            \end{subfigure}
            \begin{subfigure}{0.245\linewidth}
                \centering
                \includegraphics[width=\linewidth]{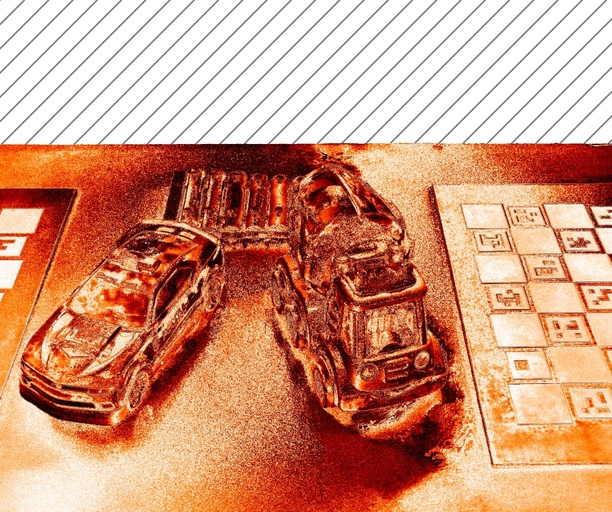}
            \end{subfigure}
            \begin{subfigure}{0.245\linewidth}
                \centering
                \includegraphics[width=\linewidth]{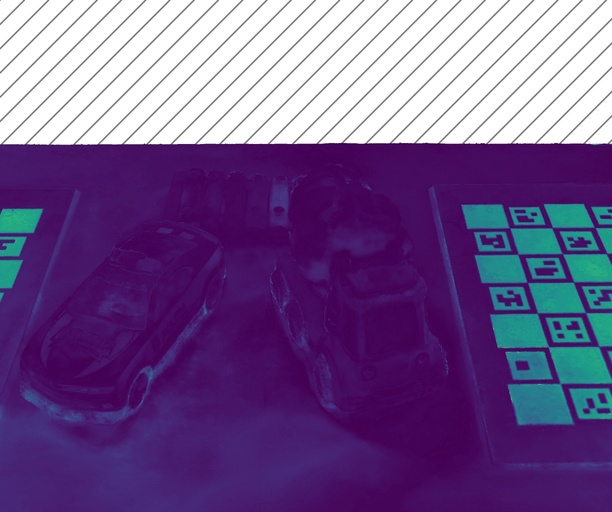}
            \end{subfigure}
            \begin{subfigure}{0.245\linewidth}
                \centering
                \includegraphics[width=\linewidth]{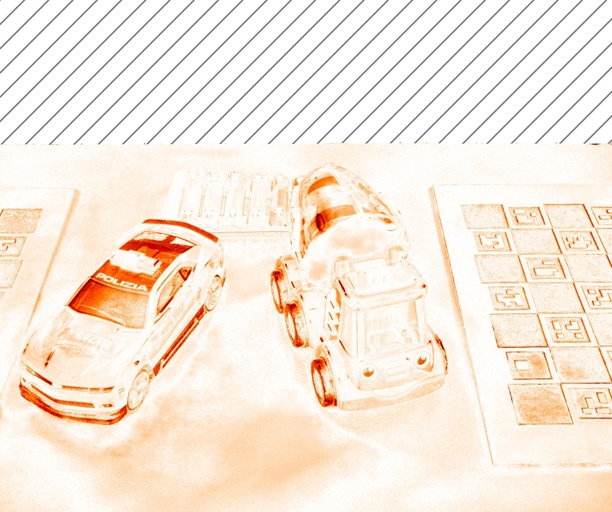}
            \end{subfigure}
        \end{minipage}
    \end{minipage}
    \begin{minipage}{\linewidth}
        \begin{minipage}{0.025\linewidth}
            \vfill
            \rotatebox[origin=cb]{90}{PA before}
            \vfill
        \end{minipage}
        \begin{minipage}{0.97\linewidth}
            \begin{subfigure}{0.245\linewidth}
                \centering
                \includegraphics[width=\linewidth]{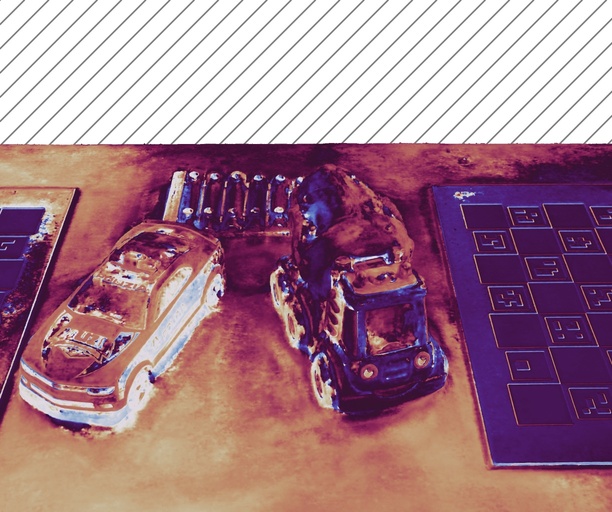}
            \end{subfigure}
            \begin{subfigure}{0.245\linewidth}
                \centering
                \includegraphics[width=\linewidth]{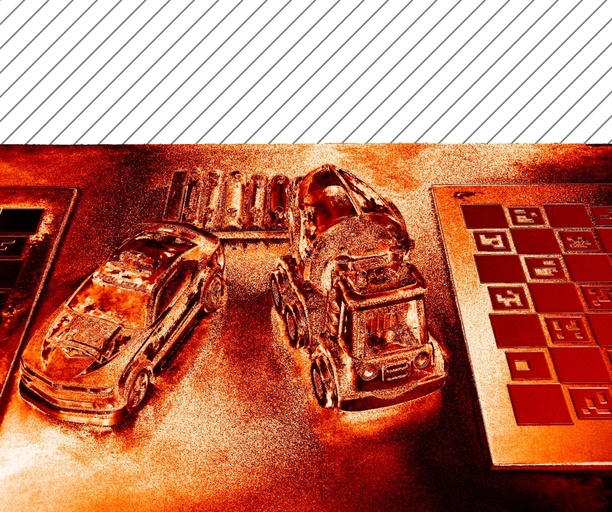}
            \end{subfigure}
            \begin{subfigure}{0.245\linewidth}
                \centering
                \includegraphics[width=\linewidth]{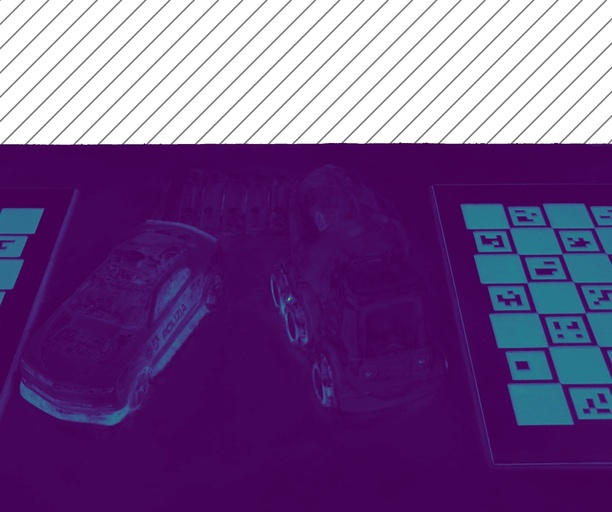}
            \end{subfigure}
            \begin{subfigure}{0.245\linewidth}
                \centering
                \includegraphics[width=\linewidth]{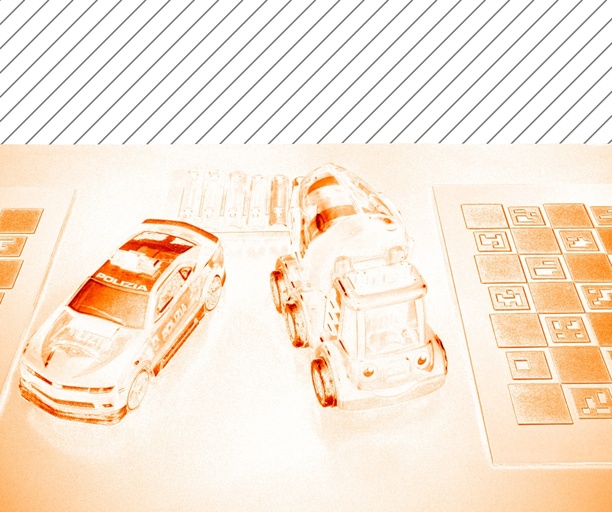}
            \end{subfigure}
        \end{minipage}
    \end{minipage}
    \begin{minipage}{\linewidth}
        \begin{minipage}{0.025\linewidth}
            \vfill
            \rotatebox[origin=cb]{90}{PA after}
            \vfill
        \end{minipage}
        \begin{minipage}{0.97\linewidth}
            \begin{subfigure}{0.245\linewidth}
                \centering
                \includegraphics[width=\linewidth]{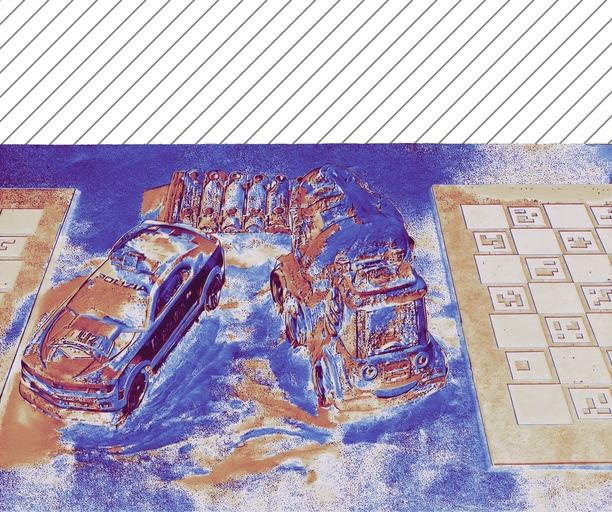}
            \end{subfigure}
            \begin{subfigure}{0.245\linewidth}
                \centering
                \includegraphics[width=\linewidth]{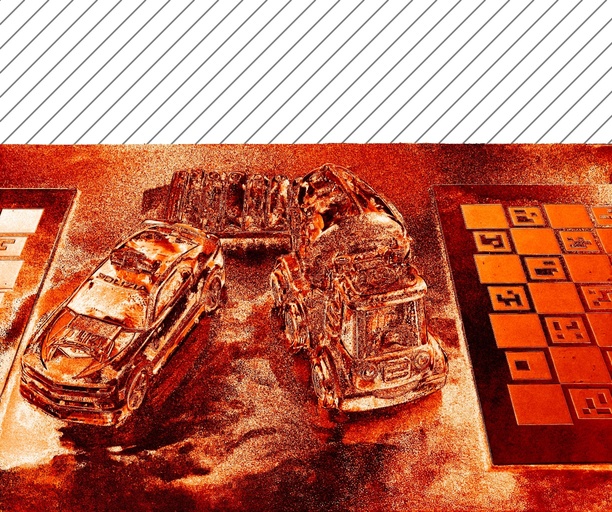}
            \end{subfigure}
            \begin{subfigure}{0.245\linewidth}
                \centering
                \includegraphics[width=\linewidth]{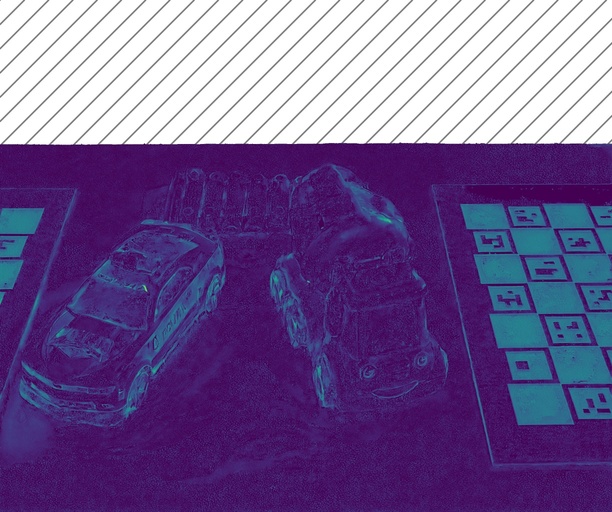}
            \end{subfigure}
            \begin{subfigure}{0.245\linewidth}
                \centering
                \includegraphics[width=\linewidth]{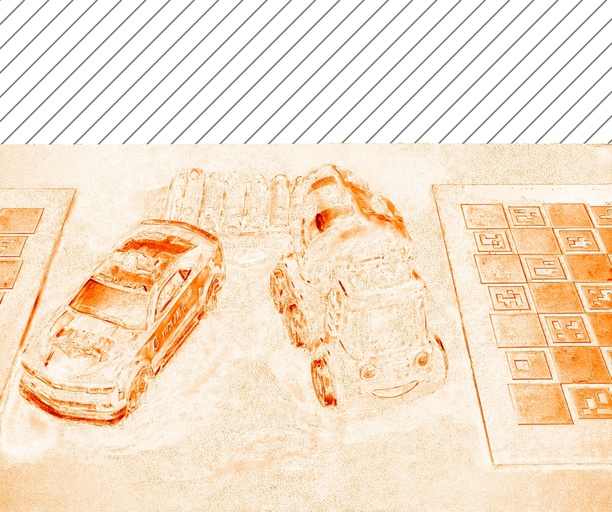}
            \end{subfigure}
        \end{minipage}
    \end{minipage}
    \begin{minipage}{\linewidth}
        \begin{minipage}{0.025\linewidth}
            \vfill
            \rotatebox[origin=cb]{90}{GT}
            \vfill
        \end{minipage}
        \begin{minipage}{0.97\linewidth}
            \begin{subfigure}{0.245\linewidth}
                \centering
                \includegraphics[width=\linewidth]{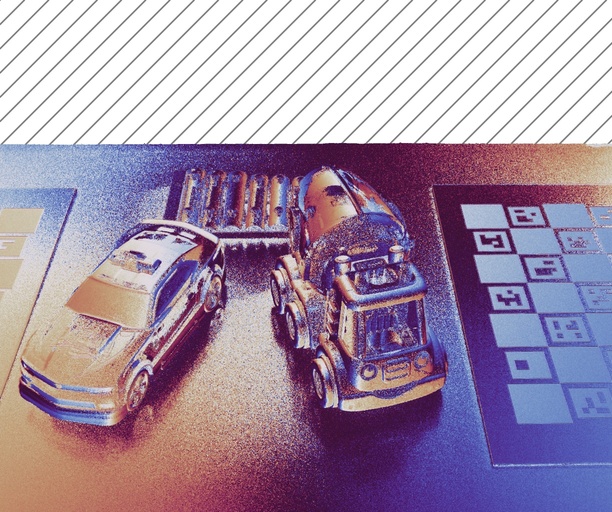}
            \end{subfigure}
            \begin{subfigure}{0.245\linewidth}
                \centering
                \hfill
            \end{subfigure}
            \begin{subfigure}{0.245\linewidth}
                \centering
                \includegraphics[width=\linewidth]{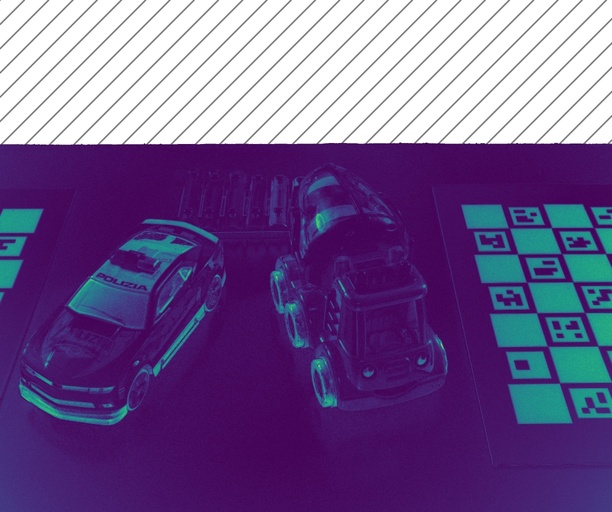}
            \end{subfigure}
            \begin{subfigure}{0.245\linewidth}
                \centering
                \hfill
            \end{subfigure}
        \end{minipage}
    \end{minipage}
    \caption{Qualitative comparison between \gls{method-name} and \gls{mms-fw} $+$ PolarAnything (PA) in terms of recovered polarization on the ``Toys'' scene. ``before'' and ``after'' refer to whether the RGB-to-Pol conversion in performed before or after the training of \gls{mms-fw}.}
    \label{fig:polar_toys}
    \vspace{-0.5cm}
\end{figure*}

\begin{figure*}[t]
    \centering
    \begin{minipage}{\linewidth}
        \centering
        \hfill
        \begin{minipage}{0.96\linewidth}
            \begin{minipage}{0.245\linewidth}
                \centering
                AoP
            \end{minipage}
            \begin{minipage}{0.245\linewidth}
                \centering
                MAngE
            \end{minipage}
            \begin{minipage}{0.245\linewidth}
                \centering
                DoP
            \end{minipage}
            \begin{minipage}{0.245\linewidth}
                \centering
                MAbsE
            \end{minipage}
        \end{minipage}
        \vspace{2pt}
    \end{minipage}
    \begin{minipage}{\linewidth}
        \begin{minipage}{0.025\linewidth}
            \vfill
            \rotatebox[origin=cb]{90}{Ours}
            \vfill
        \end{minipage}
        \begin{minipage}{0.97\linewidth}
            \begin{subfigure}{0.245\linewidth}
                \centering
                \includegraphics[width=\linewidth]{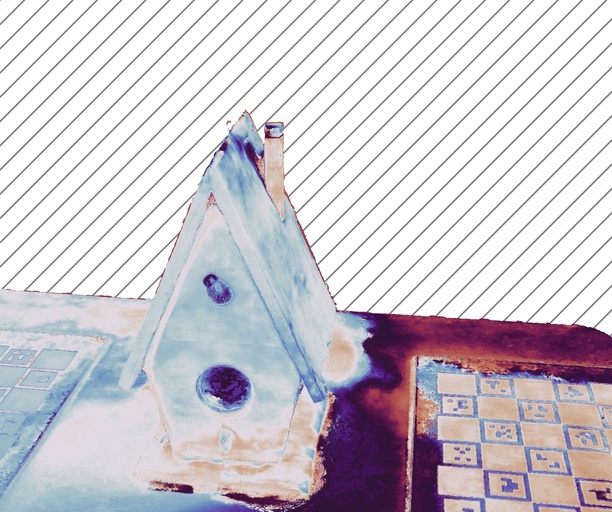}
            \end{subfigure}
            \begin{subfigure}{0.245\linewidth}
                \centering
                \includegraphics[width=\linewidth]{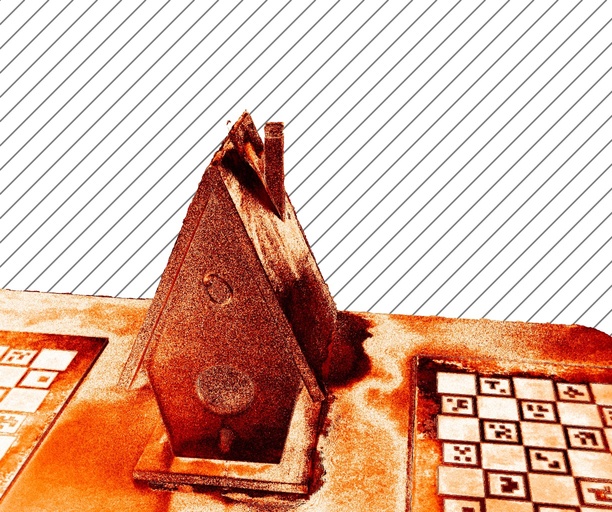}
            \end{subfigure}
            \begin{subfigure}{0.245\linewidth}
                \centering
                \includegraphics[width=\linewidth]{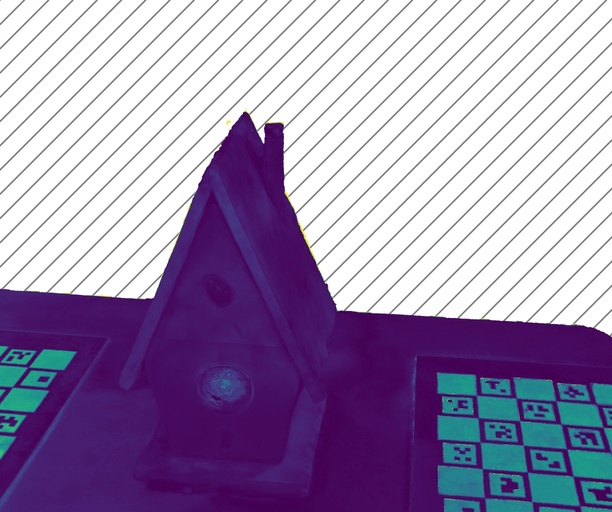}
            \end{subfigure}
            \begin{subfigure}{0.245\linewidth}
                \centering
                \includegraphics[width=\linewidth]{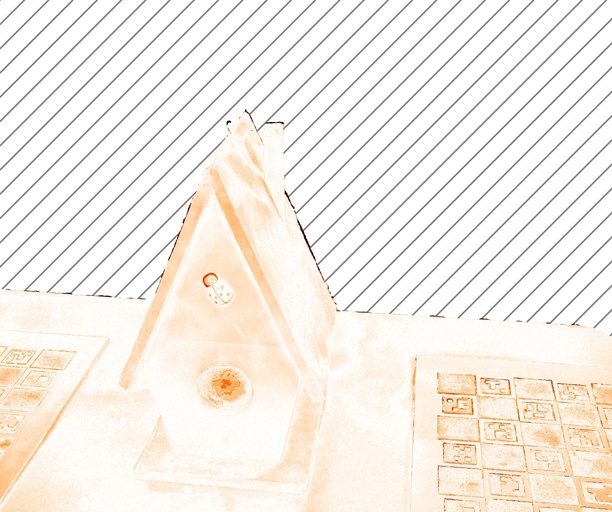}
            \end{subfigure}
        \end{minipage}
    \end{minipage}
    \begin{minipage}{\linewidth}
        \begin{minipage}{0.025\linewidth}
            \vfill
            \rotatebox[origin=cb]{90}{PA before}
            \vfill
        \end{minipage}
        \begin{minipage}{0.97\linewidth}
            \begin{subfigure}{0.245\linewidth}
                \centering
                \includegraphics[width=\linewidth]{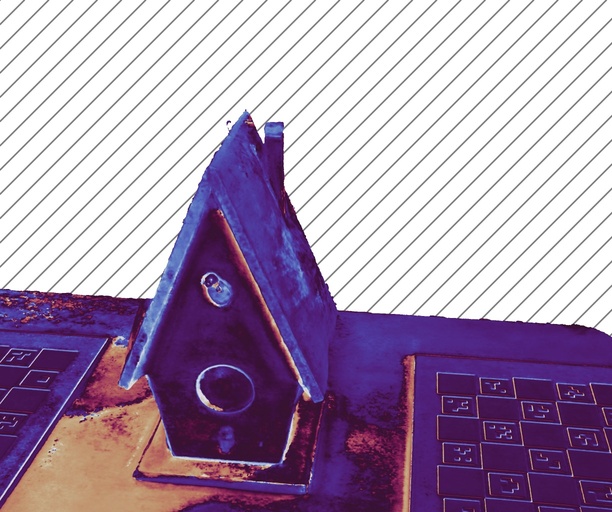}
            \end{subfigure}
            \begin{subfigure}{0.245\linewidth}
                \centering
                \includegraphics[width=\linewidth]{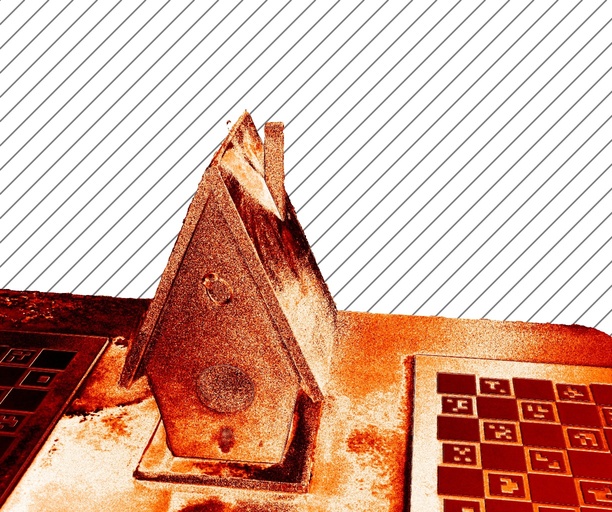}
            \end{subfigure}
            \begin{subfigure}{0.245\linewidth}
                \centering
                \includegraphics[width=\linewidth]{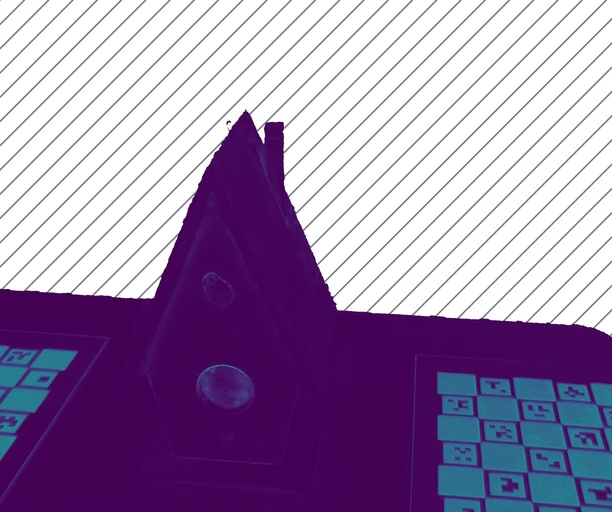}
            \end{subfigure}
            \begin{subfigure}{0.245\linewidth}
                \centering
                \includegraphics[width=\linewidth]{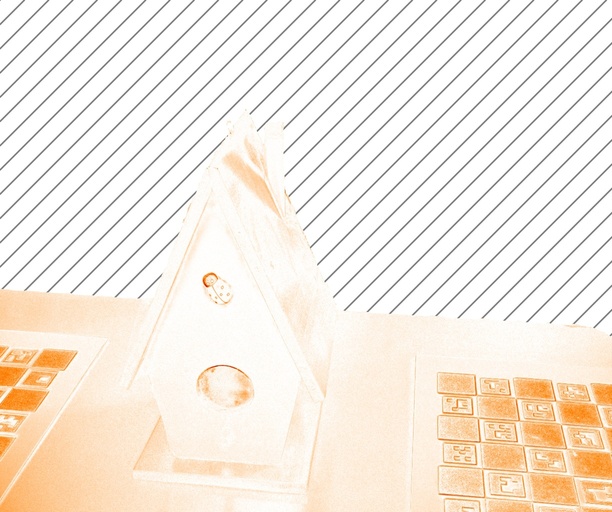}
            \end{subfigure}
        \end{minipage}
    \end{minipage}
    \begin{minipage}{\linewidth}
        \begin{minipage}{0.025\linewidth}
            \vfill
            \rotatebox[origin=cb]{90}{PA after}
            \vfill
        \end{minipage}
        \begin{minipage}{0.97\linewidth}
            \begin{subfigure}{0.245\linewidth}
                \centering
                \includegraphics[width=\linewidth]{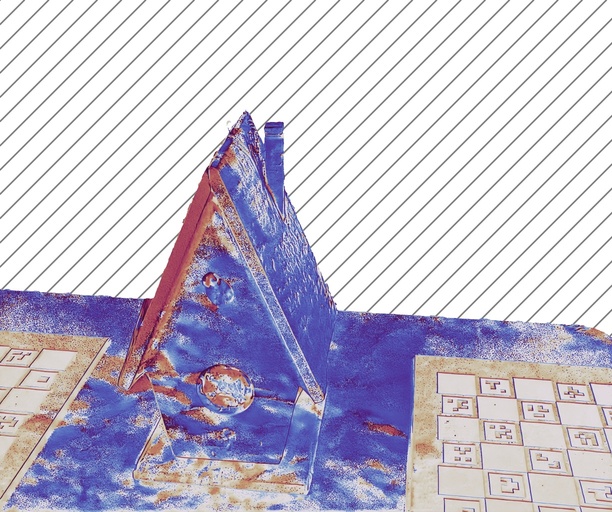}
            \end{subfigure}
            \begin{subfigure}{0.245\linewidth}
                \centering
                \includegraphics[width=\linewidth]{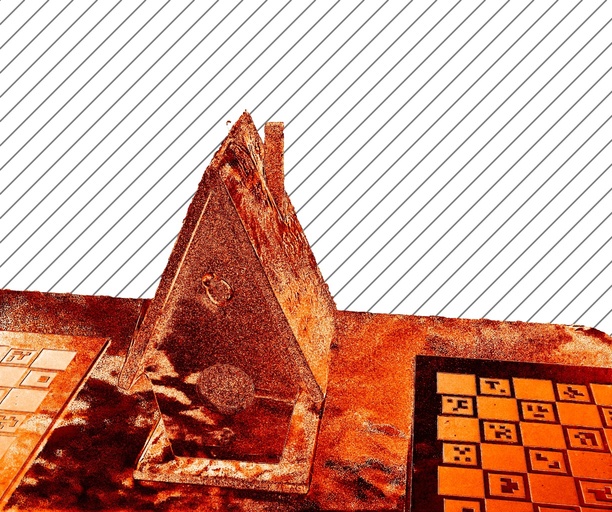}
            \end{subfigure}
            \begin{subfigure}{0.245\linewidth}
                \centering
                \includegraphics[width=\linewidth]{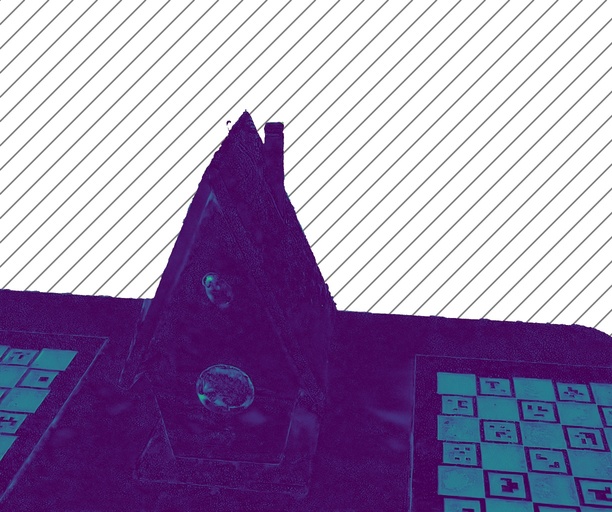}
            \end{subfigure}
            \begin{subfigure}{0.245\linewidth}
                \centering
                \includegraphics[width=\linewidth]{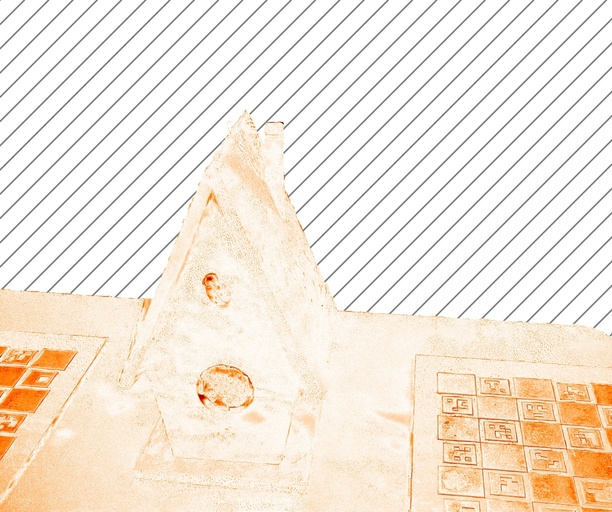}
            \end{subfigure}
        \end{minipage}
    \end{minipage}
    \begin{minipage}{\linewidth}
        \begin{minipage}{0.025\linewidth}
            \vfill
            \rotatebox[origin=cb]{90}{GT}
            \vfill
        \end{minipage}
        \begin{minipage}{0.97\linewidth}
            \begin{subfigure}{0.245\linewidth}
                \centering
                \includegraphics[width=\linewidth]{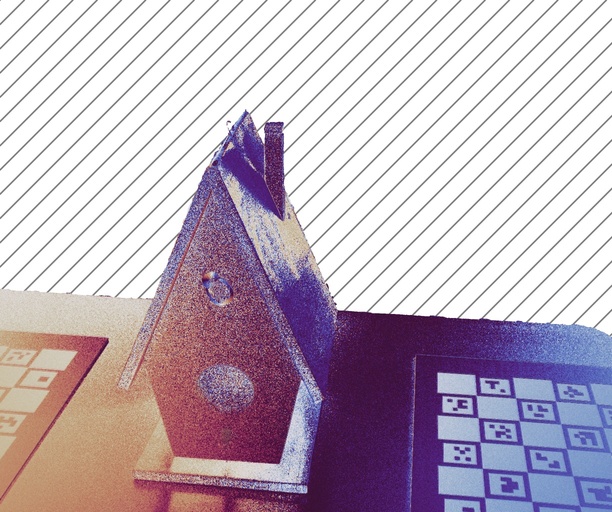}
            \end{subfigure}
            \begin{subfigure}{0.245\linewidth}
                \centering
                \hfill
            \end{subfigure}
            \begin{subfigure}{0.245\linewidth}
                \centering
                \includegraphics[width=\linewidth]{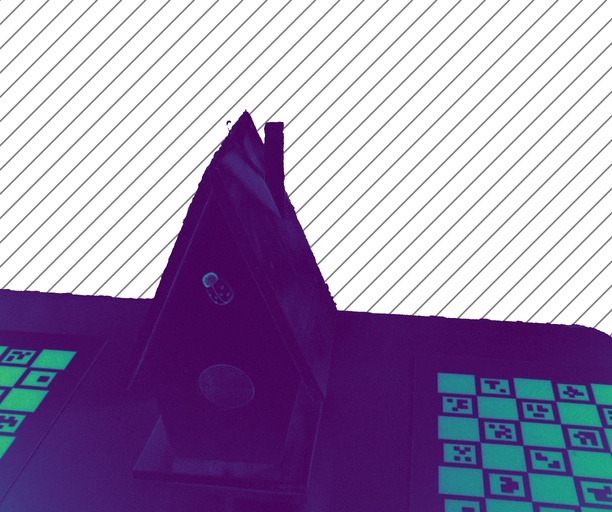}
            \end{subfigure}
            \begin{subfigure}{0.245\linewidth}
                \centering
                \hfill
            \end{subfigure}
        \end{minipage}
    \end{minipage}
    \caption{Qualitative comparison between \gls{method-name} and \gls{mms-fw} $+$ PolarAnything (PA) in terms of recovered polarization on the ``Birdhouse'' scene. ``before'' and ``after'' refer to whether the RGB-to-Pol conversion in performed before or after the training of \gls{mms-fw}.}
    \label{fig:polar_birdhouse}
    \vspace{-0.5cm}
\end{figure*}

\begin{figure*}[t]
    \centering
    \begin{minipage}{\linewidth}
        \centering
        \hfill
        \begin{minipage}{0.96\linewidth}
            \begin{minipage}{0.245\linewidth}
                \centering
                AoP
            \end{minipage}
            \begin{minipage}{0.245\linewidth}
                \centering
                MAngE
            \end{minipage}
            \begin{minipage}{0.245\linewidth}
                \centering
                DoP
            \end{minipage}
            \begin{minipage}{0.245\linewidth}
                \centering
                MAbsE
            \end{minipage}
        \end{minipage}
        \vspace{2pt}
    \end{minipage}
    \begin{minipage}{\linewidth}
        \begin{minipage}{0.025\linewidth}
            \vfill
            \rotatebox[origin=cb]{90}{Ours}
            \vfill
        \end{minipage}
        \begin{minipage}{0.97\linewidth}
            \begin{subfigure}{0.245\linewidth}
                \centering
                \includegraphics[width=\linewidth]{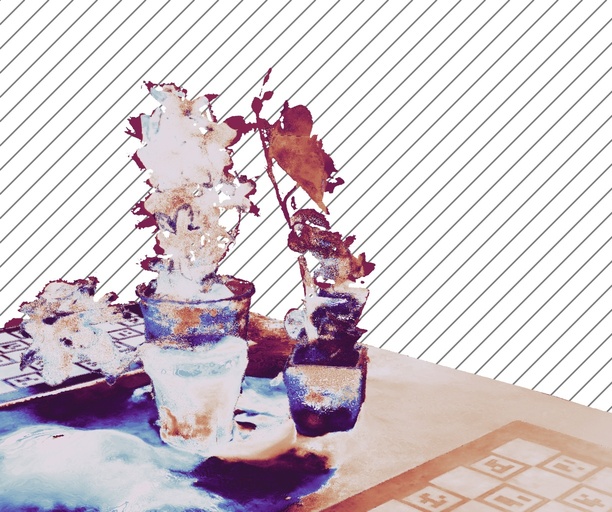}
            \end{subfigure}
            \begin{subfigure}{0.245\linewidth}
                \centering
                \includegraphics[width=\linewidth]{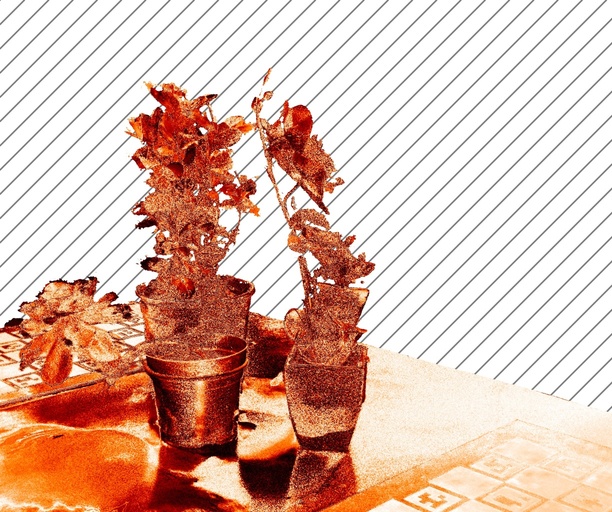}
            \end{subfigure}
            \begin{subfigure}{0.245\linewidth}
                \centering
                \includegraphics[width=\linewidth]{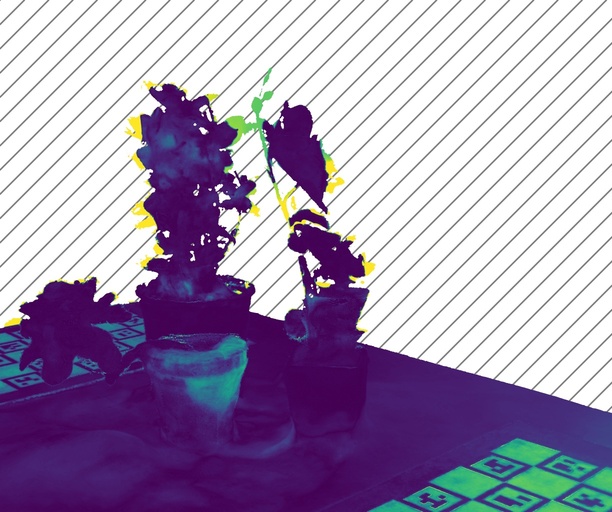}
            \end{subfigure}
            \begin{subfigure}{0.245\linewidth}
                \centering
                \includegraphics[width=\linewidth]{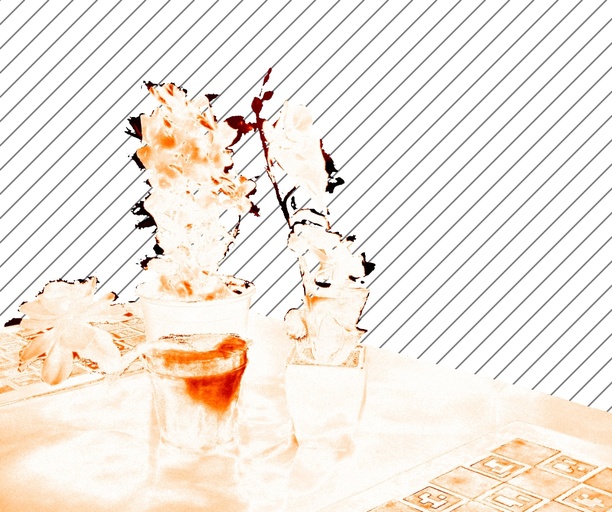}
            \end{subfigure}
        \end{minipage}
    \end{minipage}
    \begin{minipage}{\linewidth}
        \begin{minipage}{0.025\linewidth}
            \vfill
            \rotatebox[origin=cb]{90}{PA before}
            \vfill
        \end{minipage}
        \begin{minipage}{0.97\linewidth}
            \begin{subfigure}{0.245\linewidth}
                \centering
                \includegraphics[width=\linewidth]{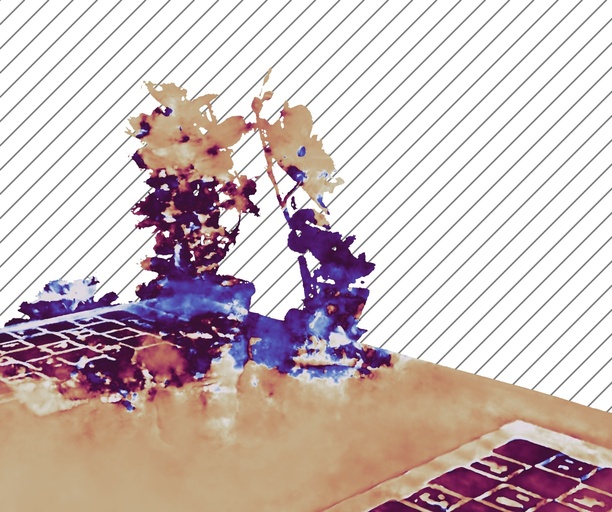}
            \end{subfigure}
            \begin{subfigure}{0.245\linewidth}
                \centering
                \includegraphics[width=\linewidth]{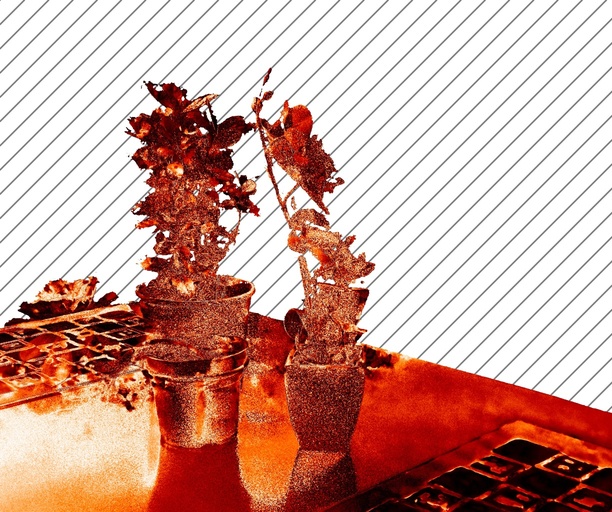}
            \end{subfigure}
            \begin{subfigure}{0.245\linewidth}
                \centering
                \includegraphics[width=\linewidth]{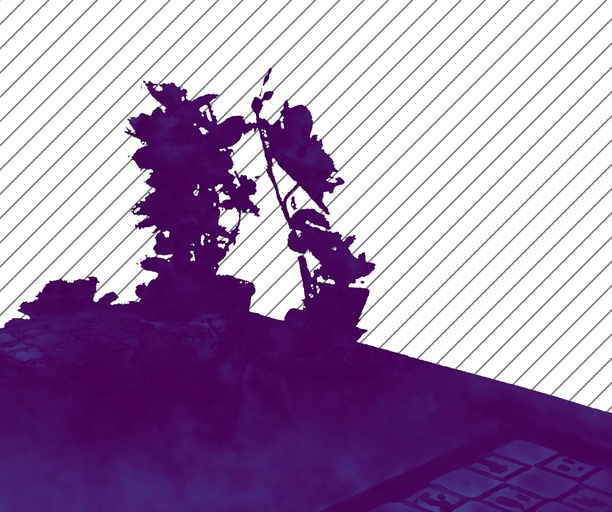}
            \end{subfigure}
            \begin{subfigure}{0.245\linewidth}
                \centering
                \includegraphics[width=\linewidth]{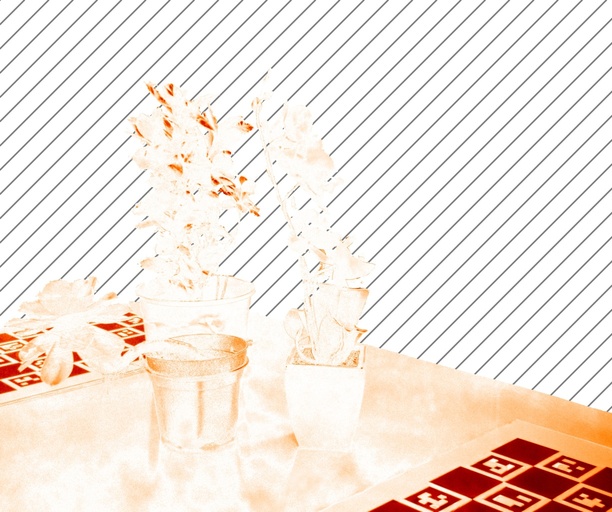}
            \end{subfigure}
        \end{minipage}
    \end{minipage}
    \begin{minipage}{\linewidth}
        \begin{minipage}{0.025\linewidth}
            \vfill
            \rotatebox[origin=cb]{90}{PA after}
            \vfill
        \end{minipage}
        \begin{minipage}{0.97\linewidth}
            \begin{subfigure}{0.245\linewidth}
                \centering
                \includegraphics[width=\linewidth]{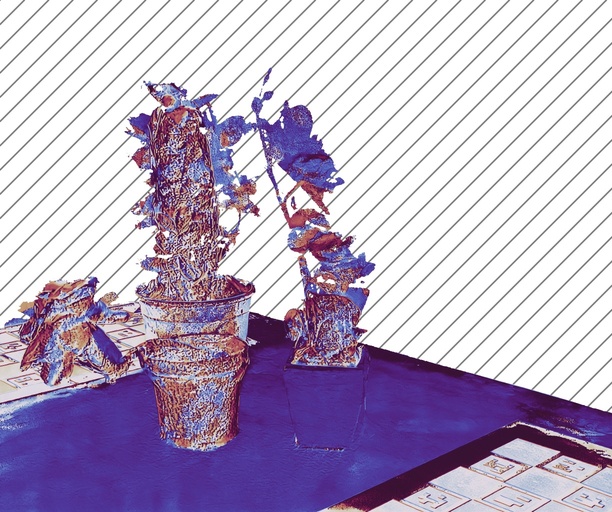}
            \end{subfigure}
            \begin{subfigure}{0.245\linewidth}
                \centering
                \includegraphics[width=\linewidth]{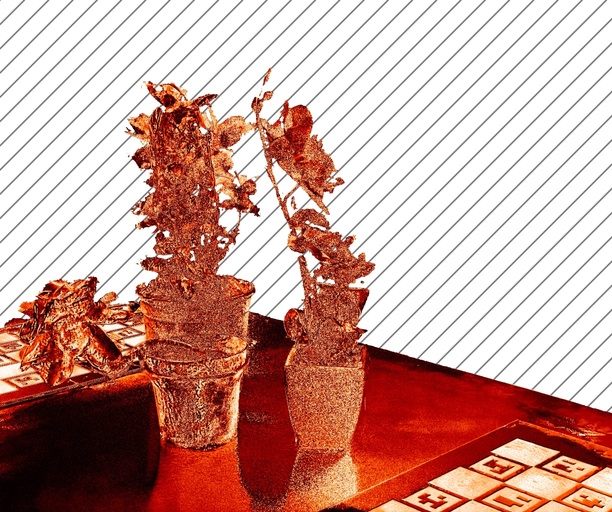}
            \end{subfigure}
            \begin{subfigure}{0.245\linewidth}
                \centering
                \includegraphics[width=\linewidth]{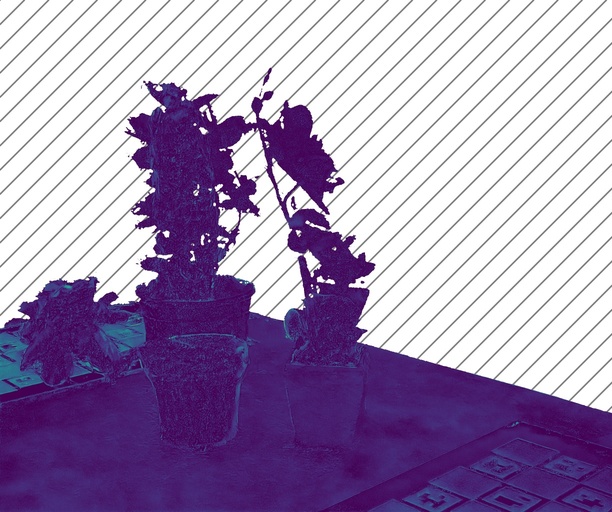}
            \end{subfigure}
            \begin{subfigure}{0.245\linewidth}
                \centering
                \includegraphics[width=\linewidth]{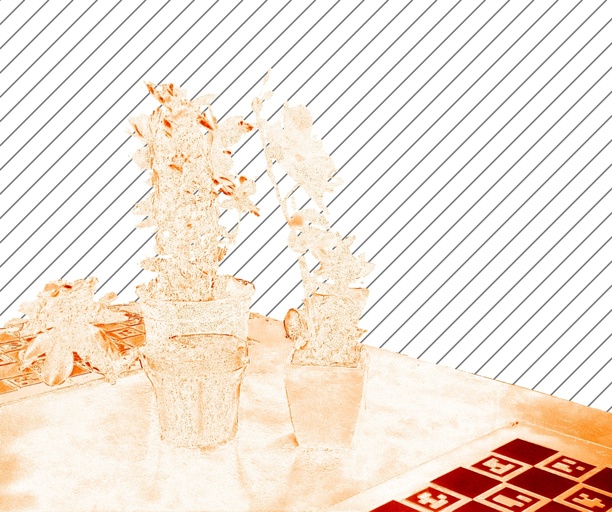}
            \end{subfigure}
        \end{minipage}
    \end{minipage}
    \begin{minipage}{\linewidth}
        \begin{minipage}{0.025\linewidth}
            \vfill
            \rotatebox[origin=cb]{90}{GT}
            \vfill
        \end{minipage}
        \begin{minipage}{0.97\linewidth}
            \begin{subfigure}{0.245\linewidth}
                \centering
                \includegraphics[width=\linewidth]{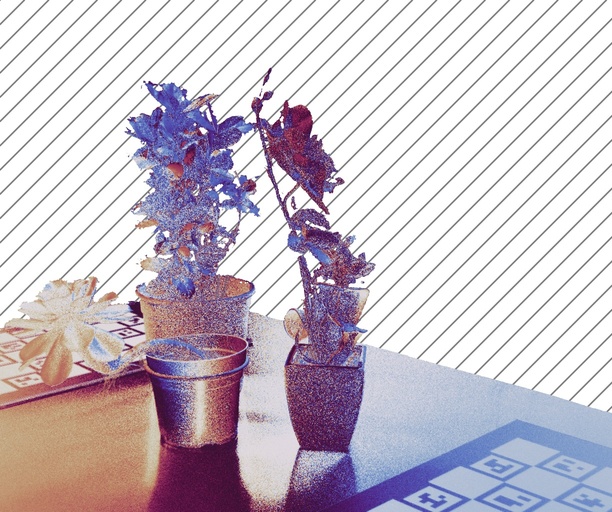}
            \end{subfigure}
            \begin{subfigure}{0.245\linewidth}
                \centering
                \hfill
            \end{subfigure}
            \begin{subfigure}{0.245\linewidth}
                \centering
                \includegraphics[width=\linewidth]{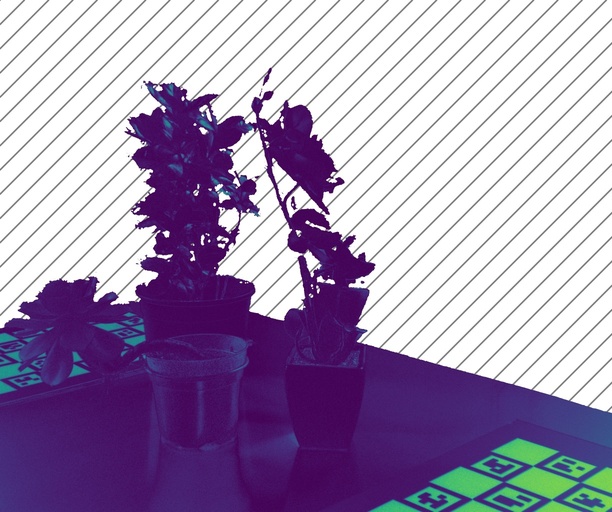}
            \end{subfigure}
            \begin{subfigure}{0.245\linewidth}
                \centering
                \hfill
            \end{subfigure}
        \end{minipage}
    \end{minipage}
    \caption{Qualitative comparison between \gls{method-name} and \gls{mms-fw} $+$ PolarAnything (PA) in terms of recovered polarization on the ``Bouquet'' scene. ``before'' and ``after'' refer to whether the RGB-to-Pol conversion in performed before or after the training of \gls{mms-fw}.}
    \label{fig:polar_bouquet}
    \vspace{-0.5cm}
\end{figure*}

In this section, we provide additional details about the modality conversion procedure with third-party methods performed in the experiments of Sec. 4.5. \\
For the \gls{ms} data, we employ
MST++ \cite{cai2022mst++}. It can convert an RGB image to a \gls{hsi} frame with 31 channels. We start by converting the RGB images of the 5 \gls{ft} scenes into \gls{hsi} images, then we map the 31 \gls{hsi} channels to the 9 channels of MMS-DATA \gls{ms} frames. For this purpose, we estimate a transformation matrix $\textbf{M}\in\mathbb{R}^{9\times31}$ by considering the quantum efficiency of the \gls{ms} sensor used in MMS-DATA, namely a Silios CMS-C1. Specifically, each row of $\textbf{M}$ contains the 31 coefficients to perform the linear combination that maps the 31 \gls{hsi} bands to each of the 9 \gls{ms} bands. The 31 row-wise coefficients, where each row corresponds to a different band of the \gls{ms} sensor, are the values of the \gls{ms} sensor quantum efficiency that correspond to the wavelength of each of the 31 \gls{hsi} bands. However, considering that MST++ does not have clues about the illumination spectrum of MMS-DATA, we refine $\textbf{M}$ with least squares by exploiting the images showing an X-Rite ColorChecker that are released along with MMS-DATA. \\
\indent For the \gls{pol} data, we instead use PolarAnything~\cite{zhang2025polaranything}. 
The procedure is analogous, except that we did not estimate any transformation to match the illumination.
Both \gls{aop} and \gls{dop} are independent of sensor exposure and, in this case, even of wavelength since the polarimetric sensor is monochromatic. \\
\indent Finally, it is important to mention that in the main paper, we first convert RGB frames to either \gls{pol} or \gls{ms} and then train \gls{mms-fw} with them. This is the most straightforward  procedure, but it is also possible to reverse the order: \ie, first train \gls{mms-fw} on RGB frames and then convert the RGB output renderings with MST++ or PolarAnything to multispectral or polarimetric data, respectively.
We evaluate both strategies and provide qualitative results in \cref{fig:mst++_birdhouse,fig:mst++_bouquet,fig:mst++_fruits,fig:mst++_teddybear,fig:mst++_toys,fig:polar_birdhouse,fig:polar_bouquet,fig:polar_fruits,fig:polar_teddybear,fig:polar_toys}. 
It is possible to observe that the results are much worse when the conversion is performed after \gls{mms-fw} training. This likely happens because both MST++ and PolarAnything are trained on real images; thus, they struggle when queried with rendered images that have different noise statistics and present artifacts typically produced by NeRF-like methods.
Moreover, this procedure cannot benefit from the multi-view consistency enforced by NeRF-like models during training, as the modality conversion is performed as a final step on each rendering independently.\\
\indent Comparing these results with \gls{method-name}, it is possible to see that \Gls{ms} data estimation from our approach is more accurate, especially for the ``Fruits'' and ``Teddybear'' scenes, where the gap is very wide.
Concerning polarization, both the AoP and DoP are better estimated by using \gls{method-name}. It achieves good results, especially in the DoP estimation. The AoP estimation appears much more challenging for all approaches; nevertheless, \gls{method-name} can produce much better estimations, while PolarAnything struggles more with this task.

\section{Comparison on the  X-NeRF Dataset}
\label{sec:xnerf}

\begin{figure}[t]
    \centering
    \includegraphics[width=\linewidth]{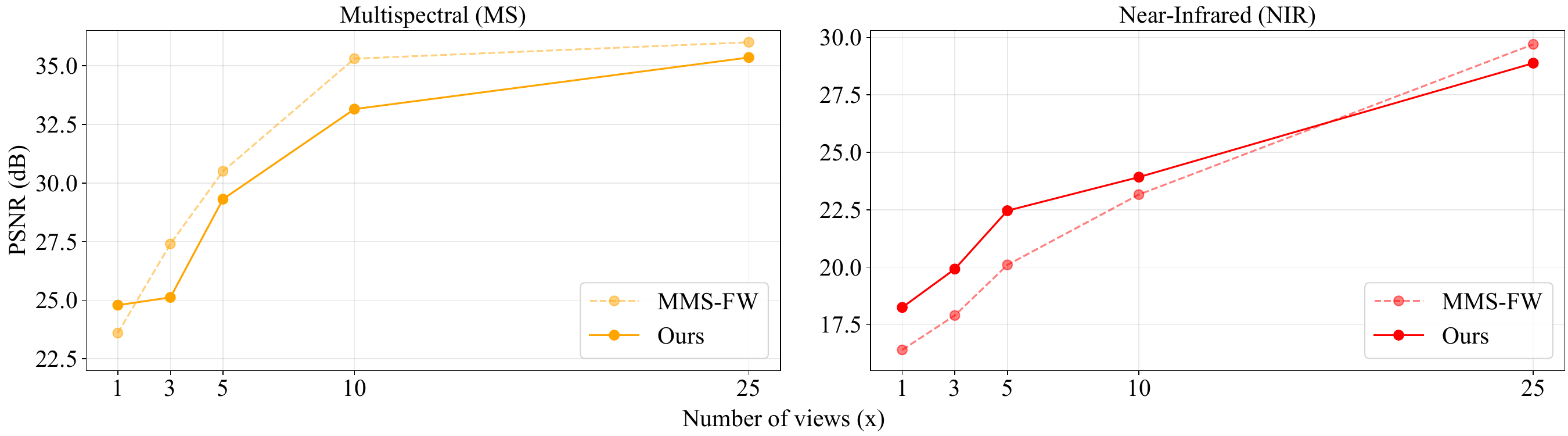}
    \caption{Tests with an unbalanced combination of modalities. The X axis corresponds to the number of additional modality (MS or NIR) frames. Comparison between SPoILeR (Ours) and MMS-FW. Results averaged on all 16 X-NeRF scenes.}
    \label{fig:xnerf_plot}
\end{figure}

As briefly mentioned in Sec. 4.4 of the main document, we also conduct experiments on the X-NeRF~\cite{poggi2022cross} dataset. This is the only other dataset available in literature that encompasses frames from multiple heterogeneous imaging modalities, even if its forward-facing viewpoint arrangement is not well suited to highlight the capabilities of our approach, as discussed later.\\
\indent The idea is to verify whether \gls{method-name} can generalize across different scenes captured with different sensors compared to what is included in MMS-DATA. The idea of this test is to consider the model pre-trained on MMS-DATA with the standard \gls{pt} phase and fine-tune it on the X-NeRF scenes. For this purpose, we must consider that since the X-NeRF sensors are different from the MMS-DATA sensors (\eg, in terms of quantum efficiency, number of channels, exposure, white balancing, \etc), during the \gls{ft} we have to re-train the modality decoders for each X-NeRF modality. 
For this reason, in addition to the coefficient field \textbf{c}, we also optimize the modality decoders $\mathcal{D}_i$. These two modules are optimized from scratch for every X-NeRF scene during each \gls{ft}, while the basis field \textbf{b}, the projection function $\mathcal{J}$, and the latent encoder $\mathcal{Z}$ are frozen.\\
\indent To enable the optimization of the modality decoders, it is required that at least one frame of the considered scene per modality is available. 
For this reason, we choose to conduct this test in the scenario with an unbalanced combination of modalities, presented in Sec. 4.4 of the main paper. 
An additional reason is that the X-NeRF dataset does not provide the \gls{nir} and \gls{ms} camera poses; thus, we need at least one frame during the model optimization to regress their pose on the camera rig.\\
\indent We conduct the experiment by using all the RGB frames and an increasing number of additional modality frames, as in Sec. 4.4. Since X-NeRF scenes have $\sim$30 images, we used 1, 3, 5, 10, and 25 views for training and 5 views for testing. In \cref{fig:xnerf_plot} we show the PSNR as a function of  the number of additional modality frames, and compare the results of \gls{method-name} with \gls{mms-fw}. As anticipated in Sec. 4.4, the models on average perform similarly: \gls{method-name} achieves slightly better results with \gls{nir} images and \gls{mms-fw} with \gls{ms} images. On the one hand, the good PSNR scores demonstrate that \gls{method-name} can generalize well on scenarios different from the \gls{pt} scenario, even when the sensors employed are not the same. 
On the other hand, the camera arrangement of the X-NeRF scenes leads \gls{method-name} and \gls{mms-fw} to achieve uniform results. This happens because X-NeRF scenes are forward-facing, and all viewpoints are close to each other. For this reason, even a few additional modality views allow covering a very large portion of the scene, and this reduces the advantage given by the \gls{pt} knowledge, leading the two models to perform on par.

\section{Pre-training Results}
\label{sec:pt}

\begin{table}[t]
    \centering
    \caption{Results of \gls{method-name} pre-training, averaged on the considered 5 \gls{pt} scenes.}
    \begin{tabular*}{0.7\linewidth}{@{\extracolsep{\fill}}cccccc}
        \toprule
             &Train Mod.& Test Mod.& PSNR$\uparrow$&SSIM$\uparrow$ &\\
        \midrule
              &\multirow{5}{*}{ALL}& \green{RGB}& 26.71& - &\\
        \arrayrulecolor{black!30}\cline{3-6}
               && \purple{Mono}& 23.63&0.80 &\\
        \cline{3-6}
               && \red{NIR}& 25.12&0.82 &\\
        \cline{3-6}
              & & \blue{Pol}& 23.02&- &\\
        \cline{3-6}
               && \orange{MS}& 24.49&- &\\
        \arrayrulecolor{black}\bottomrule
    \end{tabular*}
    \label{tab:pt}
    \vspace{-0.3cm}
\end{table}

\begin{figure*}[t]
    \centering
    \begin{minipage}{\linewidth}
        \centering
        \hfill
        \begin{minipage}{0.96\linewidth}
            \begin{minipage}{0.325\linewidth}
                \centering
                Ours
            \end{minipage}
            \begin{minipage}{0.325\linewidth}
                \centering
                GT
            \end{minipage}
            \begin{minipage}{0.325\linewidth}
                \centering
                Error
            \end{minipage}
        \end{minipage}
        \vspace{2pt}
    \end{minipage}
    \begin{minipage}{\linewidth}
        \begin{minipage}{0.025\linewidth}
            \vfill
            \rotatebox[origin=cb]{90}{RGB}
            \vfill
        \end{minipage}
        \begin{minipage}{0.97\linewidth}
            \begin{subfigure}{0.325\linewidth}
                \centering
                \includegraphics[width=\linewidth]{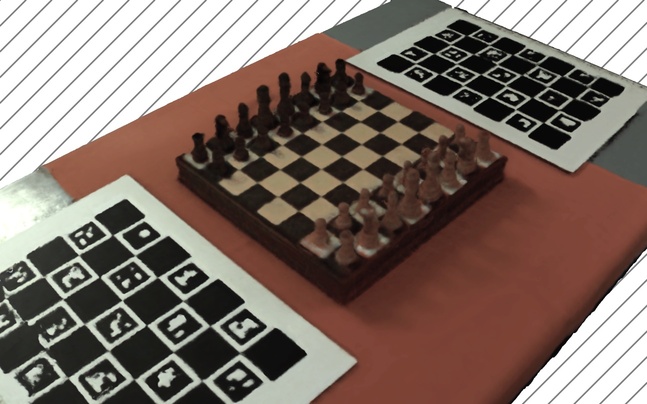}
            \end{subfigure}
            \begin{subfigure}{0.325\linewidth}
                \centering
                \includegraphics[width=\linewidth]{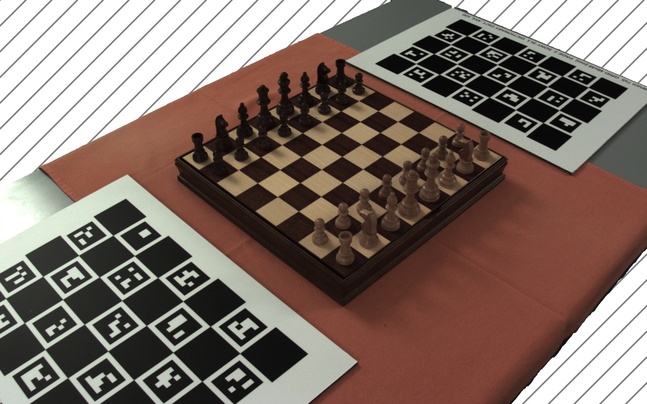}
            \end{subfigure}
            \begin{subfigure}{0.325\linewidth}
                \centering
                \includegraphics[width=\linewidth]{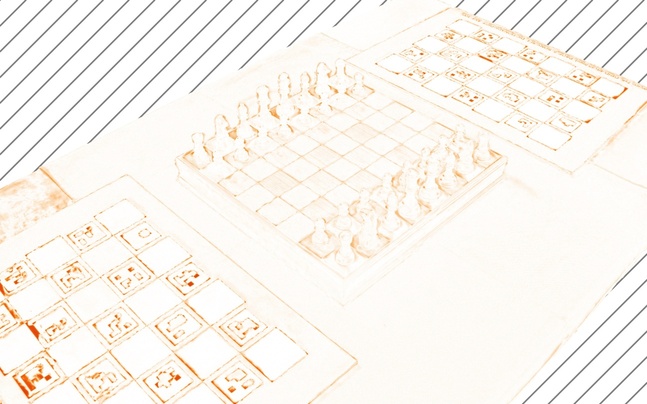}
            \end{subfigure}
        \end{minipage}
    \end{minipage}
    \begin{minipage}{\linewidth}
        \begin{minipage}{0.025\linewidth}
            \vfill
            \rotatebox[origin=cb]{90}{NIR}
            \vfill
        \end{minipage}
        \begin{minipage}{0.97\linewidth}
            \begin{subfigure}{0.325\linewidth}
                \centering
                \includegraphics[width=\linewidth]{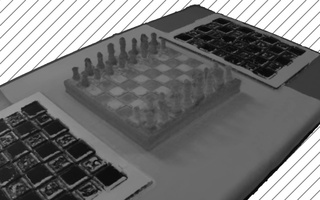}
            \end{subfigure}
            \begin{subfigure}{0.325\linewidth}
                \centering
                \includegraphics[width=\linewidth]{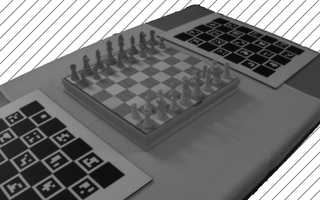}
            \end{subfigure}
            \begin{subfigure}{0.325\linewidth}
                \centering
                \includegraphics[width=\linewidth]{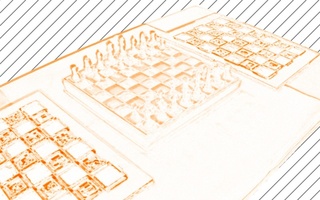}
            \end{subfigure}
        \end{minipage}
    \end{minipage}
    \begin{minipage}{\linewidth}
        \begin{minipage}{0.025\linewidth}
            \vfill
            \rotatebox[origin=cb]{90}{Mono}
            \vfill
        \end{minipage}
        \begin{minipage}{0.97\linewidth}
            \begin{subfigure}{0.325\linewidth}
                \centering
                \includegraphics[width=\linewidth]{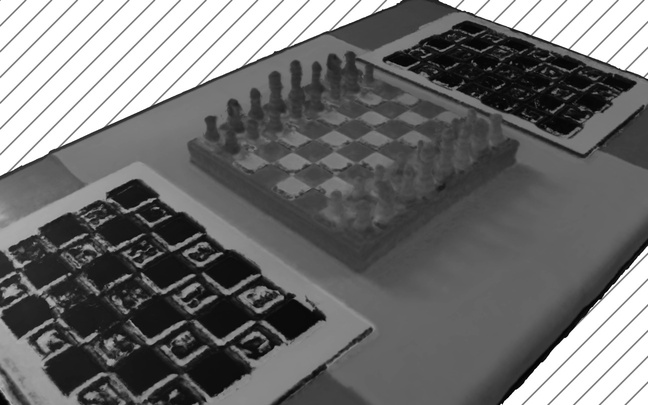}
            \end{subfigure}
            \begin{subfigure}{0.325\linewidth}
                \centering
                \includegraphics[width=\linewidth]{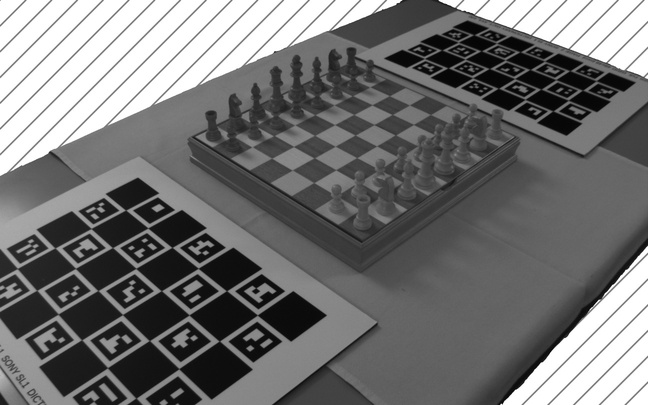}
            \end{subfigure}
            \begin{subfigure}{0.325\linewidth}
                \centering
                \includegraphics[width=\linewidth]{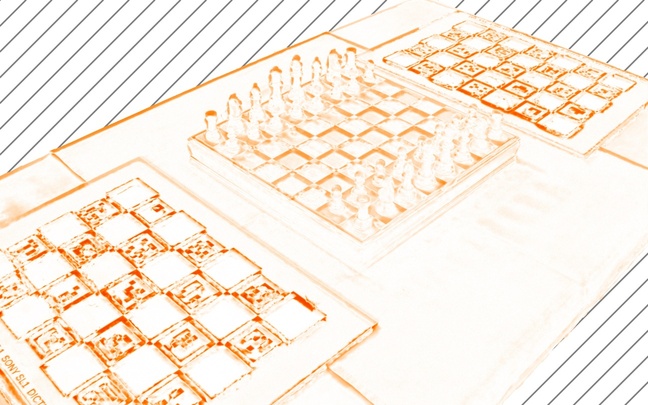}
            \end{subfigure}
        \end{minipage}
    \end{minipage}
    \begin{minipage}{\linewidth}
        \begin{minipage}{0.025\linewidth}
            \vfill
            \rotatebox[origin=cb]{90}{Pol}
            \vfill
        \end{minipage}
        \begin{minipage}{0.97\linewidth}
            \begin{subfigure}{0.325\linewidth}
                \centering
                \includegraphics[width=\linewidth]{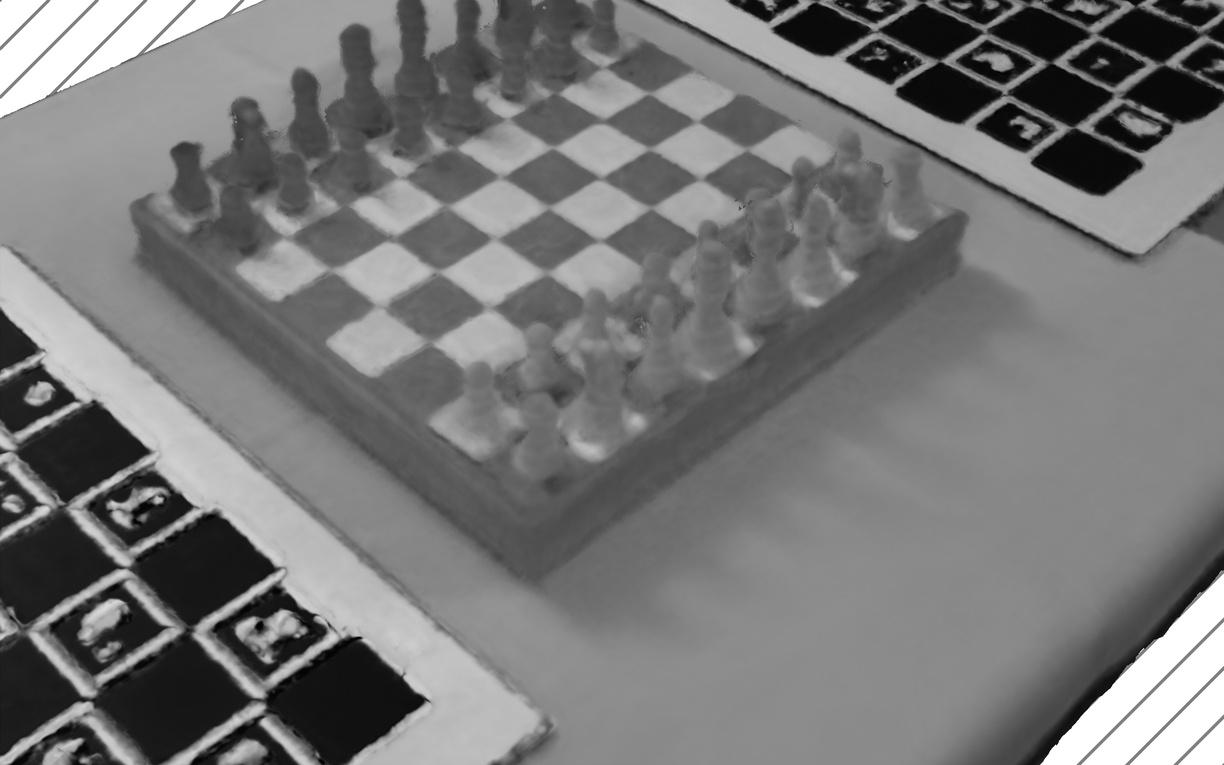}
            \end{subfigure}
            \begin{subfigure}{0.325\linewidth}
                \centering
                \includegraphics[width=\linewidth]{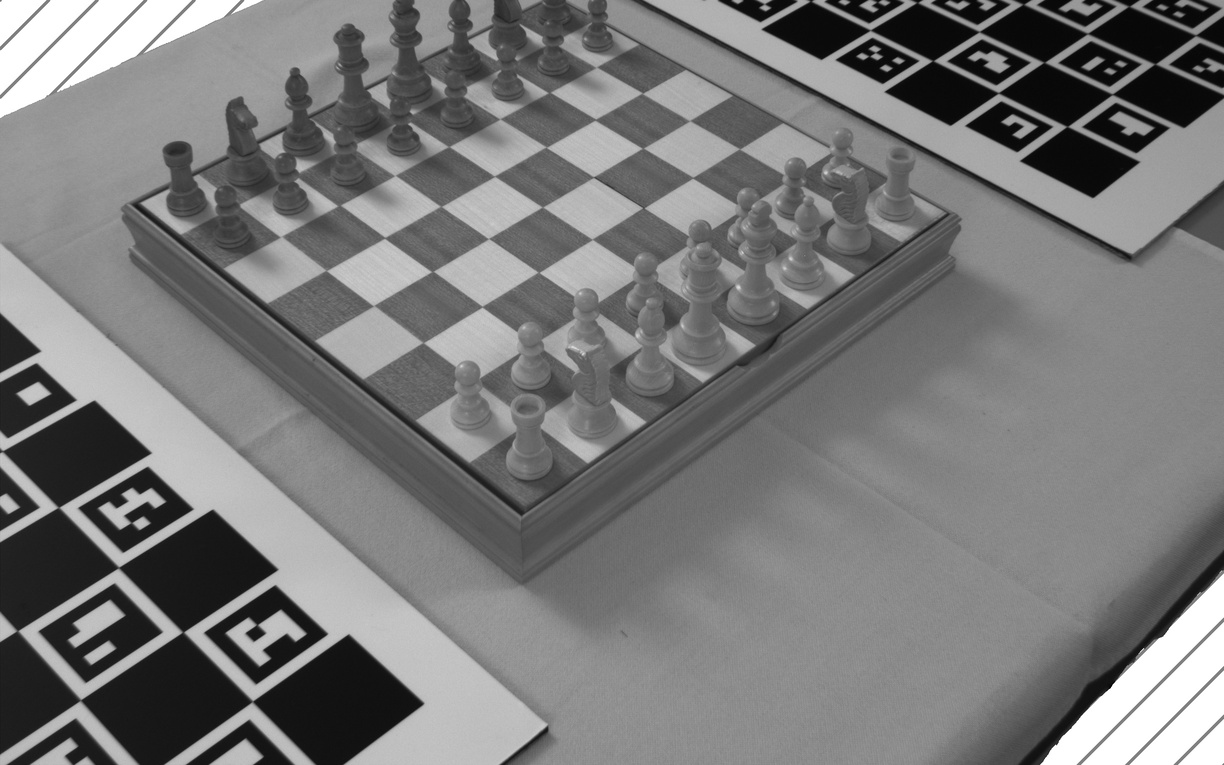}
            \end{subfigure}
            \begin{subfigure}{0.325\linewidth}
                \centering
                \includegraphics[width=\linewidth]{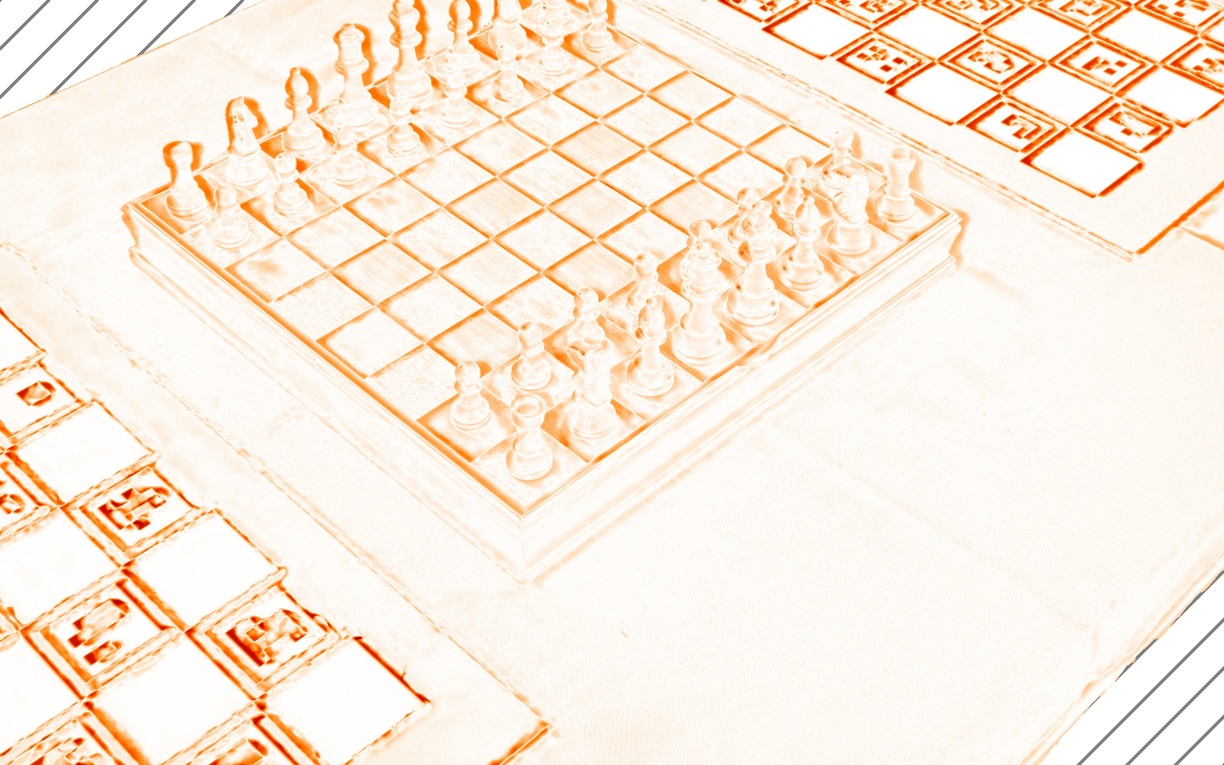}
            \end{subfigure}
        \end{minipage}
    \end{minipage}
    \begin{minipage}{\linewidth}
        \begin{minipage}{0.025\linewidth}
            \vfill
            \rotatebox[origin=cb]{90}{MS}
            \vfill
        \end{minipage}
        \begin{minipage}{0.97\linewidth}
            \begin{subfigure}{0.325\linewidth}
                \centering
                \includegraphics[width=\linewidth]{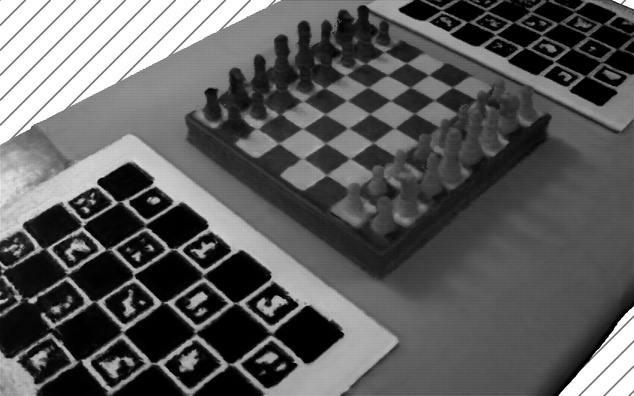}
            \end{subfigure}
            \begin{subfigure}{0.325\linewidth}
                \centering
                \includegraphics[width=\linewidth]{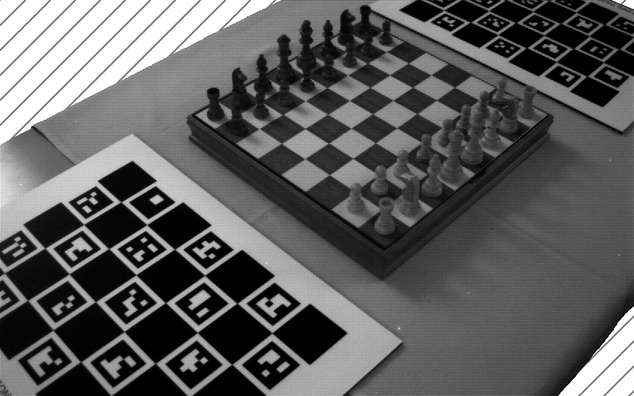}
            \end{subfigure}
            \begin{subfigure}{0.325\linewidth}
                \centering
                \includegraphics[width=\linewidth]{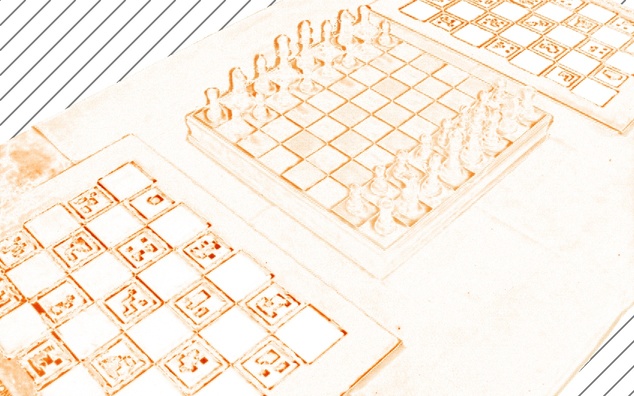}
            \end{subfigure}
        \end{minipage}
    \end{minipage}
    \caption{Qualitative renderings of the ``Chess'' scene from the pre-training step with all-modality supervision. All frames are mosaicked, except RGB frames. RGB is demosaicked only for visualization purposes.}
    \label{fig:pt_chess}
    \vspace{-0.5cm}
\end{figure*}

\begin{figure*}[t]
    \centering
    \begin{minipage}{\linewidth}
        \centering
        \hfill
        \begin{minipage}{0.96\linewidth}
            \begin{minipage}{0.325\linewidth}
                \centering
                Ours
            \end{minipage}
            \begin{minipage}{0.325\linewidth}
                \centering
                GT
            \end{minipage}
            \begin{minipage}{0.325\linewidth}
                \centering
                Error
            \end{minipage}
        \end{minipage}
        \vspace{2pt}
    \end{minipage}
    \begin{minipage}{\linewidth}
        \begin{minipage}{0.025\linewidth}
            \vfill
            \rotatebox[origin=cb]{90}{RGB}
            \vfill
        \end{minipage}
        \begin{minipage}{0.97\linewidth}
            \begin{subfigure}{0.325\linewidth}
                \centering
                \includegraphics[width=\linewidth]{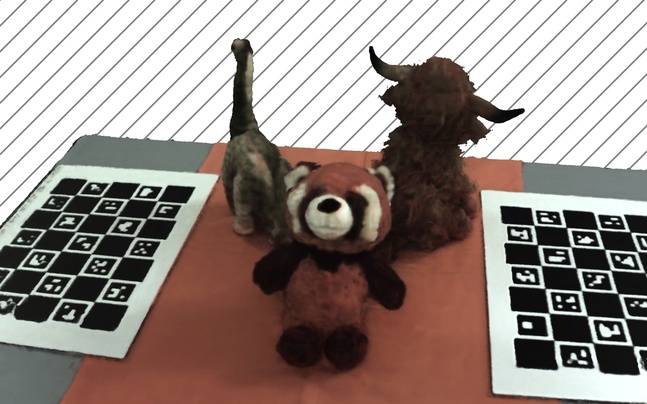}
            \end{subfigure}
            \begin{subfigure}{0.325\linewidth}
                \centering
                \includegraphics[width=\linewidth]{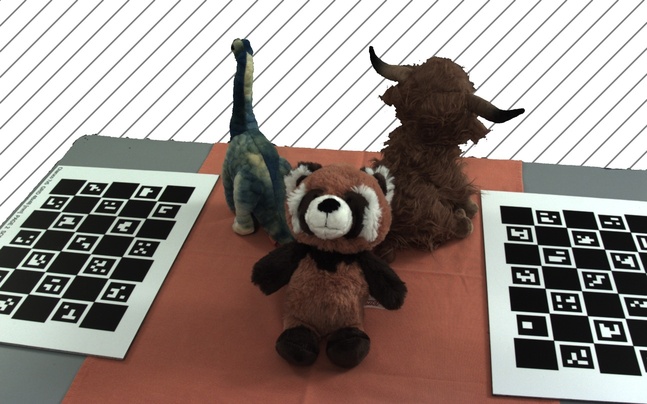}
            \end{subfigure}
            \begin{subfigure}{0.325\linewidth}
                \centering
                \includegraphics[width=\linewidth]{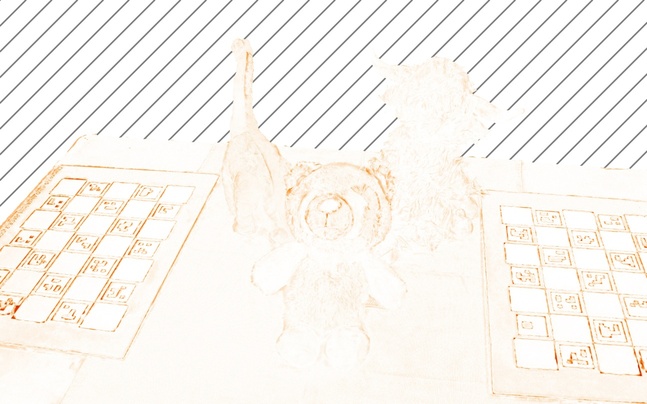}
            \end{subfigure}
        \end{minipage}
    \end{minipage}
    \begin{minipage}{\linewidth}
        \begin{minipage}{0.025\linewidth}
            \vfill
            \rotatebox[origin=cb]{90}{NIR}
            \vfill
        \end{minipage}
        \begin{minipage}{0.97\linewidth}
            \begin{subfigure}{0.325\linewidth}
                \centering
                \includegraphics[width=\linewidth]{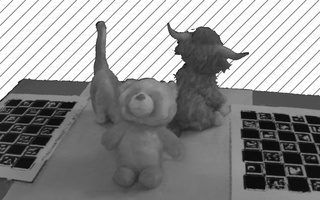}
            \end{subfigure}
            \begin{subfigure}{0.325\linewidth}
                \centering
                \includegraphics[width=\linewidth]{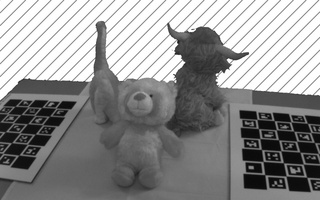}
            \end{subfigure}
            \begin{subfigure}{0.325\linewidth}
                \centering
                \includegraphics[width=\linewidth]{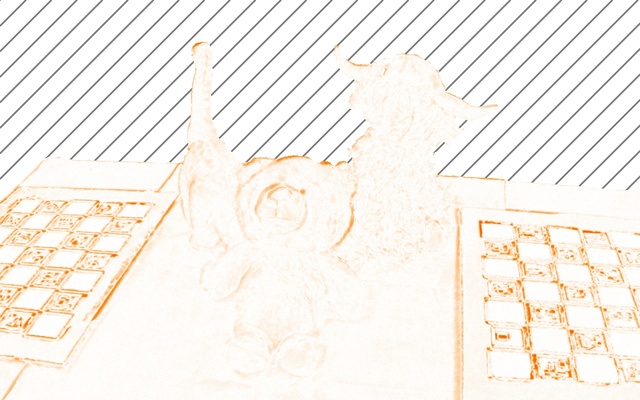}
            \end{subfigure}
        \end{minipage}
    \end{minipage}
    \begin{minipage}{\linewidth}
        \begin{minipage}{0.025\linewidth}
            \vfill
            \rotatebox[origin=cb]{90}{Mono}
            \vfill
        \end{minipage}
        \begin{minipage}{0.97\linewidth}
            \begin{subfigure}{0.325\linewidth}
                \centering
                \includegraphics[width=\linewidth]{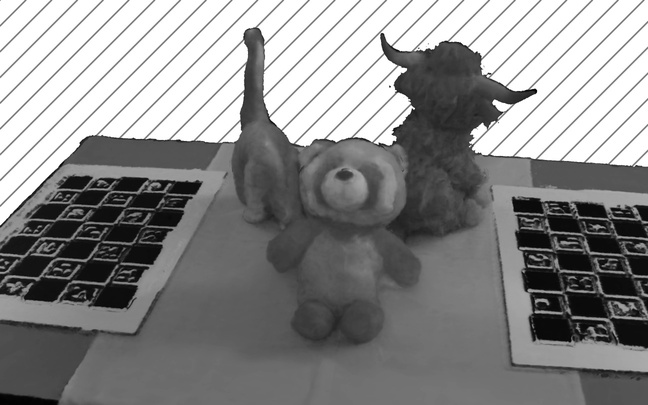}
            \end{subfigure}
            \begin{subfigure}{0.325\linewidth}
                \centering
                \includegraphics[width=\linewidth]{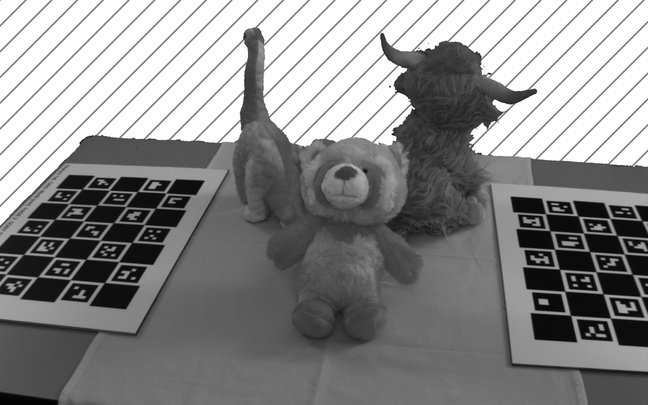}
            \end{subfigure}
            \begin{subfigure}{0.325\linewidth}
                \centering
                \includegraphics[width=\linewidth]{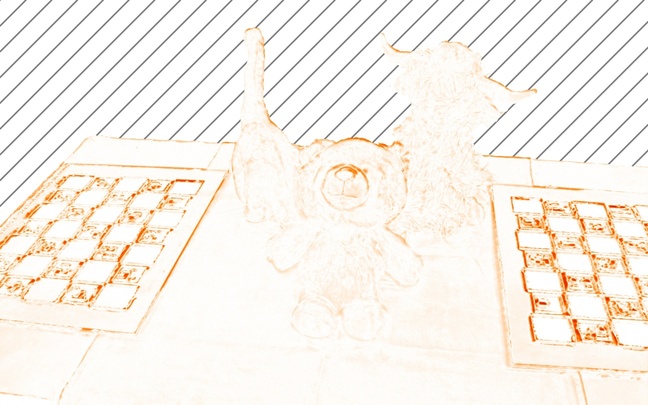}
            \end{subfigure}
        \end{minipage}
    \end{minipage}
    \begin{minipage}{\linewidth}
        \begin{minipage}{0.025\linewidth}
            \vfill
            \rotatebox[origin=cb]{90}{Pol}
            \vfill
        \end{minipage}
        \begin{minipage}{0.97\linewidth}
            \begin{subfigure}{0.325\linewidth}
                \centering
                \includegraphics[width=\linewidth]{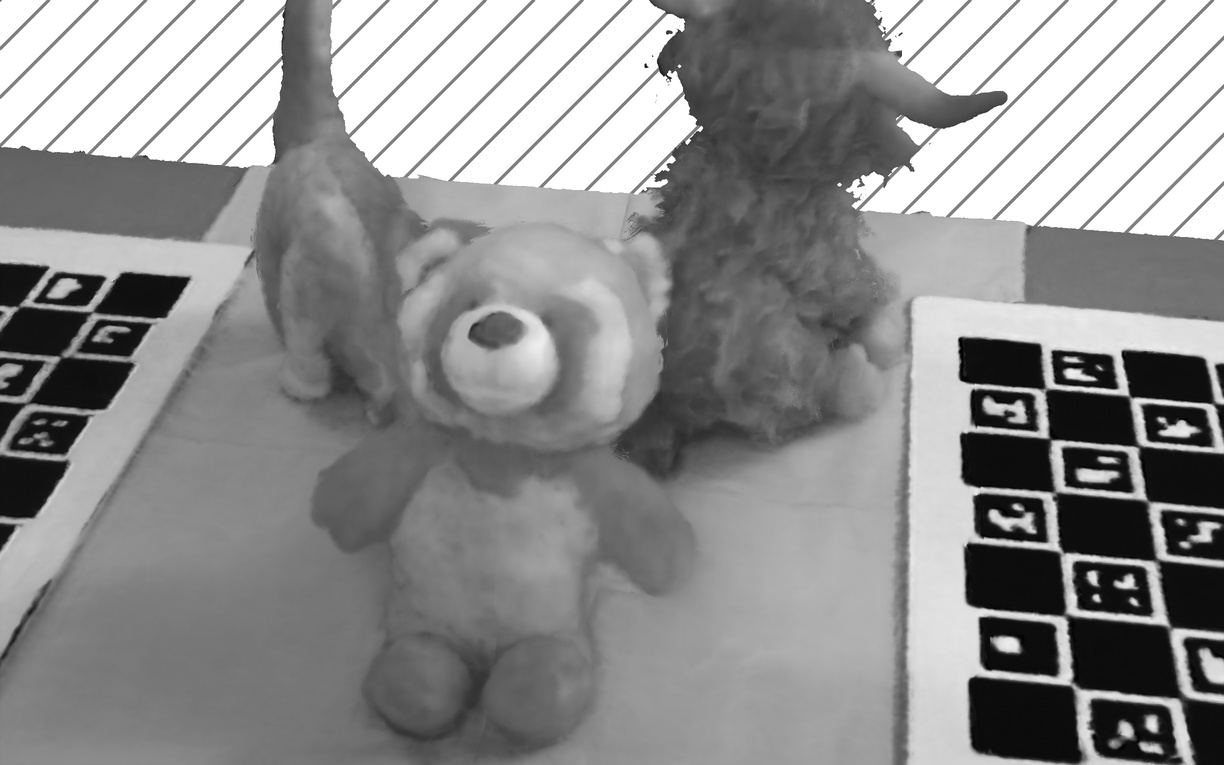}
            \end{subfigure}
            \begin{subfigure}{0.325\linewidth}
                \centering
                \includegraphics[width=\linewidth]{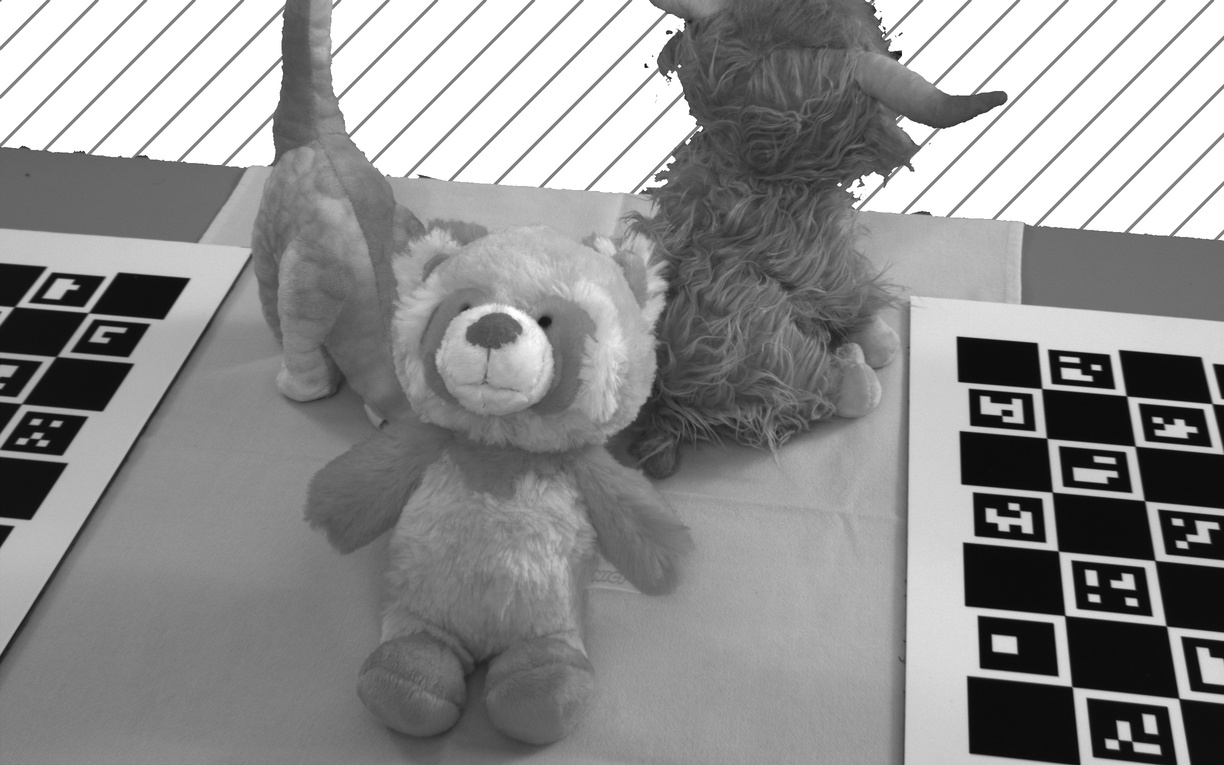}
            \end{subfigure}
            \begin{subfigure}{0.325\linewidth}
                \centering
                \includegraphics[width=\linewidth]{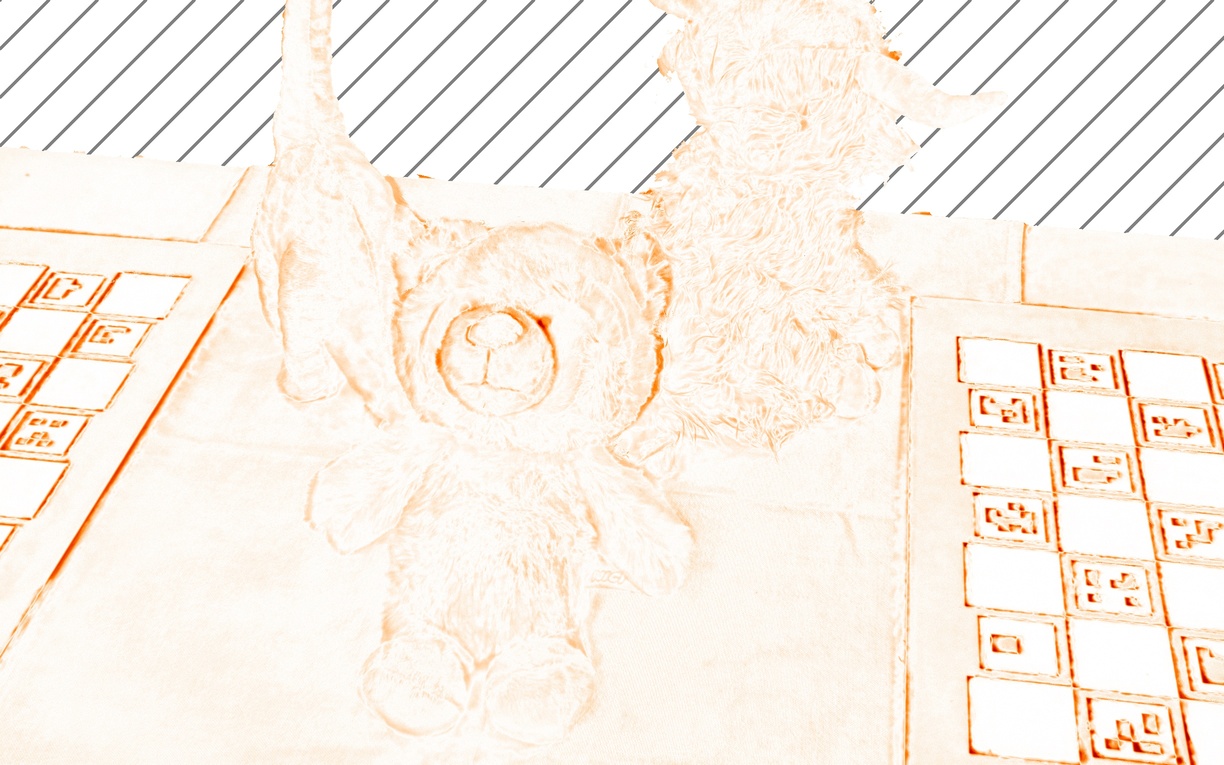}
            \end{subfigure}
        \end{minipage}
    \end{minipage}
    \begin{minipage}{\linewidth}
        \begin{minipage}{0.025\linewidth}
            \vfill
            \rotatebox[origin=cb]{90}{MS}
            \vfill
        \end{minipage}
        \begin{minipage}{0.97\linewidth}
            \begin{subfigure}{0.325\linewidth}
                \centering
                \includegraphics[width=\linewidth]{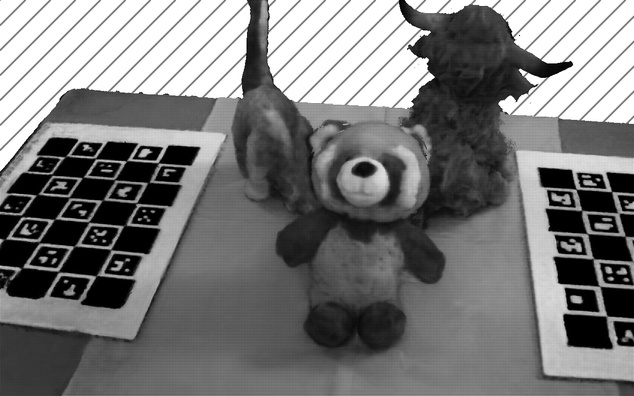}
            \end{subfigure}
            \begin{subfigure}{0.325\linewidth}
                \centering
                \includegraphics[width=\linewidth]{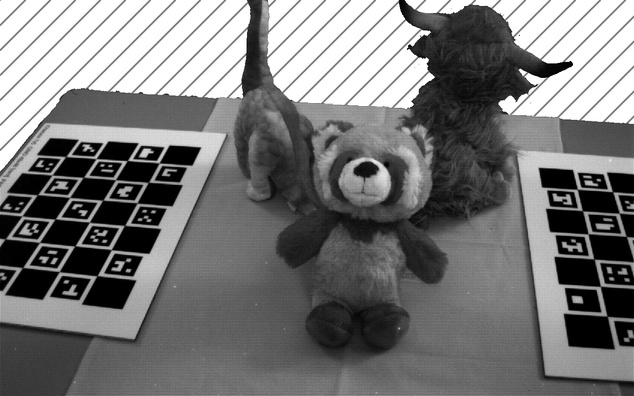}
            \end{subfigure}
            \begin{subfigure}{0.325\linewidth}
                \centering
                \includegraphics[width=\linewidth]{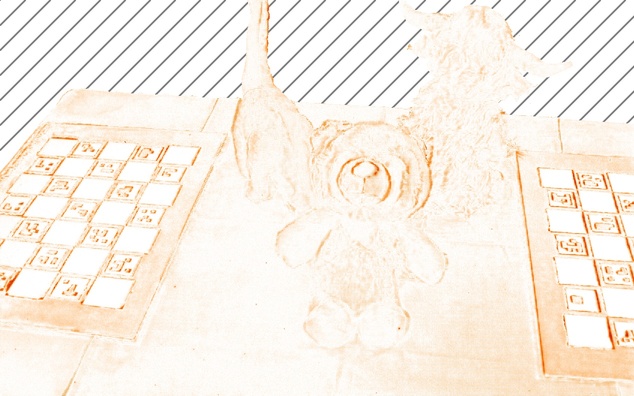}
            \end{subfigure}
        \end{minipage}
    \end{minipage}
    \caption{Qualitative renderings of the ``Forestgang 1'' scene from the pre-training step with all-modality supervision. All frames are mosaicked, except RGB frames. RGB is demosaicked only for visualization purposes.}
    \label{fig:pt_forestgang1}
    \vspace{-0.5cm}
\end{figure*}

\begin{figure*}[t]
    \centering
    \begin{minipage}{\linewidth}
        \centering
        \hfill
        \begin{minipage}{0.96\linewidth}
            \begin{minipage}{0.325\linewidth}
                \centering
                Ours
            \end{minipage}
            \begin{minipage}{0.325\linewidth}
                \centering
                GT
            \end{minipage}
            \begin{minipage}{0.325\linewidth}
                \centering
                Error
            \end{minipage}
        \end{minipage}
        \vspace{2pt}
    \end{minipage}
    \begin{minipage}{\linewidth}
        \begin{minipage}{0.025\linewidth}
            \vfill
            \rotatebox[origin=cb]{90}{RGB}
            \vfill
        \end{minipage}
        \begin{minipage}{0.97\linewidth}
            \begin{subfigure}{0.325\linewidth}
                \centering
                \includegraphics[width=\linewidth]{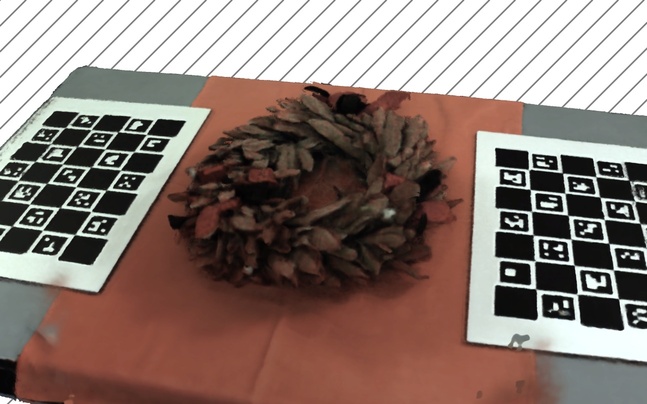}
            \end{subfigure}
            \begin{subfigure}{0.325\linewidth}
                \centering
                \includegraphics[width=\linewidth]{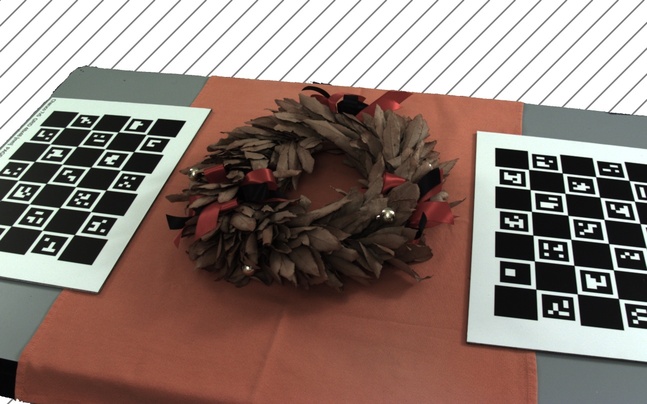}
            \end{subfigure}
            \begin{subfigure}{0.325\linewidth}
                \centering
                \includegraphics[width=\linewidth]{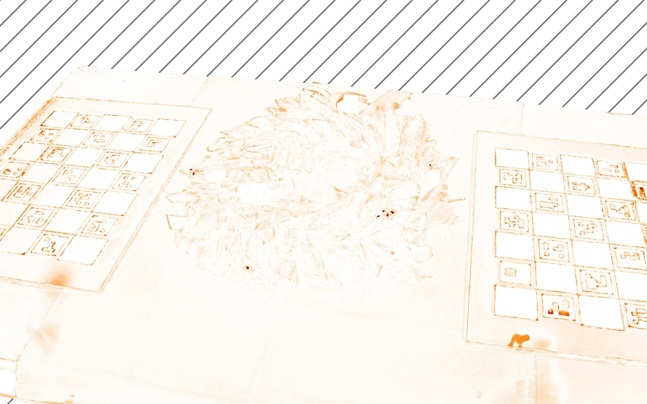}
            \end{subfigure}
        \end{minipage}
    \end{minipage}
    \begin{minipage}{\linewidth}
        \begin{minipage}{0.025\linewidth}
            \vfill
            \rotatebox[origin=cb]{90}{NIR}
            \vfill
        \end{minipage}
        \begin{minipage}{0.97\linewidth}
            \begin{subfigure}{0.325\linewidth}
                \centering
                \includegraphics[width=\linewidth]{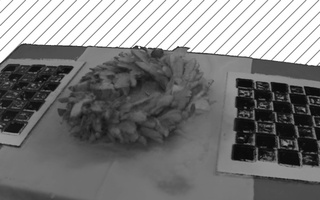}
            \end{subfigure}
            \begin{subfigure}{0.325\linewidth}
                \centering
                \includegraphics[width=\linewidth]{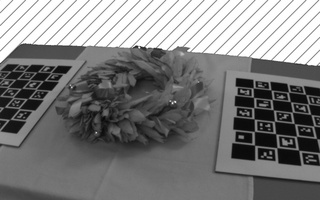}
            \end{subfigure}
            \begin{subfigure}{0.325\linewidth}
                \centering
                \includegraphics[width=\linewidth]{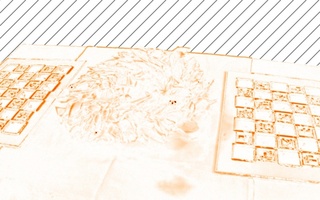}
            \end{subfigure}
        \end{minipage}
    \end{minipage}
    \begin{minipage}{\linewidth}
        \begin{minipage}{0.025\linewidth}
            \vfill
            \rotatebox[origin=cb]{90}{Mono}
            \vfill
        \end{minipage}
        \begin{minipage}{0.97\linewidth}
            \begin{subfigure}{0.325\linewidth}
                \centering
                \includegraphics[width=\linewidth]{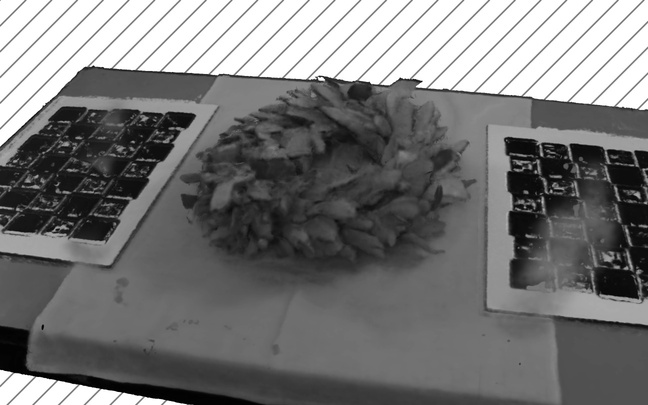}
            \end{subfigure}
            \begin{subfigure}{0.325\linewidth}
                \centering
                \includegraphics[width=\linewidth]{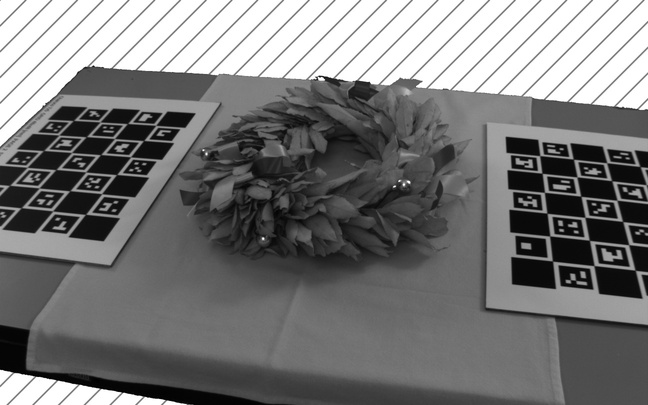}
            \end{subfigure}
            \begin{subfigure}{0.325\linewidth}
                \centering
                \includegraphics[width=\linewidth]{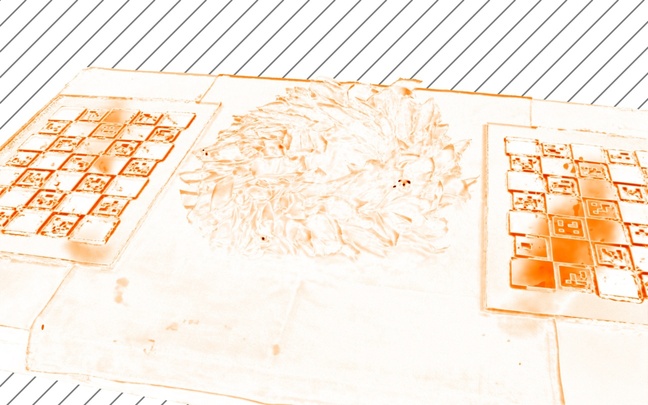}
            \end{subfigure}
        \end{minipage}
    \end{minipage}
    \begin{minipage}{\linewidth}
        \begin{minipage}{0.025\linewidth}
            \vfill
            \rotatebox[origin=cb]{90}{Pol}
            \vfill
        \end{minipage}
        \begin{minipage}{0.97\linewidth}
            \begin{subfigure}{0.325\linewidth}
                \centering
                \includegraphics[width=\linewidth]{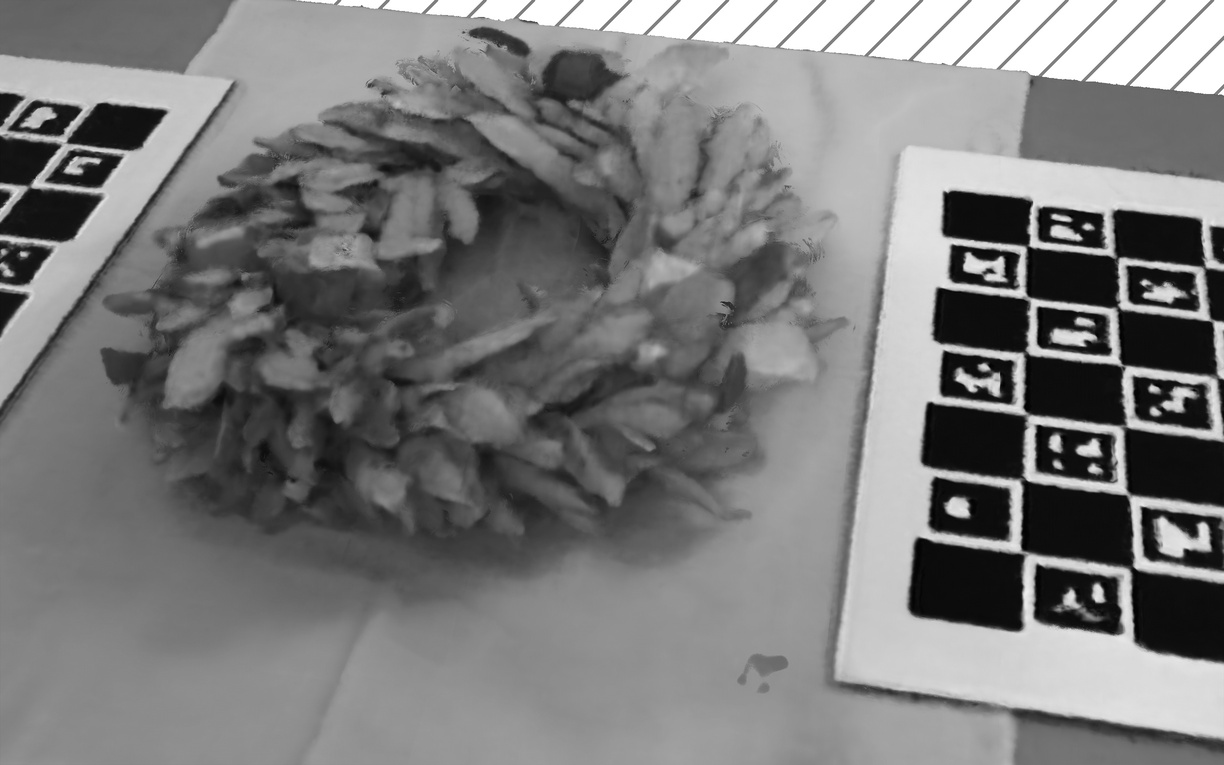}
            \end{subfigure}
            \begin{subfigure}{0.325\linewidth}
                \centering
                \includegraphics[width=\linewidth]{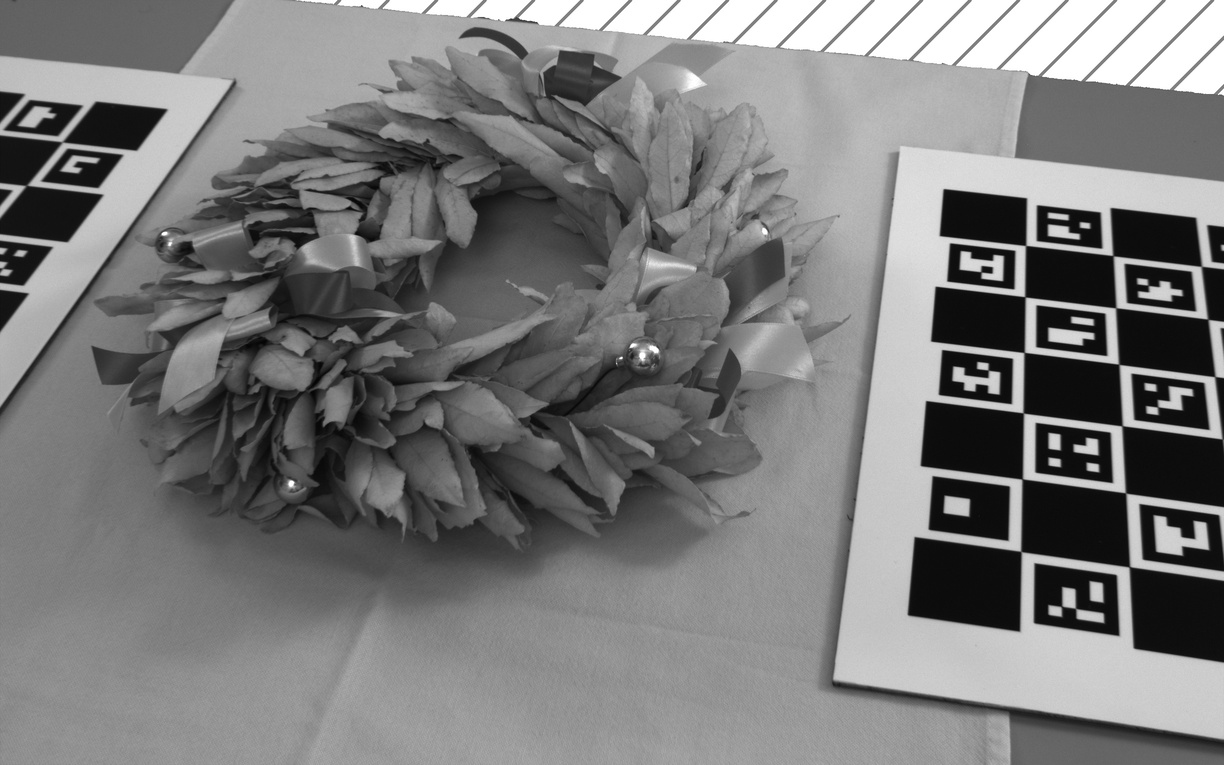}
            \end{subfigure}
            \begin{subfigure}{0.325\linewidth}
                \centering
                \includegraphics[width=\linewidth]{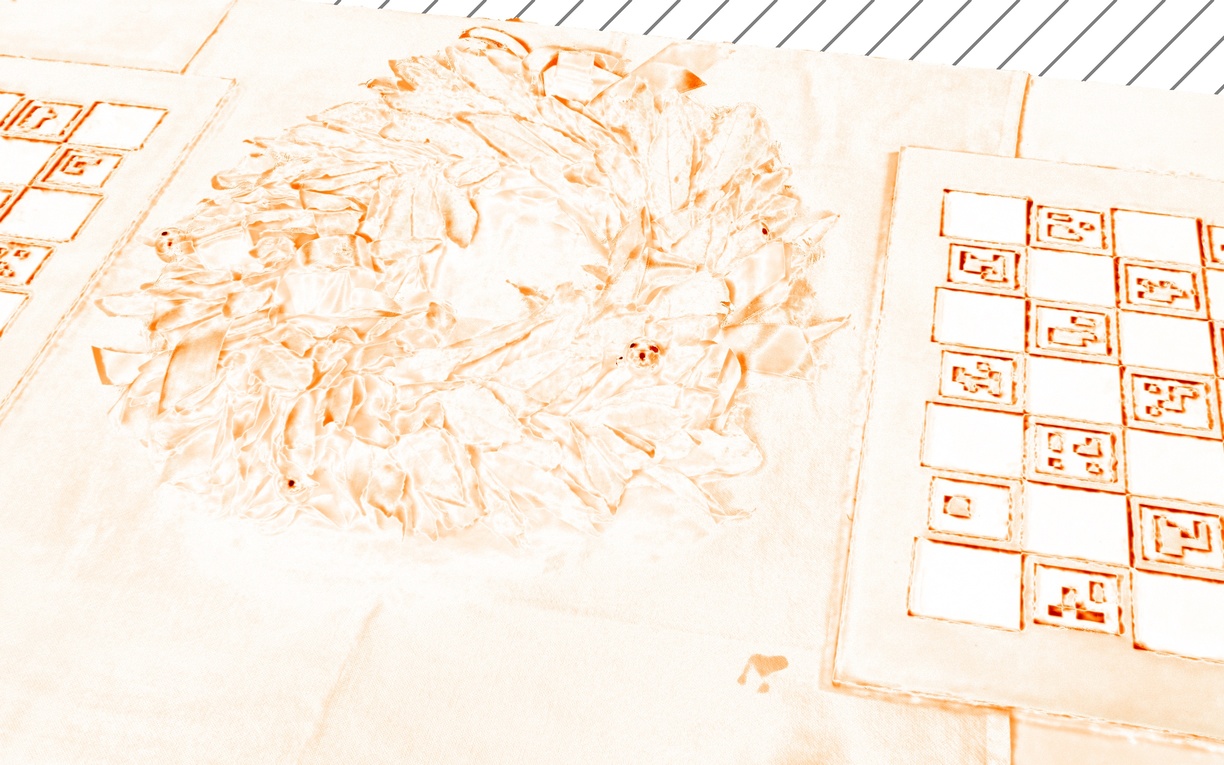}
            \end{subfigure}
        \end{minipage}
    \end{minipage}
    \begin{minipage}{\linewidth}
        \begin{minipage}{0.025\linewidth}
            \vfill
            \rotatebox[origin=cb]{90}{MS}
            \vfill
        \end{minipage}
        \begin{minipage}{0.97\linewidth}
            \begin{subfigure}{0.325\linewidth}
                \centering
                \includegraphics[width=\linewidth]{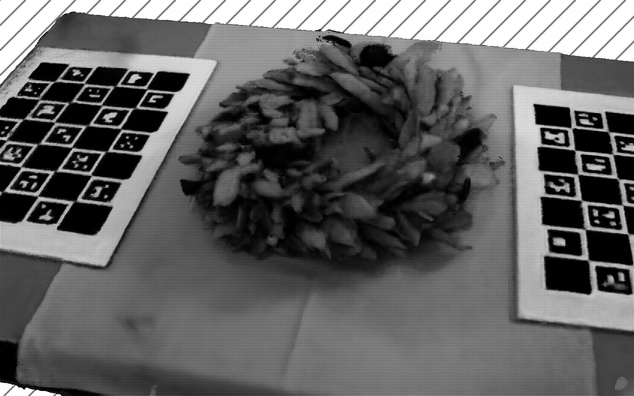}
            \end{subfigure}
            \begin{subfigure}{0.325\linewidth}
                \centering
                \includegraphics[width=\linewidth]{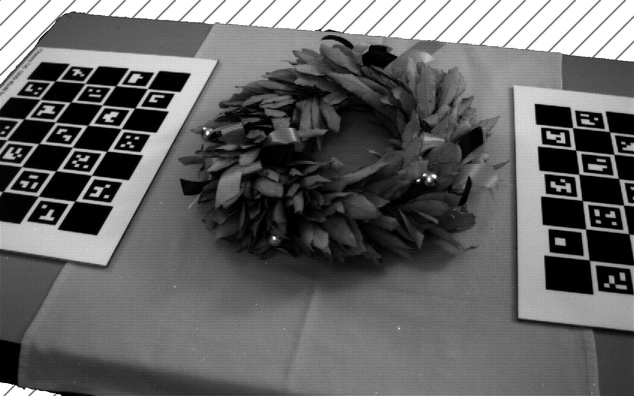}
            \end{subfigure}
            \begin{subfigure}{0.325\linewidth}
                \centering
                \includegraphics[width=\linewidth]{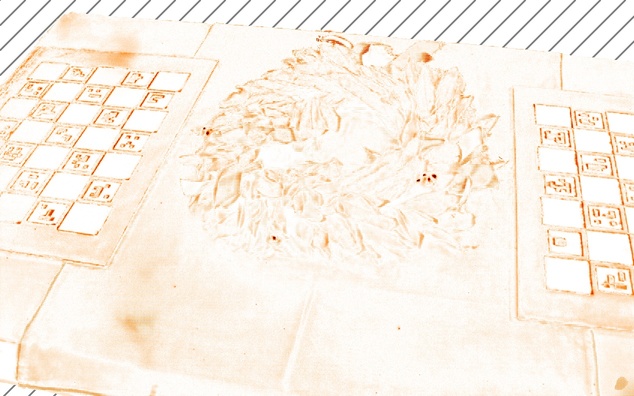}
            \end{subfigure}
        \end{minipage}
    \end{minipage}
    \caption{Qualitative renderings of the ``Laurelwreath'' scene from the pre-training step with all-modality supervision. All frames are mosaicked, except RGB frames. RGB is demosaicked only for visualization purposes.}
    \label{fig:pt_laurelwreath}
    \vspace{-0.5cm}
\end{figure*}

\begin{figure*}[t]
    \centering
    \begin{minipage}{\linewidth}
        \centering
        \hfill
        \begin{minipage}{0.96\linewidth}
            \begin{minipage}{0.325\linewidth}
                \centering
                Ours
            \end{minipage}
            \begin{minipage}{0.325\linewidth}
                \centering
                GT
            \end{minipage}
            \begin{minipage}{0.325\linewidth}
                \centering
                Error
            \end{minipage}
        \end{minipage}
        \vspace{2pt}
    \end{minipage}
    \begin{minipage}{\linewidth}
        \begin{minipage}{0.025\linewidth}
            \vfill
            \rotatebox[origin=cb]{90}{RGB}
            \vfill
        \end{minipage}
        \begin{minipage}{0.97\linewidth}
            \begin{subfigure}{0.325\linewidth}
                \centering
                \includegraphics[width=\linewidth]{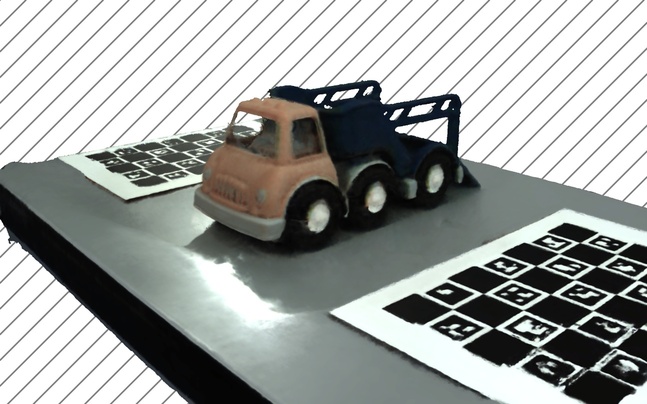}
            \end{subfigure}
            \begin{subfigure}{0.325\linewidth}
                \centering
                \includegraphics[width=\linewidth]{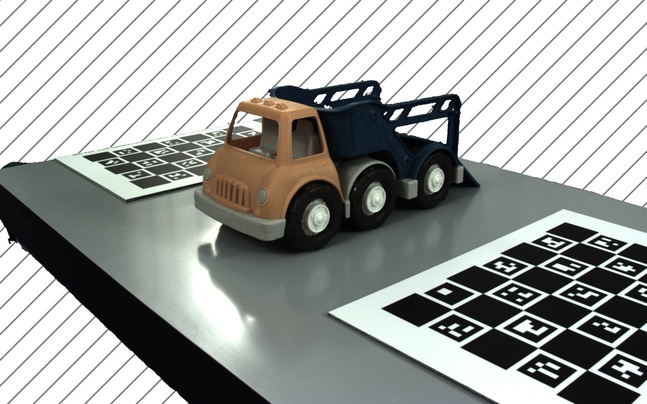}
            \end{subfigure}
            \begin{subfigure}{0.325\linewidth}
                \centering
                \includegraphics[width=\linewidth]{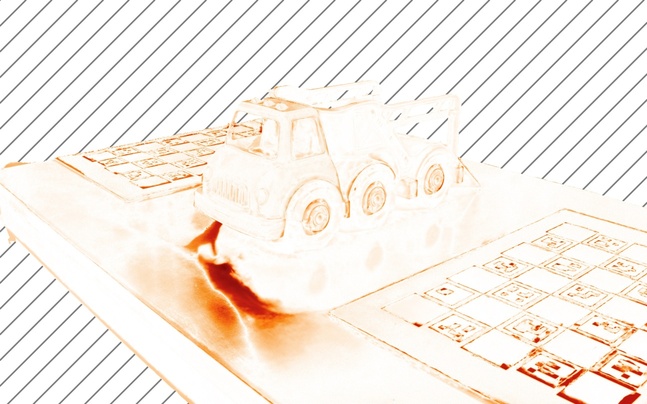}
            \end{subfigure}
        \end{minipage}
    \end{minipage}
    \begin{minipage}{\linewidth}
        \begin{minipage}{0.025\linewidth}
            \vfill
            \rotatebox[origin=cb]{90}{NIR}
            \vfill
        \end{minipage}
        \begin{minipage}{0.97\linewidth}
            \begin{subfigure}{0.325\linewidth}
                \centering
                \includegraphics[width=\linewidth]{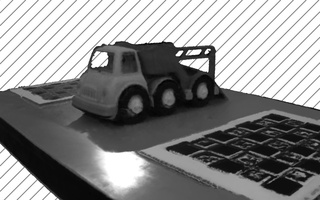}
            \end{subfigure}
            \begin{subfigure}{0.325\linewidth}
                \centering
                \includegraphics[width=\linewidth]{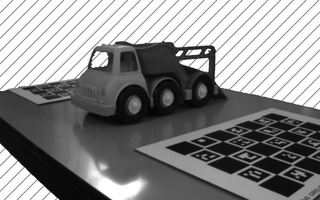}
            \end{subfigure}
            \begin{subfigure}{0.325\linewidth}
                \centering
                \includegraphics[width=\linewidth]{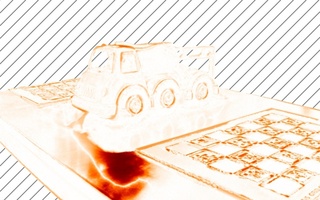}
            \end{subfigure}
        \end{minipage}
    \end{minipage}
    \begin{minipage}{\linewidth}
        \begin{minipage}{0.025\linewidth}
            \vfill
            \rotatebox[origin=cb]{90}{Mono}
            \vfill
        \end{minipage}
        \begin{minipage}{0.97\linewidth}
            \begin{subfigure}{0.325\linewidth}
                \centering
                \includegraphics[width=\linewidth]{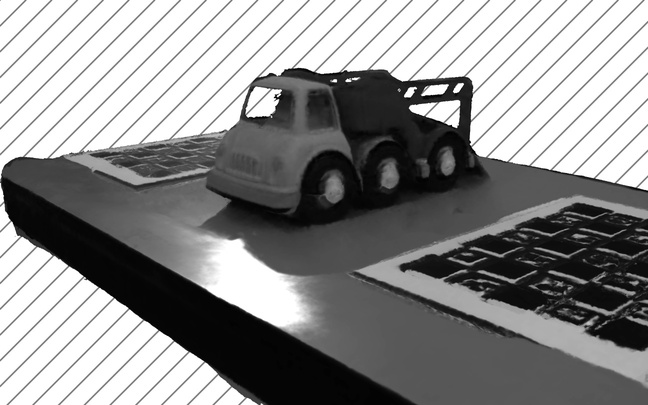}
            \end{subfigure}
            \begin{subfigure}{0.325\linewidth}
                \centering
                \includegraphics[width=\linewidth]{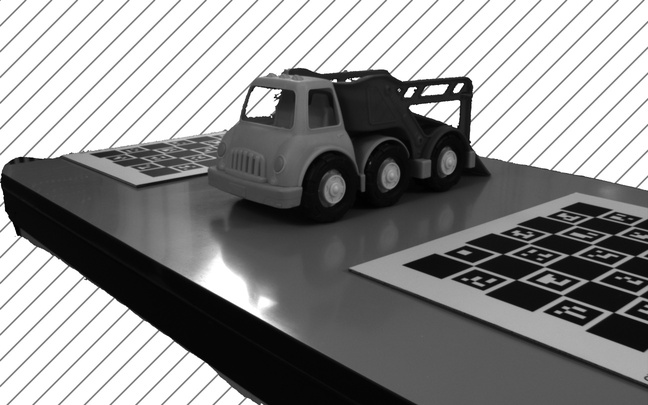}
            \end{subfigure}
            \begin{subfigure}{0.325\linewidth}
                \centering
                \includegraphics[width=\linewidth]{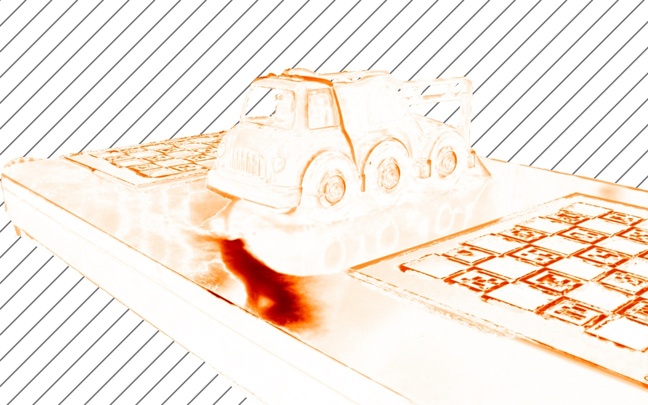}
            \end{subfigure}
        \end{minipage}
    \end{minipage}
    \begin{minipage}{\linewidth}
        \begin{minipage}{0.025\linewidth}
            \vfill
            \rotatebox[origin=cb]{90}{Pol}
            \vfill
        \end{minipage}
        \begin{minipage}{0.97\linewidth}
            \begin{subfigure}{0.325\linewidth}
                \centering
                \includegraphics[width=\linewidth]{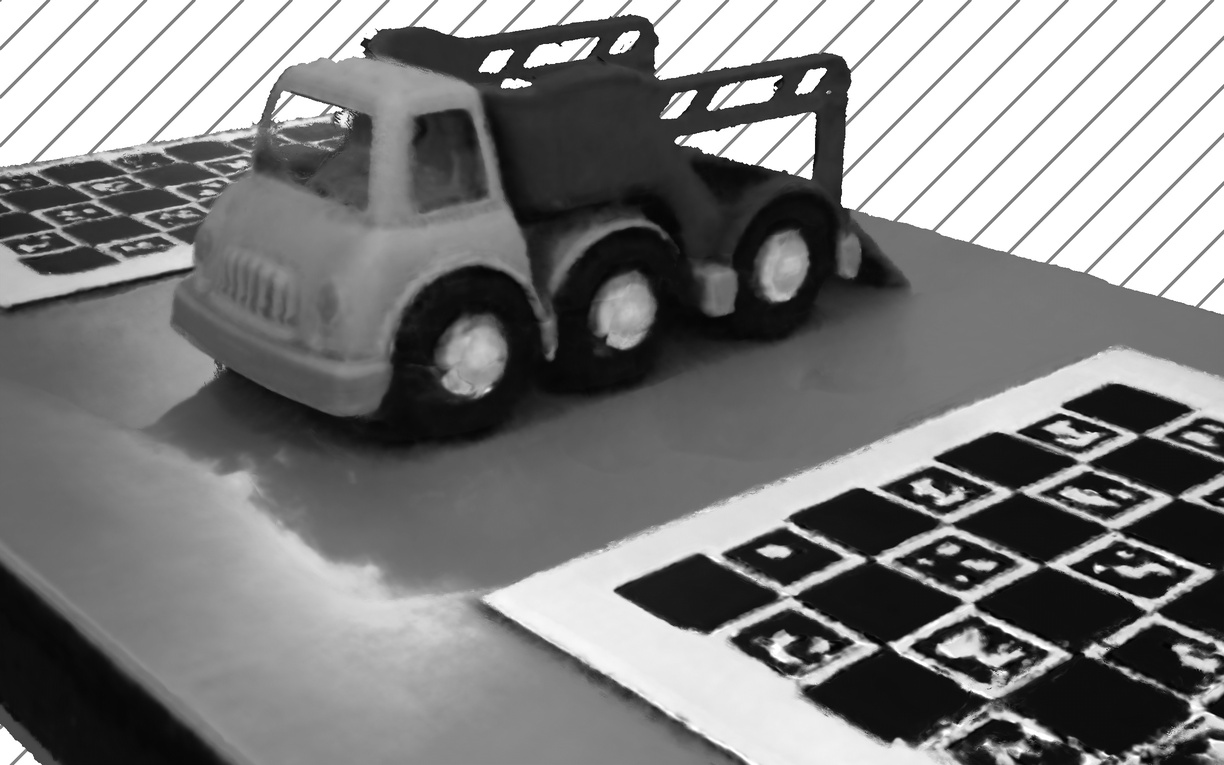}
            \end{subfigure}
            \begin{subfigure}{0.325\linewidth}
                \centering
                \includegraphics[width=\linewidth]{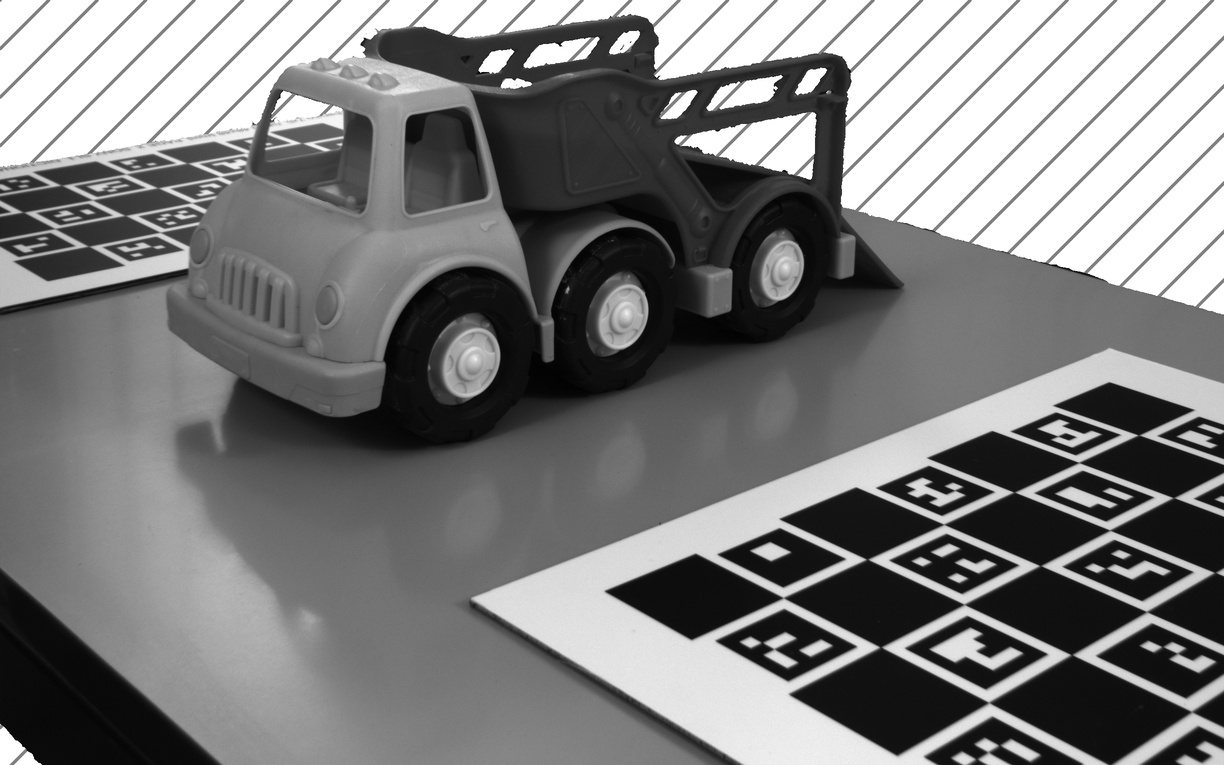}
            \end{subfigure}
            \begin{subfigure}{0.325\linewidth}
                \centering
                \includegraphics[width=\linewidth]{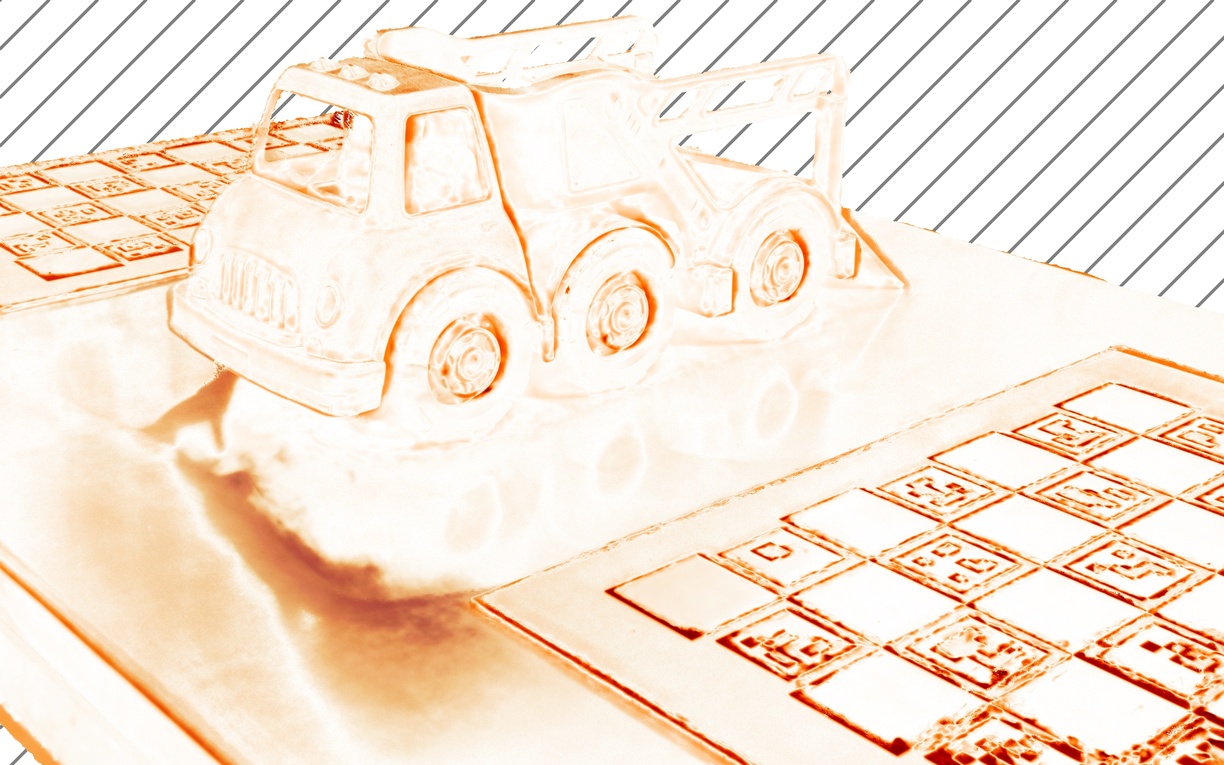}
            \end{subfigure}
        \end{minipage}
    \end{minipage}
    \begin{minipage}{\linewidth}
        \begin{minipage}{0.025\linewidth}
            \vfill
            \rotatebox[origin=cb]{90}{MS}
            \vfill
        \end{minipage}
        \begin{minipage}{0.97\linewidth}
            \begin{subfigure}{0.325\linewidth}
                \centering
                \includegraphics[width=\linewidth]{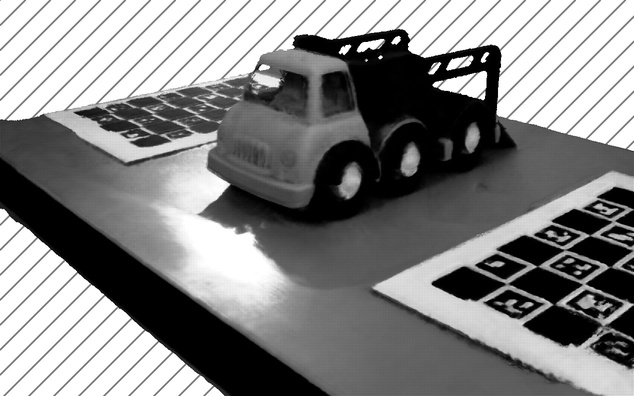}
            \end{subfigure}
            \begin{subfigure}{0.325\linewidth}
                \centering
                \includegraphics[width=\linewidth]{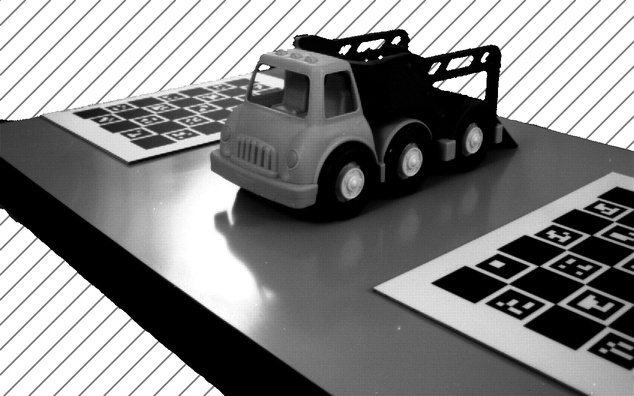}
            \end{subfigure}
            \begin{subfigure}{0.325\linewidth}
                \centering
                \includegraphics[width=\linewidth]{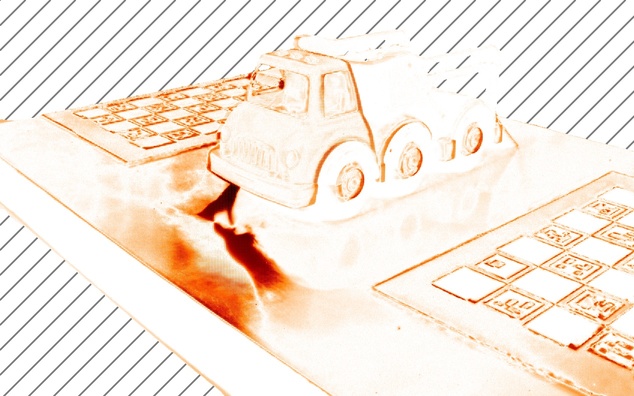}
            \end{subfigure}
        \end{minipage}
    \end{minipage}
    \caption{Qualitative renderings of the ``Truck'' scene from the pre-training step with all-modality supervision. All frames are mosaicked, except RGB frames. RGB is demosaicked only for visualization purposes.}
    \label{fig:pt_truck}
    \vspace{-0.5cm}
\end{figure*}

\begin{figure*}[t]
    \centering
    \begin{minipage}{\linewidth}
        \centering
        \hfill
        \begin{minipage}{0.96\linewidth}
            \begin{minipage}{0.325\linewidth}
                \centering
                Ours
            \end{minipage}
            \begin{minipage}{0.325\linewidth}
                \centering
                GT
            \end{minipage}
            \begin{minipage}{0.325\linewidth}
                \centering
                Error
            \end{minipage}
        \end{minipage}
        \vspace{2pt}
    \end{minipage}
    \begin{minipage}{\linewidth}
        \begin{minipage}{0.025\linewidth}
            \vfill
            \rotatebox[origin=cb]{90}{RGB}
            \vfill
        \end{minipage}
        \begin{minipage}{0.97\linewidth}
            \begin{subfigure}{0.325\linewidth}
                \centering
                \includegraphics[width=\linewidth]{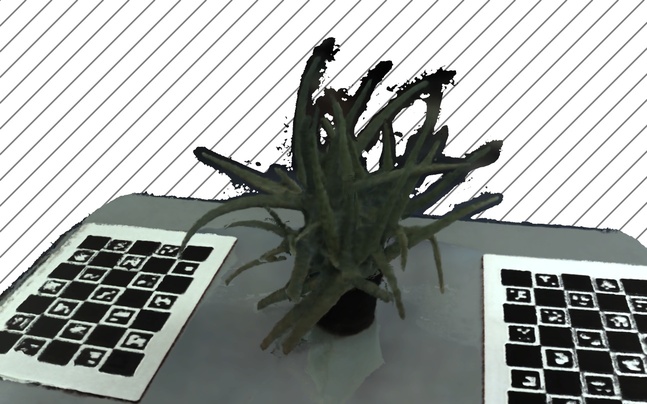}
            \end{subfigure}
            \begin{subfigure}{0.325\linewidth}
                \centering
                \includegraphics[width=\linewidth]{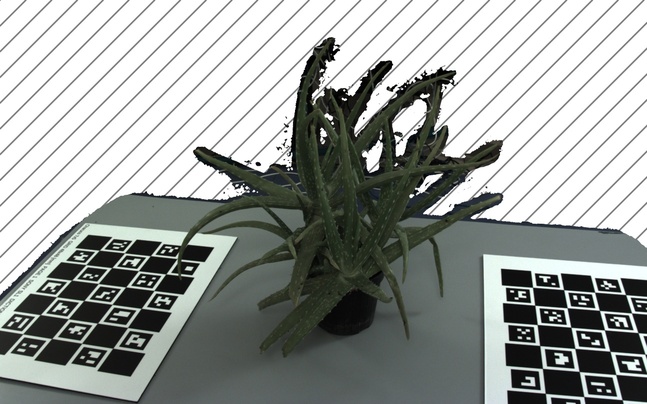}
            \end{subfigure}
            \begin{subfigure}{0.325\linewidth}
                \centering
                \includegraphics[width=\linewidth]{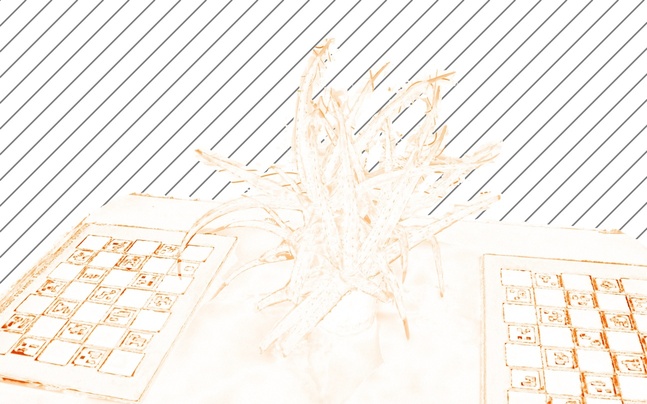}
            \end{subfigure}
        \end{minipage}
    \end{minipage}
    \begin{minipage}{\linewidth}
        \begin{minipage}{0.025\linewidth}
            \vfill
            \rotatebox[origin=cb]{90}{NIR}
            \vfill
        \end{minipage}
        \begin{minipage}{0.97\linewidth}
            \begin{subfigure}{0.325\linewidth}
                \centering
                \includegraphics[width=\linewidth]{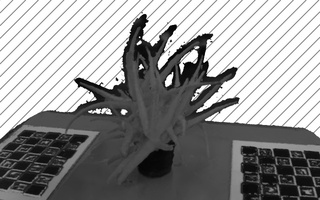}
            \end{subfigure}
            \begin{subfigure}{0.325\linewidth}
                \centering
                \includegraphics[width=\linewidth]{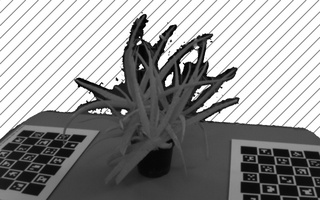}
            \end{subfigure}
            \begin{subfigure}{0.325\linewidth}
                \centering
                \includegraphics[width=\linewidth]{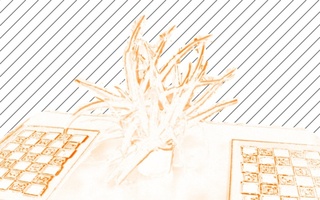}
            \end{subfigure}
        \end{minipage}
    \end{minipage}
    \begin{minipage}{\linewidth}
        \begin{minipage}{0.025\linewidth}
            \vfill
            \rotatebox[origin=cb]{90}{Mono}
            \vfill
        \end{minipage}
        \begin{minipage}{0.97\linewidth}
            \begin{subfigure}{0.325\linewidth}
                \centering
                \includegraphics[width=\linewidth]{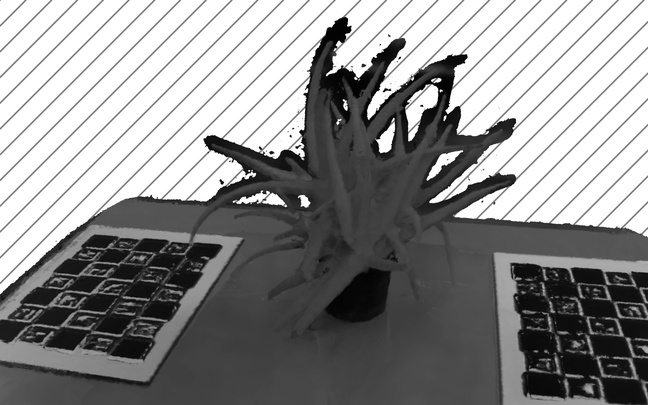}
            \end{subfigure}
            \begin{subfigure}{0.325\linewidth}
                \centering
                \includegraphics[width=\linewidth]{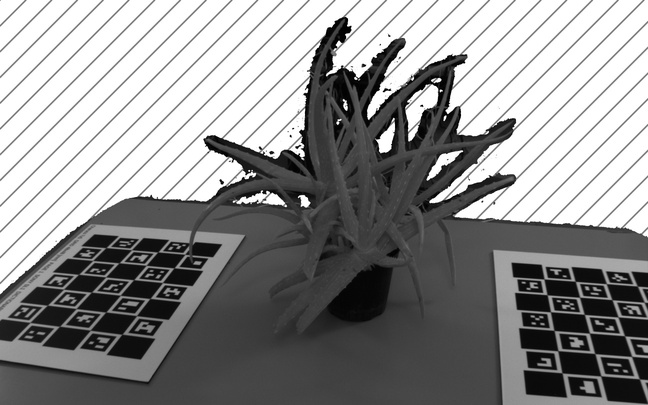}
            \end{subfigure}
            \begin{subfigure}{0.325\linewidth}
                \centering
                \includegraphics[width=\linewidth]{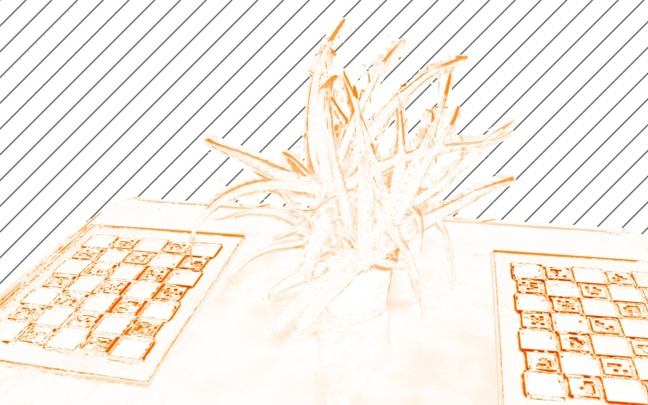}
            \end{subfigure}
        \end{minipage}
    \end{minipage}
    \begin{minipage}{\linewidth}
        \begin{minipage}{0.025\linewidth}
            \vfill
            \rotatebox[origin=cb]{90}{Pol}
            \vfill
        \end{minipage}
        \begin{minipage}{0.97\linewidth}
            \begin{subfigure}{0.325\linewidth}
                \centering
                \includegraphics[width=\linewidth]{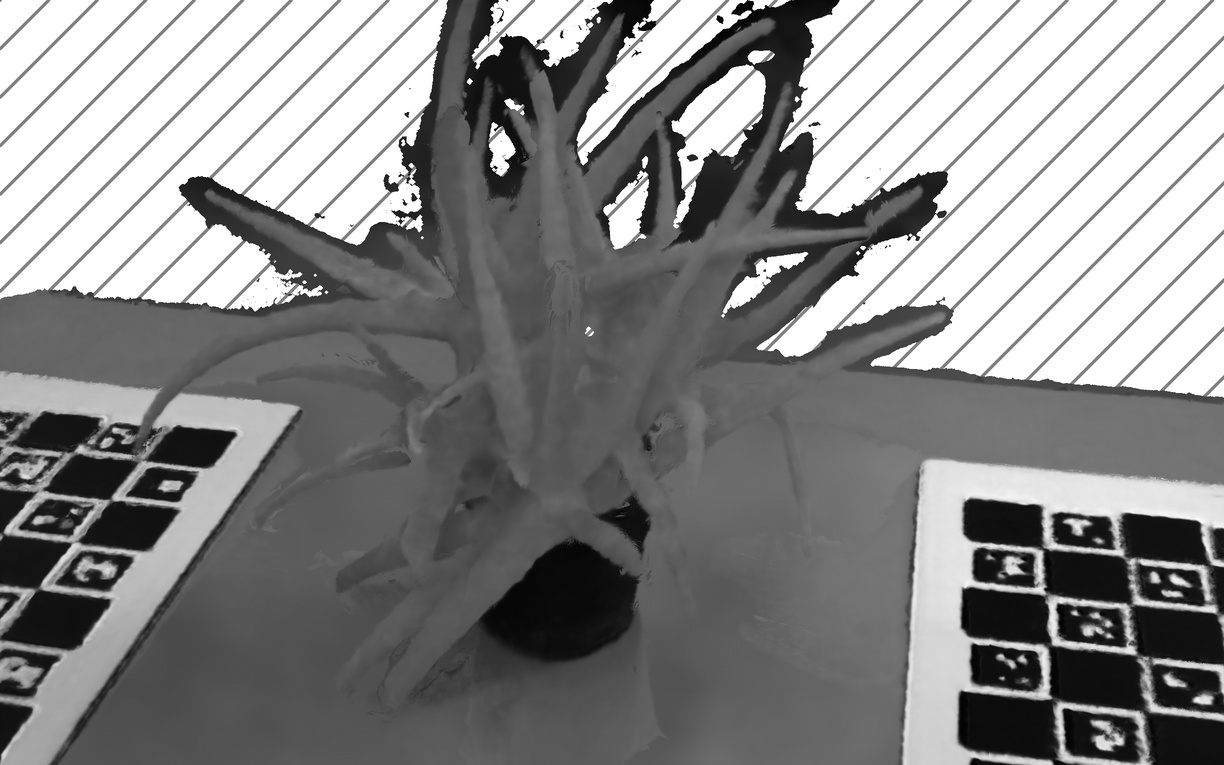}
            \end{subfigure}
            \begin{subfigure}{0.325\linewidth}
                \centering
                \includegraphics[width=\linewidth]{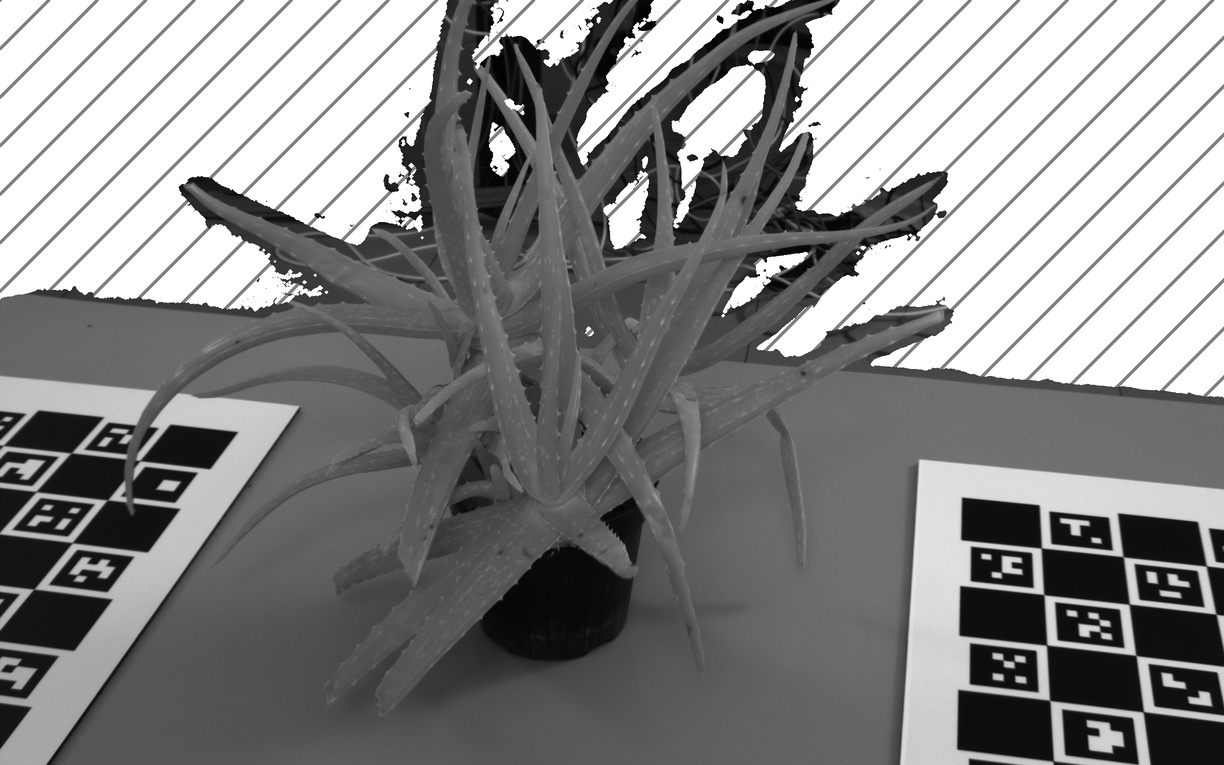}
            \end{subfigure}
            \begin{subfigure}{0.325\linewidth}
                \centering
                \includegraphics[width=\linewidth]{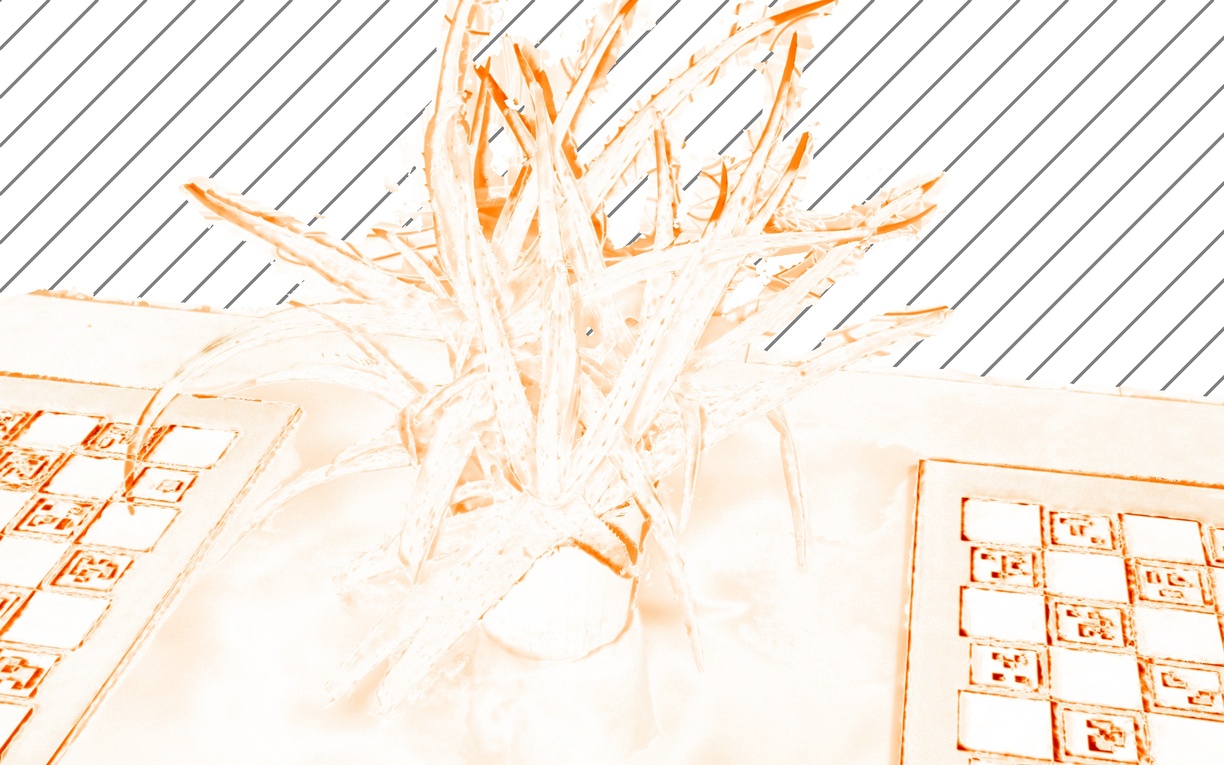}
            \end{subfigure}
        \end{minipage}
    \end{minipage}
    \begin{minipage}{\linewidth}
        \begin{minipage}{0.025\linewidth}
            \vfill
            \rotatebox[origin=cb]{90}{MS}
            \vfill
        \end{minipage}
        \begin{minipage}{0.97\linewidth}
            \begin{subfigure}{0.325\linewidth}
                \centering
                \includegraphics[width=\linewidth]{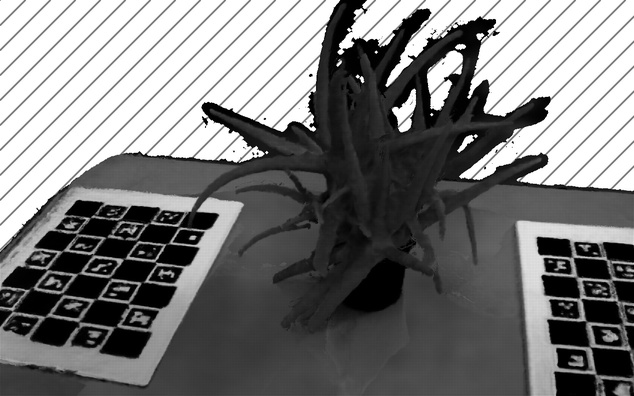}
            \end{subfigure}
            \begin{subfigure}{0.325\linewidth}
                \centering
                \includegraphics[width=\linewidth]{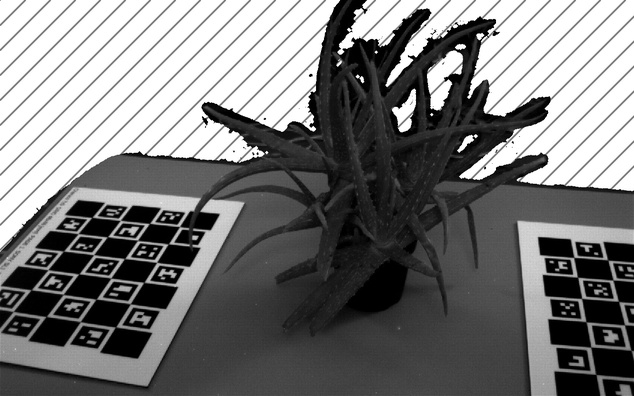}
            \end{subfigure}
            \begin{subfigure}{0.325\linewidth}
                \centering
                \includegraphics[width=\linewidth]{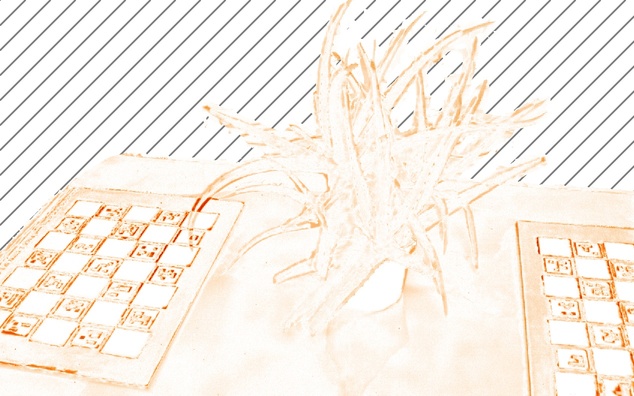}
            \end{subfigure}
        \end{minipage}
    \end{minipage}
    \caption{Qualitative renderings of the ``Aloe'' scene from the pre-training step with all-modality supervision. All frames are mosaicked, except RGB frames. RGB is demosaicked only for visualization purposes.}
    \label{fig:pt_aloe}
    \vspace{-0.5cm}
\end{figure*}

In this section, we report the results of the pre-training phase. It is performed on the 27 MMS-DATA scenes not used for the \gls{ft} step.
In \cref{tab:pt}, we show the \gls{pt} metrics averaged on 5 random \gls{pt} scenes: \ie,  the  ``Chess'', ``Forestgang 1'', ``Laurelwreath'', ``Truck'', and ``Aloe'' scenes, along with some qualitative results in \cref{fig:pt_aloe,fig:pt_chess,fig:pt_forestgang1,fig:pt_laurelwreath,fig:pt_truck}. As mentioned in the main paper, the results of the \gls{ft} step outperform the results of the \gls{pt}. 
The reason is that the focus of the \gls{pt} is to learn a generalized and robust multimodal latent space from several scenes, while the \gls{ft} is optimized to maximize the rendering quality of a single scene. Moreover, even if the per-scene model capacity during the \gls{pt} is the same as in the \gls{ft}, the fact that the shared modules are iteratively optimized during the \gls{pt} makes the model convergence more unstable, and this leads to a lower rendering accuracy.




\end{document}